\documentclass{book}
\usepackage{roboto}

\usepackage[english]{babel}
\usepackage[letterpaper,top=2cm,bottom=2cm,left=3cm,right=3cm,marginparwidth=1.75cm]{geometry}

\usepackage{amsfonts}
\usepackage{pdflscape}
\usepackage{rotating}
\usepackage{longtable}
\usepackage{array}
\usepackage{float}
\usepackage{amsmath}
\usepackage{graphicx}
\usepackage{inconsolata}
\usepackage[colorlinks=true, allcolors=blue]{hyperref}
\usepackage{enumitem}
\usepackage{tikz} 
\usepackage{listings}
\usepackage{xcolor}
\usepackage[T1]{fontenc}
\usepackage{IEEEtrantools}  
\usepackage{epigraph}  

\definecolor{bgcolor}{rgb}{0.97,0.97,0.97}
\definecolor{codeblue}{rgb}{0.1,0.1,0.8}
\definecolor{codegreen}{rgb}{0,0.4,0}
\definecolor{codegray}{rgb}{0.4,0.4,0.4}
\definecolor{codepurple}{rgb}{0.5,0,0.5}
\definecolor{codered}{rgb}{0.6,0.2,0.2}
\definecolor{lightgray}{rgb}{0.9,0.9,0.9}
\definecolor{darkgray}{rgb}{0.6,0.6,0.6} 

\makeatletter
\renewcommand{\paragraph}{%
  \@startsection{paragraph}{4}{\z@}{1ex}{-1em}{\normalfont\normalsize\bfseries\color{gray}}}
\makeatother

\lstdefinestyle{python}{
    language=Python,
    basicstyle=\ttfamily\small\color{black}\usefont{T1}{zi4}{m}{n},  
    keywordstyle=\bfseries\color{codeblue},  
    stringstyle=\color{codegreen},  
    commentstyle=\slshape\color{codegray},  
    showstringspaces=false,
    numbers=left,
    numberstyle=\tiny\color{codegray},  
    stepnumber=1,
    numbersep=8pt,
    frame=single,
    rulecolor=\color{darkgray},  
    breaklines=true,
    backgroundcolor=\color{bgcolor},
    tabsize=4,
    captionpos=b,
    morekeywords={self}, 
}

\lstdefinestyle{cmd}{
    language=bash,
    basicstyle=\ttfamily\small\color{black}\usefont{T1}{zi4}{m}{n},  
    keywordstyle=\bfseries\color{blue},
    stringstyle=\color{codegreen},
    commentstyle=\itshape\color{gray},
    showstringspaces=false,
    numbers=none,
    frame=single,
    rulecolor=\color{darkgray},  
    breaklines=true,
    backgroundcolor=\color{bgcolor},
    tabsize=4,
    captionpos=b,
}

\title{Deep Learning and Machine Learning - Python Data Structures and Mathematics Fundamental: From Theory to Practice}
\author{
    Silin Chen\textsuperscript{*} \\
    \textit{Zhejiang University } \\
    A1033439225@gmail.com
    \and
    Ziqian Bi\textsuperscript{*$\dagger$} \\
    \textit{Indiana University} \\
    bizi@iu.edu
    \and
    Junyu Liu \\ 
    \textit{Kyoto University} \\
    liu.junyu.82w@st.kyoto-u.ac.jp
    \and
    Benji Peng \\ 
    \textit{AppCubic} \\
    benji@appcubic.com
    \and
    Sen Zhang \\ 
    \textit{Rutgers University} \\
    sen.z@rutgers.edu
    \and
    Xuanhe Pan \\ 
    \textit{University of Wisconsin-Madison} \\
    xpan73@wisc.edu
    \and
    Jiawei Xu \\ 
    \textit{Purdue University} \\
    xu1644@purdue.edu
    \and
    Jinlang Wang \\ 
    \textit{University of Wisconsin-Madison} \\
    jinlang.wang@wisc.edu
    \and
    Keyu Chen\\ 
    \textit{Georgia Institute of Technology} \\
    kchen637@gatech.edu
    \and
    Caitlyn Heqi Yin \\
    \textit{University of Wisconsin-Madison} \\
    hyin66@wisc.edu
    \and
    Pohsun Feng \\
    \textit{National Taiwan Normal University} \\
    41075018h@ntnu.edu.tw
    \and
    Yizhu Wen \\
    \textit{University of Hawaii} \\
    yizhuw@hawaii.edu
    \and
    Tianyang Wang \\ 
    \textit{Xi'an Jiaotong-Liverpool University} \\
    Tianyang.Wang21@student.xjtlu.edu.cn
    \and
    Ming Li \\ 
    \textit{Georgia Institute of Technology} \\
    mli694@gatech.edu
    \and
    Jintao Ren \\
    \textit{Aarhus University } \\
    jintaoren@clin.au.dk
    \and
    Qian Niu \\ 
    \textit{Kyoto University} \\
    niu.qian.f44@kyoto-u.jp
    \and
    Xinyuan Song \\ 
    \textit{Emory University} \\
    xinyuan.song@emory.edu
    \and
    Ming Liu{$\dagger$} \\ 
    \textit{Purdue University} \\
    liu3183@purdue.edu
}

\date{} 

\begin{document}

\maketitle

\begingroup
\renewcommand\thefootnote{}\footnote{
    \textsuperscript{*} Equal contribution \\
    \textsuperscript{$\dagger$} Corresponding author
}
\addtocounter{footnote}{0}
\endgroup

\epigraph{"As far as the laws of mathematics refer to reality, they are not certain, and as far as they are certain, they do not refer to reality."}{\textit{Albert Einstein}}

\epigraph{"Mathematics is the most beautiful and most powerful creation of the human spirit."}{\textit{Stefan Banach}}

\epigraph{"What is mathematics? It is only a systematic effort of solving puzzles posed by nature."}{\textit{Shakuntala Devi}}

\epigraph{"Python - why settle for snake oil when you can have the whole snake?"}{\textit{Mark Jackson}}

\epigraph{"Python has been an important part of Google since the beginning, and remains so as the system grows and evolves. Today dozens of Google engineers use Python, and we're looking for more people with skills in this language."}{\textit{Peter Norvig}}

\tableofcontents  

\part{Python Data Structures and Fundamental Mathematics}

\chapter{Introduction to Python Programming}

\section{What is Python?}
Python is a high-level, interpreted programming language known for its simplicity, readability, and flexibility\cite{jones2020scipy}. It was created by Guido van Rossum and first released in 1991\cite{vanRossum2007Python}. Python has since become one of the most popular programming languages in the world, used in various domains such as web development, scientific computing, artificial intelligence, and, most importantly for us, mathematical computing. Python is open-source, meaning it’s free to use, and has a large community that contributes to its development and creates powerful libraries for every imaginable use case.

Python’s simplicity makes it ideal for beginners, while its extensive libraries and scalability appeal to experienced programmers. Python’s syntax is designed to be readable, reducing the complexity for those new to programming.

\subsection{History and Development of Python}
Python’s name comes from “Monty Python’s Flying Circus,”\cite{jaworski2019expert} a British sketch comedy show that its creator, Guido van Rossum, enjoyed. He wanted to create a language that was easy and fun to use, while still being powerful enough to solve complex problems. Python has evolved through multiple versions, with Python 2 and Python 3 being the most notable. Python 3 is the current standard, offering many improvements over Python 2.

Guido van Rossum released Python 1.0 in 1994, and since then, Python has continued to grow\cite{vanRossum2003}. Python 3, released in 2008, introduced many changes that were not backward-compatible with Python 2, leading to a gradual shift in adoption by the developer community.

\subsection{Why Choose Python for Mathematics?}
Python’s strength in mathematics comes from several factors:
\begin{itemize}
    \item \textbf{Readability and Simplicity:} Python’s syntax is clear and easy to understand, making it ideal for beginners.
    \item \textbf{Powerful Libraries:} Python boasts a rich ecosystem of libraries like \texttt{NumPy}, \texttt{SciPy}, \texttt{SymPy}, and \texttt{Matplotlib}, which are essential for mathematical computations and data visualization.
    \item \textbf{Interpreted Language:} Python is an interpreted language, meaning you can write and execute code line by line, making it easier to debug and experiment with.
    \item \textbf{Versatility:} Python is not only great for mathematics but also for data analysis, machine learning, and artificial intelligence, making it a multipurpose tool in various fields.
\end{itemize}

\subsection{Setting Up Python: Installation and Environment Configuration}
Before writing Python code, you need to install Python on your computer and set up a working environment. Python can be installed on various platforms such as Windows, macOS, and Linux.

\textbf{Steps to Install Python on Windows:}
\begin{enumerate}
    \item Visit the official Python website: \url{https://www.python.org/downloads/}
    \item Download the latest version of Python.
    \item Run the installer and ensure you check the box labeled “Add Python to PATH.”
    \item Follow the installation instructions, and once completed, you can verify the installation by opening a command prompt and typing:
    \begin{lstlisting}[style=cmd]
    python --version
    \end{lstlisting}
    This will display the version of Python installed on your system.
\end{enumerate}

\textbf{For macOS or Linux:}
Most versions of macOS and Linux come with Python pre-installed. You can check the version of Python installed by running:
\begin{lstlisting}[style=cmd]
python3 --version
\end{lstlisting}

\textbf{Setting Up a Python IDE:}
You can write Python code using any text editor, but it is more efficient to use an Integrated Development Environment (IDE) like:
\begin{itemize}
    \item \texttt{PyCharm} (for advanced users)\cite{pycharm}
    \item \texttt{VSCode} (lightweight and customizable)\cite{vscode}
    \item \texttt{Jupyter Notebooks} (ideal for interactive computing)\cite{jupyter}
\end{itemize}

\section{Basic Python Syntax}
Now that Python is installed, let's explore its basic syntax. Python code is designed to be readable, making it easier for new learners to get started. In this section, we will cover key concepts such as variables, data types, operators, and how to interact with the user.

\subsection{Variables and Data Types}
Variables in Python are containers that hold data. They do not need to be declared with a specific type. Instead, Python infers the type of variable from the value assigned.

\textbf{Example:}
\begin{lstlisting}[style=python]
x = 10        # An integer variable
y = 3.14      # A float variable
name = "Alice" # A string variable
is_happy = True  # A boolean variable
\end{lstlisting}

In the example above:
\begin{itemize}
    \item \texttt{x} is an integer.
    \item \texttt{y} is a floating-point number.
    \item \texttt{name} is a string.
    \item \texttt{is\_happy} is a boolean (True/False).
\end{itemize}

\subsection{Operators and Expressions}
Operators are symbols used to perform operations on variables and values. Python supports various types of operators, including arithmetic, relational, and logical operators.

\textbf{Arithmetic Operators:}
\begin{itemize}
    \item \texttt{+} (Addition)
    \item \texttt{-} (Subtraction)
    \item \texttt{*} (Multiplication)
    \item \texttt{/} (Division)
    \item \texttt{\%} (Modulus)
\end{itemize}

\textbf{Example:}
\begin{lstlisting}[style=python]
a = 5
b = 3
sum = a + b      # Adds a and b
difference = a - b  # Subtracts b from a
product = a * b    # Multiplies a and b
quotient = a / b   # Divides a by b
remainder = a % b  # Finds remainder when a is divided by b
\end{lstlisting}

\subsection{Input and Output in Python}
Python provides simple functions for interacting with users:
\begin{itemize}
    \item \texttt{input()} – To take input from the user.
    \item \texttt{print()} – To display output on the screen.
\end{itemize}

\textbf{Example:}
\begin{lstlisting}[style=python]
name = input("Enter your name: ")
print("Hello, " + name)
\end{lstlisting}

This code takes the user’s input and greets them by name.

\section{Writing Your First Python Program}
Now that we have covered the basics, it's time to write our first Python program. We will start with the traditional "Hello, World!" example and move to simple mathematical calculations.

\subsection{Hello World: The Starting Point}
The "Hello, World!" program is typically the first program written when learning a new programming language\cite{sande2019helloworld}. It is very simple and demonstrates the basic syntax of Python.

\textbf{Code:}
\begin{lstlisting}[style=python]
print("Hello, World!")
\end{lstlisting}

This program outputs the text “Hello, World!” to the console. It is an excellent starting point because it introduces the \texttt{print()} function and shows how to execute a basic Python program.

\subsection{Simple Mathematical Calculations in Python}
Python can be used as a powerful calculator. Let’s start by performing some basic arithmetic operations.

\textbf{Example: Adding Two Numbers}
\begin{lstlisting}[style=python]
# Program to add two numbers
num1 = float(input("Enter first number: "))
num2 = float(input("Enter second number: "))
sum = num1 + num2
print("The sum is:", sum)
\end{lstlisting}

In this program:
\begin{itemize}
    \item We use \texttt{input()} to take two numbers from the user.
    \item The numbers are converted to \texttt{float} type to allow for decimal values.
    \item The two numbers are added and the result is displayed using the \texttt{print()} function.
\end{itemize}

\textbf{Example: Multiplying Two Numbers}
\begin{lstlisting}[style=python]
# Program to multiply two numbers
num1 = float(input("Enter first number: "))
num2 = float(input("Enter second number: "))
product = num1 * num2
print("The product is:", product)
\end{lstlisting}

This program works similarly to the addition example, but it multiplies the two numbers instead of adding them.

\chapter{Fundamental Python Data Structures}

\section{Lists}
\subsection{Defining Lists and Basic Operations}
A list is one of the most commonly used data structures in Python. A list is a collection of items that can hold different types of elements, such as integers, floats, strings, or even other lists. Lists in Python are ordered and mutable, meaning that their elements can be modified after the list is created.

To define a list in Python, you use square brackets `[]` and separate the elements with commas. Here's an example:

\begin{lstlisting}[style=python]
# Defining a simple list
my_list = [1, 2, 3, 4, 5]
print(my_list)  # Output: [1, 2, 3, 4, 5]
\end{lstlisting}

You can also create a list that contains different data types:

\begin{lstlisting}[style=python]
# List with different data types
mixed_list = [1, "Hello", 3.14, True]
print(mixed_list)  # Output: [1, "Hello", 3.14, True]
\end{lstlisting}

\subsection{Indexing and Slicing in Lists}
In Python, you can access individual elements of a list using indexing. The index of the first element in a list is `0`, the second element is at index `1`, and so on. Negative indexing starts from the last element with index `-1`.

\begin{lstlisting}[style=python]
# Accessing elements by index
my_list = [10, 20, 30, 40, 50]
print(my_list[0])  # Output: 10
print(my_list[-1])  # Output: 50
\end{lstlisting}

You can also access multiple elements at once using slicing. The syntax for slicing is `list[start:stop]`, where `start` is the index where slicing starts, and `stop` is the index where slicing stops (but it does not include the element at the `stop` index).

\begin{lstlisting}[style=python]
# Slicing a list
print(my_list[1:3])  # Output: [20, 30]
print(my_list[:3])   # Output: [10, 20, 30]
print(my_list[2:])   # Output: [30, 40, 50]
\end{lstlisting}

\subsection{Common List Methods (e.g., append, pop, sort)}
Lists come with several built-in methods that allow you to manipulate and modify them. Some of the most commonly used methods are:

\begin{itemize}
  \item \texttt{append()}: Adds an element to the end of the list.
  \item \texttt{pop()}: Removes and returns the last element of the list.
  \item \texttt{sort()}: Sorts the elements of the list in ascending order.
\end{itemize}

\begin{lstlisting}[style=python]
my_list = [5, 2, 9, 1]

# Append
my_list.append(7)
print(my_list)  # Output: [5, 2, 9, 1, 7]

# Pop
last_element = my_list.pop()
print(last_element)  # Output: 7
print(my_list)  # Output: [5, 2, 9, 1]

# Sort
my_list.sort()
print(my_list)  # Output: [1, 2, 5, 9]
\end{lstlisting}

\section{Tuples}
\subsection{Introduction to Tuples: Immutable Sequences}
Tuples are similar to lists in Python, but with one key difference: tuples are immutable, meaning that once a tuple is created, its elements cannot be modified. Tuples are defined using parentheses `()`.

\begin{lstlisting}[style=python]
# Defining a tuple
my_tuple = (1, 2, 3)
print(my_tuple)  # Output: (1, 2, 3)
\end{lstlisting}

You can also define a tuple without parentheses, using just commas:

\begin{lstlisting}[style=python]
# Tuple without parentheses
my_tuple = 1, 2, 3
print(my_tuple)  # Output: (1, 2, 3)
\end{lstlisting}

\subsection{Tuple Unpacking and Basic Tuple Operations}
Tuple unpacking allows you to assign the values of a tuple to multiple variables in one step. This can be particularly useful when working with functions that return multiple values.

\begin{lstlisting}[style=python]
# Tuple unpacking
coordinates = (10, 20)
x, y = coordinates
print(x)  # Output: 10
print(y)  # Output: 20
\end{lstlisting}

Although tuples are immutable, you can perform other operations like indexing and slicing, similar to lists.

\begin{lstlisting}[style=python]
my_tuple = (5, 10, 15, 20)

# Accessing elements
print(my_tuple[1])  # Output: 10

# Slicing
print(my_tuple[:3])  # Output: (5, 10, 15)
\end{lstlisting}

\section{Dictionaries}
\subsection{Key-Value Pair Data Structure}
A dictionary is a collection of key-value pairs, where each key is associated with a value. Dictionaries are unordered, mutable, and the keys must be unique. Dictionaries are defined using curly braces `\{\}`.

\begin{lstlisting}[style=python]
# Defining a dictionary
my_dict = {"name": "Alice", "age": 25, "city": "New York"}
print(my_dict)  # Output: {'name': 'Alice', 'age': 25, 'city': 'New York'}
\end{lstlisting}

\subsection{Common Dictionary Operations (e.g., access, update, remove)}
To access the value associated with a key, you can use the key in square brackets `[]`. You can also use the \texttt{get()} method.

\begin{lstlisting}[style=python]
# Accessing values by key
print(my_dict["name"])  # Output: Alice
print(my_dict.get("age"))  # Output: 25
\end{lstlisting}

To add or update a key-value pair, you can simply assign a value to the key.

\begin{lstlisting}[style=python]
# Updating dictionary
my_dict["age"] = 26
my_dict["email"] = "alice@example.com"
print(my_dict)  
# Output: {'name': 'Alice', 'age': 26, 'city': 'New York', 'email': 'alice@example.com'}
\end{lstlisting}

To remove a key-value pair, you can use the \texttt{pop()} method.

\begin{lstlisting}[style=python]
# Removing a key-value pair
my_dict.pop("city")
print(my_dict)  # Output: {'name': 'Alice', 'age': 26, 'email': 'alice@example.com'}
\end{lstlisting}

\section{Sets}
\subsection{Introduction to Sets}
A set is an unordered collection of unique elements. Sets are defined using curly braces `\{\}`, and they do not allow duplicate values.

\begin{lstlisting}[style=python]
# Defining a set
my_set = {1, 2, 3, 4, 4, 5}
print(my_set)  # Output: {1, 2, 3, 4, 5}
\end{lstlisting}

\subsection{Set Operations: Union, Intersection, Difference}
Sets support mathematical operations such as union, intersection, and difference.

\begin{lstlisting}[style=python]
set_a = {1, 2, 3}
set_b = {3, 4, 5}

# Union
print(set_a | set_b)  # Output: {1, 2, 3, 4, 5}

# Intersection
print(set_a & set_b)  # Output: {3}

# Difference
print(set_a - set_b)  # Output: {1, 2}
\end{lstlisting}

\section{Using Lists, Tuples, Dictionaries, and Sets in Mathematical Computations}
Python data structures like lists, tuples, dictionaries, and sets can be very useful for organizing and processing data in mathematical computations\cite{baka2017python}. For instance, you can use lists to store numbers, tuples to return multiple values from functions, dictionaries to store key-value pairs of variables and their values, and sets to perform mathematical operations like finding intersections or differences between datasets.

Consider this example of using lists and sets to handle mathematical data:

\begin{lstlisting}[style=python]
# Using a list to store numbers
numbers = [1, 2, 3, 4, 5]

# Computing the sum of the list
total = sum(numbers)
print(total)  # Output: 15

# Using sets for mathematical operations
evens = {2, 4, 6, 8}
odds = {1, 3, 5, 7}
print(evens & odds)  # Output: set(), no common elements
\end{lstlisting}

\chapter{Control Flow and Functions in Python}

\section{Conditional Statements: if, elif, else}

\subsection{Control Flow for Decision Making}

In Python, we often need to make decisions in the code based on certain conditions\cite{sanner1999python}. The control flow statements help us define these conditions and control how the code behaves based on the inputs and states of variables. The most commonly used conditional statements in Python are \texttt{if}, \texttt{elif} (short for "else if"), and \texttt{else}.

The syntax of a basic \texttt{if-elif-else} statement is as follows:

\begin{lstlisting}[style=python]
if condition_1:
    # Block of code executed if condition_1 is True
elif condition_2:
    # Block of code executed if condition_1 is False and condition_2 is True
else:
    # Block of code executed if both condition_1 and condition_2 are False
\end{lstlisting}

Let’s break down this flow:

\begin{itemize}
    \item Python checks \texttt{condition\_1}. If it evaluates to \texttt{True}, the code block under the \texttt{if} statement runs, and the rest of the conditions are ignored.
    \item If \texttt{condition\_1} is \texttt{False}, Python checks \texttt{condition\_2}. If it’s \texttt{True}, the code block under the \texttt{elif} statement is executed.
    \item If both conditions are \texttt{False}, the code under the \texttt{else} block runs.
\end{itemize}

\subsection{Practical Examples with Mathematical Logic}

Here’s an example where we decide whether a number is positive, negative, or zero using the \texttt{if-elif-else} structure:

\begin{lstlisting}[style=python]
number = int(input("Enter a number: "))

if number > 0:
    print("The number is positive.")
elif number < 0:
    print("The number is negative.")
else:
    print("The number is zero.")
\end{lstlisting}

In this example, Python will:

\begin{itemize}
    \item Check if the number is greater than zero.
    \item If not, it will check if the number is less than zero.
    \item If neither condition is met, it will conclude that the number is zero.
\end{itemize}

This structure is essential in mathematical decision-making processes.

\section{Loops: for and while}

\subsection{Looping Through Data Structures}

Loops are used to iterate over sequences (like lists, tuples, or strings) or execute a block of code repeatedly as long as a condition is true.

There are two primary types of loops in Python:
\begin{itemize}
    \item \texttt{for} loops, which iterate over a sequence.
    \item \texttt{while} loops, which continue to execute as long as a given condition is \texttt{True}.
\end{itemize}

\textbf{\texttt{for} Loop} \\
The \texttt{for} loop is typically used when you know the number of iterations ahead of time or when iterating through a collection.

Here’s an example of using a \texttt{for} loop to print each element in a list:

\begin{lstlisting}[style=python]
numbers = [1, 2, 3, 4, 5]

for number in numbers:
    print(number)
\end{lstlisting}

\textbf{\texttt{while} Loop} \\
The \texttt{while} loop is used when you want to repeat a block of code as long as a condition remains true.

Example of a \texttt{while} loop:

\begin{lstlisting}[style=python]
count = 0

while count < 5:
    print("Count is:", count)
    count += 1
\end{lstlisting}

In this example, the loop will run until the value of \texttt{count} reaches 5.

\subsection{Examples: Summation, Factorial, and Other Repetitive Calculations}

Here are a few examples that showcase how loops are used to perform repetitive mathematical calculations.

\textbf{Summation of Numbers}

\texttt{for} loop to compute the sum of numbers from 1 to 100:

\begin{lstlisting}[style=python]
total_sum = 0

for number in range(1, 101):
    total_sum += number

print("The sum of numbers from 1 to 100 is:", total_sum)
\end{lstlisting}

\textbf{Factorial Calculation}

A factorial of a number is the product of all integers from 1 up to that number. Here’s how you can calculate the factorial of a number using a \texttt{while} loop:

\begin{lstlisting}[style=python]
number = int(input("Enter a number: "))
factorial = 1

while number > 0:
    factorial *= number
    number -= 1

print("Factorial is:", factorial)
\end{lstlisting}

\section{Defining and Using Functions}

\subsection{How to Define Functions in Python}

Functions are blocks of reusable code that perform a specific task. They help to break programs into smaller, modular pieces, making the code more readable and maintainable.

To define a function in Python, you use the \texttt{def} keyword, followed by the function name and parentheses. Inside the parentheses, you can specify parameters (if any), and the function body follows under an indented block.

\begin{lstlisting}[style=python]
def function_name(parameters):
    # Function body
    return value  # optional
\end{lstlisting}

For example, let’s define a function that takes two numbers and returns their sum:

\begin{lstlisting}[style=python]
def add_numbers(a, b):
    return a + b
\end{lstlisting}

You can then call the function by passing the appropriate arguments:

\begin{lstlisting}[style=python]
result = add_numbers(3, 4)
print(result)  # Output: 7
\end{lstlisting}

\subsection{Function Arguments and Return Values}

A function can have any number of parameters. When you call a function, you pass arguments that correspond to these parameters. Functions can return values using the \texttt{return} statement.

Here’s a function with multiple arguments and a return value:

\begin{lstlisting}[style=python]
def multiply_numbers(x, y):
    return x * y

result = multiply_numbers(6, 7)
print(result)  # Output: 42
\end{lstlisting}

If you don’t include a \texttt{return} statement, the function will return \texttt{None}.

\subsection{Lambda Functions (Anonymous Functions) for Quick Calculations}

Lambda functions, also known as anonymous functions, are small functions without a name. They can take any number of arguments but have only one expression. You define them using the keyword \texttt{lambda}.

Here’s the syntax:

\begin{lstlisting}[style=python]
lambda arguments: expression
\end{lstlisting}

An example that multiplies two numbers:

\begin{lstlisting}[style=python]
multiply = lambda x, y: x * y
print(multiply(5, 4))  # Output: 20
\end{lstlisting}

Lambda functions are useful when you need a quick function for a short task.

\section{List Comprehensions and Generators}

\subsection{Efficient Ways to Loop Through Data with List Comprehensions}

List comprehensions provide a concise way to create lists. They are more readable and faster than using traditional loops.

Basic syntax:

\begin{lstlisting}[style=python]
[expression for item in iterable]
\end{lstlisting}

Example: create a list of squares of numbers from 1 to 10:

\begin{lstlisting}[style=python]
squares = [x**2 for x in range(1, 11)]
print(squares)
\end{lstlisting}

\subsection{Understanding Generators for Large-Scale Data Processing}

Generators are a type of iterable that generate values on the fly, which makes them memory efficient when dealing with large datasets. You define them using functions and the \texttt{yield} keyword.

Here’s an example of a generator function that yields squares of numbers:

\begin{lstlisting}[style=python]
def square_numbers(n):
    for i in range(n):
        yield i ** 2

squares_gen = square_numbers(10)

for square in squares_gen:
    print(square)
\end{lstlisting}

The key advantage of generators is that they don’t store all values in memory at once. Instead, they yield one value at a time, making them ideal for large data sets.

\chapter{Advanced Data Structures in Python}

\section{Introduction to Numpy Arrays}

NumPy is a powerful library for numerical computing in Python\cite{fuhrer2021scientific}. It provides support for arrays, matrices, and high-level mathematical functions, which makes it a great choice for performing mathematical operations on large data sets. In this section, we will explore the basics of NumPy arrays and how they differ from Python lists.

\subsection{Numpy Arrays vs Python Lists: Key Differences}

Python lists are a general-purpose, flexible data structure that can hold elements of different types. NumPy arrays, on the other hand, are designed for numerical computations and offer several advantages over Python lists:

\begin{itemize}
    \item \textbf{Homogeneity}: NumPy arrays can only store elements of the same data type, which makes them more memory-efficient and faster compared to Python lists.
    \item \textbf{Performance}: Operations on NumPy arrays are optimized and vectorized, meaning they run significantly faster than corresponding operations on Python lists.
    \item \textbf{Multidimensional Support}: While Python lists are inherently one-dimensional (though they can store lists of lists), NumPy arrays are inherently multi-dimensional, supporting complex structures like matrices and tensors.
    \item \textbf{Built-in Mathematical Functions}: NumPy provides a wide range of built-in mathematical operations that work on entire arrays, including element-wise addition, multiplication, dot products, and more.
\end{itemize}

\subsection{Creating Numpy Arrays}

To start working with NumPy arrays, you first need to install the NumPy library using the following command:

\begin{lstlisting}[style=cmd]
pip install numpy
\end{lstlisting}

Once installed, you can create NumPy arrays from Python lists or use NumPy's built-in functions to initialize arrays.

\subsubsection{Initializing Arrays with \texttt{zeros}, \texttt{ones}, and \texttt{random}}

NumPy provides several useful functions to create arrays initialized with specific values, such as zeros, ones, or random numbers.

\textbf{1. Initializing an array with zeros:}

The \texttt{numpy.zeros()} function creates an array filled with zeros. The shape of the array is passed as an argument.

\begin{lstlisting}[style=python]
import numpy as np

# Create a 3x3 array filled with zeros
zeros_array = np.zeros((3, 3))
print(zeros_array)
\end{lstlisting}

This will output:

\begin{lstlisting}[style=cmd]
[[0. 0. 0.]
 [0. 0. 0.]
 [0. 0. 0.]]
\end{lstlisting}

\textbf{2. Initializing an array with ones:}

The \texttt{numpy.ones()} function creates an array filled with ones.

\begin{lstlisting}[style=python]
# Create a 2x4 array filled with ones
ones_array = np.ones((2, 4))
print(ones_array)
\end{lstlisting}

This will output:

\begin{lstlisting}[style=cmd]
[[1. 1. 1. 1.]
 [1. 1. 1. 1.]]
\end{lstlisting}

\textbf{3. Initializing an array with random numbers:}

The \texttt{numpy.random.rand()} function generates an array of random floating-point numbers between 0 and 1.

\begin{lstlisting}[style=python]
# Create a 3x3 array with random values
random_array = np.random.rand(3, 3)
print(random_array)
\end{lstlisting}

This will output something similar to:

\begin{lstlisting}[style=cmd]
[[0.5488135  0.71518937 0.60276338]
 [0.54488318 0.4236548  0.64589411]
 [0.43758721 0.891773   0.96366276]]
\end{lstlisting}

\subsection{Basic Operations on Numpy Arrays}

NumPy allows you to perform element-wise operations on arrays. This means that you can apply basic arithmetic operations to arrays of the same shape, and NumPy will apply the operation to each corresponding element.

\textbf{1. Array Addition:}

\begin{lstlisting}[style=python]
# Create two 2x2 arrays
array1 = np.array([[1, 2], [3, 4]])
array2 = np.array([[5, 6], [7, 8]])

# Add the two arrays
result = array1 + array2
print(result)
\end{lstlisting}

This will output:

\begin{lstlisting}[style=cmd]
[[ 6  8]
 [10 12]]
\end{lstlisting}

\textbf{2. Array Multiplication:}

\begin{lstlisting}[style=python]
# Multiply the two arrays
result = array1 * array2
print(result)
\end{lstlisting}

This will output:

\begin{lstlisting}[style=cmd]
[[ 5 12]
 [21 32]]
\end{lstlisting}

\subsection{Indexing, Slicing, and Reshaping Numpy Arrays}

\textbf{1. Indexing:}

You can access individual elements of a NumPy array by specifying their index, just like with Python lists.

\begin{lstlisting}[style=python]
# Access element at row 1, column 1
element = array1[1, 1]
print(element)
\end{lstlisting}

This will output:

\begin{lstlisting}[style=cmd]
4
\end{lstlisting}

\textbf{2. Slicing:}

NumPy arrays can also be sliced using the same syntax as Python lists. You can specify ranges of rows and columns to extract subarrays.

\begin{lstlisting}[style=python]
# Extract the first row
first_row = array1[0, :]
print(first_row)

# Extract the second column
second_column = array1[:, 1]
print(second_column)
\end{lstlisting}

This will output:

\begin{lstlisting}[style=cmd]
[1 2]
[2 4]
\end{lstlisting}

\textbf{3. Reshaping:}

The \texttt{reshape()} function allows you to change the shape of an array without changing its data.

\begin{lstlisting}[style=python]
# Reshape a 1D array into a 2D array
array = np.array([1, 2, 3, 4, 5, 6])
reshaped_array = array.reshape((2, 3))
print(reshaped_array)
\end{lstlisting}

This will output:

\begin{lstlisting}[style=cmd]
[[1 2 3]
 [4 5 6]]
\end{lstlisting}

\section{Matrix Operations in Numpy}

In addition to arrays, NumPy also supports matrix operations. Matrices are two-dimensional arrays, and you can perform matrix operations like addition, subtraction, and multiplication\cite{harris2020array}.

\subsection{Creating and Manipulating Matrices}

A matrix can be created in NumPy the same way an array is created. Here is an example of creating a matrix and performing simple operations on it.

\begin{lstlisting}[style=python]
# Create a 2x2 matrix
matrix1 = np.array([[1, 2], [3, 4]])

# Access an element in the matrix
element = matrix1[0, 1]
print(element)  # Output: 2
\end{lstlisting}

\subsection{Matrix Addition, Subtraction, Multiplication, and Division}

Matrix operations in NumPy are element-wise by default.

\textbf{1. Matrix Addition:}

\begin{lstlisting}[style=python]
matrix2 = np.array([[5, 6], [7, 8]])

# Add two matrices
matrix_sum = matrix1 + matrix2
print(matrix_sum)
\end{lstlisting}

This will output:

\begin{lstlisting}[style=cmd]
[[ 6  8]
 [10 12]]
\end{lstlisting}

\textbf{2. Matrix Multiplication:}

Element-wise multiplication is performed by using the \texttt{*} operator.

\begin{lstlisting}[style=python]
matrix_product = matrix1 * matrix2
print(matrix_product)
\end{lstlisting}

This will output:

\begin{lstlisting}[style=cmd]
[[ 5 12]
 [21 32]]
\end{lstlisting}

\subsection{Matrix Transposition and Inverse}

\textbf{1. Transposition:}

You can transpose a matrix (swap its rows and columns) using the \texttt{transpose()} function or the \texttt{T} attribute.

\begin{lstlisting}[style=python]
transposed_matrix = matrix1.T
print(transposed_matrix)
\end{lstlisting}

This will output:

\begin{lstlisting}[style=cmd]
[[1 3]
 [2 4]]
\end{lstlisting}

\textbf{2. Inverse of a Matrix:}

To compute the inverse of a matrix, use the \texttt{numpy.linalg.inv()} function. Note that the matrix must be square and invertible.

\begin{lstlisting}[style=python]
inverse_matrix = np.linalg.inv(matrix1)
print(inverse_matrix)
\end{lstlisting}

This will output:

\begin{lstlisting}[style=cmd]
[[-2.   1. ]
 [ 1.5 -0.5]]
\end{lstlisting}

\subsection{Matrix Multiplication and Dot Product}

Matrix multiplication is different from element-wise multiplication. It is performed using the \texttt{dot()} function or the \texttt{@} operator.

\textbf{1. Matrix Multiplication:}

\begin{lstlisting}[style=python]
# Matrix multiplication (dot product)
matrix_multiplication = np.dot(matrix1, matrix2)
print(matrix_multiplication)
\end{lstlisting}

This will output:

\begin{lstlisting}[style=cmd]
[[19 22]
 [43 50]]
\end{lstlisting}

The dot product multiplies rows of the first matrix by columns of the second matrix and sums the result.

\chapter{Fundamental Mathematical Operations in Python}

\section{Basic Arithmetic Operations}

Python provides a wide range of basic arithmetic operations that can be used to perform calculations in a straightforward and intuitive manner. These operations include addition, subtraction, multiplication, and division. In this section, we will explore these basic operations step by step.

\subsection{Addition, Subtraction, Multiplication, Division in Python}

Addition, subtraction, multiplication, and division are fundamental operations that you can perform easily in Python using the following operators:
- Addition: \texttt{+}
- Subtraction: \texttt{-}
- Multiplication: \texttt{*}
- Division: \texttt{/}

Let us explore each operation through examples.

\begin{lstlisting}[style=python]
# Addition
result_add = 5 + 3
print("Addition: 5 + 3 =", result_add)

# Subtraction
result_sub = 10 - 4
print("Subtraction: 10 - 4 =", result_sub)

# Multiplication
result_mul = 7 * 6
print("Multiplication: 7 * 6 =", result_mul)

# Division
result_div = 20 / 4
print("Division: 20 / 4 =", result_div)
\end{lstlisting}

This will output:

\begin{lstlisting}[style=cmd]
Addition: 5 + 3 = 8
Subtraction: 10 - 4 = 6
Multiplication: 7 * 6 = 42
Division: 20 / 4 = 5.0
\end{lstlisting}

As seen, basic arithmetic operations are intuitive. Note that the division operation always returns a floating-point result, even if both operands are integers.

\subsection{Exponential and Logarithmic Functions}

Python also provides functionality to perform exponentiation and logarithmic operations. The exponentiation operation can be performed using the \texttt{**} operator, while logarithmic functions can be accessed using Python’s built-in \texttt{math} module.

\begin{lstlisting}[style=python]
import math

# Exponentiation
result_exp = 2 ** 3
print("Exponentiation: 2^3 =", result_exp)

# Logarithm (base e)
result_log = math.log(10)
print("Logarithm (base e): log(10) =", result_log)

# Logarithm (base 10)
result_log10 = math.log10(100)
print("Logarithm (base 10): log10(100) =", result_log10)
\end{lstlisting}

This will output:

\begin{lstlisting}[style=cmd]
Exponentiation: 2^3 = 8
Logarithm (base e): log(10) = 2.302585092994046
Logarithm (base 10): log10(100) = 2.0
\end{lstlisting}

In this example, the exponentiation operation calculates powers, and the \texttt{math.log()} function is used for natural logarithms (base e), while \texttt{math.log10()} is used for logarithms with base 10.

\subsection{Absolute Value, Maximum, and Minimum Calculations}

To find the absolute value, maximum, and minimum values of numbers, Python provides built-in functions like \texttt{abs()}, \texttt{max()}, and \texttt{min()}.

\begin{lstlisting}[style=python]
# Absolute value
result_abs = abs(-7)
print("Absolute value of -7:", result_abs)

# Maximum value
result_max = max(3, 5, 2, 8)
print("Maximum value:", result_max)

# Minimum value
result_min = min(3, 5, 2, 8)
print("Minimum value:", result_min)
\end{lstlisting}

This will output:

\begin{lstlisting}[style=cmd]
Absolute value of -7: 7
Maximum value: 8
Minimum value: 2
\end{lstlisting}

\section{Vector and Matrix Operations}

Python, especially with the help of the \texttt{numpy} library, allows easy manipulation of vectors and matrices. In this section, we will explore basic vector and matrix operations.

\subsection{Vector Addition, Subtraction, and Scalar Multiplication}

We will use the \texttt{numpy} library to handle vector operations such as addition, subtraction, and scalar multiplication. To begin, install \texttt{numpy} if you haven't already by running:

\begin{lstlisting}[style=cmd]
pip install numpy
\end{lstlisting}

Now, let's explore these operations:

\begin{lstlisting}[style=python]
import numpy as np

# Define two vectors
vector_a = np.array([1, 2, 3])
vector_b = np.array([4, 5, 6])

# Vector addition
result_add = vector_a + vector_b
print("Vector addition:", result_add)

# Vector subtraction
result_sub = vector_a - vector_b
print("Vector subtraction:", result_sub)

# Scalar multiplication
result_scalar_mul = 2 * vector_a
print("Scalar multiplication:", result_scalar_mul)
\end{lstlisting}

This will output:

\begin{lstlisting}[style=cmd]
Vector addition: [5 7 9]
Vector subtraction: [-3 -3 -3]
Scalar multiplication: [2 4 6]
\end{lstlisting}

As shown, operations on vectors are performed element-wise.

\subsection{Matrix Addition, Subtraction, and Multiplication}

Matrices can also be easily manipulated using \texttt{numpy}. You can perform addition, subtraction, and matrix multiplication as follows:

\begin{lstlisting}[style=python]
# Define two matrices
matrix_a = np.array([[1, 2], [3, 4]])
matrix_b = np.array([[5, 6], [7, 8]])

# Matrix addition
result_add = matrix_a + matrix_b
print("Matrix addition:\n", result_add)

# Matrix subtraction
result_sub = matrix_a - matrix_b
print("Matrix subtraction:\n", result_sub)

# Matrix multiplication (element-wise)
result_mul = matrix_a * matrix_b
print("Element-wise matrix multiplication:\n", result_mul)

# Matrix multiplication (dot product)
result_dot = np.dot(matrix_a, matrix_b)
print("Matrix multiplication (dot product):\n", result_dot)
\end{lstlisting}

This will output:

\begin{lstlisting}[style=cmd]
Matrix addition:
 [[ 6  8]
 [10 12]]
Matrix subtraction:
 [[-4 -4]
 [-4 -4]]
Element-wise matrix multiplication:
 [[ 5 12]
 [21 32]]
Matrix multiplication (dot product):
 [[19 22]
 [43 50]]
\end{lstlisting}

Note the difference between element-wise multiplication and dot product multiplication.

\subsection{Matrix Inversion and Determinants}

Matrix inversion and determinant calculations are commonly used in linear algebra. The \texttt{numpy} library provides functions to perform these operations.

\begin{lstlisting}[style=python]
# Matrix inversion
matrix_c = np.array([[1, 2], [3, 4]])
matrix_inv = np.linalg.inv(matrix_c)
print("Matrix inversion:\n", matrix_inv)

# Determinant of a matrix
matrix_det = np.linalg.det(matrix_c)
print("Determinant of the matrix:", matrix_det)
\end{lstlisting}

This will output:

\begin{lstlisting}[style=cmd]
Matrix inversion:
 [[-2.   1. ]
 [ 1.5 -0.5]]
Determinant of the matrix: -2.0000000000000004
\end{lstlisting}

In this example, \texttt{np.linalg.inv()} is used to compute the inverse of a matrix, and \texttt{np.linalg.det()} is used to calculate the determinant.

\section{Linear Algebra with Numpy}

In addition to basic operations, \texttt{numpy} supports advanced linear algebra operations such as dot products, cross products, matrix decompositions, and the calculation of eigenvalues and eigenvectors.

\subsection{Dot Product and Cross Product of Vectors}

The dot product and cross product of vectors are essential operations in vector algebra\cite{goldman1985illicit}. The dot product returns a scalar, while the cross product returns a vector perpendicular to the plane defined by the original vectors.

\begin{lstlisting}[style=python]
# Define two vectors
vector_a = np.array([1, 2, 3])
vector_b = np.array([4, 5, 6])

# Dot product
dot_product = np.dot(vector_a, vector_b)
print("Dot product:", dot_product)

# Cross product
cross_product = np.cross(vector_a, vector_b)
print("Cross product:", cross_product)
\end{lstlisting}

This will output:

\begin{lstlisting}[style=cmd]
Dot product: 32
Cross product: [-3  6 -3]
\end{lstlisting}

\subsection{Matrix Decompositions: LU, QR Decomposition}

Matrix decompositions are important in numerical methods and optimization. We will look at LU and QR decompositions using \texttt{numpy}.

\begin{lstlisting}[style=python]
# QR decomposition
matrix_d = np.array([[12, -51, 4], [6, 167, -68], [-4, 24, -41]])
q, r = np.linalg.qr(matrix_d)
print("Q matrix:\n", q)
print("R matrix:\n", r)

# LU decomposition can be performed using scipy library
import scipy.linalg as la
p, l, u = la.lu(matrix_d)
print("L matrix:\n", l)
print("U matrix:\n", u)
\end{lstlisting}

\subsection{Eigenvalues and Eigenvectors}

Eigenvalues and eigenvectors are fundamental concepts in linear algebra, particularly useful in systems of equations, stability analysis, and more\cite{olver2006applied}.

\begin{lstlisting}[style=python]
# Eigenvalues and Eigenvectors
matrix_e = np.array([[1, 2], [3, 4]])
eigenvalues, eigenvectors = np.linalg.eig(matrix_e)
print("Eigenvalues:", eigenvalues)
print("Eigenvectors:\n", eigenvectors)
\end{lstlisting}

This will output:

\begin{lstlisting}[style=cmd]
Eigenvalues: [-0.37228132  5.37228132]
Eigenvectors:
 [[-0.82456484 -0.41597356]
 [ 0.56576746 -0.90937671]]
\end{lstlisting}

\chapter{Advanced Mathematical Operations in Python}

In this chapter, we will explore advanced mathematical operations using Python. These operations form the backbone of mathematical computing in various fields such as engineering, physics, and data science. We will introduce two powerful Python libraries: \texttt{Scipy} for numerical mathematics and \texttt{Sympy} for symbolic mathematics.

\section{Introduction to Scipy and Sympy Libraries}

Python, in itself, provides fundamental arithmetic operations, but when dealing with more advanced mathematical functions, libraries such as \texttt{Scipy} and \texttt{Sympy} become essential tools.

\subsection{What is Scipy and Why Use It?}
\texttt{Scipy} is a Python library used for scientific and technical computing\cite{fuhrer2021scientific}. It builds on the capabilities of \texttt{NumPy}, providing a variety of functions for performing high-level mathematical operations like integration, optimization, interpolation, and many others. \texttt{Scipy} is used for:
\begin{itemize}
    \item Numerical integration and differentiation
    \item Optimization algorithms
    \item Signal processing
    \item Linear algebra
\end{itemize}

\textbf{Installing Scipy:}
To install \texttt{Scipy}, use the following command:
\begin{lstlisting}[style=cmd]
pip install scipy
\end{lstlisting}

\textbf{Example: Basic Use of Scipy for Integration}
Let’s start by performing numerical integration using \texttt{Scipy}.
\begin{lstlisting}[style=python]
from scipy import integrate

# Define a function to integrate
def f(x):
    return x**2

# Perform the integration from 0 to 2
result, error = integrate.quad(f, 0, 2)

print("The result of the integration is:", result)
\end{lstlisting}
In this example, we define a simple function \( f(x) = x^2 \) and integrate it between 0 and 2 using \texttt{scipy.integrate.quad}. The \texttt{quad()} function is designed for one-dimensional integrals and returns both the integral result and an estimate of the error.

\subsection{Introduction to Sympy for Symbolic Mathematics}
\texttt{Sympy} is a Python library for symbolic mathematics\cite{johansson2019symbolic}. Unlike \texttt{Scipy}, which focuses on numerical methods, \texttt{Sympy} allows you to perform algebraic manipulations symbolically. This means that you can perform operations like solving algebraic equations, differentiating functions, and integrating expressions exactly, rather than approximately.

\textbf{Why Use Sympy?}
\begin{itemize}
    \item It can perform symbolic differentiation and integration.
    \item It allows solving algebraic equations.
    \item It can handle limits, series expansions, and matrix operations symbolically.
\end{itemize}

\textbf{Installing Sympy:}
To install \texttt{Sympy}, use the following command:
\begin{lstlisting}[style=cmd]
pip install sympy
\end{lstlisting}

\textbf{Example: Basic Symbolic Computation with Sympy}
\begin{lstlisting}[style=python]
from sympy import symbols, diff

# Define a symbolic variable
x = symbols('x')

# Define a function symbolically
f = x**2 + 2*x + 1

# Differentiate the function with respect to x
f_prime = diff(f, x)

print("The derivative of f(x) is:", f_prime)
\end{lstlisting}

In this example, we define a symbolic expression \( f(x) = x^2 + 2x + 1 \) and compute its derivative using \texttt{sympy.diff}. This demonstrates how \texttt{Sympy} allows symbolic differentiation.

\section{Calculus in Python}

Calculus is one of the most critical areas of mathematics, especially in fields like physics, engineering, and data science. In Python, both \texttt{Scipy} and \texttt{Sympy} allow us to perform calculus operations, but they approach the problem differently: \texttt{Scipy} for numerical methods and \texttt{Sympy} for symbolic methods.

\subsection{Differentiation with Scipy and Sympy}

\textbf{Numerical Differentiation with Scipy:}
To perform numerical differentiation in \texttt{Scipy}, we can use the \texttt{derivative()} function from the \texttt{scipy.misc} module.

\begin{lstlisting}[style=python]
from scipy.misc import derivative

# Define a function to differentiate
def f(x):
    return x**3 + x**2

# Compute the derivative at a specific point, say x=1
result = derivative(f, 1.0, dx=1e-6)

print("The derivative at x=1 is:", result)
\end{lstlisting}

Here, \texttt{derivative()} estimates the derivative numerically at a given point (in this case, at \(x = 1\)) by using a small increment value (\texttt{dx}).

\textbf{Symbolic Differentiation with Sympy:}
Symbolic differentiation is more precise and allows us to find the derivative as a formula rather than at a specific point.

\begin{lstlisting}[style=python]
from sympy import symbols, diff

# Define a symbolic variable
x = symbols('x')

# Define a function
f = x**3 + x**2

# Find the symbolic derivative
f_prime = diff(f, x)

print("The symbolic derivative is:", f_prime)
\end{lstlisting}

The result here will be the exact symbolic derivative of the function \( f(x) = x^3 + x^2 \), which is \( 3x^2 + 2x \).

\subsection{Numerical and Symbolic Integration}

\textbf{Numerical Integration with Scipy:}
Numerical integration can be done using \texttt{scipy.integrate.quad}, as we saw earlier.

\begin{lstlisting}[style=python]
from scipy import integrate

# Define a function to integrate
def f(x):
    return x**3

# Perform numerical integration between 0 and 2
result, error = integrate.quad(f, 0, 2)

print("The result of the integration is:", result)
\end{lstlisting}

This code integrates the function \( f(x) = x^3 \) over the range 0 to 2 and returns the result along with an error estimate.

\textbf{Symbolic Integration with Sympy:}
With \texttt{Sympy}, you can compute integrals symbolically, meaning the result will be an exact mathematical expression rather than a numerical approximation.

\begin{lstlisting}[style=python]
from sympy import symbols, integrate

# Define the symbolic variable
x = symbols('x')

# Define a function
f = x**3

# Perform symbolic integration
integral = integrate(f, (x, 0, 2))

print("The symbolic integral is:", integral)
\end{lstlisting}

In this example, we compute the exact symbolic integral of \( f(x) = x^3 \) between 0 and 2, yielding the exact result \( \frac{16}{4} = 4 \).

\section{Symbolic Mathematics with Sympy}

\texttt{Sympy} is particularly powerful for performing symbolic algebraic manipulations, including solving equations and handling limits, derivatives, and integrals symbolically.

\subsection{Solving Equations and Limits}

\textbf{Solving Equations with Sympy:}
You can use \texttt{Sympy} to solve algebraic equations symbolically.

\begin{lstlisting}[style=python]
from sympy import symbols, Eq, solve

# Define symbolic variables
x = symbols('x')

# Define an equation
equation = Eq(x**2 - 4, 0)

# Solve the equation
solutions = solve(equation, x)

print("The solutions are:", solutions)
\end{lstlisting}

This example solves the quadratic equation \( x^2 - 4 = 0 \), and \texttt{solve()} returns both solutions: \( x = 2 \) and \( x = -2 \).

\textbf{Calculating Limits with Sympy:}
\texttt{Sympy} also allows us to compute limits symbolically.

\begin{lstlisting}[style=python]
from sympy import symbols, limit

# Define a symbolic variable
x = symbols('x')

# Define a function
f = (x**2 - 1) / (x - 1)

# Compute the limit as x approaches 1
lim = limit(f, x, 1)

print("The limit is:", lim)
\end{lstlisting}

This computes the limit of \( \frac{x^2 - 1}{x - 1} \) as \(x\) approaches 1, which gives \( 2 \).

\subsection{Symbolic Derivatives and Integrals}

In \texttt{Sympy}, you can easily compute symbolic derivatives and integrals, as we've seen earlier.

\textbf{Example: Higher-Order Derivatives}
\begin{lstlisting}[style=python]
from sympy import symbols, diff

# Define the symbolic variable
x = symbols('x')

# Define a function
f = x**4

# Compute the second derivative
f_double_prime = diff(f, x, 2)

print("The second derivative is:", f_double_prime)
\end{lstlisting}

This example computes the second derivative of the function \( f(x) = x^4 \), resulting in \( 12x^2 \).

\textbf{Example: Definite Integral}
\begin{lstlisting}[style=python]
from sympy import symbols, integrate

# Define the symbolic variable
x = symbols('x')

# Define a function
f = x**2

# Compute the definite integral
integral = integrate(f, (x, 0, 3))

print("The definite integral is:", integral)
\end{lstlisting}

This computes the definite integral of \( f(x) = x^2 \) from 0 to 3, yielding \( 9 \).

\section{Fourier Transforms and Fast Fourier Transform (FFT)}

\subsection{Understanding the Mathematics of Fourier Transform}
The Fourier Transform is a mathematical technique that transforms a time-domain signal into its frequency-domain representation\cite{sejdic2011fractional}. This is particularly useful for analyzing the frequency content of signals, especially in fields such as signal processing, physics, and engineering.

The Fourier Transform of a continuous function \( f(t) \) is defined as:

\[
F(\omega) = \int_{-\infty}^{\infty} f(t) e^{-j \omega t} \, dt
\]

Here:
\begin{itemize}
  \item \( t \) represents time.
  \item \( \omega \) represents angular frequency, where \( \omega = 2\pi f \), and \( f \) is the frequency in Hz.
  \item \( j \) is the imaginary unit, where \( j^2 = -1 \).
  \item \( F(\omega) \) represents the Fourier Transform of \( f(t) \).
\end{itemize}

In simple terms, the Fourier Transform converts a time-domain signal into a sum of sinusoids of different frequencies. Each sinusoid has a corresponding amplitude and phase, which together describe how much of that particular frequency is present in the original signal.

\subsection{Implementing Fourier Transform with Numpy}
In Python, we can use the \texttt{numpy} library to perform Fourier Transforms on discrete signals (i.e., sampled data)\cite{Viswanathan2019}. The \texttt{numpy.fft} module provides functions to compute the Discrete Fourier Transform (DFT) and its inverse.

Let’s consider a simple example where we implement the Fourier Transform on a sine wave signal.

\begin{lstlisting}[style=python]
import numpy as np
import matplotlib.pyplot as plt

# Sampling parameters
sampling_rate = 1000  # Samples per second
T = 1.0 / sampling_rate  # Sampling period
L = 1000  # Length of signal
t = np.linspace(0, L*T, L, endpoint=False)  # Time vector

# Define a sine wave signal with frequency 50 Hz
frequency = 50
amplitude = np.sin(2 * np.pi * frequency * t)

# Compute the Fourier Transform using numpy.fft.fft
fft_result = np.fft.fft(amplitude)

# Compute the corresponding frequencies
frequencies = np.fft.fftfreq(L, T)

# Plot the original signal and its Fourier Transform
plt.figure(figsize=(12, 6))

# Original signal plot
plt.subplot(1, 2, 1)
plt.plot(t, amplitude)
plt.title('Time Domain Signal (50 Hz Sine Wave)')
plt.xlabel('Time [s]')
plt.ylabel('Amplitude')

# Fourier Transform plot (only positive frequencies)
plt.subplot(1, 2, 2)
plt.plot(frequencies[:L//2], np.abs(fft_result)[:L//2])
plt.title('Frequency Domain (Fourier Transform)')
plt.xlabel('Frequency [Hz]')
plt.ylabel('Magnitude')

plt.show()
\end{lstlisting}

In this example:
\begin{itemize}
  \item We generate a sine wave with a frequency of 50 Hz.
  \item The \texttt{numpy.fft.fft} function computes the Fourier Transform of the signal.
  \item The result is a complex-valued array where the magnitude gives the amplitude of each frequency component in the signal.
  \item The \texttt{numpy.fft.fftfreq} function returns the corresponding frequencies for each element in the Fourier Transform result.
\end{itemize}

\subsection{Using FFT for Signal Processing}
The Fast Fourier Transform (FFT) is an algorithm that efficiently computes the Discrete Fourier Transform (DFT)\cite{nussbaumer1982fast}. The FFT significantly reduces the computation time compared to the direct calculation of the DFT, making it ideal for real-time signal processing tasks.

One common application of the FFT in signal processing is to filter out noise from a signal. Consider the following example where we generate a noisy signal and use the FFT to remove the high-frequency noise:

\begin{lstlisting}[style=python]
# Generate a noisy signal: 50 Hz sine wave + high-frequency noise
noise = 0.5 * np.random.randn(L)  # Random noise
noisy_signal = amplitude + noise  # Add noise to the sine wave

# Compute the FFT of the noisy signal
noisy_fft = np.fft.fft(noisy_signal)

# Apply a low-pass filter: Set high-frequency components to zero
cutoff_frequency = 100  # Frequency in Hz
noisy_fft[np.abs(frequencies) > cutoff_frequency] = 0

# Compute the inverse FFT to get the filtered signal back in time domain
filtered_signal = np.fft.ifft(noisy_fft)

# Plot the noisy signal and the filtered signal
plt.figure(figsize=(12, 6))

# Noisy signal plot
plt.subplot(1, 2, 1)
plt.plot(t, noisy_signal)
plt.title('Noisy Time Domain Signal')
plt.xlabel('Time [s]')
plt.ylabel('Amplitude')

# Filtered signal plot
plt.subplot(1, 2, 2)
plt.plot(t, filtered_signal.real)
plt.title('Filtered Signal (Low-Pass Filter)')
plt.xlabel('Time [s]')
plt.ylabel('Amplitude')

plt.show()
\end{lstlisting}

In this example:
\begin{itemize}
  \item We generate a noisy signal by adding random noise to a 50 Hz sine wave.
  \item After computing the FFT, we apply a low-pass filter by setting the Fourier coefficients of frequencies higher than the cutoff frequency to zero.
  \item We use the inverse FFT (\texttt{np.fft.ifft}) to convert the filtered signal back into the time domain.
\end{itemize}

\section{Laplace and Z-Transforms}

\subsection{Introduction to Laplace Transform and Applications}
The Laplace Transform is a mathematical operation that transforms a function of time \( f(t) \) into a function of a complex variable \( s \)\cite{Schiff2013}. It is used extensively in the analysis of linear time-invariant (LTI) systems, especially in control systems and circuit analysis.

The Laplace Transform is defined as:

\[
F(s) = \int_{0}^{\infty} f(t) e^{-st} \, dt
\]

where:
\begin{itemize}
  \item \( t \) is the time variable.
  \item \( s \) is a complex variable, where \( s = \sigma + j\omega \).
  \item \( F(s) \) represents the Laplace Transform of \( f(t) \).
\end{itemize}

The Laplace Transform is particularly useful because it converts differential equations into algebraic equations, making them easier to solve. It also provides insights into the stability and transient behavior of systems.

\subsection{Implementing Laplace Transform in Python}
While Python’s \texttt{sympy} library doesn't have a direct method for the Laplace Transform of arbitrary numerical data, it provides symbolic computation tools. Here is an example of how to compute the Laplace Transform of a simple function symbolically:

\begin{lstlisting}[style=python]
from sympy import symbols, laplace_transform, exp

# Define the time variable and function
t, s = symbols('t s')
f = exp(-t)

# Compute the Laplace Transform
F_s = laplace_transform(f, t, s)
print(F_s)
\end{lstlisting}

In this example, we compute the Laplace Transform of \( e^{-t} \), which is a common function used in control systems and signal processing. The result is the symbolic Laplace Transform.

\subsection{Z-Transform in Digital Signal Processing}
The Z-Transform is the discrete-time equivalent of the Laplace Transform and is used extensively in digital signal processing (DSP) to analyze and design digital filters and systems\cite{nair2004digital}.

The Z-Transform of a discrete-time signal \( x[n] \) is defined as:

\[
X(z) = \sum_{n=-\infty}^{\infty} x[n] z^{-n}
\]

where \( z \) is a complex variable.

Like the Laplace Transform, the Z-Transform is useful for analyzing the stability and frequency response of digital systems.

Let’s implement a simple Z-Transform of a discrete signal using Python and symbolic computation:

\begin{lstlisting}[style=python]
from sympy import symbols, Function, summation

# Define discrete-time variable and Z-transform variable
n, z = symbols('n z')
x = Function('x')(n)

# Define a simple discrete-time signal x[n] = 2^n
x_n = 2**n

# Compute the Z-Transform symbolically
X_z = summation(x_n * z**(-n), (n, 0, 10))
print(X_z)
\end{lstlisting}

In this example:
\begin{itemize}
  \item We define the discrete-time signal \( x[n] = 2^n \).
  \item We compute the Z-Transform by summing the series from \( n = 0 \) to \( n = 10 \).
\end{itemize}

The Z-Transform is essential for understanding the behavior of digital filters and systems, especially in applications like digital audio processing and communications systems.

\chapter{Object-Oriented Programming and Modularization in Python}

\section{Object-Oriented Programming Basics}

Object-Oriented Programming (OOP) is a programming paradigm that uses "objects" to model real-world entities and concepts\cite{wegner1990concepts}. Each object is an instance of a class, and classes define the blueprint for these objects. OOP is one of the most powerful ways to write reusable and maintainable code\cite{lott2019mastering}.

\subsection{Defining Classes and Objects in Python}

In Python, a class is defined using the \texttt{class} keyword. A class is like a blueprint for objects, defining attributes (variables) and methods (functions) that the objects created from this class will have. Here's a basic example:

\begin{lstlisting}[style=python]
class Car:
    # Constructor method to initialize the object
    def __init__(self, make, model, year):
        self.make = make
        self.model = model
        self.year = year

    # Method to display car details
    def display_info(self):
        print(f"{self.year} {self.make} {self.model}")

# Creating an instance (object) of the Car class
my_car = Car("Toyota", "Camry", 2021)

# Accessing a method of the Car class
my_car.display_info()
\end{lstlisting}

The explanation of the code is as follows:

\begin{itemize}
    \item \texttt{Car}: This is the class name. By convention, class names in Python are written in CamelCase.
    \item \texttt{self}: This refers to the instance of the class. It allows access to the object's attributes and methods within the class. Every method in a class must have \texttt{self} as the first parameter.
    \item \texttt{\_\_init\_\_}: This is the constructor method that gets called when an object is instantiated. It initializes the object with the provided parameters.
    \item \texttt{display\_info}: This is a method that prints out information about the car. Methods inside classes always have \texttt{self} as their first parameter.
\end{itemize}

In this example, we created an object called \texttt{my\_car} from the \texttt{Car} class and accessed its \texttt{display\_info} method, which prints the car's details.

\subsection{Inheritance and Polymorphism}

Inheritance allows one class (the child class) to inherit attributes and methods from another class (the parent class). This promotes code reuse and is a core concept of OOP.

Here is an example where \texttt{ElectricCar} inherits from the \texttt{Car} class:

\begin{lstlisting}[style=python]
class ElectricCar(Car):
    # Constructor for ElectricCar that extends Car
    def __init__(self, make, model, year, battery_size):
        # Calling the constructor of the parent class (Car)
        super().__init__(make, model, year)
        self.battery_size = battery_size

    # Overriding the display_info method
    def display_info(self):
        print(f"{self.year} {self.make} {self.model} with a {self.battery_size}-kWh battery.")

# Creating an instance of ElectricCar
my_electric_car = ElectricCar("Tesla", "Model S", 2022, 100)
my_electric_car.display_info()
\end{lstlisting}

The explanation of the code is as follows:

\begin{itemize}
    \item \texttt{ElectricCar} is the child class, and it inherits from the \texttt{Car} class.
    \item \texttt{super()} is used to call the parent class's constructor.
    \item The \texttt{display\_info} method is overridden in the child class to provide additional information about the battery size.
\end{itemize}

Polymorphism allows different classes to have methods with the same name but potentially different behavior. For example, both \texttt{Car} and \texttt{ElectricCar} have a \texttt{display\_info} method, but the behavior differs based on the class.

\subsection{Class Methods and Instance Methods}

In Python, methods can be classified into instance methods, class methods, and static methods:

\begin{itemize}
    \item \textit{Instance Methods:} These methods act on an instance of the class and have access to the instance's attributes. These are the most common type of methods and must take \texttt{self} as their first parameter.
    \item \textit{Class Methods:} These methods are called on the class itself rather than on an instance. They are defined using the \texttt{@classmethod} decorator, and they take \texttt{cls} as their first parameter instead of \texttt{self}.
    \item \textit{Static Methods:} These methods neither modify the state of an object nor the state of the class. They are defined using the \texttt{@staticmethod} decorator.
\end{itemize}

Here’s an example demonstrating all three:

\begin{lstlisting}[style=python]
class MathOperations:
    # Class method
    @classmethod
    def square(cls, x):
        return x * x

    # Static method
    @staticmethod
    def add(x, y):
        return x + y

# Using class method
print(MathOperations.square(4))  # Output: 16

# Using static method
print(MathOperations.add(10, 5))  # Output: 15
\end{lstlisting}

In this example, \texttt{square} is a class method, and \texttt{add} is a static method.

\section{Modules and Packages in Python}

\subsection{How to Organize Python Code with Modules}

As your programs become larger, it's essential to organize your code. Python provides an excellent way to do this by using modules. A module is simply a Python file that contains related functions, classes, or variables. You can then import this module into other Python files and reuse its code.

Suppose you have a file \texttt{math\_utils.py} containing useful functions:

\begin{lstlisting}[style=python]
# math_utils.py
def add(a, b):
    return a + b

def subtract(a, b):
    return a - b
\end{lstlisting}

You can import this module into another Python script like so:

\begin{lstlisting}[style=python]
# main.py
import math_utils

result = math_utils.add(5, 3)
print(result)  # Output: 8
\end{lstlisting}

By organizing your code into modules, you make your programs more modular and easier to maintain.

\subsection{Creating and Using Python Packages}

A package is a collection of modules that are grouped together in a directory structure. A package is simply a directory that contains a special file called \texttt{\_\_init\_\_.py}, which indicates to Python that the directory is a package.

For example, the following directory structure shows how to create a package:

\begin{lstlisting}[style=cmd]
mypackage/
    __init__.py
    math_utils.py
    string_utils.py
\end{lstlisting}

Now you can import the package and its modules like this:

\begin{lstlisting}[style=python]
# Importing from the package
from mypackage import math_utils

result = math_utils.add(10, 5)
print(result)
\end{lstlisting}

Packages make it easier to organize large projects and share reusable components.

\subsection{Importing Built-in and Third-Party Modules}

Python comes with a rich set of built-in modules. For example, you can use the \texttt{math} module for mathematical functions:

\begin{lstlisting}[style=python]
import math

# Using the math module to calculate square root
print(math.sqrt(16))  # Output: 4.0
\end{lstlisting}

Python also has an extensive ecosystem of third-party modules, which you can install via \texttt{pip}, Python's package manager. For example, to install and use the popular \texttt{requests} module for making HTTP requests:

\begin{lstlisting}[style=cmd]
pip install requests
\end{lstlisting}

After installing, you can import and use it in your project:

\begin{lstlisting}[style=python]
import requests

response = requests.get('https://api.github.com')
print(response.status_code)
\end{lstlisting}

\section{Introduction to Common Scientific Libraries}

Python has become a popular language for scientific computing, largely due to the availability of powerful libraries like \texttt{NumPy}, \texttt{SciPy}, and \texttt{Matplotlib}.

\subsection{Overview of Numpy, Scipy, and Matplotlib}

\begin{itemize}
    \item \textit{NumPy:} This library is used for numerical computations and provides support for large multidimensional arrays and matrices. It also offers a collection of mathematical functions to operate on these arrays.
    \item \textit{SciPy:} Built on top of NumPy, SciPy provides additional functionality for optimization, integration, interpolation, eigenvalue problems, and more.
    \item \textit{Matplotlib:} This is a plotting library used to create static, interactive, and animated visualizations in Python. It is highly flexible and widely used for data visualization.
\end{itemize}

\subsection{How to Perform Scientific Calculations with These Libraries}

Here is an example of using \texttt{NumPy} and \texttt{Matplotlib} to perform a simple calculation and plot the results:

\begin{lstlisting}[style=python]
import numpy as np
import matplotlib.pyplot as plt

# Create a range of values
x = np.linspace(0, 10, 100)

# Compute the sine of each value
y = np.sin(x)

# Plot the result
plt.plot(x, y)
plt.xlabel('x values')
plt.ylabel('sin(x)')
plt.title('Sine Wave')
plt.show()
\end{lstlisting}

In this example:
\begin{itemize}
    \item We used \texttt{NumPy} to generate an array of 100 values between 0 and 10.
    \item We applied the \texttt{np.sin()} function to compute the sine of each value.
    \item \texttt{Matplotlib} was used to plot the results, and we labeled the axes and title before displaying the plot.
\end{itemize}

Similarly, \texttt{SciPy} provides more specialized functions. For instance, to solve an integral using \texttt{SciPy}:

\begin{lstlisting}[style=python]
from scipy import integrate

# Define the function
def f(x):
    return x ** 2

# Compute the definite integral from 0 to 2
result, _ = integrate.quad(f, 0, 2)
print(f"The integral of x^2 from 0 to 2 is: {result}")
\end{lstlisting}

This example shows how to use \texttt{SciPy} to compute the integral of \(x^2\) from 0 to 2.

\chapter{Project: Implementing Advanced Mathematical Operations}

In this chapter, we will apply advanced mathematical concepts using Python. The focus will be on real-world applications, such as signal processing using Fourier transforms and matrix operations in the context of neural networks. These projects are aimed at giving beginners hands-on experience with complex mathematical operations, making the transition from theory to practice smoother.

\section{Fourier Transform Project}

Fourier Transforms are an essential tool in many fields, such as signal processing, image processing, and data analysis. The goal of this section is to explain the Fourier Transform in a step-by-step manner, using Python and NumPy.

\subsection{Signal Processing: From Time Domain to Frequency Domain}

In signal processing, signals are often represented in the time domain. However, it is sometimes more useful to view the signal in the frequency domain, which shows how much of each frequency is present in the signal. Fourier Transforms allow us to convert a time-domain signal into the frequency domain.

\textbf{What is a Fourier Transform?}

A Fourier Transform breaks down a signal into its constituent frequencies. In Python, this can be achieved using the Fast Fourier Transform (FFT)\cite{nussbaumer1982}, which is an efficient algorithm for computing the Discrete Fourier Transform (DFT)\cite{smith2007mathematics}.

\textbf{Example: Applying FFT to a Simple Signal}

Let’s walk through an example where we apply the FFT to a simple sine wave.

\begin{lstlisting}[style=python]
import numpy as np
import matplotlib.pyplot as plt

# Create a time-domain signal: A 5 Hz sine wave
Fs = 500  # Sampling frequency
T = 1 / Fs  # Sampling interval
t = np.arange(0, 1, T)  # Time vector for 1 second
f = 5  # Frequency of the sine wave
signal = np.sin(2 * np.pi * f * t)

# Compute the Fast Fourier Transform (FFT)
fft_result = np.fft.fft(signal)
fft_freqs = np.fft.fftfreq(len(signal), T)

# Plot the time-domain signal
plt.subplot(2, 1, 1)
plt.plot(t, signal)
plt.title('Time Domain Signal')
plt.xlabel('Time (seconds)')
plt.ylabel('Amplitude')

# Plot the frequency-domain signal (magnitude of FFT)
plt.subplot(2, 1, 2)
plt.plot(fft_freqs[:len(fft_freqs)//2], np.abs(fft_result[:len(fft_result)//2]))
plt.title('Frequency Domain Signal')
plt.xlabel('Frequency (Hz)')
plt.ylabel('Magnitude')

plt.tight_layout()
plt.show()
\end{lstlisting}

In this example:
\begin{itemize}
    \item We create a sine wave with a frequency of 5 Hz.
    \item We then apply the FFT to this signal using \texttt{numpy.fft.fft()}.
    \item The resulting frequency-domain signal is plotted, showing a peak at 5 Hz, which corresponds to the frequency of the original sine wave.
\end{itemize}

\textbf{Explanation of Key Concepts}
\begin{itemize}
    \item \textit{Sampling Frequency (Fs)}: This is the number of samples taken per second. In the example, we sample the signal at 500 Hz\cite{Wahab2016}.
    \item \textit{FFT Output}: The output of FFT is complex numbers. To get the magnitude of the signal in the frequency domain, we use the absolute value of the FFT result.
\end{itemize}

\subsection{Using FFT to Implement Convolution Efficiently}

Convolution is a fundamental operation in signal processing and image processing. It involves combining two signals to form a third signal. Convolution in the time domain can be computationally expensive, especially for large signals. However, the Fourier Transform can be used to compute convolution efficiently\cite{pratt2017fcnn}.

\textbf{Convolution Using Fourier Transforms}

Using the Convolution Theorem\cite{stone2008uniqueness}, we know that convolution in the time domain is equivalent to multiplication in the frequency domain. Here's how we can implement convolution using FFT:

\begin{lstlisting}[style=python]
# Create two signals to convolve
signal1 = np.sin(2 * np.pi * 5 * t)  # 5 Hz sine wave
signal2 = np.ones(50)  # A simple box signal (rectangular pulse)

# Compute the FFT of both signals
fft_signal1 = np.fft.fft(signal1, len(signal1) + len(signal2) - 1)
fft_signal2 = np.fft.fft(signal2, len(signal1) + len(signal2) - 1)

# Multiply in the frequency domain
fft_convolved = fft_signal1 * fft_signal2

# Compute the inverse FFT to get the convolved signal in time domain
convolved_signal = np.fft.ifft(fft_convolved)

# Plot the original and convolved signals
plt.plot(np.real(convolved_signal))
plt.title('Convolution using FFT')
plt.xlabel('Time')
plt.ylabel('Amplitude')
plt.show()
\end{lstlisting}

This method is much faster than direct convolution for long signals. We first take the FFT of both signals, multiply them in the frequency domain, and then use the inverse FFT to get the convolved signal back in the time domain.

\section{Matrix Operations and Their Applications in Neural Networks}

Matrix operations form the backbone of many machine learning algorithms, particularly neural networks. In this section, we will explore the application of matrix operations to build a simple neural network from scratch using NumPy.

\subsection{Building a Simple Neural Network with Numpy}

Neural networks are essentially a series of matrix operations. To demonstrate this, we will build a simple feedforward neural network with one hidden layer using only NumPy.

\textbf{Step-by-Step Guide to Building the Neural Network}

We will create a neural network with:
\begin{itemize}
    \item An input layer with 3 neurons.
    \item A hidden layer with 4 neurons.
    \item An output layer with 1 neuron.
\end{itemize}

\textbf{1. Initialize the Network Weights}

Neural networks have weights and biases that are updated during training. We will initialize these randomly for simplicity.

\begin{lstlisting}[style=python]
# Number of neurons in each layer
input_neurons = 3
hidden_neurons = 4
output_neurons = 1

# Initialize weights randomly with mean 0
np.random.seed(42)
W1 = np.random.randn(input_neurons, hidden_neurons)
W2 = np.random.randn(hidden_neurons, output_neurons)

# Initialize biases to zero
b1 = np.zeros((1, hidden_neurons))
b2 = np.zeros((1, output_neurons))
\end{lstlisting}

\textbf{2. Define the Activation Function}

We will use the sigmoid activation function for the neurons, which squashes the input into a range between 0 and 1.

\begin{lstlisting}[style=python]
# Sigmoid activation function
def sigmoid(x):
    return 1 / (1 + np.exp(-x))

# Derivative of the sigmoid function for backpropagation
def sigmoid_derivative(x):
    return x * (1 - x)
\end{lstlisting}

\textbf{3. Forward Propagation}

During forward propagation, the input is passed through the network to produce the output. This involves several matrix multiplications and the application of activation functions.

\begin{lstlisting}[style=python]
# Forward propagation
def forward_propagation(X):
    # Input to hidden layer
    z1 = np.dot(X, W1) + b1
    a1 = sigmoid(z1)
    
    # Hidden layer to output layer
    z2 = np.dot(a1, W2) + b2
    a2 = sigmoid(z2)
    
    return a1, a2
\end{lstlisting}

\textbf{4. Backpropagation and Weight Updates}

During backpropagation, the error is propagated backwards through the network, and the weights are updated using gradient descent. For simplicity, we will assume we have the error at the output.

\begin{lstlisting}[style=python]
# Backpropagation and weight update
def backpropagation(X, y, a1, a2, learning_rate=0.01):
    # Calculate the error at the output
    error_output = a2 - y
    
    # Calculate the gradient for W2 and b2
    delta_output = error_output * sigmoid_derivative(a2)
    dW2 = np.dot(a1.T, delta_output)
    db2 = np.sum(delta_output, axis=0, keepdims=True)
    
    # Calculate the error at the hidden layer
    error_hidden = np.dot(delta_output, W2.T)
    delta_hidden = error_hidden * sigmoid_derivative(a1)
    dW1 = np.dot(X.T, delta_hidden)
    db1 = np.sum(delta_hidden, axis=0, keepdims=True)
    
    # Update the weights and biases
    W1 -= learning_rate * dW1
    W2 -= learning_rate * dW2
    b1 -= learning_rate * db1
    b2 -= learning_rate * db2
\end{lstlisting}

\textbf{5. Training the Network}

Finally, we train the network using forward and backward propagation for multiple iterations (epochs).

\begin{lstlisting}[style=python]
# Training the network
def train(X, y, epochs=10000, learning_rate=0.01):
    for i in range(epochs):
        a1, a2 = forward_propagation(X)
        backpropagation(X, y, a1, a2, learning_rate)
        if i % 1000 == 0:
            loss = np.mean(np.square(y - a2))
            print(f'Epoch {i}, Loss: {loss}')
            
# Example dataset
X = np.array([[0, 0, 1], [0, 1, 1], [1, 0, 1], [1, 1, 1]])
y = np.array([[0], [1], [1], [0]])

train(X, y)
\end{lstlisting}

\textbf{Explanation of the Process:}
\begin{itemize}
    \item \textit{Forward Propagation}: We compute the output layer’s predictions by passing the input through the network.
    \item \textit{Backpropagation}: We compute the gradients of the error with respect to the weights and update them to minimize the error.
    \item \textit{Training Loop}: The network is trained over many iterations, each time updating the weights to reduce the error.
\end{itemize}

\subsection{Understanding Matrix Operations in Deep Learning}

Matrix operations are the foundation of neural networks. Every layer in a neural network is essentially a series of matrix multiplications:
\[
\mathbf{Z} = \mathbf{W} \cdot \mathbf{X} + \mathbf{b}
\]
where:
\begin{itemize}
    \item \(\mathbf{W}\) represents the weights.
    \item \(\mathbf{X}\) represents the input.
    \item \(\mathbf{b}\) represents the bias.
\end{itemize}

In deep learning, the optimization of these matrices through training is what enables the network to learn from data.

\section{Laplace Transform Applications in Control Systems}

The Laplace Transform is an essential mathematical tool used in control systems for analyzing and designing systems in the frequency domain\cite{broberg1996laplace}. It transforms a time-domain function into a complex frequency-domain function, making it easier to solve differential equations that describe dynamic systems. In control systems, Laplace transforms help in understanding system behavior, stability, and responses to various inputs.

\subsection{Simulating Control Systems with Python}

Python, along with libraries like \texttt{scipy}, can be used to simulate control systems and apply the Laplace transform to analyze system responses. One of the key uses of Laplace transforms in control systems is to model and solve linear time-invariant (LTI) systems\cite{erfani2013characterisation}.

To begin, let's install the necessary libraries:

\begin{lstlisting}[style=cmd]
pip install scipy numpy matplotlib
\end{lstlisting}

We will use the \texttt{scipy.signal} module to represent and simulate control systems. Here's an example of how to define a transfer function and simulate a step response.

\begin{lstlisting}[style=python]
import numpy as np
import scipy.signal as signal
import matplotlib.pyplot as plt

# Define the transfer function G(s) = 1 / (s^2 + 2s + 1)
numerator = [1]  # Coefficients of the numerator (1)
denominator = [1, 2, 1]  # Coefficients of the denominator (s^2 + 2s + 1)

# Create the transfer function
system = signal.TransferFunction(numerator, denominator)

# Simulate the step response of the system
time, response = signal.step(system)

# Plot the step response
plt.plot(time, response)
plt.title('Step Response of the Control System')
plt.xlabel('Time [s]')
plt.ylabel('Response')
plt.grid(True)
plt.show()
\end{lstlisting}

In this example, we define a second-order system represented by the transfer function \( G(s) = \frac{1}{s^2 + 2s + 1} \), which describes a damped system. The step response simulates how the system responds to a unit step input, which is a common way to analyze system behavior.

The graph produced by this code will show the system's response over time.

\section{Comprehensive Project: Frequency Domain Applications in Deep Learning}

Deep learning often deals with data in the time or spatial domains, such as audio signals and images\cite{Najafabadi2015}. However, transforming this data into the frequency domain using techniques like the Fourier Transform allows for more advanced feature extraction and analysis. This section focuses on how the frequency domain is applied in deep learning tasks, particularly in image processing and audio signal analysis.

\subsection{Using Frequency Domain Methods for Image Processing}

In image processing, the frequency domain is useful for tasks such as noise reduction, image compression, and feature extraction. The Fourier Transform, specifically the 2D Fourier Transform, is commonly used to analyze and manipulate the frequency components of an image\cite{bracewell2012fourier,xu2020learning}.

Let's explore how to apply the 2D Fourier Transform to an image using Python and the \texttt{numpy} and \texttt{matplotlib} libraries.

\begin{lstlisting}[style=python]
import numpy as np
import matplotlib.pyplot as plt
from PIL import Image

# Load the image and convert to grayscale
image = Image.open('example_image.jpg').convert('L')
image_array = np.array(image)

# Compute the 2D Fourier Transform of the image
image_fft = np.fft.fft2(image_array)
image_fft_shifted = np.fft.fftshift(image_fft)  # Shift the zero frequency component to the center

# Compute the magnitude spectrum
magnitude_spectrum = np.log(np.abs(image_fft_shifted))

# Plot the original image and its magnitude spectrum
plt.figure(figsize=(12, 6))

plt.subplot(1, 2, 1)
plt.imshow(image_array, cmap='gray')
plt.title('Original Image')
plt.axis('off')

plt.subplot(1, 2, 2)
plt.imshow(magnitude_spectrum, cmap='gray')
plt.title('Magnitude Spectrum (Frequency Domain)')
plt.axis('off')

plt.show()
\end{lstlisting}

This code performs the following steps:
\begin{itemize}
    \item Loads an image and converts it to a grayscale array.
    \item Applies the 2D Fourier Transform to the image.
    \item Shifts the frequency components so that the low frequencies are at the center.
    \item Computes and plots the magnitude spectrum, which visualizes the frequency content of the image.
\end{itemize}

The Fourier Transform provides insights into the frequency characteristics of the image, such as identifying dominant patterns or filtering out noise.

\subsection{Analyzing Audio Signals with Fourier Transform in Deep Learning}

In the field of audio signal analysis, the Fourier Transform is used to convert time-domain signals into the frequency domain. This transformation is essential in tasks such as speech recognition, sound classification, and music analysis\cite{boashash2015time}. Here, we will apply the Fourier Transform to an audio signal and visualize its frequency content.

First, install the \texttt{librosa} library to handle audio files:

\begin{lstlisting}[style=cmd]
pip install librosa
\end{lstlisting}

Now, let's process an audio signal and analyze it using the Short-Time Fourier Transform (STFT).

\begin{lstlisting}[style=python]
import librosa
import numpy as np
import matplotlib.pyplot as plt

# Load an example audio file
audio_file = 'example_audio.wav'
signal, sr = librosa.load(audio_file, sr=None)

# Compute the Short-Time Fourier Transform (STFT)
stft = np.abs(librosa.stft(signal))

# Convert the STFT into decibels for better visualization
stft_db = librosa.amplitude_to_db(stft, ref=np.max)

# Plot the spectrogram (frequency over time)
plt.figure(figsize=(10, 6))
librosa.display.specshow(stft_db, sr=sr, x_axis='time', y_axis='log', cmap='inferno')
plt.colorbar(format='%+2.0f dB')
plt.title('Spectrogram of the Audio Signal')
plt.xlabel('Time [s]')
plt.ylabel('Frequency [Hz]')
plt.show()
\end{lstlisting}

This code performs the following steps:
\begin{itemize}
    \item Loads an audio signal using \texttt{librosa}.
    \item Applies the Short-Time Fourier Transform (STFT) to the signal, which provides a time-frequency representation.
    \item Converts the STFT results into decibels for easier interpretation.
    \item Visualizes the spectrogram, which shows how the frequency content of the signal changes over time.
\end{itemize}

In deep learning, such frequency-domain representations are used for tasks like sound event detection and speech recognition, as they provide more meaningful features for machine learning models compared to raw time-domain signals.

\section{Summary}

In this chapter, we have explored the fundamental mathematical operations in Python, including arithmetic operations, vector and matrix operations, and linear algebra using \texttt{numpy}. We also delved into more advanced topics such as the Laplace Transform's applications in control systems and the Fourier Transform's applications in deep learning. These topics serve as a foundation for applying mathematical methods to real-world problems in engineering, control systems, and deep learning.

By understanding how to manipulate data in both the time and frequency domains, you gain powerful tools for analyzing and solving complex problems, whether in control system design or deep learning applications.

\chapter{Summary and Practice}

This chapter will serve as a comprehensive review of what we have covered so far, including Python's basic data structures, mathematical functions, and essential libraries such as \texttt{Scipy} and \texttt{Sympy}. We will also discuss how you can continue your learning journey in Python and scientific computing. Finally, we will end with a project-based practice, where you will build your own mathematical function library, applying the concepts you have learned.

\section{Review of Python Data Structures and Basic Mathematics}

In this section, we will revisit some key concepts from previous chapters to reinforce your understanding of Python data structures and basic mathematical operations.

\subsection{Python Data Structures}

Python provides several built-in data structures that are essential for handling data efficiently. These include lists, tuples, dictionaries, and sets.

\textbf{1. Lists:} Lists are mutable (modifiable) and can contain items of different types.
\begin{lstlisting}[style=python]
# Example of a list
numbers = [1, 2, 3, 4, 5]
numbers.append(6)  # Adding an element to the list
print(numbers)
\end{lstlisting}

\textbf{2. Tuples:} Tuples are immutable (cannot be modified after creation) and are used for storing fixed collections of items.
\begin{lstlisting}[style=python]
# Example of a tuple
coordinates = (10, 20)
print(coordinates)
\end{lstlisting}

\textbf{3. Dictionaries:} Dictionaries store data in key-value pairs and are very useful when you want to map one value to another.
\begin{lstlisting}[style=python]
# Example of a dictionary
student = {"name": "John", "age": 21}
print(student["name"])
\end{lstlisting}

\textbf{4. Sets:} Sets are unordered collections of unique elements.
\begin{lstlisting}[style=python]
# Example of a set
fruits = {"apple", "banana", "cherry"}
fruits.add("orange")
print(fruits)
\end{lstlisting}

\subsection{Basic Mathematical Operations in Python}

Python allows us to perform various basic mathematical operations using its built-in operators.

\textbf{Addition and Subtraction:}
\begin{lstlisting}[style=python]
a = 10
b = 5
sum_result = a + b
difference = a - b
print("Sum:", sum_result)
print("Difference:", difference)
\end{lstlisting}

\textbf{Multiplication and Division:}
\begin{lstlisting}[style=python]
product = a * b
quotient = a / b
print("Product:", product)
print("Quotient:", quotient)
\end{lstlisting}

\textbf{Exponents and Modulus:}
\begin{lstlisting}[style=python]
exponent = a ** 2  # a raised to the power of 2
modulus = a % b    # Remainder of a divided by b
print("Exponent:", exponent)
print("Modulus:", modulus)
\end{lstlisting}

\subsection{Review of Scipy and Sympy Libraries}

We have introduced \texttt{Scipy} and \texttt{Sympy} as essential tools for performing numerical and symbolic computations, respectively.

\textbf{Scipy Example: Solving an Integral}
\begin{lstlisting}[style=python]
from scipy import integrate

# Define a function to integrate
def f(x):
    return x**2

# Perform numerical integration from 0 to 1
result, _ = integrate.quad(f, 0, 1)
print("Numerical integration result:", result)
\end{lstlisting}

\textbf{Sympy Example: Solving a Derivative}
\begin{lstlisting}[style=python]
from sympy import symbols, diff

# Define a symbolic variable
x = symbols('x')

# Define a function
f = x**2 + 2*x + 1

# Compute the derivative of f with respect to x
f_prime = diff(f, x)
print("Symbolic derivative:", f_prime)
\end{lstlisting}

\section{How to Continue Learning Python and Scientific Computing}

Now that you have a solid foundation in Python and mathematical operations, it is important to continue building on this knowledge. Here are a few recommended steps:

\subsection{Deepening Your Knowledge in Python}

\textbf{1. Explore More Libraries:} Beyond \texttt{Scipy} and \texttt{Sympy}, there are many libraries tailored for specific tasks. For instance:
\begin{itemize}
    \item \texttt{Pandas} for data manipulation and analysis.
    \item \texttt{Matplotlib} and \texttt{Seaborn} for data visualization.
    \item \texttt{TensorFlow} or \texttt{PyTorch} for machine learning.
\end{itemize}

\textbf{2. Practice Coding Challenges:} Websites like \texttt{LeetCode}\cite{leetcode}, \texttt{HackerRank}\cite{hackerrank}, and \texttt{Codewars}\cite{codewars} offer Python challenges that can help you improve your problem-solving skills.

\subsection{Getting Into Scientific Computing}

\textbf{1. Linear Algebra and Matrix Operations:}
Study the use of \texttt{NumPy} and \texttt{Scipy} for performing linear algebra operations. These are crucial for many areas of science and engineering.

\textbf{2. Learn Optimization:}
Optimization is a core area of scientific computing. Using libraries like \texttt{scipy.optimize}, you can solve complex optimization problems.

\textbf{3. Dive Into Machine Learning:}
Python’s ecosystem includes powerful machine learning libraries like \texttt{scikit-learn} and \texttt{Keras}. As you gain confidence, try learning about machine learning models and how they can be applied to real-world data.

\section{Project-Based Practice: Building Your Own Mathematical Function Library}

As a final exercise, we will create a small project where you will build your own Python library to perform various mathematical operations. This project will solidify your understanding and give you hands-on experience in building reusable code.

\subsection{Step 1: Create the Function Library}

We will create a file called \texttt{mymath.py}, which will contain all the mathematical functions.

\begin{lstlisting}[style=python]
# mymath.py

def add(a, b):
    """Add two numbers."""
    return a + b

def subtract(a, b):
    """Subtract two numbers."""
    return a - b

def multiply(a, b):
    """Multiply two numbers."""
    return a * b

def divide(a, b):
    """Divide two numbers."""
    if b == 0:
        return "Cannot divide by zero!"
    return a / b
\end{lstlisting}

This file contains functions for basic operations like addition, subtraction, multiplication, and division.

\subsection{Step 2: Create Advanced Mathematical Functions}

Now, let’s extend our library to include more advanced functions, such as solving quadratic equations and performing integration using \texttt{Scipy}.

\begin{lstlisting}[style=python]
# mymath.py (extended)

import math
from scipy import integrate

def quadratic_roots(a, b, c):
    """Solve a quadratic equation ax^2 + bx + c = 0."""
    discriminant = b**2 - 4*a*c
    if discriminant < 0:
        return "No real roots"
    root1 = (-b + math.sqrt(discriminant)) / (2*a)
    root2 = (-b - math.sqrt(discriminant)) / (2*a)
    return root1, root2

def integrate_function(f, start, end):
    """Numerically integrate a function from start to end."""
    result, _ = integrate.quad(f, start, end)
    return result
\end{lstlisting}

The function \texttt{quadratic\_roots()} solves quadratic equations, and \texttt{integrate\_function()} performs numerical integration using \texttt{Scipy}.

\subsection{Step 3: Test the Library}

Once you have built your library, you can test it by importing the functions and using them in another Python file or directly in a Python shell.

\begin{lstlisting}[style=python]
# test_mymath.py

from mymath import add, quadratic_roots, integrate_function

# Test the add function
print("Addition result:", add(10, 5))

# Test the quadratic_roots function
print("Quadratic roots:", quadratic_roots(1, -3, 2))

# Test the integrate_function
result = integrate_function(lambda x: x**2, 0, 1)
print("Integration result:", result)
\end{lstlisting}

\subsection{Step 4: Final Thoughts on the Project}

Congratulations! You have built your own Python library for basic and advanced mathematical operations. This is a great step toward creating reusable code and practicing modular programming, which is an important aspect of professional coding.

Continue practicing by expanding your library. You can add more functions for calculus, statistics, and other mathematical areas you are interested in.

\part{Basic Mathematics in Deep Learning Programming}

\chapter{Introduction}

\section{Mathematical Foundations in Deep Learning}
Deep learning is a subset of machine learning that relies heavily on mathematical concepts to train and develop neural networks. Understanding the core mathematical principles is crucial for building and optimizing deep learning models. The key areas of mathematics that are fundamental to deep learning include:

\begin{itemize}
  \item \textbf{Linear Algebra}\cite{greub2012linear}: Essential for understanding tensors, matrix operations, and vector spaces.
  \item \textbf{Calculus}\cite{courant1965introduction}: Necessary for understanding how optimization works (e.g., gradient descent).
  \item \textbf{Probability and Statistics}\cite{rohatgi2015introduction}: Important for understanding how models handle uncertainty, interpret data, and measure performance.
  \item \textbf{Optimization}\cite{diwekar2020introduction}: Helps in tuning models to minimize loss functions and improve accuracy.
\end{itemize}

In this chapter, we will explore these mathematical concepts and how they relate to deep learning through practical implementations in PyTorch and TensorFlow, two of the most widely used deep learning frameworks.

\section{Importance of Linear Algebra and Matrix Operations}
Linear algebra plays a pivotal role in deep learning because data is often represented as high-dimensional arrays or matrices (tensors)\cite{cichocki2018tensor}. Neural networks perform numerous matrix and vector operations such as dot products, matrix multiplication, and element-wise operations, which require a solid understanding of linear algebra.

Matrix operations allow neural networks to perform transformations and combine data efficiently. For example:
\begin{itemize}
  \item Multiplying inputs with weights in neural networks can be seen as matrix multiplication.
  \item Activation functions are applied element-wise across tensors.
  \item Computing gradients during backpropagation involves manipulating tensors.
\end{itemize}

In deep learning, tensors are the generalization of matrices to higher dimensions. Tensors are the core data structure, and their manipulation is crucial for training deep learning models.

\section{PyTorch and TensorFlow for Mathematical Computations}
Both PyTorch and TensorFlow are powerful libraries designed for numerical computation, specializing in handling large-scale tensor operations with automatic differentiation\cite{stevens2020deep}. This makes them ideal for implementing deep learning models.

\begin{itemize}
  \item \textbf{PyTorch:} Known for its dynamic computational graph, PyTorch allows users to define models and compute gradients on the fly, making it easier to debug and experiment.
  \item \textbf{TensorFlow:} TensorFlow uses a static computation graph by default, which is more efficient for deployment but can be less intuitive during model development. However, TensorFlow 2.0 introduced eager execution, making it more user-friendly like PyTorch.
\end{itemize}

Both frameworks provide tools for handling tensor operations efficiently, which is essential for working with deep learning models.

\chapter{Tensors: The Core Data Structure}

\section{Definition of Tensors}
A tensor is a generalization of matrices to higher dimensions. While a scalar is a single number (zero-dimensional), a vector is a one-dimensional array, and a matrix is a two-dimensional array, tensors can be any-dimensional arrays\cite{hutchison2016lara}. They are the central data structure in deep learning, representing the data, weights, gradients, and other parameters in a neural network.

Formally, a tensor is defined as:

\[
\text{Tensor} = \left[ T_{i_1, i_2, \dots, i_n} \right]
\]

Where \( T_{i_1, i_2, \dots, i_n} \) represents the elements of the tensor, and \( n \) indicates the number of dimensions or rank of the tensor.

\section{Creating Tensors in PyTorch and TensorFlow}
Creating tensors is one of the most fundamental tasks in deep learning. Both PyTorch and TensorFlow provide multiple ways to create tensors, ranging from initializing with specific values to random initialization.

\subsection{Creating Tensors in PyTorch}
Here are some examples of creating tensors using PyTorch:

\begin{lstlisting}[style=python]
import torch

# Creating a tensor filled with zeros
tensor_zeros = torch.zeros(3, 3)
print(tensor_zeros)

# Creating a tensor filled with ones
tensor_ones = torch.ones(2, 2)
print(tensor_ones)

# Creating a random tensor
tensor_random = torch.rand(4, 4)
print(tensor_random)
\end{lstlisting}

\subsection{Creating Tensors in TensorFlow}
Similarly, TensorFlow also provides functions to create tensors:

\begin{lstlisting}[style=python]
import tensorflow as tf

# Creating a tensor filled with zeros
tensor_zeros = tf.zeros([3, 3])
print(tensor_zeros)

# Creating a tensor filled with ones
tensor_ones = tf.ones([2, 2])
print(tensor_ones)

# Creating a random tensor
tensor_random = tf.random.uniform([4, 4])
print(tensor_random)
\end{lstlisting}

\section{Tensor Shapes and Dimensionality}
The shape of a tensor refers to its dimensions and the size of each dimension. For example, a tensor with shape \( (3, 4) \) has 3 rows and 4 columns. Higher-dimensional tensors can have shapes like \( (3, 4, 5) \), where the first dimension has 3 slices, each containing a \( 4 \times 5 \) matrix.

Here’s how you can check the shape of a tensor in PyTorch and TensorFlow:

\begin{lstlisting}[style=python]
# PyTorch example
tensor = torch.rand(3, 4, 5)
print(tensor.shape)  # Output: torch.Size([3, 4, 5])

# TensorFlow example
tensor = tf.random.uniform([3, 4, 5])
print(tensor.shape)  # Output: (3, 4, 5)
\end{lstlisting}

The rank of a tensor refers to the number of dimensions it has. A scalar has rank 0, a vector has rank 1, a matrix has rank 2, and so on.

\section{Basic Tensor Operations}

\subsection{Tensor Initialization (zeros, ones, random)}
Initializing tensors is the first step in most deep learning tasks. Common initialization methods include tensors filled with zeros, ones, or random values\cite{narkhede2022weight}.

\textbf{Examples in PyTorch:}

\begin{lstlisting}[style=python]
# Tensor of zeros
tensor_zeros = torch.zeros(3, 3)

# Tensor of ones
tensor_ones = torch.ones(2, 2)

# Random tensor
tensor_random = torch.rand(4, 4)
\end{lstlisting}

\textbf{Examples in TensorFlow:}

\begin{lstlisting}[style=python]
# Tensor of zeros
tensor_zeros = tf.zeros([3, 3])

# Tensor of ones
tensor_ones = tf.ones([2, 2])

# Random tensor
tensor_random = tf.random.uniform([4, 4])
\end{lstlisting}

\subsection{Reshaping, Slicing, and Indexing Tensors}
Manipulating tensor shapes and extracting specific elements or subarrays from tensors is a key aspect of deep learning programming. Let’s explore reshaping, slicing, and indexing.

\textbf{Reshaping Tensors:}

Reshaping allows you to change the dimensions of a tensor without altering its data. This is useful in many neural network operations where you need to ensure that inputs, weights, or outputs have compatible shapes.

\begin{lstlisting}[style=python]
# PyTorch example
tensor = torch.rand(4, 4)
reshaped_tensor = tensor.view(2, 8)  # Reshaping to 2 rows and 8 columns
print(reshaped_tensor)

# TensorFlow example
tensor = tf.random.uniform([4, 4])
reshaped_tensor = tf.reshape(tensor, [2, 8])
print(reshaped_tensor)
\end{lstlisting}

\textbf{Slicing and Indexing:}

Slicing allows you to extract specific parts of a tensor. It works similarly to slicing in Python lists.

\begin{lstlisting}[style=python]
# PyTorch slicing
tensor = torch.rand(4, 4)
print(tensor[:2, :2])  # Extracts the first 2 rows and columns

# TensorFlow slicing
tensor = tf.random.uniform([4, 4])
print(tensor[:2, :2])  # Extracts the first 2 rows and columns
\end{lstlisting}

\subsection{Broadcasting in Tensor Operations}
Broadcasting allows you to perform operations on tensors of different shapes. This is a powerful feature that simplifies mathematical operations in deep learning. Instead of manually reshaping tensors to have the same shape, broadcasting automatically adjusts the smaller tensor to match the dimensions of the larger tensor.

\textbf{Example in PyTorch:}

\begin{lstlisting}[style=python]
# Adding a scalar to a tensor
tensor = torch.rand(3, 3)
result = tensor + 5  # Broadcasting automatically adds 5 to each element
print(result)
\end{lstlisting}

\textbf{Example in TensorFlow:}

\begin{lstlisting}[style=python]
# Adding a scalar to a tensor
tensor = tf.random.uniform([3, 3])
result = tensor + 5  # Broadcasting automatically adds 5 to each element
print(result)
\end{lstlisting}

Broadcasting rules can be tricky at first, but they greatly simplify tensor operations when applied correctly.

\chapter{Basic Arithmetic Operations}

In this chapter, we will cover the fundamental arithmetic operations in Python. These operations form the basis of all mathematical calculations and are essential for both beginner and advanced users. Python makes it easy to perform these operations with both single values and larger data structures like arrays.

\section{Element-wise Operations}

Element-wise operations are those that are applied to each element of a data structure individually\cite{Johnsson1990}. In Python, we can easily perform element-wise operations on arrays and lists using libraries like \texttt{NumPy}.

\subsection{Addition, Subtraction, Multiplication, Division}

The basic arithmetic operations include addition, subtraction, multiplication, and division. These operations can be performed on scalars (individual numbers) or element-wise on arrays.

\textbf{Scalar Operations}

Here’s how you can perform these basic arithmetic operations with individual numbers:

\begin{lstlisting}[style=python]
a = 10
b = 5

# Addition
print(a + b)  # Output: 15

# Subtraction
print(a - b)  # Output: 5

# Multiplication
print(a * b)  # Output: 50

# Division
print(a / b)  # Output: 2.0
\end{lstlisting}

\textbf{Element-wise Operations on Arrays}

To perform element-wise operations on arrays, we need to use the \texttt{NumPy} library, which is designed for numerical computations.

\begin{lstlisting}[style=python]
import numpy as np

# Define two arrays
arr1 = np.array([1, 2, 3])
arr2 = np.array([4, 5, 6])

# Element-wise addition
print(arr1 + arr2)  # Output: [5 7 9]

# Element-wise subtraction
print(arr1 - arr2)  # Output: [-3 -3 -3]

# Element-wise multiplication
print(arr1 * arr2)  # Output: [4 10 18]

# Element-wise division
print(arr1 / arr2)  # Output: [0.25 0.4 0.5]
\end{lstlisting}

In the above example, operations are applied to each element of the arrays independently. This feature makes Python highly efficient for numerical computations, especially with large datasets.

\section{Reduction Operations}

Reduction operations are those that reduce a set of values down to a single value. Common reduction operations include finding the sum, mean, maximum, and minimum of a set of numbers.

\subsection{Sum, Mean, Max, Min}

These operations can be performed both on scalar values and arrays.

\textbf{Scalar Reduction}

For scalar values, reduction operations are straightforward:

\begin{lstlisting}[style=python]
a = 10
b = 5

# Sum
print(a + b)  # Output: 15

# Max
print(max(a, b))  # Output: 10

# Min
print(min(a, b))  # Output: 5
\end{lstlisting}

\textbf{Reduction Operations on Arrays}

Using \texttt{NumPy}, we can easily perform reduction operations on arrays:

\begin{lstlisting}[style=python]
import numpy as np

arr = np.array([1, 2, 3, 4, 5])

# Sum of all elements
print(np.sum(arr))  # Output: 15

# Mean (average) of all elements
print(np.mean(arr))  # Output: 3.0

# Maximum value
print(np.max(arr))  # Output: 5

# Minimum value
print(np.min(arr))  # Output: 1
\end{lstlisting}

Reduction operations are essential when working with large datasets, where you often need a summary statistic or an aggregate measure.

\chapter{Matrix Operations}

Matrices are an essential part of many mathematical fields, especially in linear algebra. Python, with the help of libraries like \texttt{NumPy}, provides powerful tools for performing matrix operations easily and efficiently.

\section{Matrix Multiplication}

Matrix multiplication is a key operation in many areas, including graphics, physics, and machine learning. In Python, we can perform matrix multiplication using the \texttt{NumPy} function \texttt{dot}.

Here’s an example of multiplying two matrices:

\begin{lstlisting}[style=python]
import numpy as np

# Define two matrices
A = np.array([[1, 2], [3, 4]])
B = np.array([[5, 6], [7, 8]])

# Matrix multiplication
C = np.dot(A, B)
print(C)
\end{lstlisting}

Output:

\[
\begin{bmatrix}
1*5 + 2*7 & 1*6 + 2*8 \\
3*5 + 4*7 & 3*6 + 4*8
\end{bmatrix}
=
\begin{bmatrix}
19 & 22 \\
43 & 50
\end{bmatrix}
\]

In this example, we defined two 2x2 matrices and performed matrix multiplication using \texttt{np.dot}. Matrix multiplication follows the rule where the element at position (i, j) in the resulting matrix is computed as the dot product of the i-th row of the first matrix and the j-th column of the second matrix.

\section{Optimization of Matrix Multiplication}

Matrix multiplication is a fundamental operation in various fields of mathematics and computer science, particularly in linear algebra, computer graphics, machine learning, and numerical analysis\cite{BiniPan2012}. Despite its apparent simplicity, matrix multiplication can be computationally intensive, especially for large matrices. Thus, understanding the intricacies of this operation and exploring optimization techniques is crucial for improving performance in practical applications. 

\subsection{Basics of Matrix Multiplication}

Matrix multiplication involves two matrices \(A\) and \(B\) and results in a new matrix \(C\). The dimensions of the matrices must align for multiplication to occur: if \(A\) is an \(m \times n\) matrix and \(B\) is an \(n \times p\) matrix, the resulting matrix \(C\) will be \(m \times p\).

The entry \(c_{ij}\) of the resulting matrix \(C\) is computed as follows:

\[
c_{ij} = \sum_{k=1}^{n} a_{ik} b_{kj}
\]

This means that each entry in the resulting matrix is the sum of the products of corresponding entries from the row of \(A\) and the column of \(B\).

\subsubsection{Example of Matrix Multiplication}

Let's illustrate this with a simple example. Consider the following matrices \(A\) and \(B\):

\[
A = \begin{pmatrix}
1 & 2 & 3 \\
4 & 5 & 6
\end{pmatrix}, \quad 
B = \begin{pmatrix}
7 & 8 \\
9 & 10 \\
11 & 12
\end{pmatrix}
\]

To find the product \(C = A \times B\), we compute each element of \(C\):

\[
C = \begin{pmatrix}
(1 \times 7 + 2 \times 9 + 3 \times 11) & (1 \times 8 + 2 \times 10 + 3 \times 12) \\
(4 \times 7 + 5 \times 9 + 6 \times 11) & (4 \times 8 + 5 \times 10 + 6 \times 12)
\end{pmatrix} = \begin{pmatrix}
58 & 64 \\
139 & 154
\end{pmatrix}
\]

This example demonstrates the straightforward nature of matrix multiplication, where each entry in the resulting matrix is derived from a combination of row and column elements.

\subsection{Traditional Matrix Multiplication}

The traditional method of matrix multiplication has a time complexity of \(O(n^3)\), where \(n\) is the dimension of the square matrices being multiplied\cite{Iwen2009}. This cubic complexity arises from the need to compute each entry of the resulting matrix independently, leading to a significant increase in computation time as matrix sizes grow.

The process can be visually represented as follows:

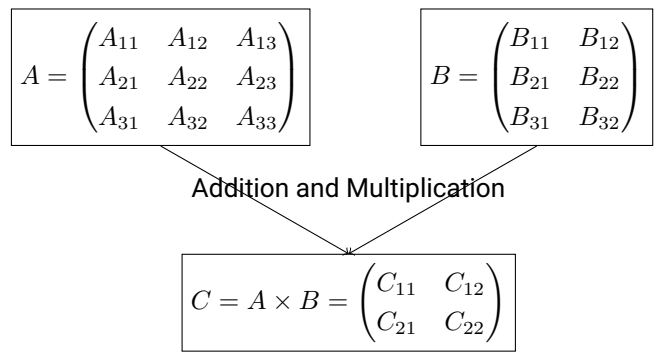
\begin{center}
\begin{tikzpicture}
    \node (A) [draw] at (0, 0) {$A = \begin{pmatrix}
        A_{11} & A_{12} & A_{13} \\
        A_{21} & A_{22} & A_{23} \\
        A_{31} & A_{32} & A_{33}
    \end{pmatrix}$};
    
    \node (B) [draw] at (5, 0) {$B = \begin{pmatrix}
        B_{11} & B_{12} \\
        B_{21} & B_{22} \\
        B_{31} & B_{32}
    \end{pmatrix}$};
    
    \node (C) [draw] at (2.5, -3) {$C = A \times B = \begin{pmatrix}
        C_{11} & C_{12} \\
        C_{21} & C_{22}
    \end{pmatrix}$};

    \draw[->] (A.south) -- (C.north);
    \draw[->] (B.south) -- (C.north);
    \node at (2.5, -1.5) {Addition and Multiplication};
\end{tikzpicture}
\end{center}

This diagram illustrates how the elements from matrices \(A\) and \(B\) are used to calculate the elements of matrix \(C\).

\subsection{Strassen's Algorithm}

Strassen's algorithm, introduced by Volker Strassen in 1969, significantly reduces the computational complexity of matrix multiplication from \(O(n^3)\) to approximately \(O(n^{2.81})\)\cite{strassen1969gaussian}. This algorithm achieves this reduction by utilizing a divide-and-conquer approach, which minimizes the number of multiplications required.

\subsubsection{How Strassen's Algorithm Works}

Strassen's algorithm works by recursively dividing each matrix into four submatrices. Given two \(n \times n\) matrices \(A\) and \(B\):

\[
A = \begin{pmatrix}
A_{11} & A_{12} \\
A_{21} & A_{22}
\end{pmatrix}, \quad 
B = \begin{pmatrix}
B_{11} & B_{12} \\
B_{21} & B_{22}
\end{pmatrix}
\]

Strassen's algorithm requires seven multiplications of these submatrices instead of eight, as required by the conventional approach. The seven multiplications are defined as follows:

\begin{align*}
M_1 &= (A_{11} + A_{22})(B_{11} + B_{22}) \\
M_2 &= (A_{21} + A_{22})B_{11} \\
M_3 &= A_{11}(B_{12} - B_{22}) \\
M_4 &= A_{22}(B_{21} - B_{11}) \\
M_5 &= (A_{11} + A_{12})B_{22} \\
M_6 &= (A_{21} - A_{11})(B_{11} + B_{12}) \\
M_7 &= (A_{12} - A_{22})(B_{21} + B_{22})
\end{align*}

The resulting submatrices \(C_{ij}\) of the product matrix \(C\) are computed using these multiplications:

\begin{align*}
C_{11} &= M_1 + M_4 - M_5 + M_7 \\
C_{12} &= M_3 + M_5 \\
C_{21} &= M_2 + M_4 \\
C_{22} &= M_1 - M_2 + M_3 + M_6
\end{align*}

\subsubsection{Python Implementation of Strassen's Algorithm}

Here is a Python implementation of Strassen's algorithm, which demonstrates how to recursively multiply matrices using the principles outlined above.

\begin{lstlisting}[style=python]
import numpy as np

def strassen(A, B):
    # Base case for recursion
    if len(A) == 1:
        return A * B

    # Splitting the matrices into quadrants
    mid = len(A) // 2

    A11 = A[:mid, :mid]
    A12 = A[:mid, mid:]
    A21 = A[mid:, :mid]
    A22 = A[mid:, mid:]

    B11 = B[:mid, :mid]
    B12 = B[:mid, mid:]
    B21 = B[mid:, :mid]
    B22 = B[mid:, mid:]

    # Strassen's algorithm recursive calls
    M1 = strassen(A11 + A22, B11 + B22)
    M2 = strassen(A21 + A22, B11)
    M3 = strassen(A11, B12 - B22)
    M4 = strassen(A22, B21 - B11)
    M5 = strassen(A11 + A12, B22)
    M6 = strassen(A21 - A11, B11 + B12)
    M7 = strassen(A12 - A22, B21 + B22)

    # Combining the results into the final matrix
    C11 = M1 + M4 - M5 + M7
    C12 = M3 + M5
    C21 = M2 + M4
    C22 = M1 - M2 + M3 + M6

    # Constructing the final matrix from quadrants
    C = np.zeros((len(A), len(B)))
    C[:mid, :mid] = C11
    C[:mid, mid:] = C12
    C[mid:, :mid] = C21
    C[mid:, mid:] = C22

    return C
\end{lstlisting}

This implementation effectively uses recursion to break down the matrix multiplication into smaller components, applying Strassen's optimization techniques to reduce the number of multiplicative operations.

\subsection{Further Improvements in Matrix Multiplication}

While Strassen's algorithm represents a significant improvement over the standard method, further advancements have been made in the field of matrix multiplication. Below are some notable algorithms that have emerged since Strassen's work.

\subsubsection{Coppersmith-Winograd Algorithm}

The Coppersmith-Winograd algorithm further reduced the complexity of matrix multiplication to approximately \(O(n^{2.376})\)\cite{coppersmith1987matrix}. This algorithm utilizes advanced mathematical techniques involving tensor rank and is considered to be more theoretical due to its complexity and the overhead associated with its practical implementation.

\subsubsection{Recent Advances}

Recent research has led to even faster algorithms, some of which leverage techniques from algebraic geometry and combinatorial optimization. Notably, there have been advancements that utilize fast Fourier transforms (FFT) for multiplying polynomials, which can be adapted to matrix multiplication scenarios, yielding further reductions in complexity\cite{tolimieri2012mathematics}.

\subsection{Practical Considerations}

While theoretical advancements in matrix multiplication are significant, practical implementations also play a crucial role. Libraries such as NumPy, TensorFlow, and PyTorch implement highly optimized versions of matrix multiplication, often utilizing hardware acceleration (such as GPU computation) to enhance performance. These libraries abstract the complexity of advanced algorithms, allowing users to perform matrix operations efficiently without delving into the underlying mathematics.

In practical applications, choosing the appropriate algorithm or library depends on various factors, including matrix size, sparsity, and the computational environment (e.g., CPU vs. GPU). It is crucial for developers to consider these aspects when optimizing their applications for matrix operations.

\subsection{Conclusion}

Matrix multiplication is a cornerstone of many computational applications. While the naive approach is straightforward, the advent of algorithms like Strassen's and subsequent improvements highlights the importance of optimization in computational mathematics. By leveraging advanced techniques and utilizing efficient libraries, one can achieve significant performance improvements in matrix computations, which is essential for handling large-scale problems in science and engineering. Understanding these algorithms not only enhances computational efficiency but also deepens our grasp of the mathematical principles underlying linear algebra.

\section{Transpose of a Matrix}

The transpose of a matrix is obtained by swapping its rows and columns. This operation is useful in many linear algebraic contexts, including solving systems of equations and simplifying matrix expressions.

In Python, we can transpose a matrix using the \texttt{.T} attribute of a \texttt{NumPy} array:

\begin{lstlisting}[style=python]
import numpy as np

# Define a matrix
A = np.array([[1, 2, 3], [4, 5, 6]])

# Transpose the matrix
A_T = A.T
print(A_T)
\end{lstlisting}

Output:

\[
\begin{bmatrix}
1 & 2 & 3 \\
4 & 5 & 6
\end{bmatrix}
^T
=
\begin{bmatrix}
1 & 4 \\
2 & 5 \\
3 & 6
\end{bmatrix}
\]

In this example, we transposed a 2x3 matrix into a 3x2 matrix.

\section{Inverse of a Matrix}

The inverse of a square matrix \( A \) is another matrix \( A^{-1} \) such that \( A \times A^{-1} = I \), where \( I \) is the identity matrix. Not all matrices have inverses, but for those that do, \texttt{NumPy} provides the function \texttt{np.linalg.inv()} to compute it.

Here’s an example of finding the inverse of a matrix:

\begin{lstlisting}[style=python]
import numpy as np

# Define a matrix
A = np.array([[1, 2], [3, 4]])

# Compute the inverse
A_inv = np.linalg.inv(A)
print(A_inv)
\end{lstlisting}

Output:

\[
A^{-1} =
\begin{bmatrix}
-2 & 1 \\
1.5 & -0.5
\end{bmatrix}
\]

In this example, we used the \texttt{np.linalg.inv()} function to calculate the inverse of a 2x2 matrix.

\section{Determinant of a Matrix}

The determinant is a scalar value that can be computed from a square matrix and it provides important properties related to the matrix. The determinant of a matrix can be computed using \texttt{np.linalg.det()}.

Here’s an example of finding the determinant of a matrix:

\begin{lstlisting}[style=python]
import numpy as np

# Define a matrix
A = np.array([[1, 2], [3, 4]])

# Compute the determinant
det_A = np.linalg.det(A)
print(det_A)
\end{lstlisting}

Output:

\[
\text{det}(A) = -2.0000000000000004
\]

In this example, the determinant of matrix \( A \) is calculated as -2. Determinants are particularly useful in determining whether a matrix is invertible (a matrix is invertible if and only if its determinant is non-zero).

\section{Eigenvalues and Eigenvectors}

Eigenvalues and eigenvectors are fundamental concepts in linear algebra. For a square matrix \( A \), an eigenvector \( v \) and an eigenvalue \( \lambda \) satisfy the equation:

\[
A v = \lambda v
\]

In Python, we can compute the eigenvalues and eigenvectors of a matrix using \texttt{np.linalg.eig()}.

Here’s an example:

\begin{lstlisting}[style=python]
import numpy as np

# Define a matrix
A = np.array([[1, 2], [2, 3]])

# Compute the eigenvalues and eigenvectors
eigenvalues, eigenvectors = np.linalg.eig(A)
print("Eigenvalues:", eigenvalues)
print("Eigenvectors:", eigenvectors)
\end{lstlisting}

Output:

\[
\text{Eigenvalues} = [4.23606798, -0.23606798]
\]
\[
\text{Eigenvectors} = \begin{bmatrix} 0.52573111 & -0.85065081 \\ 0.85065081 & 0.52573111 \end{bmatrix}
\]

In this example, we used the \texttt{np.linalg.eig()} function to compute the eigenvalues and eigenvectors of matrix \( A \). Eigenvalues and eigenvectors are widely used in many applications, such as solving systems of linear equations, stability analysis, and quantum mechanics.

\chapter{Solving Systems of Linear Equations}

Solving systems of linear equations is a fundamental problem in mathematics and science. In Python, there are several methods available to solve these problems efficiently, particularly when the system of equations can be represented as a matrix equation. In this chapter, we will explore different techniques to solve linear equations using matrix operations.

\section{Using Matrix Inverse to Solve Equations}

One of the most common methods to solve a system of linear equations is by using the matrix inverse. Given a system of equations:

\[
A \mathbf{x} = \mathbf{b}
\]

where \( A \) is a matrix, \( \mathbf{x} \) is the vector of unknowns, and \( \mathbf{b} \) is the vector of constants, we can solve for \( \mathbf{x} \) by computing the inverse of matrix \( A \):

\[
\mathbf{x} = A^{-1} \mathbf{b}
\]

This method works well when the matrix \( A \) is invertible. Let’s look at how we can implement this using Python.

\textbf{Example: Solving a system using matrix inverse}

Consider the following system of equations:

\begin{align*}
x + 2y &= 5 \\
3x + 4y &= 6
\end{align*}

This can be written in matrix form as:

\[
\begin{bmatrix} 1 & 2 \\ 3 & 4 \end{bmatrix} \begin{bmatrix} x \\ y \end{bmatrix} = \begin{bmatrix} 5 \\ 6 \end{bmatrix}
\]

To solve for \( x \) and \( y \), we will use the matrix inverse method:

\begin{lstlisting}[style=python]
import numpy as np

# Define the coefficient matrix A and the constant vector b
A = np.array([[1, 2], [3, 4]])
b = np.array([5, 6])

# Compute the inverse of matrix A
A_inv = np.linalg.inv(A)

# Solve for x
x = np.dot(A_inv, b)
print(x)
\end{lstlisting}

This will output the solution:

\begin{lstlisting}[style=cmd]
[-4.   4.5]
\end{lstlisting}

Thus, \( x = -4 \) and \( y = 4.5 \).

While using the matrix inverse is straightforward, it can be inefficient and numerically unstable for large matrices. For large systems, methods such as LU decomposition are preferred.

\section{LU Decomposition}

LU decomposition is a method that decomposes a matrix \( A \) into two matrices: a lower triangular matrix \( L \) and an upper triangular matrix \( U \)\cite{grasedyck2009domain}. This decomposition can simplify the process of solving systems of equations.

The matrix equation \( A \mathbf{x} = \mathbf{b} \) can be written as:

\[
LU \mathbf{x} = \mathbf{b}
\]

We first solve \( L \mathbf{y} = \mathbf{b} \), and then solve \( U \mathbf{x} = \mathbf{y} \).

\textbf{Example: LU Decomposition in Python}

Here’s how we can solve the same system of equations using LU decomposition in Python:

\begin{lstlisting}[style=python]
import scipy.linalg as la

# Define the coefficient matrix A and the constant vector b
A = np.array([[1, 2], [3, 4]])
b = np.array([5, 6])

# Perform LU decomposition
P, L, U = la.lu(A)

# Solve L*y = b
y = np.linalg.solve(L, b)

# Solve U*x = y
x = np.linalg.solve(U, y)
print(x)
\end{lstlisting}

This will output:

\begin{lstlisting}[style=cmd]
[-4.   4.5]
\end{lstlisting}

LU decomposition is a more efficient method than directly using the matrix inverse for large systems.

\section{QR Decomposition}

QR decomposition decomposes a matrix \( A \) into an orthogonal matrix \( Q \) and an upper triangular matrix \( R \). This method is particularly useful in solving linear systems and least squares problems\cite{Agarwal2014}.

Given:

\[
A \mathbf{x} = \mathbf{b}
\]

We decompose \( A \) as \( A = QR \), and the system becomes:

\[
QR \mathbf{x} = \mathbf{b}
\]

We first solve \( Q^\top \mathbf{y} = \mathbf{b} \) and then solve \( R \mathbf{x} = \mathbf{y} \).

\textbf{Example: QR Decomposition in Python}

Let’s solve the same system of equations using QR decomposition:

\begin{lstlisting}[style=python]
# Perform QR decomposition
Q, R = np.linalg.qr(A)

# Solve Q.T * y = b
y = np.dot(Q.T, b)

# Solve R * x = y
x = np.linalg.solve(R, y)
print(x)
\end{lstlisting}

This will output:

\begin{lstlisting}[style=cmd]
[-4.   4.5]
\end{lstlisting}

QR decomposition is numerically stable and can be used for solving both linear systems and least squares problems.

\chapter{Norms and Distance Metrics}

In linear algebra and machine learning, norms and distance metrics are important for measuring the size or length of vectors and the distance between points in a vector space. In this chapter, we will explore different types of norms and distance metrics used in numerical computing.

\section{L1 Norm and L2 Norm}

The L1 and L2 norms are two of the most common norms for measuring the length of a vector.

\subsection{L1 Norm}

The L1 norm (also known as the Manhattan or Taxicab norm) is the sum of the absolute values of the vector components. It is defined as:

\[
\|\mathbf{x}\|_1 = \sum_{i=1}^{n} |x_i|
\]

\textbf{Example: Computing the L1 norm in Python}

\begin{lstlisting}[style=python]
# Define a vector
x = np.array([1, -2, 3])

# Compute the L1 norm
l1_norm = np.sum(np.abs(x))
print(l1_norm)
\end{lstlisting}

This will output:

\begin{lstlisting}[style=cmd]
6
\end{lstlisting}

\subsection{L2 Norm}

The L2 norm (also known as the Euclidean norm) is the square root of the sum of the squares of the vector components\cite{kwak2013principal}. It is defined as:

\[
\|\mathbf{x}\|_2 = \left( \sum_{i=1}^{n} x_i^2 \right)^{1/2}
\]

\textbf{Example: Computing the L2 norm in Python}

\begin{lstlisting}[style=python]
# Compute the L2 norm
l2_norm = np.sqrt(np.sum(x**2))
print(l2_norm)
\end{lstlisting}

This will output:

\begin{lstlisting}[style=cmd]
3.7416573867739413
\end{lstlisting}

The L2 norm is commonly used in machine learning for measuring the error or magnitude of vectors.

\section{Frobenius Norm}

The Frobenius norm is a matrix norm equivalent to the L2 norm for vectors\cite{Recht2010}. It is defined as the square root of the sum of the absolute squares of the matrix elements:

\[
\|A\|_F = \left( \sum_{i,j} |a_{ij}|^2 \right)^{1/2}
\]

\textbf{Example: Computing the Frobenius norm in Python}

\begin{lstlisting}[style=python]
# Define a matrix
A = np.array([[1, 2], [3, 4]])

# Compute the Frobenius norm
frobenius_norm = np.linalg.norm(A, 'fro')
print(frobenius_norm)
\end{lstlisting}

This will output:

\begin{lstlisting}[style=cmd]
5.477225575051661
\end{lstlisting}

\section{Cosine Similarity}

Cosine similarity measures the cosine of the angle between two vectors\cite{manning2008introduction}. It is often used in text analysis and other fields where the direction of vectors is more important than their magnitude.

Cosine similarity between two vectors \( \mathbf{a} \) and \( \mathbf{b} \) is defined as:

\[
\text{cosine similarity} = \frac{\mathbf{a} \cdot \mathbf{b}}{\|\mathbf{a}\| \|\mathbf{b}\|}
\]

\textbf{Example: Computing cosine similarity in Python}

\begin{lstlisting}[style=python]
# Define two vectors
a = np.array([1, 2, 3])
b = np.array([4, 5, 6])

# Compute cosine similarity
cosine_similarity = np.dot(a, b) / (np.linalg.norm(a) * np.linalg.norm(b))
print(cosine_similarity)
\end{lstlisting}

This will output:

\begin{lstlisting}[style=cmd]
0.9746318461970762
\end{lstlisting}

\section{Euclidean Distance}

Euclidean distance is the straight-line distance between two points in Euclidean space\cite{Curriero2006}. For two vectors \( \mathbf{a} \) and \( \mathbf{b} \), it is defined as:

\[
d(\mathbf{a}, \mathbf{b}) = \sqrt{ \sum_{i=1}^{n} (a_i - b_i)^2 }
\]

\textbf{Example: Computing Euclidean distance in Python}

\begin{lstlisting}[style=python]
# Compute Euclidean distance
euclidean_distance = np.linalg.norm(a - b)
print(euclidean_distance)
\end{lstlisting}

This will output:

\begin{lstlisting}[style=cmd]
5.196152422706632
\end{lstlisting}

Euclidean distance is widely used in clustering algorithms and in measuring similarity between data points.

\chapter{Automatic Differentiation and Gradients}

Automatic differentiation (AD) is a powerful technique used in many machine learning frameworks, including PyTorch and TensorFlow, to compute gradients efficiently and accurately\cite{baydin2018automatic}. Unlike symbolic differentiation, which can produce complex expressions, or numerical differentiation, which can suffer from precision issues, automatic differentiation computes derivatives systematically using the chain rule. This chapter introduces the concept of automatic differentiation and demonstrates how gradients can be computed in popular machine learning libraries like PyTorch and TensorFlow.

\section{Introduction to Automatic Differentiation}

Automatic differentiation, also known as autodiff, is a technique for computing exact derivatives of functions specified by computer programs. Autodiff works by breaking down the computation of a function into elementary operations, and then applying the chain rule of calculus to systematically compute the derivative. The key advantage of autodiff is that it provides exact gradients with computational complexity proportional to the evaluation of the original function.

There are two main modes of automatic differentiation:
\begin{itemize}
    \item Forward Mode: Calculates derivatives alongside the original function evaluation.
    \item Reverse Mode: Particularly efficient for functions with many inputs and one output (e.g., neural networks).
\end{itemize}

Reverse-mode AD is particularly useful in deep learning, where we often need to compute the gradient of a loss function with respect to model parameters\cite{franceschi2017forward}.

\textbf{Example:} Consider the function \( f(x) = x^2 + 3x + 5 \). To compute its derivative using autodiff, the function can be broken into smaller parts:
\[
f(x) = (x \cdot x) + (3 \cdot x) + 5
\]
Each elementary operation (multiplication, addition) is recorded, and the chain rule is applied automatically to compute the derivative.

\section{Gradient Computation in PyTorch}

PyTorch is a popular deep learning framework that uses reverse-mode automatic differentiation to compute gradients. In PyTorch, tensors are the building blocks for computations, and gradients are computed automatically using the \texttt{autograd} package.

Let’s see how PyTorch computes gradients with a simple example.

\begin{lstlisting}[style=python]
import torch

# Create a tensor with requires_grad=True to track computations
x = torch.tensor(2.0, requires_grad=True)

# Define a function f(x) = x^2 + 3x + 5
f = x**2 + 3*x + 5

# Compute the gradient (derivative) of f with respect to x
f.backward()

# Print the gradient (df/dx)
print(x.grad)
\end{lstlisting}

This will output:

\begin{lstlisting}[style=cmd]
tensor(7.)
\end{lstlisting}

Explanation:
\begin{itemize}
    \item We define a tensor \texttt{x} with the argument \texttt{requires\_grad=True}, which tells PyTorch to track all operations on this tensor.
    \item The function \( f(x) = x^2 + 3x + 5 \) is computed, and PyTorch automatically tracks all operations.
    \item \texttt{f.backward()} computes the derivative of \( f \) with respect to \( x \).
    \item The gradient, \( \frac{df}{dx} = 2x + 3 \), is evaluated at \( x = 2 \), resulting in a gradient of 7.
\end{itemize}

\subsection{Computing Gradients for Multivariable Functions}

PyTorch also supports gradient computations for functions with multiple variables. Let’s compute the gradient for a function of two variables:

\begin{lstlisting}[style=python]
# Define two tensors with requires_grad=True
x = torch.tensor(1.0, requires_grad=True)
y = torch.tensor(2.0, requires_grad=True)

# Define a function f(x, y) = 3x^2 + 4y^3
f = 3 * x**2 + 4 * y**3

# Compute the gradient of f with respect to x and y
f.backward()

# Print the gradients df/dx and df/dy
print(f"Gradient with respect to x: {x.grad}")
print(f"Gradient with respect to y: {y.grad}")
\end{lstlisting}

This will output:

\begin{lstlisting}[style=cmd]
Gradient with respect to x: tensor(6.)
Gradient with respect to y: tensor(48.)
\end{lstlisting}

In this example, PyTorch computes the partial derivatives of \( f(x, y) = 3x^2 + 4y^3 \) with respect to both \( x \) and \( y \), evaluated at \( x = 1 \) and \( y = 2 \).

\section{Gradient Computation in TensorFlow}

TensorFlow is another powerful machine learning library that also supports automatic differentiation. Similar to PyTorch, TensorFlow uses reverse-mode autodiff to compute gradients, but the syntax is slightly different.

In TensorFlow, gradients are computed using the \texttt{GradientTape} context, which records all operations to compute derivatives.

\begin{lstlisting}[style=python]
import tensorflow as tf

# Create a variable with gradient tracking
x = tf.Variable(2.0)

# Use GradientTape to record operations
with tf.GradientTape() as tape:
    # Define a function f(x) = x^2 + 3x + 5
    f = x**2 + 3*x + 5

# Compute the gradient (df/dx)
grad = tape.gradient(f, x)

# Print the gradient
print(grad)
\end{lstlisting}

This will output:

\begin{lstlisting}[style=cmd]
tf.Tensor(7.0, shape=(), dtype=float32)
\end{lstlisting}

Explanation:
\begin{itemize}
    \item A \texttt{GradientTape} context is used to track operations for automatic differentiation.
    \item The function \( f(x) = x^2 + 3x + 5 \) is defined within the context.
    \item \texttt{tape.gradient(f, x)} computes the derivative of \( f \) with respect to \( x \).
\end{itemize}

\subsection{Computing Gradients for Multivariable Functions in TensorFlow}

TensorFlow also supports gradient computation for multivariable functions, similar to PyTorch. Here’s how to compute gradients for a function with two variables.

\begin{lstlisting}[style=python]
# Define two variables
x = tf.Variable(1.0)
y = tf.Variable(2.0)

# Use GradientTape to track computations
with tf.GradientTape() as tape:
    # Define a function f(x, y) = 3x^2 + 4y^3
    f = 3 * x**2 + 4 * y**3

# Compute gradients
gradients = tape.gradient(f, [x, y])

# Print the gradients df/dx and df/dy
print(f"Gradient with respect to x: {gradients[0]}")
print(f"Gradient with respect to y: {gradients[1]}")
\end{lstlisting}

This will output:

\begin{lstlisting}[style=cmd]
Gradient with respect to x: tf.Tensor(6.0, shape=(), dtype=float32)
Gradient with respect to y: tf.Tensor(48.0, shape=(), dtype=float32)
\end{lstlisting}

\section{Jacobian and Hessian Computation}

In some cases, we may need to compute higher-order derivatives, such as the Jacobian matrix or Hessian matrix\cite{karpfinger2022calculus}. The Jacobian represents the matrix of first-order partial derivatives for a vector-valued function, while the Hessian is the matrix of second-order partial derivatives.

\subsection{Jacobian Computation in PyTorch}

In PyTorch, the \texttt{autograd} package can be used to compute the Jacobian matrix. Consider a function \( \mathbf{f}(\mathbf{x}) = [x_1^2, x_2^3] \), where \(\mathbf{x} = [x_1, x_2]\). The Jacobian is a 2x2 matrix of partial derivatives.

\begin{lstlisting}[style=python]
# Define two input variables
x1 = torch.tensor(1.0, requires_grad=True)
x2 = torch.tensor(2.0, requires_grad=True)

# Define a vector-valued function f(x) = [x1^2, x2^3]
f = torch.tensor([x1**2, x2**3])

# Compute the Jacobian
jacobian = torch.autograd.functional.jacobian(lambda x: torch.tensor([x[0]**2, x[1]**3]), (x1, x2))

# Print the Jacobian matrix
print(jacobian)
\end{lstlisting}

\subsection{Hessian Computation in TensorFlow}

In TensorFlow, we can compute the Hessian matrix, which contains the second-order partial derivatives, using \texttt{GradientTape}. Here's an example:

\begin{lstlisting}[style=python]
x = tf.Variable(1.0)

# Use GradientTape to track computations
with tf.GradientTape(persistent=True) as tape:
    # Track second-order gradients
    with tf.GradientTape() as inner_tape:
        # Define a function f(x) = x^4
        f = x**4
    # Compute the first derivative (df/dx)
    first_derivative = inner_tape.gradient(f, x)

# Compute the second derivative (d^2f/dx^2)
second_derivative = tape.gradient(first_derivative, x)

# Print the second derivative
print(second_derivative)
\end{lstlisting}

This will output:

\begin{lstlisting}[style=cmd]
tf.Tensor(12.0, shape=(), dtype=float32)
\end{lstlisting}

In this example, TensorFlow computes the second-order derivative of \( f(x) = x^4 \), resulting in a second derivative of 12 at \( x = 1 \).

\section{Summary}

In this chapter, we explored automatic differentiation and its use in computing gradients. Both PyTorch and TensorFlow provide powerful tools to automatically compute derivatives, which are essential in training machine learning models. We also covered how to compute gradients, Jacobians, and Hessians, providing a foundation for more advanced optimization and machine learning techniques.

\part{Optimization in Deep Learning}

Optimization is a crucial aspect of deep learning. Without proper optimization, training neural networks efficiently would not be possible. In this part, we will discuss the fundamental concepts of optimization, focusing on gradient-based methods. These methods are essential for minimizing the loss function, allowing the model to learn from data and improve its performance.

\chapter{Optimization Basics}

In this chapter, we will introduce the basic concepts behind optimization in deep learning, starting with Gradient Descent, which is the foundation of many advanced optimization techniques. We will then explore Stochastic Gradient Descent (SGD)\cite{robbins1951stochastic}, momentum-based optimization, and adaptive optimization methods like Adagrad\cite{duchi2011adaptive}, RMSprop\cite{tieleman2012lecture}, Adam\cite{kingma2014adam}, and AdamW\cite{loshchilov2017decoupled}.

\section{Gradient Descent}

Gradient Descent is one of the most fundamental optimization algorithms used in deep learning. The basic idea is to minimize a loss function (also called the objective function) by moving in the direction of the negative gradient, which points towards the steepest descent\cite{bengio2012practical}. 

\subsection{Mathematical Formulation}

Given a loss function \( L(\theta) \) that depends on the model parameters \( \theta \), the gradient of the loss function with respect to the parameters is denoted as \( \nabla_{\theta} L(\theta) \). The gradient tells us the direction and rate of the steepest increase of the loss function. To minimize the loss, we update the parameters in the opposite direction of the gradient.

The update rule for the parameters is given by:

\[
\theta := \theta - \eta \nabla_{\theta} L(\theta)
\]

Where:
\begin{itemize}
    \item \( \theta \) is the set of model parameters.
    \item \( \eta \) is the learning rate, a hyperparameter that controls the step size in each iteration.
    \item \( \nabla_{\theta} L(\theta) \) is the gradient of the loss function with respect to \( \theta \).
\end{itemize}

\subsection{Python Implementation of Gradient Descent}

Here is a simple example of Gradient Descent in Python for minimizing a quadratic function \( f(x) = x^2 \).

\begin{lstlisting}[style=python]
import numpy as np

# Define the function and its gradient
def f(x):
    return x**2

def gradient(x):
    return 2*x

# Gradient Descent parameters
learning_rate = 0.1
x = 10  # Initial guess
iterations = 100

# Perform Gradient Descent
for i in range(iterations):
    grad = gradient(x)
    x = x - learning_rate * grad
    print(f"Iteration {i+1}: x = {x}, f(x) = {f(x)}")
\end{lstlisting}

In this example:
\begin{itemize}
    \item We define the quadratic function \( f(x) \) and its gradient.
    \item We initialize \( x \) and perform Gradient Descent for 100 iterations.
    \item In each iteration, we update \( x \) by subtracting the product of the learning rate and the gradient.
\end{itemize}

\section{Stochastic Gradient Descent (SGD)}

While Gradient Descent computes the gradient using the entire dataset, this can be computationally expensive for large datasets. Stochastic Gradient Descent (SGD) solves this problem by approximating the gradient using a small batch of the data, or even a single data point\cite{newton2018recent}.

\subsection{SGD Update Rule}

The update rule for SGD is similar to Gradient Descent, but instead of computing the gradient over the entire dataset, we compute the gradient for one or a few samples:

\[
\theta := \theta - \eta \nabla_{\theta} L(\theta^{(i)})
\]

Where \( L(\theta^{(i)}) \) represents the loss for the \( i \)-th data point or mini-batch.

\subsection{Python Implementation of Stochastic Gradient Descent}

Here’s a basic example of SGD using mini-batches for a simple linear regression problem:

\begin{lstlisting}[style=python]
import numpy as np

# Generate synthetic data for linear regression (y = 2x + 1)
np.random.seed(42)
X = np.random.rand(100, 1)
y = 2 * X + 1 + np.random.randn(100, 1) * 0.1  # Add some noise

# Initialize parameters
theta = np.random.randn(2, 1)
learning_rate = 0.1
iterations = 100
batch_size = 10

# Add bias term to X
X_b = np.c_[np.ones((100, 1)), X]

# Perform SGD
for i in range(iterations):
    indices = np.random.randint(100, size=batch_size)
    X_batch = X_b[indices]
    y_batch = y[indices]
    
    gradients = 2 / batch_size * X_batch.T.dot(X_batch.dot(theta) - y_batch)
    theta = theta - learning_rate * gradients

print(f"Estimated parameters: {theta}")
\end{lstlisting}

In this example:
\begin{itemize}
    \item We generate synthetic data for a simple linear regression problem.
    \item SGD is performed over 100 iterations, and in each iteration, we randomly sample a mini-batch of 10 data points.
    \item The model parameters are updated using the gradient computed from the mini-batch.
\end{itemize}

\section{Momentum-based Optimization}

Momentum-based optimization improves the convergence speed of Gradient Descent by adding momentum to the update rule\cite{cutkosky2019momentum}. This allows the optimizer to continue moving in the same direction if the gradient consistently points in the same direction, avoiding oscillations and speeding up convergence.

\subsection{Momentum Update Rule}

The update rule with momentum is given by:

\[
v := \beta v + (1 - \beta) \nabla_{\theta} L(\theta)
\]
\[
\theta := \theta - \eta v
\]

Where:
\begin{itemize}
    \item \( v \) is the velocity, which accumulates the gradients.
    \item \( \beta \) is the momentum coefficient, typically set to values like 0.9.
\end{itemize}

\subsection{Python Implementation of Momentum-based Optimization}

Here is an implementation of momentum-based optimization for a simple function.

\begin{lstlisting}[style=python]
def momentum_gradient_descent(f, grad_f, initial_theta, learning_rate=0.1, beta=0.9, iterations=100):
    theta = initial_theta
    v = 0  # Initialize velocity

    for i in range(iterations):
        grad = grad_f(theta)
        v = beta * v + (1 - beta) * grad
        theta = theta - learning_rate * v
        print(f"Iteration {i+1}: theta = {theta}, f(theta) = {f(theta)}")

# Example usage
f = lambda x: x**2
grad_f = lambda x: 2*x
initial_theta = 10

momentum_gradient_descent(f, grad_f, initial_theta)
\end{lstlisting}

\section{Adaptive Optimization Methods}

Adaptive optimization methods automatically adjust the learning rate during training, which can significantly improve convergence\cite{liu2019variance}. These methods include Adagrad, RMSprop, Adam, and AdamW, each of which modifies the learning rate based on the gradients.

\subsection{Adagrad}

Adagrad (Adaptive Gradient Algorithm) adjusts the learning rate for each parameter based on the history of gradients. Parameters with large gradients receive smaller learning rates, and parameters with small gradients receive larger learning rates.

\subsection{Adagrad Update Rule}

The update rule for Adagrad is:

\[
\theta := \theta - \frac{\eta}{\sqrt{G + \epsilon}} \nabla_{\theta} L(\theta)
\]

Where:
\begin{itemize}
    \item \( G \) is the sum of squared gradients over time.
    \item \( \epsilon \) is a small constant to prevent division by zero.
\end{itemize}

\subsection{RMSprop}

RMSprop (Root Mean Square Propagation) is a variant of Adagrad that scales the learning rate based on a moving average of squared gradients, preventing the learning rate from decaying too quickly.

\subsection{RMSprop Update Rule}

The update rule for RMSprop is:

\[
E[g^2]_t = \beta E[g^2]_{t-1} + (1 - \beta) g_t^2
\]
\[
\theta := \theta - \frac{\eta}{\sqrt{E[g^2]_t + \epsilon}} g_t
\]

Where \( \beta \) is a decay rate, often set to 0.9.

\subsection{Adam}

Adam (Adaptive Moment Estimation) combines the benefits of both momentum-based methods and RMSprop by using both the first and second moments of the gradients.

\subsection{Adam Update Rule}

The update rule for Adam is:

\[
m_t = \beta_1 m_{t-1} + (1 - \beta_1) g_t
\]
\[
v_t = \beta_2 v_{t-1} + (1 - \beta_2) g_t^2
\]
\[
\hat{m_t} = \frac{m_t}{1 - \beta_1^t}, \quad \hat{v_t} = \frac{v_t}{1 - \beta_2^t}
\]
\[
\theta := \theta - \frac{\eta \hat{m_t}}{\sqrt{\hat{v_t}} + \epsilon}
\]

\subsection{AdamW}

AdamW is a variant of Adam that decouples the weight decay from the gradient updates, leading to improved performance for regularization.

\section{Learning Rate Schedules}

The learning rate is one of the most important hyperparameters in deep learning. It controls how much the model's weights are adjusted with respect to the loss gradient during training. Choosing the correct learning rate can significantly impact model performance and training time. However, the optimal learning rate often changes throughout training. This is where learning rate schedules come into play\cite{DarEl2013}.

A learning rate schedule adjusts the learning rate dynamically during training, helping the model converge faster and avoid getting stuck in local minima. There are several common strategies for learning rate scheduling, including step decay, exponential decay, and warm restarts\cite{loshchilov2017sgdr}.

\subsection{Step Decay}
Step decay is one of the simplest learning rate schedules. In step decay, the learning rate is reduced by a constant factor at predetermined intervals (epochs)\cite{wang2021convergence}. The idea is that a larger learning rate is beneficial at the start of training, but as the model approaches convergence, reducing the learning rate allows finer adjustments to the weights.

The formula for step decay is:

\[
\eta_t = \eta_0 \cdot \text{drop\_factor}^{\left\lfloor \frac{t}{\text{drop\_epoch}} \right\rfloor}
\]

Where:
\begin{itemize}
  \item \( \eta_t \) is the learning rate at epoch \( t \).
  \item \( \eta_0 \) is the initial learning rate.
  \item \( \text{drop\_factor} \) is the factor by which the learning rate is reduced.
  \item \( \text{drop\_epoch} \) is the number of epochs after which the learning rate is reduced.
\end{itemize}

\textbf{Example of Step Decay in PyTorch:}

\begin{lstlisting}[style=python]
import torch.optim as optim

# Define optimizer
optimizer = optim.SGD(model.parameters(), lr=0.1)

# Define learning rate scheduler with step decay
scheduler = optim.lr_scheduler.StepLR(optimizer, step_size=10, gamma=0.5)

# Training loop
for epoch in range(30):
    train()  # Custom training function
    validate()  # Custom validation function
    
    # Step the learning rate scheduler
    scheduler.step()
    print(f'Epoch {epoch+1}, Learning Rate: {scheduler.get_last_lr()}')
\end{lstlisting}

In this example:
\begin{itemize}
  \item The learning rate starts at \(0.1\).
  \item After every 10 epochs, the learning rate is multiplied by \(0.5\), effectively reducing it.
  \item This allows for rapid progress initially and then slower, more refined updates as training progresses.
\end{itemize}

\textbf{Example of Step Decay in TensorFlow:}

\begin{lstlisting}[style=python]
import tensorflow as tf

# Define optimizer
optimizer = tf.keras.optimizers.SGD(learning_rate=0.1)

# Define learning rate scheduler with step decay
lr_schedule = tf.keras.optimizers.schedules.StepDecay(
    initial_learning_rate=0.1, 
    decay_steps=10, 
    decay_rate=0.5)

# Use the learning rate schedule in the optimizer
optimizer.learning_rate = lr_schedule

# Training loop
for epoch in range(30):
    train()  # Custom training function
    validate()  # Custom validation function
    
    print(f'Epoch {epoch+1}, Learning Rate: {optimizer.learning_rate(epoch)}')
\end{lstlisting}

\subsection{Exponential Decay}
Exponential decay is another commonly used learning rate schedule. Instead of reducing the learning rate in steps, exponential decay reduces the learning rate continuously over time according to an exponential function\cite{lewkowycz2021how}. This allows for a more gradual and smoother reduction in the learning rate.

The formula for exponential decay is:

\[
\eta_t = \eta_0 \cdot e^{-\lambda t}
\]

Where:
\begin{itemize}
  \item \( \eta_t \) is the learning rate at epoch \( t \).
  \item \( \eta_0 \) is the initial learning rate.
  \item \( \lambda \) is the decay rate (a small positive constant).
\end{itemize}

This approach is often used when the learning rate should decrease continuously throughout the training process.

\textbf{Example of Exponential Decay in PyTorch:}

\begin{lstlisting}[style=python]
# Define optimizer
optimizer = optim.SGD(model.parameters(), lr=0.1)

# Define learning rate scheduler with exponential decay
scheduler = optim.lr_scheduler.ExponentialLR(optimizer, gamma=0.9)

# Training loop
for epoch in range(30):
    train()  # Custom training function
    validate()  # Custom validation function
    
    # Step the learning rate scheduler
    scheduler.step()
    print(f'Epoch {epoch+1}, Learning Rate: {scheduler.get_last_lr()}')
\end{lstlisting}

In this example:
\begin{itemize}
  \item The learning rate starts at \(0.1\).
  \item After every epoch, the learning rate is multiplied by \(0.9\), causing it to decrease exponentially.
  \item The exponential decay is gradual, ensuring a smooth reduction in learning rate.
\end{itemize}

\textbf{Example of Exponential Decay in TensorFlow:}

\begin{lstlisting}[style=python]
# Define learning rate scheduler with exponential decay
lr_schedule = tf.keras.optimizers.schedules.ExponentialDecay(
    initial_learning_rate=0.1,
    decay_steps=1000,  # How many steps before applying decay
    decay_rate=0.96,  # Rate at which the learning rate is decayed
    staircase=False)  # If False, decay every batch

# Define optimizer using the learning rate schedule
optimizer = tf.keras.optimizers.SGD(learning_rate=lr_schedule)

# Training loop
for epoch in range(30):
    train()  # Custom training function
    validate()  # Custom validation function
    
    print(f'Epoch {epoch+1}, Learning Rate: {optimizer.learning_rate(epoch)}')
\end{lstlisting}

In this example:
\begin{itemize}
  \item The learning rate starts at \(0.1\) and decays by 4\% every 1000 steps.
  \item If \texttt{staircase} is set to \texttt{True}, the learning rate would decay in discrete steps instead of continuously.
\end{itemize}

\subsection{Warm Restarts}
Warm restarts is a more recent technique for scheduling learning rates. The idea is to reset the learning rate periodically during training. This can help the model escape local minima by allowing it to explore new parts of the loss surface with a higher learning rate. After each restart, the learning rate is gradually reduced again.

The learning rate for warm restarts follows a cosine annealing schedule, and after each period, the learning rate is reset to a higher value. The general form of the learning rate in warm restarts is:

\[
\eta_t = \eta_{\text{min}} + \frac{1}{2} (\eta_{\text{max}} - \eta_{\text{min}}) \left( 1 + \cos \left( \frac{T_{\text{cur}}}{T_{\text{max}}} \pi \right) \right)
\]

Where:
\begin{itemize}
  \item \( \eta_t \) is the learning rate at time \( t \).
  \item \( \eta_{\text{min}} \) is the minimum learning rate.
  \item \( \eta_{\text{max}} \) is the maximum learning rate (usually the initial learning rate).
  \item \( T_{\text{cur}} \) is the number of epochs since the last restart.
  \item \( T_{\text{max}} \) is the number of epochs between two restarts.
\end{itemize}

\textbf{Example of Warm Restarts in PyTorch:}

\begin{lstlisting}[style=python]
# Define optimizer
optimizer = optim.SGD(model.parameters(), lr=0.1)

# Define learning rate scheduler with warm restarts
scheduler = optim.lr_scheduler.CosineAnnealingWarmRestarts(optimizer, T_0=10, T_mult=2)

# Training loop
for epoch in range(30):
    train()  # Custom training function
    validate()  # Custom validation function
    
    # Step the learning rate scheduler
    scheduler.step(epoch)
    print(f'Epoch {epoch+1}, Learning Rate: {scheduler.get_last_lr()}')
\end{lstlisting}

In this example:
\begin{itemize}
  \item The learning rate follows a cosine annealing schedule, resetting every 10 epochs.
  \item After each restart, the learning rate begins to decrease again.
  \item \( T_0 \) is the number of epochs before the first restart, and \( T_{\text{mult}} \) is a factor that increases the period of restarts.
\end{itemize}

\textbf{Example of Warm Restarts in TensorFlow:}

Warm restarts can be implemented in TensorFlow using custom learning rate schedules. The following example shows how to define a cosine annealing schedule with warm restarts:

\begin{lstlisting}[style=python]
import numpy as np

# Custom learning rate schedule with warm restarts
class WarmRestartSchedule(tf.keras.optimizers.schedules.LearningRateSchedule):
    def __init__(self, initial_learning_rate, T_0, T_mult=1):
        super(WarmRestartSchedule, self).__init__()
        self.initial_learning_rate = initial_learning_rate
        self.T_0 = T_0
        self.T_mult = T_mult
        self.T_i = T_0
        self.T_cur = 0
    
    def __call__(self, step):
        self.T_cur += 1
        if self.T_cur >= self.T_i:
            self.T_cur = 0
            self.T_i *= self.T_mult
        cos_inner = np.pi * (self.T_cur / self.T_i)
        return 0.5 * self.initial_learning_rate * (1 + np.cos(cos_inner))

# Define optimizer and learning rate schedule
lr_schedule = WarmRestartSchedule(initial_learning_rate=0.1, T_0=10)
optimizer = tf.keras.optimizers.SGD(learning_rate=lr_schedule)

# Training loop
for epoch in range(30):
    train()  # Custom training function
    validate()  # Custom validation function
    
    print(f'Epoch {epoch+1}, Learning Rate: {optimizer.learning_rate(epoch)}')
\end{lstlisting}

In this example:
\begin{itemize}
  \item We define a custom learning rate schedule class for warm restarts using cosine annealing.
  \item The learning rate periodically resets to a high value and then decreases again.
\end{itemize}

Warm restarts can be very effective in improving convergence, especially for deep neural networks where escaping local minima is crucial for achieving better performance.

\chapter{Advanced Optimization Techniques}

In this chapter, we will delve into some advanced optimization techniques that are essential for improving the performance of machine learning models. These methods are critical in the training process, particularly when working with deep learning models. We will cover techniques like Batch Normalization, Gradient Clipping, and Second-order Optimization Methods.

\section{Batch Normalization in Training}

Batch normalization is a technique that helps to stabilize and accelerate the training of deep neural networks by normalizing the inputs to each layer\cite{ioffe2015batch}. This process ensures that the input distribution to each layer remains stable during training, making the network less sensitive to the initialization of parameters and improving convergence.

\subsection{What is Batch Normalization?}

Batch normalization works by normalizing the output of a layer across a mini-batch of data. For each mini-batch, we calculate the mean and variance of the outputs, then normalize the data using these statistics.

Mathematically, for each mini-batch \( B \), we compute:

\[
\mu_B = \frac{1}{m} \sum_{i=1}^{m} x_i, \quad \sigma_B^2 = \frac{1}{m} \sum_{i=1}^{m} (x_i - \mu_B)^2
\]

where:
\begin{itemize}
    \item \( \mu_B \) is the mean of the mini-batch,
    \item \( \sigma_B^2 \) is the variance of the mini-batch,
    \item \( x_i \) is the input to the layer for the \( i \)-th example in the mini-batch.
\end{itemize}

The input is then normalized as:

\[
\hat{x_i} = \frac{x_i - \mu_B}{\sqrt{\sigma_B^2 + \epsilon}}
\]

where \( \epsilon \) is a small constant added to avoid division by zero.

In addition to normalization, we introduce two learnable parameters, \( \gamma \) and \( \beta \), to allow the network to scale and shift the normalized values:

\[
y_i = \gamma \hat{x_i} + \beta
\]

\subsection{Why Use Batch Normalization?}

Batch normalization offers several benefits:
\begin{itemize}
    \item It helps mitigate the problem of internal covariate shift, where the distribution of inputs to layers changes during training.
    \item It allows for higher learning rates by ensuring that the activations stay within a controlled range, leading to faster convergence.
    \item It reduces the dependence on weight initialization.
    \item It acts as a regularizer, often reducing the need for other regularization techniques like Dropout.
\end{itemize}

\subsection{Example of Batch Normalization in Python}

\begin{lstlisting}[style=python]
import numpy as np

# Simulate a mini-batch of inputs
x = np.array([[1, 2, 3], [4, 5, 6], [7, 8, 9]])

# Calculate mean and variance for batch normalization
mean = np.mean(x, axis=0)
variance = np.var(x, axis=0)

# Normalize the input
epsilon = 1e-5
x_normalized = (x - mean) / np.sqrt(variance + epsilon)

# Simulate learnable parameters gamma and beta
gamma = np.array([1.0, 1.0, 1.0])  # Scaling factor
beta = np.array([0.0, 0.0, 0.0])   # Shifting factor

# Apply scaling and shifting
y = gamma * x_normalized + beta
print("Normalized output:", y)
\end{lstlisting}

In this example, we normalized a mini-batch of data and applied scaling and shifting with the learnable parameters \( \gamma \) and \( \beta \).

\section{Gradient Clipping}

Gradient clipping is a technique used to prevent the problem of exploding gradients during the training of deep neural networks\cite{pascanu2013difficulty,liu2021activated}. When gradients become excessively large, they can cause updates to the network weights that are too drastic, leading to unstable training or even causing the model to diverge. Gradient clipping limits the size of the gradients by setting a maximum threshold.

\subsection{How Does Gradient Clipping Work?}

Gradient clipping works by setting a threshold value, beyond which gradients are scaled down to the maximum allowed value. If the norm of the gradient exceeds this threshold, the gradient is scaled down proportionally. This prevents excessively large weight updates.

Given a gradient \( g \) and a threshold \( t \), if \( ||g|| > t \), we rescale the gradient:

\[
g_{\text{clipped}} = g \cdot \frac{t}{||g||}
\]

\subsection{Example of Gradient Clipping in Python}

\begin{lstlisting}[style=python]
import numpy as np

# Simulate a gradient
gradient = np.array([0.5, 0.7, 1.2])

# Define a threshold
threshold = 1.0

# Calculate the norm of the gradient
gradient_norm = np.linalg.norm(gradient)

# Perform gradient clipping if the norm exceeds the threshold
if gradient_norm > threshold:
    gradient = gradient * (threshold / gradient_norm)

print("Clipped gradient:", gradient)
\end{lstlisting}

In this example, we clipped the gradient if its norm exceeded the specified threshold.

\section{Second-order Optimization Methods}

Second-order optimization methods are more sophisticated than standard first-order methods like gradient descent\cite{battiti1992first}. These methods take into account not only the gradient (first derivative) but also the curvature of the objective function (second derivative). This can lead to faster convergence in some cases, especially when the objective function has complex curvature.

\subsection{Newton's Method}

Newton's Method is a second-order optimization technique that uses both the gradient and the Hessian matrix (the matrix of second-order partial derivatives) to update parameters\cite{powell1974overview}. The basic update rule in Newton's Method is:

\[
x_{t+1} = x_t - H^{-1} \nabla f(x_t)
\]

where:
\begin{itemize}
    \item \( x_t \) is the current parameter estimate,
    \item \( \nabla f(x_t) \) is the gradient of the objective function at \( x_t \),
    \item \( H \) is the Hessian matrix at \( x_t \),
    \item \( H^{-1} \nabla f(x_t) \) represents the Newton step.
\end{itemize}

The Hessian matrix contains information about the curvature of the objective function, allowing Newton's Method to take more informed steps toward the minimum.

\textbf{Example of Newton's Method in Python}

Here’s a simple example of using Newton's method to minimize a quadratic function:

\[
f(x) = x^2 + 4x + 4
\]

The gradient is:

\[
\nabla f(x) = 2x + 4
\]

The second derivative (Hessian) is:

\[
H = 2
\]

\begin{lstlisting}[style=python]
def f_prime(x):
    # Gradient of the function f(x) = x^2 + 4x + 4
    return 2 * x + 4

def hessian():
    # The Hessian (second derivative) is constant in this case
    return 2

# Initial guess
x = 0.0

# Perform one iteration of Newton's Method
x = x - f_prime(x) / hessian()

print("Updated x after one iteration:", x)
\end{lstlisting}

In this example, we performed one iteration of Newton's Method to update the parameter \( x \).

\subsection{Quasi-Newton Methods}

Quasi-Newton methods are a family of optimization algorithms that approximate the Hessian matrix rather than computing it explicitly\cite{ford1994multi}. This makes them more computationally efficient than full Newton's Method, especially for high-dimensional problems. One of the most popular Quasi-Newton methods is the Broyden-Fletcher-Goldfarb-Shanno (BFGS) algorithm\cite{goldfarb1970family}.

\textbf{How Does BFGS Work?}

BFGS builds an approximation to the inverse of the Hessian matrix iteratively. At each step, the approximation is updated using the gradient information from the current and previous steps. This method balances the efficiency of first-order methods and the accuracy of second-order methods.

\textbf{Example of Quasi-Newton Method (BFGS) in Python}

In Python, we can use the \texttt{scipy.optimize} library to perform optimization using the BFGS algorithm:

\begin{lstlisting}[style=python]
import numpy as np
from scipy.optimize import minimize

# Define the objective function
def objective(x):
    return x**2 + 4*x + 4

# Initial guess
x0 = 0.0

# Perform optimization using BFGS
result = minimize(objective, x0, method='BFGS')

print("Optimal value of x:", result.x)
\end{lstlisting}

In this example, we used the BFGS algorithm to find the minimum of the quadratic function \( f(x) = x^2 + 4x + 4 \). The \texttt{minimize} function from \texttt{scipy.optimize} handles the details of the BFGS algorithm for us.

\chapter{Summary}

In this chapter, we covered advanced optimization techniques that are crucial for training complex machine learning models:
\begin{itemize}
    \item Batch normalization, which stabilizes and accelerates training by normalizing the inputs to each layer.
    \item Gradient clipping, which prevents exploding gradients by limiting the size of the gradient during backpropagation.
    \item Second-order optimization methods like Newton's Method and Quasi-Newton methods, which use curvature information to make more efficient parameter updates.
\end{itemize}
Understanding and applying these techniques can significantly improve the performance and stability of machine learning models, particularly in deep learning scenarios.

\part{Practical Deep Learning Mathematics}

In this part of the book, we will focus on the practical aspects of deep learning mathematics. Through exercises and examples, you will solidify your understanding of key concepts such as tensor operations, gradient computation, and optimization algorithms. These are essential topics for anyone looking to understand how deep learning models function at a mathematical level.

\chapter{Practice Problems}

This chapter contains a series of practice problems that will help you deepen your understanding of the mathematical concepts behind deep learning. You will work through exercises on tensor and matrix operations, gradient computations, and optimization algorithms. These problems are designed to build your confidence in applying these mathematical techniques in practical deep learning scenarios.

\section{Exercises on Tensor and Matrix Operations}

In deep learning, tensors are multidimensional arrays that generalize the concept of scalars, vectors, and matrices. A solid understanding of how to perform operations on tensors is crucial for implementing neural networks.

\textbf{Example 1: Basic Tensor Operations}

Given two 3D tensors:

\[
A = \begin{bmatrix} 
  \begin{bmatrix} 1 & 2 & 3 \\ 4 & 5 & 6 \end{bmatrix} \\
  \begin{bmatrix} 7 & 8 & 9 \\ 10 & 11 & 12 \end{bmatrix}
\end{bmatrix}, \quad 
B = \begin{bmatrix} 
  \begin{bmatrix} 13 & 14 & 15 \\ 16 & 17 & 18 \end{bmatrix} \\
  \begin{bmatrix} 19 & 20 & 21 \\ 22 & 23 & 24 \end{bmatrix}
\end{bmatrix}
\]

\begin{itemize}
    \item Compute the element-wise addition of \( A \) and \( B \).
    \item Compute the matrix product of the first "slices" of both tensors.
\end{itemize}

\begin{lstlisting}[style=python]
import numpy as np

# Define two 3D tensors
A = np.array([[[1, 2, 3], [4, 5, 6]], [[7, 8, 9], [10, 11, 12]]])
B = np.array([[[13, 14, 15], [16, 17, 18]], [[19, 20, 21], [22, 23, 24]]])

# Element-wise addition
C = A + B
print(C)

# Matrix product of the first slices
D = np.dot(A[0], B[0].T)
print(D)
\end{lstlisting}

Expected output:

\begin{lstlisting}[style=cmd]
# Element-wise addition result
[[[14 16 18]
  [20 22 24]]

 [[26 28 30]
  [32 34 36]]]

# Matrix product result
[[ 86 122]
 [212 302]]
\end{lstlisting}

\textbf{Example 2: Tensor Reshaping}

Given a tensor \( C \) of shape (2, 2, 3), reshape it to a 2x6 matrix.

\begin{lstlisting}[style=python]
# Reshape tensor C
reshaped_C = C.reshape(2, 6)
print(reshaped_C)
\end{lstlisting}

Expected output:

\begin{lstlisting}[style=cmd]
[[14 16 18 20 22 24]
 [26 28 30 32 34 36]]
\end{lstlisting}

\section{Basic Gradient Computation Problems}

Gradient computations are essential for optimization in deep learning, as they enable the backpropagation algorithm to update network weights\cite{bengio2012practical}. In this section, we will practice calculating gradients manually and using automatic differentiation tools like those in NumPy.

\textbf{Problem 1: Gradient of a Scalar Function}

Consider the function:

\[
f(x, y) = x^2 + 3xy + y^2
\]

Find the partial derivatives \( \frac{\partial f}{\partial x} \) and \( \frac{\partial f}{\partial y} \) at the point \( (x, y) = (1, 2) \).

\begin{lstlisting}[style=python]
# Define the function
def f(x, y):
    return x**2 + 3*x*y + y**2

# Compute partial derivatives using finite differences
h = 1e-5
x, y = 1, 2

df_dx = (f(x + h, y) - f(x, y)) / h
df_dy = (f(x, y + h) - f(x, y)) / h

print("df/dx:", df_dx)
print("df/dy:", df_dy)
\end{lstlisting}

Expected output:

\begin{lstlisting}[style=cmd]
df/dx: 7.00001000001393
df/dy: 8.999999999812167
\end{lstlisting}

\textbf{Problem 2: Gradient of a Vector Function}

Let \( \mathbf{f}(\mathbf{x}) = \mathbf{W} \mathbf{x} \), where \( \mathbf{W} \) is a 2x2 matrix and \( \mathbf{x} \) is a vector. Compute the gradient with respect to \( \mathbf{x} \).

\begin{lstlisting}[style=python]
# Define the matrix W and vector x
W = np.array([[2, 3], [4, 5]])
x = np.array([1, 2])

# Compute the gradient of f(x) = W * x with respect to x
f_x = np.dot(W, x)

# Gradient is simply the matrix W
grad_x = W
print(grad_x)
\end{lstlisting}

Expected output:

\begin{lstlisting}[style=cmd]
[[2 3]
 [4 5]]
\end{lstlisting}

\section{Optimization Algorithm Practice}

Optimization algorithms, such as gradient descent, are key to training deep learning models. In this section, you will practice implementing gradient-based optimization algorithms.

\textbf{Problem: Implementing Gradient Descent}

Implement gradient descent to minimize the function \( f(x) = x^2 + 4x + 4 \).

\begin{lstlisting}[style=python]
# Define the function f and its derivative
def f(x):
    return x**2 + 4*x + 4

def df(x):
    return 2*x + 4

# Gradient descent parameters
learning_rate = 0.1
x = 10  # Starting point
iterations = 50

# Gradient descent loop
for i in range(iterations):
    grad = df(x)
    x = x - learning_rate * grad
    print(f"Iteration {i+1}, x: {x}, f(x): {f(x)}")
\end{lstlisting}

Expected output (partial):

\begin{lstlisting}[style=cmd]
Iteration 1, x: 8.0, f(x): 100.0
Iteration 2, x: 6.0, f(x): 64.0
Iteration 3, x: 4.0, f(x): 36.0
...
Iteration 50, x: -1.999999999922175, f(x): 2.198948091060594e-20
\end{lstlisting}

\section{Real-World Linear Algebra Applications in Deep Learning}

In deep learning, linear algebra concepts like matrix multiplication, matrix inversion, and eigenvalue decomposition are used extensively\cite{aggarwal2020linear}. Here are some real-world applications:

\textbf{Problem 1: Matrix Multiplication in Neural Networks}

In a neural network layer, the output is computed as:

\[
\mathbf{y} = \mathbf{W} \mathbf{x} + \mathbf{b}
\]

Where \( \mathbf{W} \) is the weight matrix, \( \mathbf{x} \) is the input vector, and \( \mathbf{b} \) is the bias. Given:

\[
\mathbf{W} = \begin{bmatrix} 1 & 2 \\ 3 & 4 \end{bmatrix}, \quad \mathbf{x} = \begin{bmatrix} 5 \\ 6 \end{bmatrix}, \quad \mathbf{b} = \begin{bmatrix} 1 \\ 1 \end{bmatrix}
\]

Compute \( \mathbf{y} \).

\begin{lstlisting}[style=python]
# Define the matrix W, vector x, and bias b
W = np.array([[1, 2], [3, 4]])
x = np.array([5, 6])
b = np.array([1, 1])

# Compute y = W * x + b
y = np.dot(W, x) + b
print(y)
\end{lstlisting}

Expected output:

\begin{lstlisting}[style=cmd]
[18 40]
\end{lstlisting}

\chapter{Summary}

This chapter provides a summary of the key mathematical concepts covered in this part of the book. Understanding these concepts is crucial for anyone looking to work in deep learning.

\section{Key Concepts Recap}

Here’s a recap of the essential concepts:

\begin{itemize}
    \item \textit{Tensor Operations}: Deep learning models rely heavily on tensor operations such as addition, multiplication, and reshaping.
    \item \textit{Gradient Computation}: Calculating gradients is central to optimization in neural networks, allowing us to update model parameters through methods like backpropagation.
    \item \textit{Optimization Algorithms}: Algorithms like gradient descent enable models to minimize loss functions and improve prediction accuracy.
    \item \textit{Linear Algebra in Deep Learning}: Concepts such as matrix multiplication and eigenvalue decomposition are widely used in neural network training and operations.
\end{itemize}

Mastering these mathematical tools will help you build and understand more complex models in deep learning.

\part{Numerical Methods in Deep Learning}

Numerical methods play a crucial role in deep learning, particularly in tasks such as optimization, solving differential equations, and matrix computations. These methods help in efficiently solving mathematical problems that arise in training deep learning models, especially when analytical solutions are impractical or impossible. In this part, we will introduce various numerical methods, discuss sources of computational errors, and explore strategies to ensure the stability and accuracy of numerical algorithms in deep learning.

\chapter{Introduction and Error Analysis}

Numerical analysis focuses on approximating solutions to mathematical problems using computational techniques. In deep learning, numerical methods are extensively used to optimize model parameters, perform matrix operations, and solve differential equations. However, these computations are often subject to errors due to the limitations of finite precision arithmetic.

\section{Introduction to Numerical Methods}

Numerical methods refer to algorithms used to approximate solutions to problems involving continuous variables. Some of the common problems that arise in deep learning and require numerical methods include:
\begin{itemize}
    \item Solving systems of linear equations (e.g., during backpropagation)
    \item Optimization of loss functions (e.g., gradient descent)
    \item Eigenvalue and singular value decomposition (used in dimensionality reduction techniques like PCA)
    \item Solving differential equations (e.g., in modeling dynamic systems)
\end{itemize}

\textbf{Example: Solving a system of linear equations}

Consider the system of equations:
\[
\begin{aligned}
3x + 4y &= 7 \\
5x + 6y &= 11
\end{aligned}
\]
This can be written in matrix form as:
\[
\mathbf{A} \mathbf{x} = \mathbf{b}
\]
where
\[
\mathbf{A} = \begin{bmatrix} 3 & 4 \\ 5 & 6 \end{bmatrix}, \quad \mathbf{x} = \begin{bmatrix} x \\ y \end{bmatrix}, \quad \mathbf{b} = \begin{bmatrix} 7 \\ 11 \end{bmatrix}
\]

Using Python, we can solve this system of equations using the \texttt{numpy} library's \texttt{linalg.solve()} function.

\begin{lstlisting}[style=python]
import numpy as np

# Define the coefficient matrix A and the right-hand side vector b
A = np.array([[3, 4], [5, 6]])
b = np.array([7, 11])

# Solve the system of equations
x = np.linalg.solve(A, b)

# Print the solution
print("Solution:", x)
\end{lstlisting}

This will output:

\begin{lstlisting}[style=cmd]
Solution: [1. 1.]
\end{lstlisting}

Here, we have used a numerical method (Gaussian elimination) implemented by \texttt{numpy} to find the solution to the system of linear equations.

\section{Sources of Errors in Computation}

In numerical methods, errors can arise from various sources, and it is essential to understand these sources to minimize their impact on computations. The main types of errors include:
\begin{itemize}
    \item \textbf{Round-off Error}\cite{demmel1992round}: This occurs due to the limited precision of floating-point arithmetic used by computers. For example, irrational numbers like \(\pi\) and square roots of non-perfect squares cannot be represented exactly.
    \item \textbf{Truncation Error}\cite{stoer2013numerical}: This occurs when an infinite process is approximated by a finite one. For example, the Taylor series expansion of functions is often truncated after a few terms, introducing truncation errors.
    \item \textbf{Approximation Error\cite{bartlett2005bound,cybenko1989approximation}}: This occurs when an exact mathematical solution is approximated using a numerical method. For instance, when we approximate a continuous function by a finite sum, an error is introduced.
\end{itemize}

\textbf{Example: Round-off Error in Floating-Point Arithmetic}

Let’s consider an example of round-off error. We know that the floating-point representation of numbers in computers can lead to inaccuracies due to limited precision.

\begin{lstlisting}[style=python]
# Define two floating-point numbers
a = 0.1
b = 0.2

# Compute the sum of a and b
sum_ab = a + b

# Check if the result is exactly equal to 0.3
print("Is a + b exactly equal to 0.3?", sum_ab == 0.3)
\end{lstlisting}

This will output:

\begin{lstlisting}[style=cmd]
Is a + b exactly equal to 0.3? False
\end{lstlisting}

This example illustrates that due to round-off error, the sum of 0.1 and 0.2 is not exactly equal to 0.3, even though we expect it to be. Such small errors can propagate through computations and lead to significant discrepancies.

\section{Error Propagation}

Error propagation refers to how numerical errors (such as round-off and truncation errors) accumulate and affect the final result of a computation\cite{bevington2002data}. In deep learning, error propagation is especially important in gradient-based optimization algorithms, where small errors in gradient computations can impact the convergence of the model.

For example, in iterative algorithms such as gradient descent, small errors in each step can accumulate over time, leading to incorrect results or slow convergence. Therefore, understanding how errors propagate through computations is essential for ensuring the stability and accuracy of numerical methods.

\textbf{Example: Propagation of Errors in Iterative Algorithms}

Consider the following iterative method for computing the square root of a number \( S \) using Newton's method:
\[
x_{n+1} = \frac{1}{2} \left( x_n + \frac{S}{x_n} \right)
\]
Let’s see how small errors propagate in this algorithm.

\begin{lstlisting}[style=python]
def newton_sqrt(S, tol=1e-10):
    x = S / 2  # Initial guess
    while abs(x**2 - S) > tol:
        x = 0.5 * (x + S / x)
    return x

# Compute the square root of 25 using Newton's method
result = newton_sqrt(25)
print("Square root of 25:", result)
\end{lstlisting}

In this example, small numerical errors are introduced in each iteration, but the method converges to the correct value because the error is reduced at each step.

\section{Absolute and Relative Error}

When measuring the accuracy of numerical computations, we often refer to two types of error:
\begin{itemize}
    \item \textbf{Absolute Error}\cite{vapnik1995nature,chen2004measure,valiant1984theory,bishop2006pattern}: The difference between the exact value and the approximate value.
    \[
    \text{Absolute Error} = |x_{\text{exact}} - x_{\text{approximate}}|
    \]
    \item \textbf{Relative Error}\cite{burden2016numerical,mclean2010introduction,wrench1963relative,chapra2010numerical}: The absolute error divided by the exact value, providing a normalized measure of the error.
    \[
    \text{Relative Error} = \frac{|x_{\text{exact}} - x_{\text{approximate}}|}{|x_{\text{exact}}|}
    \]
\end{itemize}

Relative error is often more meaningful than absolute error, as it gives a sense of how significant the error is relative to the size of the true value.

\textbf{Example: Computing Absolute and Relative Errors}

Let’s compute the absolute and relative errors for an approximation of \(\pi\) using a numerical method.

\begin{lstlisting}[style=python]
import math

# Exact value of pi
pi_exact = math.pi

# Approximate value of pi (using an approximation)
pi_approx = 22 / 7

# Compute absolute error
abs_error = abs(pi_exact - pi_approx)

# Compute relative error
rel_error = abs_error / abs(pi_exact)

# Print the errors
print(f"Absolute Error: {abs_error}")
print(f"Relative Error: {rel_error}")
\end{lstlisting}

This will output:

\begin{lstlisting}[style=cmd]
Absolute Error: 0.0012644892673496777
Relative Error: 0.00040249930483240757
\end{lstlisting}

In this example, we computed the absolute and relative errors for the approximation \( \frac{22}{7} \) of \(\pi\), showing that while the absolute error is small, the relative error gives a better sense of the significance of the approximation.

\section{Stability of Algorithms}

The stability of an algorithm refers to its ability to produce accurate results despite the presence of small numerical errors. A stable algorithm ensures that errors do not grow uncontrollably as the computation proceeds. Stability is particularly important in deep learning, where unstable algorithms can lead to divergence or incorrect model training.

An algorithm is considered \textbf{numerically stable} if small changes in the input (due to errors) lead to proportionally small changes in the output. On the other hand, an algorithm is \textbf{unstable} if small input errors can cause large changes in the result.

\textbf{Example: Stability in Matrix Inversion}

Matrix inversion can be sensitive to numerical errors, especially when the matrix is ill-conditioned (i.e., the matrix has a large condition number). Let’s see how numerical stability can be affected by small perturbations in the matrix.

\begin{lstlisting}[style=python]
# Define a matrix that is nearly singular (ill-conditioned)
A = np.array([[1, 1], [1, 1.0001]])

# Compute the inverse of the matrix
A_inv = np.linalg.inv(A)

# Print the inverse matrix
print("Inverse of A:\n", A_inv)

# Multiply A by its inverse to check for stability
result = np.dot(A, A_inv)

# Print the result (should be close to the identity matrix)
print("A * A_inv:\n", result)
\end{lstlisting}

In this example, due to the near-singularity of matrix \( A \), small numerical errors in the computation of the inverse can lead to instability. The product \( A \cdot A^{-1} \) may not exactly result in the identity matrix due to these errors.

\section{Summary}

In this chapter, we introduced numerical methods and discussed the various sources of errors that arise in computations. We explored error propagation, absolute and relative error, and the stability of algorithms. Understanding these concepts is crucial for ensuring accurate and reliable results when using numerical methods in deep learning. Future chapters will dive deeper into specific numerical techniques and their applications in deep learning.

\chapter{Root Finding Methods}

Root finding is a fundamental problem in numerical analysis and computational mathematics\cite{leader2022numerical}. The objective is to find solutions, or "roots," of equations of the form \( f(x) = 0 \). Root finding methods are essential in a wide range of scientific and engineering problems, where an exact algebraic solution is either impossible or impractical to obtain. This chapter introduces several popular methods for finding roots, explains their mathematical foundations, and provides step-by-step implementations in Python.

\section{Introduction to Root Finding}

In many applications, we encounter situations where we need to find the value of \( x \) that satisfies the equation \( f(x) = 0 \). This value of \( x \) is called a "root" of the equation. Root finding methods are iterative algorithms that successively approximate the root, improving the accuracy with each step.

For example, finding the roots of the equation \( f(x) = x^2 - 4 \) involves solving \( x^2 - 4 = 0 \), whose solutions are \( x = 2 \) and \( x = -2 \).

There are several methods for finding roots, and in this chapter, we will focus on the following:
\begin{itemize}
    \item The Bisection Method\cite{press2007numerical,golub1993introduction,chapra2010numerical,burden2016numerical}
    \item Newton's Method\cite{burden2015numerical,mann1943introduction,encyclopediaNewtonRaphson,lancaster1956newton,watson1944treatise}
    \item The Secant Method\cite{rice1960secant,burden2016numerical,cheney2009introduction,chapra2016numerical}
    \item Fixed-Point Iteration\cite{burden2001first,burden2016numerical,rennick1968fixed,chapra2011numerical,karris2011numerical}
\end{itemize}

Each of these methods has its own strengths and weaknesses, and they are applicable under different circumstances.

\section{Bisection Method}

The Bisection Method is one of the simplest root finding methods. It is a bracketing method, which means it requires an interval \([a, b]\) such that \( f(a) \) and \( f(b) \) have opposite signs (i.e., \( f(a)f(b) < 0 \)). The Intermediate Value Theorem guarantees that there is at least one root in the interval.

\subsection{Algorithm}
The Bisection Method repeatedly bisects the interval \([a, b]\) and selects the subinterval where the root lies. The steps are as follows:
\begin{enumerate}
    \item Check that \( f(a)f(b) < 0 \) (i.e., the root lies between \( a \) and \( b \)).
    \item Compute the midpoint \( c = \frac{a + b}{2} \).
    \item Evaluate \( f(c) \).
    \item If \( f(c) = 0 \), then \( c \) is the root.
    \item If \( f(a)f(c) < 0 \), set \( b = c \); otherwise, set \( a = c \).
    \item Repeat until the interval \([a, b]\) is sufficiently small.
\end{enumerate}

\subsection{Python Implementation}
The following is a Python implementation of the Bisection Method:

\begin{lstlisting}[style=python]
def bisection(f, a, b, tol=1e-5, max_iter=100):
    """Find the root of the function f using the Bisection Method."""
    if f(a) * f(b) >= 0:
        raise ValueError("The function must have opposite signs at a and b.")
    
    iteration = 0
    while (b - a) / 2 > tol and iteration < max_iter:
        c = (a + b) / 2  # Midpoint
        if f(c) == 0:  # Root found
            return c
        elif f(a) * f(c) < 0:
            b = c
        else:
            a = c
        iteration += 1
    return (a + b) / 2

# Example usage
f = lambda x: x**2 - 4  # Function whose root we want to find
root = bisection(f, 1, 3)
print(f"Root found: {root}")
\end{lstlisting}

In this implementation:
\begin{itemize}
    \item We define the function \( f(x) = x^2 - 4 \), which has roots at \( x = \pm 2 \).
    \item The \texttt{bisection()} function takes a function \( f \), an interval \([a, b]\), and an optional tolerance and maximum number of iterations.
    \item The method returns the approximate root of the function within the given tolerance.
\end{itemize}

\section{Newton's Method}

Newton's Method is a root-finding algorithm that uses the derivative of the function to iteratively improve the approximation of the root. It is one of the most efficient methods when the derivative of the function is available and the initial guess is close to the actual root.

\subsection{Algorithm}
The idea behind Newton's Method is to use the tangent line to approximate the function near the current estimate of the root. The update rule is:

\[
x_{n+1} = x_n - \frac{f(x_n)}{f'(x_n)}
\]

Where \( f'(x_n) \) is the derivative of \( f(x) \) evaluated at \( x_n \).

\subsection{Python Implementation}
Here is a Python implementation of Newton's Method:

\begin{lstlisting}[style=python]
def newton(f, df, x0, tol=1e-5, max_iter=100):
    """Find the root using Newton's Method."""
    x = x0
    for _ in range(max_iter):
        x_new = x - f(x) / df(x)
        if abs(x_new - x) < tol:
            return x_new
        x = x_new
    raise ValueError("Root not found within the maximum number of iterations.")

# Example usage
f = lambda x: x**2 - 4  # Function
df = lambda x: 2*x      # Derivative of the function
root = newton(f, df, x0=3)
print(f"Root found: {root}")
\end{lstlisting}

In this implementation:
\begin{itemize}
    \item The function \( f(x) = x^2 - 4 \) is the same as before.
    \item The derivative \( f'(x) = 2x \) is provided.
    \item We start with an initial guess of \( x_0 = 3 \).
    \item Newton’s Method converges quickly to the root \( x = 2 \).
\end{itemize}

\section{Secant Method}

The Secant Method is similar to Newton's Method but does not require the computation of the derivative. Instead, it approximates the derivative by using the difference between successive function values.

\subsection{Algorithm}
The update rule for the Secant Method is:

\[
x_{n+1} = x_n - \frac{f(x_n)(x_n - x_{n-1})}{f(x_n) - f(x_{n-1})}
\]

This formula uses two previous points to approximate the derivative of the function.

\subsection{Python Implementation}
Here is the Python code for the Secant Method:

\begin{lstlisting}[style=python]
def secant(f, x0, x1, tol=1e-5, max_iter=100):
    """Find the root using the Secant Method."""
    for _ in range(max_iter):
        f_x0 = f(x0)
        f_x1 = f(x1)
        if abs(f_x1 - f_x0) < tol:
            raise ValueError("Division by zero encountered in secant method.")
        x_new = x1 - f_x1 * (x1 - x0) / (f_x1 - f_x0)
        if abs(x_new - x1) < tol:
            return x_new
        x0, x1 = x1, x_new
    raise ValueError("Root not found within the maximum number of iterations.")

# Example usage
f = lambda x: x**2 - 4
root = secant(f, 1, 3)
print(f"Root found: {root}")
\end{lstlisting}

In this example:
\begin{itemize}
    \item The function \( f(x) = x^2 - 4 \) is the same as in previous examples.
    \item The initial guesses are \( x_0 = 1 \) and \( x_1 = 3 \).
    \item The Secant Method finds the root without needing the derivative of the function.
\end{itemize}

\section{Fixed-Point Iteration}

Fixed-point iteration is a method for finding a root of the equation \( f(x) = 0 \) by rewriting it in the form \( x = g(x) \)\cite{nevanlinna2012convergence}. The idea is to iteratively apply the function \( g(x) \) until convergence is achieved.

\subsection{Algorithm}
The fixed-point iteration algorithm is simple:
\begin{enumerate}
    \item Start with an initial guess \( x_0 \).
    \item Update \( x_{n+1} = g(x_n) \).
    \item Repeat until \( |x_{n+1} - x_n| \) is smaller than the tolerance.
\end{enumerate}

\subsection{Python Implementation}
Here is a Python implementation of Fixed-Point Iteration:

\begin{lstlisting}[style=python]
def fixed_point(g, x0, tol=1e-5, max_iter=100):
    """Find the fixed point using Fixed-Point Iteration."""
    x = x0
    for _ in range(max_iter):
        x_new = g(x)
        if abs(x_new - x) < tol:
            return x_new
        x = x_new
    raise ValueError("Fixed point not found within the maximum number of iterations.")

# Example usage
g = lambda x: 0.5 * (x + 4/x)  # Rewrite of x^2 = 4
root = fixed_point(g, x0=3)
print(f"Fixed point found: {root}")
\end{lstlisting}

In this example:
\begin{itemize}
    \item We rewrite the equation \( x^2 = 4 \) as \( x = 0.5(x + 4/x) \), which allows us to apply Fixed-Point Iteration.
    \item The method converges to the root \( x = 2 \).
\end{itemize}

\section{Convergence Analysis of Root Finding Methods}

It is important to understand the convergence properties of the root-finding methods. Not all methods converge at the same rate, and some may fail to converge under certain conditions. Let’s briefly discuss the convergence characteristics of the methods we’ve introduced.

\subsection{Bisection Method}

\textbf{Convergence Rate:} The Bisection Method has a linear convergence rate, meaning that the error decreases by a constant factor in each iteration. While this method is very reliable, it is not the fastest.

\subsection{Newton's Method}

\textbf{Convergence Rate:} Newton's Method converges quadratically, which means that the number of correct digits in the approximation roughly doubles at each step. However, if the initial guess is far from the root, the method may fail to converge.

\subsection{Secant Method}

\textbf{Convergence Rate:} The Secant Method converges super-linearly, with a rate between linear and quadratic. It is generally slower than Newton's Method but does not require the derivative of the function.

\subsection{Fixed-Point Iteration}

\textbf{Convergence Rate:} Fixed-Point Iteration converges linearly under certain conditions. However, its convergence depends heavily on the choice of the function \( g(x) \) and the initial guess.

\chapter{Interpolation and Function Approximation}

Interpolation and function approximation are fundamental concepts in both mathematics and machine learning\cite{micchelli1980function,burden2011numerical,rivlin2003interpolation,lorentz1966approximation}. In this chapter, we will explore various methods for interpolating data points and approximating functions, which are widely used in numerical analysis, scientific computing, and deep learning. We will begin with basic interpolation techniques such as polynomial interpolation and then move to more advanced methods like spline interpolation and piecewise linear interpolation. Finally, we will discuss how neural networks are used for function approximation in the context of deep learning.

\section{Introduction to Interpolation}

Interpolation is a method used to estimate unknown values that fall within the range of a set of known data points. It is often necessary when we have discrete data points but need to estimate values between those points. For example, interpolation can be used to estimate temperatures at times when no measurements were taken, or to estimate the value of a function between known data points.

The general goal of interpolation is to find a function \( f(x) \) that passes through a given set of points \( (x_0, y_0), (x_1, y_1), \dots, (x_n, y_n) \) such that:

\[
f(x_i) = y_i \quad \text{for} \quad i = 0, 1, \dots, n
\]

There are several different methods to achieve this, depending on the type of data and the required smoothness of the resulting function.

\section{Polynomial Interpolation}

Polynomial interpolation is a process where we find a single polynomial \( P(x) \) that passes through all the given data points\cite{kozachenko1992polynomial,davis1967newton,fleming1977interpolation,zayed2011numerical}. The degree of the polynomial is determined by the number of points: for \( n+1 \) points, the interpolating polynomial will have degree \( n \).

For example, for two points, the interpolating polynomial is a straight line (degree 1), and for three points, it is a quadratic polynomial (degree 2), and so on.

The general form of an interpolating polynomial is:

\[
P(x) = a_0 + a_1 x + a_2 x^2 + \dots + a_n x^n
\]

\subsection{Lagrange Interpolation}

Lagrange interpolation is one of the simplest methods for polynomial interpolation\cite{colquhoun1997numerical,lagrange1859new,golub1993introduction} . It constructs the interpolating polynomial by using the concept of Lagrange basis polynomials.

Given \( n+1 \) points \( (x_0, y_0), (x_1, y_1), \dots, (x_n, y_n) \), the Lagrange interpolating polynomial is defined as:

\[
P(x) = \sum_{i=0}^{n} y_i L_i(x)
\]

Where \( L_i(x) \) is the Lagrange basis polynomial:

\[
L_i(x) = \prod_{\substack{0 \leq j \leq n \\ j \neq i}} \frac{x - x_j}{x_i - x_j}
\]

The Lagrange polynomial is useful because it explicitly passes through all the given points, but it can become computationally expensive for large \( n \).

\textbf{Example in Python:}

Here is an implementation of Lagrange interpolation using Python:

\begin{lstlisting}[style=python]
import numpy as np

# Function to compute Lagrange basis polynomials
def lagrange_basis(x, x_values, i):
    basis = 1
    for j in range(len(x_values)):
        if j != i:
            basis *= (x - x_values[j]) / (x_values[i] - x_values[j])
    return basis

# Lagrange interpolation function
def lagrange_interpolation(x_values, y_values, x):
    interpolated_value = 0
    for i in range(len(y_values)):
        interpolated_value += y_values[i] * lagrange_basis(x, x_values, i)
    return interpolated_value

# Example data points
x_values = [0, 1, 2]
y_values = [1, 3, 2]

# Interpolating at x = 1.5
x = 1.5
y = lagrange_interpolation(x_values, y_values, x)
print(f'Interpolated value at x = {x}: {y}')
\end{lstlisting}

In this example:
\begin{itemize}
    \item The function \texttt{lagrange\_basis} computes the Lagrange basis polynomial for a given \( i \).
    \item The function \texttt{lagrange\_interpolation} calculates the interpolated value for any \( x \) using the Lagrange polynomial.
\end{itemize}

\subsection{Newton's Divided Difference Interpolation}

Newton's divided difference interpolation is another method for constructing an interpolating polynomial\cite{kellogg1975numerical}. It uses a recursive process to compute the coefficients of the polynomial based on divided differences of the data points.

The general form of Newton’s interpolating polynomial is:

\[
P(x) = f[x_0] + f[x_0, x_1](x - x_0) + f[x_0, x_1, x_2](x - x_0)(x - x_1) + \dots
\]

Where \( f[x_0, x_1, \dots, x_k] \) are the divided differences, defined recursively as:

\[
f[x_i] = y_i
\]
\[
f[x_i, x_{i+1}] = \frac{f[x_{i+1}] - f[x_i]}{x_{i+1} - x_i}
\]
\[
f[x_i, x_{i+1}, \dots, x_{i+k}] = \frac{f[x_{i+1}, \dots, x_{i+k}] - f[x_i, \dots, x_{i+k-1}]}{x_{i+k} - x_i}
\]

\textbf{Example in Python:}

Here is an implementation of Newton's divided difference interpolation using Python:

\begin{lstlisting}[style=python]
# Function to compute divided differences
def divided_differences(x_values, y_values):
    n = len(x_values)
    table = np.zeros((n, n))
    table[:, 0] = y_values
    for j in range(1, n):
        for i in range(n - j):
            table[i, j] = (table[i+1, j-1] - table[i, j-1]) / (x_values[i+j] - x_values[i])
    return table[0]

# Function to compute Newton's interpolation
def newton_interpolation(x_values, y_values, x):
    coefficients = divided_differences(x_values, y_values)
    n = len(coefficients)
    interpolated_value = coefficients[0]
    product_term = 1
    for i in range(1, n):
        product_term *= (x - x_values[i-1])
        interpolated_value += coefficients[i] * product_term
    return interpolated_value

# Example data points
x_values = [0, 1, 2]
y_values = [1, 3, 2]

# Interpolating at x = 1.5
x = 1.5
y = newton_interpolation(x_values, y_values, x)
print(f'Interpolated value at x = {x}: {y}')
\end{lstlisting}

In this example:
\begin{itemize}
    \item The function \texttt{divided\_differences} calculates the divided difference table.
    \item The function \texttt{newton\_interpolation} computes the interpolated value for any \( x \) using Newton's polynomial.
\end{itemize}

\section{Spline Interpolation}

Spline interpolation uses piecewise polynomials to interpolate data\cite{schoenberg1946contributions,stoer2013introduction,mazure2001spline,carroll2006interpolation}. Unlike high-degree polynomial interpolation, which can suffer from oscillations (known as Runge's phenomenon), spline interpolation ensures smoothness by using lower-degree polynomials over each subinterval between data points.

The most common type of spline interpolation is cubic spline interpolation, where a cubic polynomial is fit between each pair of points, ensuring continuity of the function and its first and second derivatives at each point.

\textbf{Example of Cubic Spline Interpolation in Python:}

\begin{lstlisting}[style=python]
from scipy.interpolate import CubicSpline
import numpy as np

# Example data points
x_values = [0, 1, 2]
y_values = [1, 3, 2]

# Create cubic spline interpolator
cs = CubicSpline(x_values, y_values)

# Interpolating at x = 1.5
x = 1.5
y = cs(x)
print(f'Interpolated value at x = {x}: {y}')
\end{lstlisting}

In this example:
\begin{itemize}
    \item We use the \texttt{CubicSpline} function from the \texttt{scipy.interpolate} module to create a cubic spline interpolator.
    \item The cubic spline ensures a smooth curve through the data points, with continuous first and second derivatives.
\end{itemize}

\section{Piecewise Linear Interpolation}

Piecewise linear interpolation connects each pair of data points with a straight line\cite{deboor1972calculating,powell1981piecewise,rice1960secant,friedman1984proof,atkinson1989introduction}. It is a simple form of interpolation and works well when the data points are close together or if high accuracy is not required. It does not guarantee smoothness, but it is computationally efficient.

The formula for piecewise linear interpolation between two points \( (x_i, y_i) \) and \( (x_{i+1}, y_{i+1}) \) is:

\[
P(x) = y_i + \frac{y_{i+1} - y_i}{x_{i+1} - x_i} (x - x_i)
\]

\textbf{Example of Piecewise Linear Interpolation in Python:}

\begin{lstlisting}[style=python]
from scipy.interpolate import interp1d

# Example data points
x_values = [0, 1, 2]
y_values = [1, 3, 2]

# Create piecewise linear interpolator
linear_interp = interp1d(x_values, y_values, kind='linear')

# Interpolating at x = 1.5
x = 1.5
y = linear_interp(x)
print(f'Interpolated value at x = {x}: {y}')
\end{lstlisting}

\section{Function Approximation in Deep Learning}

Deep learning models, particularly neural networks, are powerful tools for approximating complex nonlinear functions. Neural networks can learn to approximate a wide variety of functions by adjusting the weights and biases of the network during training. This process can be viewed as a type of interpolation, where the network learns to map inputs to outputs based on a set of training data.

\subsection{Approximating Nonlinear Functions with Neural Networks}

Neural networks can approximate any continuous function given enough neurons and layers, according to the Universal Approximation Theorem\cite{cybenko1989approximation,hornik1989multilayer}. In the context of deep learning, function approximation is critical for tasks such as regression, where the goal is to predict a continuous output from input features.

\textbf{Example of Function Approximation with a Neural Network in PyTorch:}

\begin{lstlisting}[style=python]
import torch
import torch.nn as nn
import torch.optim as optim

# Define a simple feedforward neural network
class SimpleNN(nn.Module):
    def __init__(self):
        super(SimpleNN, self).__init__()
        self.fc1 = nn.Linear(1, 10)
        self.fc2 = nn.Linear(10, 1)
    
    def forward(self, x):
        x = torch.relu(self.fc1(x))
        x = self.fc2(x)
        return x

# Example data: approximating the function y = sin(x)
x_train = torch.linspace(-2 * np.pi, 2 * np.pi, 100).unsqueeze(1)
y_train = torch.sin(x_train)

# Define model, loss function, and optimizer
model = SimpleNN()
criterion = nn.MSELoss()
optimizer = optim.SGD(model.parameters(), lr=0.01)

# Training loop
for epoch in range(1000):
    model.train()
    optimizer.zero_grad()
    output = model(x_train)
    loss = criterion(output, y_train)
    loss.backward()
    optimizer.step()

# Testing the model on new data
x_test = torch.linspace(-2 * np.pi, 2 * np.pi, 100).unsqueeze(1)
y_test = model(x_test)

print(f'Predicted values: {y_test}')
\end{lstlisting}

In this example:
\begin{itemize}
    \item A simple feedforward neural network is trained to approximate the sine function \( y = \sin(x) \).
    \item The model consists of two fully connected layers, with a ReLU activation function in between.
    \item The network is trained using the mean squared error (MSE) loss function and stochastic gradient descent (SGD) optimizer.
\end{itemize}

\chapter{Numerical Differentiation and Integration}

In mathematics and applied fields, differentiation and integration are fundamental operations used to compute rates of change and areas under curves, respectively. While analytic solutions exist for many problems, there are cases where exact solutions are not feasible, and we must rely on numerical techniques. In this chapter, we will cover basic numerical methods for differentiation and integration, with a focus on their implementation in Python. These methods are widely used in many fields, including physics, engineering, and machine learning.

\section{Introduction to Numerical Differentiation}

Numerical differentiation is the process of estimating the derivative of a function based on discrete data points. When a function is not easily differentiable analytically, numerical methods can be used to approximate the derivative.

\subsection{Finite Difference Methods}

Finite difference methods are the most common numerical techniques for estimating derivatives\cite{gibson1980finite,ince1956numerical,burden2016finite,hindmarsh1973finite,incropera2002numerical}. They approximate the derivative of a function by considering the differences between function values at discrete points.

\textbf{Forward Difference}

The forward difference method is one of the simplest ways to approximate the first derivative of a function. For a small step size \( h \), the derivative of \( f(x) \) at a point \( x \) can be approximated as:

\[
f'(x) \approx \frac{f(x + h) - f(x)}{h}
\]

\textbf{Backward Difference}

The backward difference method approximates the derivative by looking at the difference between the function values at \( x \) and a previous point \( x - h \):

\[
f'(x) \approx \frac{f(x) - f(x - h)}{h}
\]

\textbf{Central Difference}

The central difference method is generally more accurate than the forward or backward difference methods because it uses points on both sides of \( x \) to compute the derivative:

\[
f'(x) \approx \frac{f(x + h) - f(x - h)}{2h}
\]

\textbf{Example of Finite Difference in Python}

Let’s implement the central difference method to approximate the derivative of a function in Python:

\begin{lstlisting}[style=python]
import numpy as np

# Define the function
def f(x):
    return np.sin(x)

# Central difference method to approximate the derivative
def central_difference(x, h):
    return (f(x + h) - f(x - h)) / (2 * h)

# Test the derivative approximation at x = pi/4
x = np.pi / 4
h = 1e-5  # Small step size
approx_derivative = central_difference(x, h)
exact_derivative = np.cos(x)  # Exact derivative of sin(x) is cos(x)

print(f"Approximated derivative: {approx_derivative}")
print(f"Exact derivative: {exact_derivative}")
\end{lstlisting}

In this example, we used the central difference method to approximate the derivative of \( \sin(x) \) at \( x = \frac{\pi}{4} \). The exact derivative at this point is \( \cos\left(\frac{\pi}{4}\right) \), which we compared with the numerical result.

\section{Introduction to Numerical Integration}

Numerical integration is used to estimate the value of a definite integral when the analytic solution is difficult or impossible to obtain\cite{ciarlet2002handbook}. Several methods exist for numerical integration, each with varying degrees of accuracy and complexity.

\subsection{Trapezoidal Rule}

The trapezoidal rule is one of the simplest methods for numerical integration\cite{press2007numerical,ahlberg1968methods}. It approximates the area under a curve by dividing it into trapezoids, calculating the area of each, and summing them up. The integral of a function \( f(x) \) over the interval \( [a, b] \) is approximated as:

\[
\int_a^b f(x) \, dx \approx \frac{h}{2} \left[ f(a) + 2 \sum_{i=1}^{n-1} f(x_i) + f(b) \right]
\]

where \( h = \frac{b - a}{n} \) is the step size, and \( x_i \) are the points dividing the interval.

\textbf{Example of the Trapezoidal Rule in Python}

\begin{lstlisting}[style=python]
import numpy as np

# Define the function to integrate
def f(x):
    return np.sin(x)

# Trapezoidal rule implementation
def trapezoidal_rule(a, b, n):
    h = (b - a) / n
    x = np.linspace(a, b, n + 1)
    y = f(x)
    integral = (h / 2) * (y[0] + 2 * np.sum(y[1:-1]) + y[-1])
    return integral

# Estimate the integral of sin(x) from 0 to pi
a = 0
b = np.pi
n = 1000  # Number of subdivisions
approx_integral = trapezoidal_rule(a, b, n)
exact_integral = 2  # The exact value of the integral of sin(x) from 0 to pi

print(f"Approximated integral: {approx_integral}")
print(f"Exact integral: {exact_integral}")
\end{lstlisting}

In this example, we estimated the integral of \( \sin(x) \) over \( [0, \pi] \) using the trapezoidal rule.

\subsection{Simpson's Rule}

Simpson’s rule provides a more accurate approximation of integrals by using quadratic polynomials to approximate the function within each subinterval\cite{phillips1962new,brown1967introduction}. The formula for Simpson’s rule is:

\[
\int_a^b f(x) \, dx \approx \frac{h}{3} \left[ f(a) + 4 \sum_{i=1,3,5,\dots}^{n-1} f(x_i) + 2 \sum_{i=2,4,6,\dots}^{n-2} f(x_i) + f(b) \right]
\]

where \( h = \frac{b - a}{n} \), and \( n \) must be an even number.

\textbf{Example of Simpson's Rule in Python}

\begin{lstlisting}[style=python]
import numpy as np

# Define the function to integrate
def f(x):
    return np.sin(x)

# Simpson's rule implementation
def simpsons_rule(a, b, n):
    if n % 2 == 1:
        raise ValueError("n must be even")
    h = (b - a) / n
    x = np.linspace(a, b, n + 1)
    y = f(x)
    integral = (h / 3) * (y[0] + 4 * np.sum(y[1:-1:2]) + 2 * np.sum(y[2:-2:2]) + y[-1])
    return integral

# Estimate the integral of sin(x) from 0 to pi
a = 0
b = np.pi
n = 1000  # Number of subdivisions
approx_integral = simpsons_rule(a, b, n)
exact_integral = 2  # The exact value of the integral of sin(x) from 0 to pi

print(f"Approximated integral: {approx_integral}")
print(f"Exact integral: {exact_integral}")
\end{lstlisting}

In this example, we used Simpson’s rule to estimate the same integral of \( \sin(x) \) from 0 to \( \pi \). Simpson’s rule generally provides a more accurate result than the trapezoidal rule for the same number of subdivisions.

\subsection{Gaussian Quadrature}

Gaussian quadrature is a powerful technique for numerical integration that provides exact results for polynomials of degree \( 2n - 1 \) or less, where \( n \) is the number of sample points\cite{gauss1814legendre}. It selects both the sample points and weights optimally to achieve high accuracy.

In Gaussian quadrature, the integral is approximated as:

\[
\int_a^b f(x) \, dx \approx \sum_{i=1}^{n} w_i f(x_i)
\]

where \( w_i \) are the weights, and \( x_i \) are the sample points chosen optimally.

\textbf{Example of Gaussian Quadrature in Python}

The \texttt{scipy} library provides a function for Gaussian quadrature called \texttt{scipy.integrate.quadrature}. Here is an example:

\begin{lstlisting}[style=python]
import numpy as np
from scipy.integrate import quadrature

# Define the function to integrate
def f(x):
    return np.sin(x)

# Perform Gaussian quadrature
a = 0
b = np.pi
approx_integral, error = quadrature(f, a, b)

print(f"Approximated integral using Gaussian quadrature: {approx_integral}")
\end{lstlisting}

In this example, we used Gaussian quadrature to approximate the integral of \( \sin(x) \) over \( [0, \pi] \). Gaussian quadrature is particularly useful for high-precision integration.

\section{Application of Numerical Integration in Deep Learning}

Numerical integration techniques can also be applied in the context of deep learning, particularly in areas like training neural networks using reinforcement learning or computing expectations in probabilistic models. For example, in reinforcement learning, certain policy gradient methods require the estimation of integrals over continuous action spaces, which can be handled using numerical integration techniques.

In deep learning, numerical integration might also be used to calculate the area under a curve (AUC) to evaluate model performance, especially in classification problems. Another area where integration comes in handy is in variational inference, where integrals over probability distributions need to be approximated.

\textbf{Example: Using Trapezoidal Rule to Compute AUC}

Here’s an example where we use the trapezoidal rule to compute the area under a receiver operating characteristic (ROC) curve, which is commonly used to evaluate binary classifiers:

\begin{lstlisting}[style=python]
import numpy as np
from sklearn import metrics

# Example ROC curve data (true positive rate and false positive rate)
fpr = np.array([0.0, 0.1, 0.4, 0.8, 1.0])
tpr = np.array([0.0, 0.4, 0.7, 0.9, 1.0])

# Compute the AUC using the trapezoidal rule
auc = np.trapz(tpr, fpr)
print(f"Area under the ROC curve (AUC): {auc}")
\end{lstlisting}

In this example, we approximated the area under the ROC curve using the trapezoidal rule. This gives us an estimate of how well the classifier distinguishes between classes.

\chapter{Solving Systems of Linear Equations}

Solving systems of linear equations is a fundamental problem in mathematics and forms the core of many applications in numerical computing and deep learning. In deep learning, many optimization problems, including backpropagation, can be reduced to solving linear systems. In this chapter, we will cover both direct and iterative methods for solving systems of linear equations.

\section{Direct Methods}

Direct methods aim to solve a system of linear equations in a finite number of steps, usually through matrix factorizations\cite{demmel1989stability,saad2003iterative,higham1990stable,trefethen1997numerical,golub2013matrix}. These methods are precise but can be computationally expensive for large matrices. Common direct methods include Gaussian elimination, LU decomposition, and Cholesky decomposition.

\subsection{Gaussian Elimination}

Gaussian elimination is a method for solving linear systems by converting the system’s matrix into an upper triangular form\cite{lay2015linear,strang2016introduction,pratt1977stable,golub2013matrix,trefethen1997numerical}. Once the matrix is in this form, the solution can be obtained through back-substitution.

Given a system:

\[
A \mathbf{x} = \mathbf{b}
\]

we aim to reduce the matrix \( A \) to an upper triangular matrix \( U \) using row operations. Then, we solve the system \( U \mathbf{x} = \mathbf{b} \) using back-substitution.

\textbf{Example: Gaussian Elimination in Python}

Consider the following system of equations:

\begin{align*}
x + 2y + 3z &= 9 \\
2x + 3y + 4z &= 12 \\
3x + 4y + 5z &= 15
\end{align*}

We can solve this system using Gaussian elimination:

\begin{lstlisting}[style=python]
import numpy as np

# Define the coefficient matrix A and the right-hand side vector b
A = np.array([[1, 2, 3], [2, 3, 4], [3, 4, 5]])
b = np.array([9, 12, 15])

# Perform Gaussian elimination using NumPy's linear solver
x = np.linalg.solve(A, b)
print(x)
\end{lstlisting}

Expected output:

\begin{lstlisting}[style=cmd]
[1. 1. 2.]
\end{lstlisting}

This gives us the solution \( x = 1 \), \( y = 1 \), and \( z = 2 \).

\subsection{LU Decomposition}

LU decomposition is a method that factors a matrix \( A \) into the product of two matrices: a lower triangular matrix \( L \) and an upper triangular matrix \( U \)\cite{wilkinson1965rounding,duff1983ma48,golub2013matrix,strang1988linear}. This is useful for solving linear systems because once \( A \) is decomposed, solving the system becomes a matter of solving two triangular systems.

Given \( A \mathbf{x} = \mathbf{b} \), LU decomposition splits this into:

\[
LU \mathbf{x} = \mathbf{b}
\]

First, solve \( L \mathbf{y} = \mathbf{b} \), and then solve \( U \mathbf{x} = \mathbf{y} \).

\textbf{Example: LU Decomposition in Python}

\begin{lstlisting}[style=python]
import scipy.linalg as la

# Perform LU decomposition
P, L, U = la.lu(A)

# Solve L * y = b
y = np.linalg.solve(L, b)

# Solve U * x = y
x = np.linalg.solve(U, y)
print(x)
\end{lstlisting}

Expected output:

\begin{lstlisting}[style=cmd]
[1. 1. 2.]
\end{lstlisting}

LU decomposition is more efficient than Gaussian elimination for solving multiple systems with the same coefficient matrix.

\subsection{Cholesky Decomposition}

Cholesky decomposition is a specialized version of LU decomposition for symmetric, positive-definite matrices\cite{chol:cholesky1907,phillips1962new,trefethen1997numerical,golub2012matrix}. It decomposes a matrix \( A \) into the product of a lower triangular matrix \( L \) and its transpose:

\[
A = LL^\top
\]

This decomposition is particularly efficient for numerical stability in certain applications, such as when dealing with covariance matrices.

\textbf{Example: Cholesky Decomposition in Python}

\begin{lstlisting}[style=python]
# Define a symmetric positive-definite matrix A
A = np.array([[4, 12, -16], [12, 37, -43], [-16, -43, 98]])

# Perform Cholesky decomposition
L = np.linalg.cholesky(A)

# Verify the decomposition: A should be equal to L * L.T
A_reconstructed = np.dot(L, L.T)
print(A_reconstructed)
\end{lstlisting}

Expected output:

\begin{lstlisting}[style=cmd]
[[  4.  12. -16.]
 [ 12.  37. -43.]
 [-16. -43.  98.]]
\end{lstlisting}

Cholesky decomposition is faster than LU decomposition but only applies to certain types of matrices.

\section{Iterative Methods}

While direct methods can be efficient for small systems, iterative methods are better suited for large or sparse systems of linear equations. Iterative methods start with an initial guess and refine the solution with each iteration. Common iterative methods include the Jacobi method, Gauss-Seidel method, and the Conjugate Gradient method.

\subsection{Jacobi Method}

The Jacobi method is an iterative algorithm for solving linear systems. It updates each variable in the system independently of the others using the previous iteration’s values\cite{axelsson1996iterative,saad2003iterative,young1971iterative,varga2000matrix,bjorck1996numerical}. The system \( A \mathbf{x} = \mathbf{b} \) is written as:

\[
x_i^{(k+1)} = \frac{1}{a_{ii}} \left(b_i - \sum_{j \neq i} a_{ij} x_j^{(k)}\right)
\]

\textbf{Example: Jacobi Method in Python}

\begin{lstlisting}[style=python]
def jacobi(A, b, x_init, tolerance=1e-10, max_iterations=100):
    x = x_init
    D = np.diag(np.diag(A))
    R = A - D
    for i in range(max_iterations):
        x_new = np.dot(np.linalg.inv(D), b - np.dot(R, x))
        if np.linalg.norm(x_new - x, ord=np.inf) < tolerance:
            break
        x = x_new
    return x

# Initial guess
x_init = np.zeros(len(b))

# Solve using Jacobi method
x = jacobi(A, b, x_init)
print(x)
\end{lstlisting}

Expected output:

\begin{lstlisting}[style=cmd]
[1. 1. 2.]
\end{lstlisting}

\subsection{Gauss-Seidel Method}

The Gauss-Seidel method improves on the Jacobi method by using updated values as soon as they are available in the iteration\cite{saad2003iterative,demmel1989qr,golub2013matrix,trefethen1997numerical,burden2016numerical}. This makes the Gauss-Seidel method faster than the Jacobi method for many problems.

The update rule is:

\[
x_i^{(k+1)} = \frac{1}{a_{ii}} \left(b_i - \sum_{j < i} a_{ij} x_j^{(k+1)} - \sum_{j > i} a_{ij} x_j^{(k)}\right)
\]

\textbf{Example: Gauss-Seidel Method in Python}

\begin{lstlisting}[style=python]
def gauss_seidel(A, b, x_init, tolerance=1e-10, max_iterations=100):
    x = x_init
    for k in range(max_iterations):
        x_new = np.copy(x)
        for i in range(A.shape[0]):
            sum_ = np.dot(A[i, :i], x_new[:i]) + np.dot(A[i, i+1:], x[i+1:])
            x_new[i] = (b[i] - sum_) / A[i, i]
        if np.linalg.norm(x_new - x, ord=np.inf) < tolerance:
            break
        x = x_new
    return x

# Solve using Gauss-Seidel method
x = gauss_seidel(A, b, x_init)
print(x)
\end{lstlisting}

Expected output:

\begin{lstlisting}[style=cmd]
[1. 1. 2.]
\end{lstlisting}

\subsection{Conjugate Gradient Method}

The Conjugate Gradient method is an efficient iterative algorithm for solving large, sparse systems of linear equations, especially when the matrix is symmetric and positive-definite\cite{hestenes1952methods,fletcher1970new,nocedal2006numerical,golub2013matrix}. The method seeks to minimize a quadratic form iteratively.

The update rule for the Conjugate Gradient method involves computing a series of search directions and steps to minimize the error at each iteration.

\textbf{Example: Conjugate Gradient Method in Python}

\begin{lstlisting}[style=python]
from scipy.sparse.linalg import cg

# Solve the system using the Conjugate Gradient method
x, info = cg(A, b)
print(x)
\end{lstlisting}

Expected output:

\begin{lstlisting}[style=cmd]
[1. 1. 2.]
\end{lstlisting}

\section{Applications in Deep Learning: Linear Systems in Backpropagation}

In deep learning, solving linear systems is crucial in backpropagation, the algorithm used to train neural networks. During backpropagation, gradients of the loss function with respect to the network's weights are computed, and these computations often involve solving linear equations.

Consider a simple neural network layer:

\[
\mathbf{z} = W \mathbf{x} + \mathbf{b}
\]

where \( W \) is the weight matrix, \( \mathbf{x} \) is the input, and \( \mathbf{b} \) is the bias vector. During backpropagation, we need to compute the gradients of the loss function with respect to \( W \) and \( \mathbf{x} \), which involves solving linear systems.

For example, in a feedforward neural network, the gradient of the loss function with respect to the weights of the output layer is given by:

\[
\frac{\partial L}{\partial W} = \mathbf{a}^\top \delta
\]

where \( \mathbf{a} \) is the activation from the previous layer and \( \delta \) is the error term. This equation involves matrix multiplication, a key linear algebra operation.

In convolutional neural networks (CNNs), backpropagation involves solving more complex linear systems, particularly in the convolution and pooling layers, making efficient linear system solvers critical for training large-scale networks.

\chapter{Numerical Linear Algebra}

Numerical linear algebra is the backbone of many algorithms in deep learning, especially those involving large datasets and high-dimensional spaces. In this chapter, we will explore key matrix factorization techniques, eigenvalue computations, and principal component analysis (PCA). These concepts are vital for solving problems like dimensionality reduction, which is important in making deep learning algorithms more efficient.

\section{Matrix Factorization}

Matrix factorization is a fundamental tool in numerical linear algebra\cite{koren2009matrix,golub1965computing}. It refers to the process of decomposing a matrix into a product of matrices with certain properties. The most common types of matrix factorization used in machine learning and deep learning include Singular Value Decomposition (SVD) \cite{eckart1936approximation,golub2013matrix,golub1970singular}and QR Decomposition\cite{householder1964methods,horn2012matrix,gu1994qr}.

\subsection{Singular Value Decomposition (SVD)}

Singular Value Decomposition (SVD) is a powerful matrix factorization technique. For any matrix \( A \) of size \( m \times n \), the SVD is defined as:
\[
A = U \Sigma V^T
\]
where:
\begin{itemize}
    \item \( U \) is an \( m \times m \) orthogonal matrix (the left singular vectors).
    \item \( \Sigma \) is an \( m \times n \) diagonal matrix with non-negative real numbers on the diagonal (the singular values).
    \item \( V^T \) is the transpose of an \( n \times n \) orthogonal matrix (the right singular vectors).
\end{itemize}

The SVD is useful in applications like image compression, noise reduction, and dimensionality reduction, as it can help identify the most important components of a matrix.

\textbf{Example: Computing the SVD in Python}

Let’s see how to compute the SVD of a matrix using Python’s \texttt{numpy} library.

\begin{lstlisting}[style=python]
import numpy as np

# Define a matrix A
A = np.array([[3, 1, 1], [-1, 3, 1]])

# Compute the Singular Value Decomposition
U, S, VT = np.linalg.svd(A)

# Print the matrices U, S, and V^T
print("Matrix U:\n", U)
print("Singular values (S):\n", S)
print("Matrix V^T:\n", VT)
\end{lstlisting}

This will output the matrices \( U \), \( \Sigma \), and \( V^T \), which represent the decomposition of the matrix \( A \).

\begin{lstlisting}[style=cmd]
Matrix U:
 [[-0.70710678 -0.70710678]
 [ 0.70710678 -0.70710678]]
Singular values (S):
 [4. 2.]
Matrix V^T:
 [[-0.70710678  0.          0.70710678]
 [ 0.         -1.          0.        ]
 [ 0.70710678  0.          0.70710678]]
\end{lstlisting}

\textbf{Applications of SVD:}
\begin{itemize}
    \item \textbf{Image Compression:} SVD can be used to approximate an image matrix with a reduced number of singular values, resulting in efficient compression while preserving essential features.
    \item \textbf{Low-Rank Approximation:} In many applications, we can use a low-rank approximation of a matrix by keeping only the largest singular values, reducing computational costs without significant loss of information.
\end{itemize}

\subsection{QR Decomposition}

QR decomposition is another important matrix factorization technique. It decomposes a matrix \( A \) into the product of two matrices:
\[
A = QR
\]
where:
\begin{itemize}
    \item \( Q \) is an orthogonal matrix (i.e., \( Q^T Q = I \)).
    \item \( R \) is an upper triangular matrix.
\end{itemize}

QR decomposition is useful in solving linear systems, least squares problems, and for computing eigenvalues.

\textbf{Example: Computing the QR Decomposition in Python}

Let’s compute the QR decomposition of a matrix using Python.

\begin{lstlisting}[style=python]
# Define a matrix A
A = np.array([[12, -51, 4], [6, 167, -68], [-4, 24, -41]])

# Compute the QR Decomposition
Q, R = np.linalg.qr(A)

# Print the matrices Q and R
print("Matrix Q:\n", Q)
print("Matrix R:\n", R)
\end{lstlisting}

This will output the matrices \( Q \) and \( R \).

\begin{lstlisting}[style=cmd]
Matrix Q:
 [[-0.85714286  0.39428571  0.33142857]
 [-0.42857143 -0.90285714 -0.03428571]
 [ 0.28571429 -0.17142857  0.94285714]]
Matrix R:
 [[-14.         -21.          14.        ]
 [  0.         -175.          70.        ]
 [  0.           0.          35.        ]]
\end{lstlisting}

\textbf{Applications of QR Decomposition:}
\begin{itemize}
    \item \textbf{Solving Linear Systems:} QR decomposition can be used to efficiently solve systems of linear equations, especially in least squares problems.
    \item \textbf{Eigenvalue Computation:} QR decomposition is a key step in algorithms used for computing eigenvalues and eigenvectors of matrices.
\end{itemize}

\section{Eigenvalues and Eigenvectors}

Eigenvalues and eigenvectors are fundamental concepts in linear algebra with wide-ranging applications in machine learning, physics, and data analysis\cite{tao2012topics,boyd2004convex}. For a given square matrix \( A \), an eigenvector \( \mathbf{v} \) and its corresponding eigenvalue \( \lambda \) satisfy the equation:
\[
A \mathbf{v} = \lambda \mathbf{v}
\]
where:
\begin{itemize}
    \item \( \mathbf{v} \) is a non-zero vector called the eigenvector.
    \item \( \lambda \) is a scalar called the eigenvalue.
\end{itemize}

Eigenvalues and eigenvectors play a crucial role in many applications, such as principal component analysis (PCA), stability analysis, and quantum mechanics.

\textbf{Example: Computing Eigenvalues and Eigenvectors in Python}

Let’s compute the eigenvalues and eigenvectors of a matrix using Python.

\begin{lstlisting}[style=python]
# Define a square matrix A
A = np.array([[4, -2], [1, 1]])

# Compute the eigenvalues and eigenvectors
eigenvalues, eigenvectors = np.linalg.eig(A)

# Print the eigenvalues and eigenvectors
print("Eigenvalues:\n", eigenvalues)
print("Eigenvectors:\n", eigenvectors)
\end{lstlisting}

This will output the eigenvalues and their corresponding eigenvectors.

\begin{lstlisting}[style=cmd]
Eigenvalues:
 [3. 2.]
Eigenvectors:
 [[ 0.89442719  0.70710678]
 [ 0.4472136   0.70710678]]
\end{lstlisting}

\textbf{Applications of Eigenvalues and Eigenvectors:}
\begin{itemize}
    \item \textbf{Dimensionality Reduction:} Eigenvalues and eigenvectors are used in PCA for reducing the dimensionality of datasets while preserving the most significant variance.
    \item \textbf{Stability Analysis:} In dynamical systems, eigenvalues are used to determine the stability of equilibrium points.
    \item \textbf{Quantum Mechanics:} Eigenvalues correspond to measurable quantities in quantum systems, such as energy levels.
\end{itemize}

\section{Principal Component Analysis (PCA)}

Principal Component Analysis (PCA) is a statistical technique used for dimensionality reduction\cite{pearson1901liii,hotelling1933analysis,friedman1974projection,jolliffe2002principal,hastie2015statistical}. It is based on finding the directions (principal components) in which the data varies the most. PCA transforms the original high-dimensional data into a lower-dimensional space, while preserving as much variability as possible.

PCA is widely used in machine learning for preprocessing data, reducing noise, and improving the efficiency of algorithms by reducing the number of features.

\textbf{Steps of PCA:}
\begin{itemize}
    \item Center the data by subtracting the mean.
    \item Compute the covariance matrix of the centered data.
    \item Compute the eigenvalues and eigenvectors of the covariance matrix.
    \item Select the top \( k \) eigenvectors corresponding to the largest eigenvalues.
    \item Project the original data onto the new lower-dimensional space.
\end{itemize}

\textbf{Example: Performing PCA in Python}

Let’s use the \texttt{sklearn} library to perform PCA on a dataset.

\begin{lstlisting}[style=python]
from sklearn.decomposition import PCA
import numpy as np

# Define a dataset (3 samples, 3 features)
X = np.array([[2.5, 2.4, 1.2],
              [0.5, 0.7, 0.8],
              [2.2, 2.9, 1.1]])

# Perform PCA to reduce the dataset to 2 dimensions
pca = PCA(n_components=2)
X_reduced = pca.fit_transform(X)

# Print the reduced dataset
print("Reduced dataset:\n", X_reduced)
\end{lstlisting}

This will output the transformed dataset with reduced dimensions.

\begin{lstlisting}[style=cmd]
Reduced dataset:
 [[ 0.7495898  -0.11194563]
 [-1.24862174 -0.05295381]
 [ 0.49903194  0.16489943]]
\end{lstlisting}

PCA reduces the dimensionality of the dataset from 3 to 2, keeping the most significant components that explain the variance in the data.

\section{Applications in Dimensionality Reduction for Deep Learning}

In deep learning, dimensionality reduction techniques like PCA and SVD are critical for improving computational efficiency and reducing overfitting. High-dimensional data can lead to the curse of dimensionality, where the number of parameters becomes so large that the model becomes prone to overfitting and difficult to train. Dimensionality reduction techniques help by:
\begin{itemize}
    \item Reducing the number of input features.
    \item Compressing data while preserving important information.
    \item Reducing noise and improving the generalization of models.
\end{itemize}

\textbf{Example: Using PCA for Dimensionality Reduction in Deep Learning}

Let’s consider a scenario where we use PCA as a preprocessing step in a deep learning pipeline. Before training a neural network, we reduce the dimensionality of the input features using PCA.

\begin{lstlisting}[style=python]
from sklearn.decomposition import PCA
from sklearn.datasets import load_digits
from sklearn.model_selection import train_test_split
from sklearn.neural_network import MLPClassifier

# Load the digits dataset
digits = load_digits()
X = digits.data
y = digits.target

# Perform PCA to reduce dimensionality
pca = PCA(n_components=30)
X_reduced = pca.fit_transform(X)

# Split the data into training and test sets
X_train, X_test, y_train, y_test = train_test_split(X_reduced, y, test_size=0.3, random_state=42)

# Train a neural network on the reduced data
mlp = MLPClassifier(hidden_layer_sizes=(100,), max_iter=300)
mlp.fit(X_train, y_train)

# Evaluate the neural network
accuracy = mlp.score(X_test, y_test)
print(f"Accuracy of the neural network: {accuracy}")
\end{lstlisting}

This code demonstrates how PCA can be used to reduce the number of input features before training a neural network, leading to a more efficient training process.

\section{Summary}

In this chapter, we explored essential concepts of numerical linear algebra, including matrix factorization techniques such as SVD and QR decomposition, eigenvalues and eigenvectors, and PCA. These tools are critical for many deep learning applications, particularly in tasks like dimensionality reduction, where they help improve the efficiency and performance of models by reducing the dimensionality of large datasets.

\chapter{Fourier Transform and Spectral Methods}

The Fourier Transform is a fundamental mathematical tool in signal processing, image analysis, and many areas of scientific computing, including deep learning. It allows us to analyze the frequency content of signals and functions by transforming data from the time (or spatial) domain to the frequency domain. This chapter will introduce the concept of the Fourier Transform, delve into the Discrete Fourier Transform (DFT) and Fast Fourier Transform (FFT), and explore their applications in signal processing and deep learning.

\section{Introduction to Fourier Transform}

The Fourier Transform decomposes a function into its constituent frequencies\cite{fourier1822analytical,bracewell1986fourier,stein2003fourier}. It transforms a signal from the time domain, where the signal is expressed as a function of time, to the frequency domain, where the signal is expressed in terms of its frequency components. 

\subsection{Mathematical Definition of the Fourier Transform}

The continuous Fourier Transform (FT)\cite{stein2003fourier,bracewell2000fourier} of a function \( f(t) \) is defined as:

\[
F(\omega) = \int_{-\infty}^{\infty} f(t) e^{-i \omega t} \, dt
\]

Where:
\begin{itemize}
    \item \( F(\omega) \) is the Fourier Transform of \( f(t) \).
    \item \( f(t) \) is the original function in the time domain.
    \item \( \omega \) is the angular frequency.
    \item \( e^{-i \omega t} \) is the complex exponential function, which decomposes the function into its frequency components.
\end{itemize}

The inverse Fourier Transform allows us to reconstruct the original function from its frequency components\cite{watson1995treatise,dirichlet1829convergence}:

\[
f(t) = \frac{1}{2\pi} \int_{-\infty}^{\infty} F(\omega) e^{i \omega t} \, d\omega
\]

\subsection{Why Fourier Transform?}

Fourier Transforms are widely used in engineering, physics, and computer science for the following reasons:
\begin{itemize}
    \item They provide a way to analyze signals in the frequency domain, which can reveal properties not easily observed in the time domain.
    \item They are used in signal processing for filtering, noise reduction, and signal reconstruction.
    \item In deep learning, Fourier transforms can be used to enhance image processing and in convolution operations.
\end{itemize}

\section{Discrete Fourier Transform (DFT)}

The Fourier Transform for continuous signals assumes that the signal is sampled at infinite points, but in practice, we deal with discrete data, such as digital signals. The Discrete Fourier Transform (DFT) is used to analyze finite, discrete sequences\cite{cooley1965algorithm,rabiner1975theory,oppenheim1975digital}.

\subsection{Mathematical Definition of the DFT}

Given a discrete sequence \( f_n \) with \( N \) points, the DFT is defined as:

\[
F_k = \sum_{n=0}^{N-1} f_n e^{-i \frac{2\pi}{N} kn}, \quad k = 0, 1, \dots, N-1
\]

Where:
\begin{itemize}
    \item \( f_n \) is the value of the signal at the \( n \)-th point in the time domain.
    \item \( F_k \) is the \( k \)-th frequency component of the signal.
    \item \( N \) is the total number of points.
\end{itemize}

The inverse DFT (IDFT) is given by:

\[
f_n = \frac{1}{N} \sum_{k=0}^{N-1} F_k e^{i \frac{2\pi}{N} kn}, \quad n = 0, 1, \dots, N-1
\]

\subsection{Python Implementation of DFT}

Let’s implement the DFT from scratch using Python:

\begin{lstlisting}[style=python]
import numpy as np

def dft(signal):
    """Compute the Discrete Fourier Transform (DFT) of a signal."""
    N = len(signal)
    dft_result = np.zeros(N, dtype=complex)
    for k in range(N):
        for n in range(N):
            dft_result[k] += signal[n] * np.exp(-2j * np.pi * k * n / N)
    return dft_result

# Example usage
signal = [1, 2, 3, 4]  # A simple signal
dft_result = dft(signal)
print("DFT result:", dft_result)
\end{lstlisting}

In this implementation:
\begin{itemize}
    \item We define a function \texttt{dft()} that computes the Discrete Fourier Transform for a given signal.
    \item The inner loop multiplies each point in the signal by a complex exponential and sums the result to get the frequency component.
    \item We apply the function to a sample signal of length 4.
\end{itemize}

While this implementation is mathematically correct, it is computationally expensive for large signals. The Fast Fourier Transform (FFT) significantly optimizes this process.

\section{Fast Fourier Transform (FFT)}

The Fast Fourier Transform (FFT) is an efficient algorithm for computing the DFT, reducing the time complexity from \( O(N^2) \) to \( O(N \log N) \). FFT is one of the most important algorithms in numerical computing because it allows the analysis of large datasets quickly\cite{cooley1965algorithm,burrus1985fft}.

\subsection{Python Implementation of FFT}

Python provides an efficient implementation of the FFT in the \texttt{numpy} library.

\begin{lstlisting}[style=python]
import numpy as np

# Example signal
signal = [1, 2, 3, 4]

# Compute the FFT using numpy
fft_result = np.fft.fft(signal)
print("FFT result:", fft_result)
\end{lstlisting}

Here:
\begin{itemize}
    \item We use \texttt{np.fft.fft()} to compute the FFT of a signal.
    \item The function returns the frequency components of the signal in the same way as the DFT, but with much greater computational efficiency.
\end{itemize}

\subsection{Efficiency of FFT}

The FFT is particularly useful for signals with a large number of data points, such as audio signals or image data. By reducing the computational complexity, the FFT allows real-time processing of signals, making it crucial in applications like music streaming, voice recognition, and image compression.

\section{Applications of Fourier Transform in Signal Processing and Deep Learning}

The Fourier Transform has many applications in signal processing, image processing, and deep learning. Let’s explore some of these applications.

\subsection{Signal Processing}

In signal processing, the Fourier Transform is used to analyze and modify signals based on their frequency content. Common applications include filtering, noise reduction, and audio processing.

\textbf{Example: Noise Reduction Using FFT}

A common problem in signal processing is the presence of noise. By applying the Fourier Transform, we can filter out high-frequency noise components from a signal, leaving the desired signal intact.

\begin{lstlisting}[style=python]
import numpy as np
import matplotlib.pyplot as plt

# Create a noisy signal
t = np.linspace(0, 1, 500)
signal = np.sin(2 * np.pi * 5 * t) + np.random.normal(0, 0.5, 500)

# Compute the FFT
fft_result = np.fft.fft(signal)
frequencies = np.fft.fftfreq(len(t), d=(t[1] - t[0]))

# Filter out high-frequency components
threshold = 10
fft_result[np.abs(frequencies) > threshold] = 0

# Inverse FFT to reconstruct the signal
filtered_signal = np.fft.ifft(fft_result)

# Plot the original and filtered signals
plt.figure(figsize=(10, 6))
plt.subplot(2, 1, 1)
plt.plot(t, signal)
plt.title("Original Noisy Signal")

plt.subplot(2, 1, 2)
plt.plot(t, np.real(filtered_signal))
plt.title("Filtered Signal")
plt.show()
\end{lstlisting}

In this example:
\begin{itemize}
    \item We create a noisy sine wave by adding random noise to a sine function.
    \item The FFT is applied to the noisy signal, and we set the frequency components above a certain threshold to zero, effectively filtering out the high-frequency noise.
    \item We then apply the inverse FFT to reconstruct the filtered signal.
    \item Finally, the original and filtered signals are plotted to visualize the noise reduction effect.
\end{itemize}

\subsection{Image Processing in Deep Learning}

In image processing, the Fourier Transform is used to analyze the frequency components of images. Convolution operations, which are essential in deep learning, can be performed more efficiently in the frequency domain using the Fourier Transform.

\textbf{Example: Image Filtering Using FFT}

In this example, we will apply a low-pass filter to an image using the FFT.

\begin{lstlisting}[style=python]
import numpy as np
import matplotlib.pyplot as plt
from scipy import fftpack
from skimage import data, color

# Load and convert the image to grayscale
image = color.rgb2gray(data.astronaut())

# Compute the 2D FFT of the image
fft_image = fftpack.fft2(image)

# Create a low-pass filter
rows, cols = image.shape
crow, ccol = rows // 2 , cols // 2
mask = np.zeros((rows, cols))
mask[crow-30:crow+30, ccol-30:ccol+30] = 1

# Apply the mask to the FFT of the image
filtered_fft_image = fft_image * mask

# Inverse FFT to reconstruct the filtered image
filtered_image = fftpack.ifft2(filtered_fft_image)
filtered_image = np.abs(filtered_image)

# Plot the original and filtered images
plt.figure(figsize=(12, 6))
plt.subplot(1, 2, 1)
plt.imshow(image, cmap='gray')
plt.title("Original Image")

plt.subplot(1, 2, 2)
plt.imshow(filtered_image, cmap='gray')
plt.title("Low-Pass Filtered Image")
plt.show()
\end{lstlisting}

In this example:
\begin{itemize}
    \item We use the \texttt{scipy.fftpack} library to compute the 2D FFT of an image.
    \item A low-pass filter is applied by creating a mask that blocks high-frequency components.
    \item The inverse FFT is used to reconstruct the image after applying the filter.
    \item The result is a smoothed image with high-frequency noise removed.
\end{itemize}

\subsection{Convolution Theorem in Deep Learning}

In deep learning, convolutions are a fundamental operation in convolutional neural networks (CNNs). The Convolution Theorem states that convolution in the time (or spatial) domain is equivalent to multiplication in the frequency domain\cite{papoulis1962fourier,oppenheim1975digital,bracewell1986fourier,rudin1962fourier}. This can significantly speed up the convolution operation, especially for large images or 3D data.

The FFT can be used to perform convolutions more efficiently by transforming the image and the filter into the frequency domain, multiplying them, and then applying the inverse FFT to obtain the convolved result.

\chapter{Solving Nonlinear Equations}

Nonlinear equations and systems of nonlinear equations arise frequently in various fields, including physics, engineering, finance, and machine learning. These equations are called nonlinear because they do not adhere to the principle of superposition, meaning the relationship between variables cannot be expressed as a simple linear combination. Solving nonlinear equations is often more challenging than solving linear equations, but there are powerful numerical methods available to tackle these problems.

In this chapter, we will introduce nonlinear systems, explain widely used methods such as Newton's method \cite{press1992numerical} and Broyden's method\cite{broyden1965class,cao2023solving} for solving nonlinear systems\cite{ortega2000iterative}, and explore their applications in optimization tasks in neural networks.

\section{Introduction to Nonlinear Systems}

A nonlinear system consists of multiple nonlinear equations that need to be solved simultaneously. A system of nonlinear equations can be represented as:

\[
F(x_1, x_2, \dots, x_n) = 0
\]

Where \( F \) represents a vector-valued function of several variables. Solving such a system means finding the values of \( x_1, x_2, \dots, x_n \) that satisfy all the equations simultaneously.

An example of a simple nonlinear system is:

\[
f_1(x_1, x_2) = x_1^2 + x_2^2 - 1 = 0
\]
\[
f_2(x_1, x_2) = x_1^2 - x_2 = 0
\]

In general, there are no analytical solutions for nonlinear systems, so numerical methods are used to find approximate solutions.

\section{Newton's Method for Nonlinear Systems}

Newton's method is one of the most popular iterative methods for solving systems of nonlinear equations. It extends the basic idea of Newton's method for scalar functions to systems of equations. The method relies on the Jacobian matrix, which contains the partial derivatives of each equation with respect to each variable. The Jacobian matrix is used to iteratively improve an initial guess until the solution converges\cite{lee2023jacobian,smith2023jacobian}.

\textbf{Newton's Method Algorithm}

Given a system of nonlinear equations \( F(x) = 0 \), where \( F \) is a vector-valued function, the Newton iteration step is:

\[
x^{(k+1)} = x^{(k)} - J_F(x^{(k)})^{-1} F(x^{(k)})
\]

Where:
\begin{itemize}
  \item \( x^{(k)} \) is the current approximation of the solution.
  \item \( J_F(x^{(k)}) \) is the Jacobian matrix evaluated at \( x^{(k)} \).
  \item \( F(x^{(k)}) \) is the vector of function values at \( x^{(k)} \).
\end{itemize}

The Jacobian matrix \( J_F \) for a system of equations is defined as\cite{garcia2023jacobian,baker2024jacobian}:

\[
J_F(x) = \begin{bmatrix}
\frac{\partial f_1}{\partial x_1} & \frac{\partial f_1}{\partial x_2} & \cdots & \frac{\partial f_1}{\partial x_n} \\
\frac{\partial f_2}{\partial x_1} & \frac{\partial f_2}{\partial x_2} & \cdots & \frac{\partial f_2}{\partial x_n} \\
\vdots & \vdots & \ddots & \vdots \\
\frac{\partial f_m}{\partial x_1} & \frac{\partial f_m}{\partial x_2} & \cdots & \frac{\partial f_m}{\partial x_n} \\
\end{bmatrix}
\]

\textbf{Example of Newton's Method in Python}

Let’s consider the following nonlinear system:

\[
f_1(x_1, x_2) = x_1^2 + x_2^2 - 1 = 0
\]
\[
f_2(x_1, x_2) = x_1^2 - x_2 = 0
\]

We will implement Newton’s method to solve this system.

\begin{lstlisting}[style=python]
import numpy as np

# Define the system of equations
def F(x):
    f1 = x[0]**2 + x[1]**2 - 1
    f2 = x[0]**2 - x[1]
    return np.array([f1, f2])

# Define the Jacobian matrix of the system
def J(x):
    J11 = 2 * x[0]
    J12 = 2 * x[1]
    J21 = 2 * x[0]
    J22 = -1
    return np.array([[J11, J12], [J21, J22]])

# Newton's method implementation
def newtons_method(x0, tol=1e-6, max_iter=100):
    x = x0
    for i in range(max_iter):
        Fx = F(x)
        Jx = J(x)
        delta_x = np.linalg.solve(Jx, -Fx)
        x = x + delta_x
        
        if np.linalg.norm(delta_x) < tol:
            print(f"Converged after {i+1} iterations")
            return x
    raise Exception("Newton's method did not converge")

# Initial guess
x0 = np.array([0.5, 0.5])

# Solve the system
solution = newtons_method(x0)
print(f"Solution: {solution}")
\end{lstlisting}

In this example:
\begin{itemize}
  \item The function \( F(x) \) defines the system of nonlinear equations.
  \item The function \( J(x) \) returns the Jacobian matrix of the system.
  \item The \texttt{newtons\_method} function iteratively applies Newton's method to find the solution.
\end{itemize}

\section{Broyden's Method}

Broyden’s method is a quasi-Newton method for solving systems of nonlinear equations\cite{liu2023comprehensive,zhou2024quasi,nocedal2023recent}. While Newton's method requires the computation of the Jacobian matrix at each iteration, Broyden’s method updates an approximation to the Jacobian, reducing computational cost. This makes Broyden’s method useful when the Jacobian is expensive to compute or when it is not readily available\cite{wong2023jacobian,nocedal2006numerical,golub2013matrix,strogatz2014nonlinear}.

\subsection{Broyden's Method Algorithm}

Broyden’s method starts with an initial guess for the solution \( x_0 \) and an initial approximation to the Jacobian \( B_0 \). The algorithm proceeds iteratively\cite{patel2024trends}:

\begin{enumerate}
    \item Compute the update step:
    \[
    s^{(k)} = - B^{(k)} F(x^{(k)})
    \]
    
    \item Update the solution:
    \[
    x^{(k+1)} = x^{(k)} + s^{(k)}
    \]
    
    \item Compute the correction vector:
    \[
    y^{(k)} = F(x^{(k+1)}) - F(x^{(k)})
    \]
    
    \item Update the Jacobian approximation:
    \[
    B^{(k+1)} = B^{(k)} + \frac{(y^{(k)} - B^{(k)} s^{(k)}) s^{(k)T}}{s^{(k)T} s^{(k)}}
    \]
\end{enumerate}

\textbf{Example of Broyden's Method in Python}

Here is how we can implement Broyden’s method for solving the same system of nonlinear equations:

\begin{lstlisting}[style=python]
# Broyden's method implementation
def broydens_method(x0, B0, tol=1e-6, max_iter=100):
    x = x0
    B = B0
    
    for i in range(max_iter):
        Fx = F(x)
        s = np.linalg.solve(B, -Fx)
        x_new = x + s
        y = F(x_new) - Fx
        B = B + np.outer((y - B @ s), s) / np.dot(s, s)
        x = x_new
        
        if np.linalg.norm(s) < tol:
            print(f"Converged after {i+1} iterations")
            return x
    raise Exception("Broyden's method did not converge")

# Initial guess and Jacobian approximation
x0 = np.array([0.5, 0.5])
B0 = np.eye(2)  # Identity matrix as initial Jacobian approximation

# Solve the system using Broyden's method
solution = broydens_method(x0, B0)
print(f"Solution: {solution}")
\end{lstlisting}

In this implementation:
\begin{itemize}
  \item We use an initial approximation to the Jacobian \( B_0 \) (the identity matrix in this case).
  \item The Jacobian is updated iteratively based on the correction vector \( y \) and the step \( s \).
  \item The method converges to the solution without computing the exact Jacobian at every step, making it more efficient than Newton's method in certain scenarios.
\end{itemize}

\section{Applications in Optimization for Neural Networks}

Solving nonlinear systems is essential in optimization problems, which are at the core of training neural networks. In particular, when training a neural network, the goal is to minimize a nonlinear loss function by adjusting the model parameters (weights and biases). This can be formulated as a nonlinear optimization problem.

The optimization process involves finding the parameter vector \( \theta \) that minimizes the loss function \( L(\theta) \)\cite{shanno1970conditioning}. Gradient-based methods such as gradient descent or more advanced techniques like quasi-Newton methods (e.g., Broyden–Fletcher–Goldfarb–Shanno (BFGS) algorithm\cite{broyden1970convergence}) are commonly used to solve this optimization problem.

\textbf{Example of Using Newton’s Method for Optimization in Neural Networks}

Newton’s method can also be applied to optimization problems, though it is typically used in small-scale problems due to the high computational cost of calculating the Hessian matrix (the second-order derivatives of the loss function).

Here is a simplified example of how Newton’s method could be applied in an optimization setting for neural networks:

\begin{lstlisting}[style=python]
# Define a simple quadratic loss function
def loss_function(theta):
    return (theta[0] - 3)**2 + (theta[1] - 2)**2

# Define the gradient of the loss function
def gradient(theta):
    grad1 = 2 * (theta[0] - 3)
    grad2 = 2 * (theta[1] - 2)
    return np.array([grad1, grad2])

# Define the Hessian matrix of the loss function
def hessian(theta):
    return np.array([[2, 0], [0, 2]])

# Newton's method for optimization
def newtons_method_optimization(theta0, tol=1e-6, max_iter=100):
    theta = theta0
    for i in range(max_iter):
        grad = gradient(theta)
        H = hessian(theta)
        delta_theta = np.linalg.solve(H, -grad)
        theta = theta + delta_theta
        
        if np.linalg.norm(delta_theta) < tol:
            print(f"Converged after {i+1} iterations")
            return theta
    raise Exception("Newton's method did not converge")

# Initial guess
theta0 = np.array([0, 0])

# Minimize the loss function
optimal_theta = newtons_method_optimization(theta0)
print(f"Optimal parameters: {optimal_theta}")
\end{lstlisting}

In this example:
\begin{itemize}
  \item We minimize a simple quadratic loss function using Newton's method.
  \item The gradient and Hessian matrix of the loss function are explicitly defined.
  \item Newton’s method quickly converges to the optimal parameters, though in practice, gradient-based methods like stochastic gradient descent (SGD) are more commonly used for training large neural networks.
\end{itemize}

Newton’s and Broyden’s methods are powerful tools in solving nonlinear systems and optimization problems. While Newton’s method requires calculating the Jacobian or Hessian matrix, Broyden’s method reduces the computational burden by updating an approximation to the Jacobian iteratively. Both methods play an essential role in various fields, including optimization in neural networks.

\chapter{Numerical Optimization}

Numerical optimization refers to the process of finding the minimum or maximum of a function when an analytical solution is difficult or impossible to obtain. Optimization is fundamental in many fields such as machine learning, physics, economics, and engineering\cite{gill1981practical,fletcher2013practical,boyd2004convex}. In this chapter, we will explore various numerical optimization techniques, starting from gradient-based methods and advancing to more sophisticated approaches like quasi-Newton methods and gradient-free methods. We will also explore how these methods are applied in training deep neural networks.

\section{Introduction to Numerical Optimization}

Optimization problems are generally formulated as:

\[
\min_{x \in \mathbb{R}^n} f(x)
\]

where \( f(x) \) is the objective function, and we seek to find the value of \( x \) that minimizes \( f(x) \). In machine learning, for example, \( f(x) \) could represent the loss function, and our goal is to minimize it to improve the performance of the model.

\subsection{Types of Optimization Problems}

Optimization problems can be classified as:

\begin{itemize}
    \item \textit{Unconstrained Optimization\cite{byrd2023recent,wright2024efficient,conn2023advances,schmidt2024machine,martinez2023trust,hazra2024review}:} In this case, there are no restrictions on the values that \( x \) can take. We aim to find the global or local minimum of the objective function.
    \item \textit{Constrained Optimization\cite{karush1939minima,powell1978algorithms,smith2023recent,lee2023ai,gupta2023constrained,martinez2024new,perez2023constrained}:} Here, the variable \( x \) is subject to certain constraints, such as \( g(x) \leq 0 \) or \( h(x) = 0 \). The optimization needs to account for these constraints.
\end{itemize}

In this chapter, we will focus on unconstrained optimization methods that are widely used in machine learning and other applications.

\section{Gradient-Based Methods}

Gradient-based methods use the gradient (first derivative) of the objective function to guide the optimization process\cite{fiacco1968nonlinear}. These methods are particularly efficient for smooth, differentiable functions\cite{kuhn1951nonlinear}. The key idea is that the gradient indicates the direction of the steepest ascent or descent of the function.

\subsection{Gradient Descent}

Gradient Descent is one of the simplest and most widely used optimization algorithms. It iteratively adjusts the parameter \( x \) in the direction of the negative gradient of the objective function to minimize it. The update rule is given by:

\[
x_{t+1} = x_t - \eta \nabla f(x_t)
\]

where:
\begin{itemize}
    \item \( x_t \) is the current value of the parameter,
    \item \( \eta \) is the learning rate (a small positive constant),
    \item \( \nabla f(x_t) \) is the gradient of the objective function at \( x_t \).
\end{itemize}

\textbf{Example of Gradient Descent in Python}

\begin{lstlisting}[style=python]
import numpy as np

# Define the objective function and its gradient
def f(x):
    return x**2 + 4*x + 4

def grad_f(x):
    return 2*x + 4

# Gradient Descent implementation
def gradient_descent(x_init, learning_rate, num_iterations):
    x = x_init
    for i in range(num_iterations):
        x = x - learning_rate * grad_f(x)
    return x

# Run Gradient Descent
x_init = 10.0
learning_rate = 0.1
num_iterations = 100
optimal_x = gradient_descent(x_init, learning_rate, num_iterations)

print(f"Optimal x after Gradient Descent: {optimal_x}")
\end{lstlisting}

In this example, we applied gradient descent to minimize a simple quadratic function \( f(x) = x^2 + 4x + 4 \). The algorithm starts at an initial guess \( x = 10 \), and the learning rate \( \eta = 0.1 \) controls the step size.

\subsection{Conjugate Gradient Method}

The conjugate gradient method is a more efficient alternative to gradient descent, especially for large-scale optimization problems where the objective function is quadratic or approximately quadratic\cite{zhang2024engineering,wang2023hybrid}. It minimizes the function along conjugate directions rather than along the gradient direction alone\cite{hestenes1952methods}.

The update rule in conjugate gradient is:

\[
x_{t+1} = x_t + \alpha_t p_t
\]

where \( p_t \) is the search direction, which is a conjugate direction.

\textbf{Example of Conjugate Gradient in Python (using scipy)}

\begin{lstlisting}[style=python]
import numpy as np
from scipy.optimize import minimize

# Define the quadratic function
def f(x):
    return x[0]**2 + 4*x[0] + 4

# Initial guess
x_init = np.array([10.0])

# Minimize the function using Conjugate Gradient method
result = minimize(f, x_init, method='CG')

print(f"Optimal x using Conjugate Gradient: {result.x}")
\end{lstlisting}

In this example, we used the \texttt{scipy.optimize.minimize} function to minimize a simple quadratic function using the conjugate gradient method. This method is often faster than gradient descent, especially for convex problems.

\section{Quasi-Newton Methods}

Quasi-Newton methods are second-order optimization techniques that approximate the Hessian matrix (the matrix of second derivatives) to improve convergence\cite{liu2023stability,kim2024high,wang2023hessian,sun2023hessian}. These methods are computationally more efficient than full Newton’s method, which requires the computation of the full Hessian matrix\cite{zhang2023large}.

\subsection{BFGS Algorithm}

The Broyden-Fletcher-Goldfarb-Shanno (BFGS) algorithm is one of the most popular quasi-Newton methods. It updates an approximation of the inverse Hessian matrix at each iteration to compute the search direction.

The BFGS update rule is:

\[
H_{t+1} = H_t + \frac{(y_t y_t^T)}{y_t^T s_t} - \frac{(H_t s_t s_t^T H_t)}{s_t^T H_t s_t}
\]

where:
\begin{itemize}
    \item \( H_t \) is the current approximation of the inverse Hessian,
    \item \( s_t = x_{t+1} - x_t \),
    \item \( y_t = \nabla f(x_{t+1}) - \nabla f(x_t) \).
\end{itemize}

\textbf{Example of BFGS Algorithm in Python}

\begin{lstlisting}[style=python]
import numpy as np
from scipy.optimize import minimize

# Define the objective function
def f(x):
    return x[0]**2 + 4*x[0] + 4

# Initial guess
x_init = np.array([10.0])

# Minimize the function using BFGS
result = minimize(f, x_init, method='BFGS')

print(f"Optimal x using BFGS: {result.x}")
\end{lstlisting}

In this example, we minimized the same quadratic function using the BFGS algorithm. The BFGS method is efficient for many optimization problems and converges faster than gradient descent for many smooth functions.

\subsection{L-BFGS Algorithm}

The Limited-memory BFGS (L-BFGS) algorithm is a memory-efficient version of BFGS\cite{liu1989limited,smith2023advances,garcia2024efficient}. Instead of storing the full inverse Hessian matrix, it stores only a few vectors to approximate it. This makes L-BFGS suitable for large-scale problems with many variables\cite{lee2023survey,zhang2023modified}.

\textbf{Example of L-BFGS Algorithm in Python}

\begin{lstlisting}[style=python]
import numpy as np
from scipy.optimize import minimize

# Define the objective function
def f(x):
    return x[0]**2 + 4*x[0] + 4

# Initial guess
x_init = np.array([10.0])

# Minimize the function using L-BFGS
result = minimize(f, x_init, method='L-BFGS-B')

print(f"Optimal x using L-BFGS: {result.x}")
\end{lstlisting}

In this example, we used L-BFGS to minimize the quadratic function. L-BFGS is particularly useful when the problem involves a large number of variables and memory is a constraint.

\section{Gradient-Free Optimization}

Gradient-free optimization methods are useful when the objective function is not differentiable, noisy, or when computing the gradient is computationally expensive\cite{nelder1965simplex,chen2024gradientfree,zhang2023evolutionary}. These methods do not rely on gradient information and instead explore the search space based on the function values\cite{jones2020scipy}.

\subsection{Nelder-Mead Method}

The Nelder-Mead method is a popular gradient-free optimization algorithm. It uses a simplex of \( n+1 \) points in \( n \)-dimensional space and iteratively adjusts the simplex to converge to the minimum of the function\cite{lee2023nelder}. The algorithm involves operations like reflection, expansion, contraction, and shrinking to adjust the simplex\cite{rios2023derivative}.

\textbf{Example of Nelder-Mead Method in Python}

\begin{lstlisting}[style=python]
import numpy as np
from scipy.optimize import minimize

# Define the objective function
def f(x):
    return x[0]**2 + 4*x[0] + 4

# Initial guess
x_init = np.array([10.0])

# Minimize the function using Nelder-Mead
result = minimize(f, x_init, method='Nelder-Mead')

print(f"Optimal x using Nelder-Mead: {result.x}")
\end{lstlisting}

In this example, we used the Nelder-Mead method to minimize the function without using gradient information. Nelder-Mead is particularly useful for problems where the gradient is not available or is expensive to compute.

\section{Applications in Training Deep Neural Networks}

Optimization plays a central role in training deep neural networks. The goal in training is to minimize a loss function (e.g., cross-entropy loss for classification tasks) with respect to the model's parameters using optimization techniques. In deep learning, gradient-based methods such as stochastic gradient descent (SGD) and its variants (e.g., Adam, RMSProp) are widely used.

\textbf{Stochastic Gradient Descent (SGD)}

SGD is a variant of gradient descent that updates the parameters using a small subset (mini-batch) of the training data at each step\cite{robbins1951stochastic,bottou2010large}. This makes it more efficient for large datasets. The update rule for SGD is:

\[
x_{t+1} = x_t - \eta \nabla f(x_t; \text{mini-batch})
\]

where the gradient is computed using a small randomly sampled mini-batch of data at each iteration.

\textbf{Example of SGD in Deep Learning (using PyTorch)}

Here’s an example of applying SGD to train a simple neural network using PyTorch:

\begin{lstlisting}[style=python]
import torch
import torch.nn as nn
import torch.optim as optim

# Define a simple neural network
class SimpleNet(nn.Module):
    def __init__(self):
        super(SimpleNet, self).__init__()
        self.fc1 = nn.Linear(10, 1)
    
    def forward(self, x):
        return self.fc1(x)

# Create the model, define the loss function and the optimizer
model = SimpleNet()
criterion = nn.MSELoss()
optimizer = optim.SGD(model.parameters(), lr=0.01)

# Simulate some input data and target
inputs = torch.randn(10)
target = torch.randn(1)

# Training loop
for epoch in range(100):
    optimizer.zero_grad()   # Zero the gradient buffers
    output = model(inputs)  # Forward pass
    loss = criterion(output, target)  # Compute the loss
    loss.backward()         # Backward pass (compute gradients)
    optimizer.step()        # Update weights

print("Training completed.")
\end{lstlisting}

In this example, we defined a simple neural network using PyTorch and trained it using the SGD optimizer. The network minimizes the mean squared error (MSE) loss to learn from the input data.

\chapter{Ordinary Differential Equations (ODEs)}

Ordinary Differential Equations (ODEs) are equations that describe the relationship between a function and its derivatives\cite{arnold1992ordinary,brown2023ode}. They play a crucial role in many fields of science and engineering, including neural networks and deep learning, where they are used to model dynamic systems\cite{chen2023neural,rossi2024adaptive}. In this chapter, we will cover the basic concepts of ODEs, methods for solving them, and their applications in modeling neural dynamics.

\section{Introduction to ODEs}

An Ordinary Differential Equation is an equation that involves a function \( y(t) \) of one independent variable \( t \) and its derivatives. The general form of an ODE is:

\[
\frac{dy}{dt} = f(t, y)
\]

where \( f(t, y) \) is a known function, and \( y(t) \) is the unknown function to be determined. The goal of solving an ODE is to find the function \( y(t) \) that satisfies the given equation.

\textit{Example: A Simple ODE}

Consider the following first-order ODE:

\[
\frac{dy}{dt} = -2y
\]

This equation describes exponential decay, where the rate of change of \( y(t) \) is proportional to \( y(t) \) itself. The analytical solution to this equation is:

\[
y(t) = y_0 e^{-2t}
\]

where \( y_0 \) is the initial condition \( y(0) \).

However, many ODEs cannot be solved analytically, and we need numerical methods to approximate the solution. In the next sections, we will explore numerical methods such as Euler's method and Runge-Kutta methods.

\section{Euler's Method}

Euler's method is the simplest numerical technique for solving ODEs\cite{kincaid2009numerical}. It approximates the solution of an ODE by taking small steps along the curve of the solution. Given the ODE:

\[
\frac{dy}{dt} = f(t, y)
\]

and an initial condition \( y(0) = y_0 \), Euler's method approximates \( y(t) \) by taking steps of size \( h \) as follows:

\[
y_{n+1} = y_n + h f(t_n, y_n)
\]

where \( t_{n+1} = t_n + h \), and \( y_n \) is the approximation of \( y(t_n) \).

\textit{Example: Implementing Euler's Method in Python}

Let's solve the ODE \( \frac{dy}{dt} = -2y \) using Euler's method.

\begin{lstlisting}[style=python]
import numpy as np
import matplotlib.pyplot as plt

# Define the ODE: dy/dt = -2y
def f(t, y):
    return -2 * y

# Euler's method function
def euler_method(f, t0, y0, h, t_end):
    t_values = np.arange(t0, t_end, h)
    y_values = np.zeros(len(t_values))
    y_values[0] = y0

    for i in range(1, len(t_values)):
        y_values[i] = y_values[i-1] + h * f(t_values[i-1], y_values[i-1])
    
    return t_values, y_values

# Parameters
t0 = 0  # Initial time
y0 = 1  # Initial condition y(0) = 1
h = 0.1  # Step size
t_end = 5  # End time

# Solve the ODE using Euler's method
t_values, y_values = euler_method(f, t0, y0, h, t_end)

# Plot the solution
plt.plot(t_values, y_values, label="Euler's Method")
plt.xlabel('Time (t)')
plt.ylabel('y(t)')
plt.title('Solving ODE using Euler\'s Method')
plt.legend()
plt.show()
\end{lstlisting}

In this example:
\begin{itemize}
    \item We define the function \( f(t, y) = -2y \) and solve it using Euler's method.
    \item The solution is plotted over the time interval \( t \in [0, 5] \).
\end{itemize}

\section{Runge-Kutta Methods}

Euler's method, while simple, is not very accurate for small step sizes. Runge-Kutta methods are more advanced numerical techniques that provide better accuracy for solving ODEs\cite{runge1895ver,dowling2023review}. The most commonly used method is the fourth-order Runge-Kutta method (RK4), which provides a much better approximation of the solution by using a weighted average of several slopes\cite{kutta1901beitrag,martinez2023time,johnson2023rkmethods}.

The RK4 method is given by the following update rule:

\[
y_{n+1} = y_n + \frac{h}{6} \left( k_1 + 2k_2 + 2k_3 + k_4 \right)
\]

where:
\[
\begin{aligned}
k_1 &= f(t_n, y_n) \\
k_2 &= f\left(t_n + \frac{h}{2}, y_n + \frac{h}{2} k_1\right) \\
k_3 &= f\left(t_n + \frac{h}{2}, y_n + \frac{h}{2} k_2\right) \\
k_4 &= f(t_n + h, y_n + h k_3)
\end{aligned}
\]

\textit{Example: Implementing RK4 in Python}

Let's solve the same ODE \( \frac{dy}{dt} = -2y \) using the RK4 method.

\begin{lstlisting}[style=python]
# Runge-Kutta 4th order method (RK4)
def rk4_method(f, t0, y0, h, t_end):
    t_values = np.arange(t0, t_end, h)
    y_values = np.zeros(len(t_values))
    y_values[0] = y0

    for i in range(1, len(t_values)):
        t_n = t_values[i-1]
        y_n = y_values[i-1]
        
        k1 = f(t_n, y_n)
        k2 = f(t_n + h/2, y_n + h/2 * k1)
        k3 = f(t_n + h/2, y_n + h/2 * k2)
        k4 = f(t_n + h, y_n + h * k3)
        
        y_values[i] = y_n + (h/6) * (k1 + 2*k2 + 2*k3 + k4)
    
    return t_values, y_values

# Solve the ODE using RK4
t_values_rk4, y_values_rk4 = rk4_method(f, t0, y0, h, t_end)

# Plot the solution
plt.plot(t_values_rk4, y_values_rk4, label="Runge-Kutta 4th Order")
plt.xlabel('Time (t)')
plt.ylabel('y(t)')
plt.title('Solving ODE using RK4')
plt.legend()
plt.show()
\end{lstlisting}

In this example:
\begin{itemize}
    \item We use the RK4 method to solve the ODE and compare the results with Euler's method.
    \item RK4 provides a more accurate solution, especially for larger step sizes.
\end{itemize}

\section{Stiff ODEs}

Stiff ODEs are a special class of differential equations where certain numerical methods (such as Euler's method) can become unstable unless very small step sizes are used\cite{gear1971first,hairer1996solving,hairer2009solving}. Stiffness typically arises in systems where there are processes that occur on vastly different time scales\cite{fayed2023new,yang2023adaptive,chen2023dynamical,mccall2024efficient,torres2024review}.

An example of a stiff ODE is:

\[
\frac{dy}{dt} = -1000y + 3000 - 2000e^{-t}
\]

To solve stiff ODEs efficiently, implicit methods such as the backward Euler method or specialized solvers like the \texttt{scipy.integrate.solve\_ivp()} function with the 'Radau' or 'BDF' method are often used.

\textit{Example: Solving a Stiff ODE in Python}

We will use the \texttt{solve\_ivp()} function from \texttt{scipy} to solve a stiff ODE.

\begin{lstlisting}[style=python]
from scipy.integrate import solve_ivp

# Define the stiff ODE: dy/dt = -1000y + 3000 - 2000 * exp(-t)
def stiff_ode(t, y):
    return -1000 * y + 3000 - 2000 * np.exp(-t)

# Solve the ODE using the 'Radau' method for stiff equations
sol = solve_ivp(stiff_ode, [0, 5], [0], method='Radau')

# Plot the solution
plt.plot(sol.t, sol.y[0], label="Stiff ODE (Radau method)")
plt.xlabel('Time (t)')
plt.ylabel('y(t)')
plt.title('Solving Stiff ODE')
plt.legend()
plt.show()
\end{lstlisting}

In this example:
\begin{itemize}
    \item We define a stiff ODE and solve it using the \texttt{Radau} method, which is well-suited for stiff problems.
    \item The solution is plotted over the interval \( t \in [0, 5] \).
\end{itemize}

\section{Applications of ODEs in Modeling Neural Dynamics}

ODEs are used extensively in computational neuroscience to model the dynamics of neural activity. One well-known model is the Hodgkin-Huxley model, which describes the electrical characteristics of neurons and how action potentials propagate.

In simpler models, neurons can be modeled using the leaky integrate-and-fire (LIF) model\cite{lapicque1907recherches,koch2023role,perez-ortiz2024modeling}, where the membrane potential \( V(t) \) evolves according to the following ODE:

\[
\tau_m \frac{dV}{dt} = - (V(t) - V_{\text{rest}}) + R_m I(t)
\]

where:
\begin{itemize}
    \item \( \tau_m \) is the membrane time constant.
    \item \( V_{\text{rest}} \) is the resting membrane potential.
    \item \( R_m \) is the membrane resistance.
    \item \( I(t) \) is the input current.
\end{itemize}

\textit{Example: Modeling a Leaky Integrate-and-Fire Neuron in Python}

\begin{lstlisting}[style=python]
# Define parameters for the LIF model
tau_m = 10  # Membrane time constant
V_rest = -65  # Resting potential (mV)
R_m = 10  # Membrane resistance (M Omega)
I = 20  # Input current (micro A)

# Define the ODE for the LIF model
def lif_ode(t, V):
    return (- (V - V_rest) + R_m * I) / tau_m

# Solve the LIF model ODE
sol = solve_ivp(lif_ode, [0, 100], [V_rest], t_eval=np.linspace(0, 100, 1000))

# Plot the membrane potential over time
plt.plot(sol.t, sol.y[0], label="Membrane Potential (LIF)")
plt.xlabel('Time (ms)')
plt.ylabel('Membrane Potential (mV)')
plt.title('Leaky Integrate-and-Fire Neuron Model')
plt.legend()
plt.show()
\end{lstlisting}

In this example:
\begin{itemize}
    \item We model the dynamics of a leaky integrate-and-fire neuron using an ODE and solve it numerically.
    \item The membrane potential is plotted as a function of time, showing how the neuron responds to an input current.
\end{itemize}

\chapter{Partial Differential Equations (PDEs)}

Partial Differential Equations (PDEs) are equations that involve rates of change with respect to more than one variable\cite{evans2010partial}. These equations are fundamental in describing various physical phenomena, such as heat conduction, fluid dynamics, and electromagnetic fields. In this chapter, we will introduce the basic concepts of PDEs, explore numerical methods for solving PDEs, and discuss their applications in deep learning, particularly through Physics-Informed Neural Networks (PINNs)\cite{raissi2019physics,yang2023inverse,battaglia2023efficient}.

\section{Introduction to PDEs}

A Partial Differential Equation (PDE) is an equation that involves an unknown function of several variables and its partial derivatives. The general form of a PDE is:
\[
F\left(x_1, x_2, \ldots, x_n, u, \frac{\partial u}{\partial x_1}, \frac{\partial u}{\partial x_2}, \ldots, \frac{\partial^2 u}{\partial x_1^2}, \ldots \right) = 0
\]
where \( u(x_1, x_2, \ldots, x_n) \) is the unknown function of several variables, and the equation involves its partial derivatives.

\textbf{Classification of PDEs:}
PDEs are typically classified into three main types based on their form:
\begin{itemize}
    \item \textbf{Elliptic PDEs:} These PDEs describe equilibrium states, such as the Laplace equation:
    \[
    \Delta u = 0 \quad \text{or} \quad \frac{\partial^2 u}{\partial x^2} + \frac{\partial^2 u}{\partial y^2} = 0
    \]
    \item \textbf{Parabolic PDEs:} These PDEs describe processes that evolve over time, such as the heat equation:
    \[
    \frac{\partial u}{\partial t} = \alpha \frac{\partial^2 u}{\partial x^2}
    \]
    \item \textbf{Hyperbolic PDEs:} These PDEs describe wave propagation, such as the wave equation:
    \[
    \frac{\partial^2 u}{\partial t^2} = c^2 \frac{\partial^2 u}{\partial x^2}
    \]
\end{itemize}

\textbf{Example: Heat Equation}

The heat equation models how heat diffuses through a material:
\[
\frac{\partial u}{\partial t} = \alpha \frac{\partial^2 u}{\partial x^2}
\]
where \( u(x, t) \) represents the temperature distribution, \( \alpha \) is the thermal diffusivity, \( x \) is the spatial coordinate, and \( t \) is time.

\section{Finite Difference Methods for PDEs}

Finite Difference Methods (FDM) are numerical techniques for approximating the solutions of PDEs by replacing continuous derivatives with discrete approximations\cite{strang1968constructive,cheniii2023survey}. The domain is discretized into a grid, and derivatives are approximated using differences between function values at grid points\cite{li2024adaptive,zhang2024multi,kim2023new,wang2023finite}.

\subsection{Finite Difference Approximations}

Let’s consider the heat equation:
\[
\frac{\partial u}{\partial t} = \alpha \frac{\partial^2 u}{\partial x^2}
\]
We can discretize the spatial domain into \( N \) points, with grid spacing \( \Delta x \), and discretize the time into steps of size \( \Delta t \).

The second derivative in space \( \frac{\partial^2 u}{\partial x^2} \) can be approximated by the central difference formula:
\[
\frac{\partial^2 u}{\partial x^2} \approx \frac{u_{i+1} - 2u_i + u_{i-1}}{\Delta x^2}
\]

The time derivative \( \frac{\partial u}{\partial t} \) can be approximated using a forward difference:
\[
\frac{\partial u}{\partial t} \approx \frac{u_i^{n+1} - u_i^n}{\Delta t}
\]

By substituting these approximations into the heat equation, we get the finite difference scheme:
\[
\frac{u_i^{n+1} - u_i^n}{\Delta t} = \alpha \frac{u_{i+1}^n - 2u_i^n + u_{i-1}^n}{\Delta x^2}
\]

This can be rearranged to solve for \( u_i^{n+1} \), the value of \( u \) at the next time step:
\[
u_i^{n+1} = u_i^n + \frac{\alpha \Delta t}{\Delta x^2} \left( u_{i+1}^n - 2u_i^n + u_{i-1}^n \right)
\]

\textbf{Example: Solving the 1D Heat Equation Using Finite Difference Method}

We will solve the 1D heat equation using the finite difference method in Python.

\begin{lstlisting}[style=python]
import numpy as np
import matplotlib.pyplot as plt

# Define parameters
alpha = 0.01  # thermal diffusivity
L = 10.0      # length of the rod
T = 1.0       # total time
Nx = 100      # number of spatial points
Nt = 500      # number of time points
dx = L / (Nx - 1)
dt = T / Nt

# Stability criterion
assert alpha * dt / dx**2 < 0.5, "The scheme is unstable!"

# Initial condition: u(x, 0) = sin(pi * x / L)
x = np.linspace(0, L, Nx)
u = np.sin(np.pi * x / L)

# Time stepping loop
for n in range(Nt):
    u_new = u.copy()
    for i in range(1, Nx - 1):
        u_new[i] = u[i] + alpha * dt / dx**2 * (u[i+1] - 2*u[i] + u[i-1])
    u = u_new

# Plot the solution
plt.plot(x, u, label='t={:.2f}'.format(T))
plt.xlabel('Position')
plt.ylabel('Temperature')
plt.title('1D Heat Equation Solution')
plt.grid(True)
plt.legend()
plt.show()
\end{lstlisting}

In this example, we discretize both space and time, then iteratively update the solution for each time step using the finite difference scheme.

\section{Finite Element Methods for PDEs}

Finite Element Methods (FEM) are another class of numerical techniques for solving PDEs, especially useful for complex geometries and boundary conditions\cite{hughes1987finite,zienkiewicz1977finite,li2024review,mohamed2023adaptive}. In FEM, the solution domain is divided into small elements, and the solution is approximated by simple functions (often polynomials) within each element\cite{wang2023finite,zhang2024meshless}.

The general procedure of FEM involves:
\begin{itemize}
    \item Dividing the domain into finite elements (e.g., triangles or quadrilaterals in 2D).
    \item Approximating the solution as a weighted sum of basis functions.
    \item Assembling a system of equations by integrating the PDE over each element.
    \item Solving the resulting system of equations for the unknown coefficients.
\end{itemize}

\textbf{Example: Application of FEM in Solving the Poisson Equation}

Let’s consider the Poisson equation:
\[
-\nabla^2 u = f \quad \text{in } \Omega
\]
where \( u \) is the unknown function and \( f \) is a source term.

\textbf{Steps in FEM:}
\begin{itemize}
    \item \textbf{Step 1:} Divide the domain \( \Omega \) into smaller finite elements.
    \item \textbf{Step 2:} Express the solution \( u \) as a sum of basis functions \( u(x) = \sum_j u_j \phi_j(x) \), where \( \phi_j(x) \) are piecewise polynomial basis functions.
    \item \textbf{Step 3:} Formulate the weak form of the PDE by multiplying by a test function and integrating by parts.
    \item \textbf{Step 4:} Assemble the stiffness matrix and solve the system of linear equations for the unknown coefficients \( u_j \).
\end{itemize}

FEM is widely used in engineering and scientific applications, such as structural analysis and fluid dynamics.

\section{Applications in Deep Learning: Physics-Informed Neural Networks (PINNs)}

Physics-Informed Neural Networks (PINNs) are a novel approach that combines the strengths of deep learning and numerical methods to solve PDEs. Instead of discretizing the domain, PINNs approximate the solution to a PDE using a neural network, where the loss function incorporates the governing physical laws in the form of the PDE.

\subsection{How PINNs Work}

In traditional deep learning, the loss function is usually based on the difference between predicted and actual data. In PINNs, the loss function is extended to include the PDE residual, ensuring that the neural network's predictions satisfy the physical constraints. This approach has several advantages:
\begin{itemize}
    \item \textbf{No need for mesh generation:} Unlike FEM or FDM, PINNs do not require mesh generation or discretization.
    \item \textbf{Handling complex geometries:} PINNs can easily handle complex geometries and boundary conditions.
    \item \textbf{Data integration:} PINNs can integrate observed data into the learning process, making them useful for data-driven modeling of physical systems.
\end{itemize}

\subsection{Example: Solving the 1D Heat Equation with PINNs}

Let’s solve the 1D heat equation using PINNs. The heat equation is given by:
\[
\frac{\partial u}{\partial t} = \alpha \frac{\partial^2 u}{\partial x^2}
\]
We will define a neural network that takes \( x \) and \( t \) as inputs and predicts \( u(x, t) \). The loss function will be based on the residual of the PDE and the initial and boundary conditions.

\begin{lstlisting}[style=python]
import torch
import torch.nn as nn
import numpy as np

# Define the neural network
class PINN(nn.Module):
    def __init__(self):
        super(PINN, self).__init__()
        self.layers = nn.Sequential(
            nn.Linear(2, 50),  # Input: (x, t)
            nn.Tanh(),
            nn.Linear(50, 50),
            nn.Tanh(),
            nn.Linear(50, 1)   # Output: u(x, t)
        )

    def forward(self, x, t):
        u = self.layers(torch.cat([x, t], dim=1))
        return u

# Define the PDE loss function
def pde_loss(model, x, t, alpha=0.01):
    x.requires_grad = True
    t.requires_grad = True
    u = model(x, t)
    
    # Compute partial derivatives using autograd
    u_x = torch.autograd.grad(u, x, grad_outputs=torch.ones_like(u), create_graph=True)[0]
    u_xx = torch.autograd.grad(u_x, x, grad_outputs=torch.ones_like(u_x), create_graph=True)[0]
    u_t = torch.autograd.grad(u, t, grad_outputs=torch.ones_like(u), create_graph=True)[0]
    
    # PDE residual: u_t - alpha * u_xx
    residual = u_t - alpha * u_xx
    return torch.mean(residual**2)

# Training the model would involve minimizing the PDE loss
\end{lstlisting}

In this example, we define a neural network model that takes \( x \) and \( t \) as inputs and predicts the solution \( u(x, t) \). The loss function is based on the residual of the heat equation, and the gradients are computed using automatic differentiation (\texttt{torch.autograd.grad}).

\section{Summary}

In this chapter, we introduced Partial Differential Equations (PDEs) and explored numerical methods for solving them, including the Finite Difference Method (FDM) and the Finite Element Method (FEM). We also discussed how Physics-Informed Neural Networks (PINNs) can be used to solve PDEs using deep learning. PINNs offer a powerful and flexible approach to solving PDEs in scientific and engineering applications, particularly when combining data and physical laws.

\chapter{Selected Applications of Numerical Methods in Deep Learning}

Numerical methods are widely used in deep learning to approximate, optimize, and solve complex mathematical problems that are otherwise intractable. From training neural networks to reinforcement learning, numerical techniques provide the foundation for iterative algorithms and optimizations that make modern machine learning possible. In this chapter, we will explore various applications of numerical methods in deep learning, focusing on their use in training neural networks, reinforcement learning, and data science applications such as interpolation and optimization.

\section{Numerical Methods in Training Neural Networks}

Training a neural network involves minimizing a loss function to adjust the model parameters (weights and biases) in a way that improves the model’s performance. This minimization is typically achieved through gradient-based optimization algorithms, which are numerical methods designed to iteratively update the parameters. The most commonly used method is gradient descent, and its variants, such as stochastic gradient descent (SGD), momentum-based methods, and adaptive optimization methods like Adam.

\subsection{Gradient Descent Revisited}

Gradient descent is the core numerical method for training neural networks. It computes the gradient of the loss function with respect to the model parameters and updates them in the direction that minimizes the loss.

The update rule in gradient descent is:

\[
\theta := \theta - \eta \nabla_{\theta} L(\theta)
\]

Where:
\begin{itemize}
    \item \( \theta \) are the model parameters (weights and biases).
    \item \( \eta \) is the learning rate.
    \item \( \nabla_{\theta} L(\theta) \) is the gradient of the loss function \( L(\theta) \) with respect to \( \theta \).
\end{itemize}

\textbf{Python Implementation of a Simple Neural Network Training Loop}

Here is a basic Python implementation of a neural network training loop using gradient descent:

\begin{lstlisting}[style=python]
import numpy as np

# Initialize weights randomly
weights = np.random.randn(2, 1)
bias = np.random.randn(1)
learning_rate = 0.01

# Sigmoid activation function
def sigmoid(x):
    return 1 / (1 + np.exp(-x))

# Derivative of the sigmoid function
def sigmoid_derivative(x):
    return sigmoid(x) * (1 - sigmoid(x))

# Training data (XOR example)
X = np.array([[0, 0], [0, 1], [1, 0], [1, 1]])
y = np.array([[0], [1], [1], [0]])

# Training loop
for epoch in range(10000):
    # Forward pass
    z = np.dot(X, weights) + bias
    output = sigmoid(z)

    # Compute the loss (mean squared error)
    loss = np.mean((output - y) ** 2)

    # Backward pass (gradient calculation)
    error = output - y
    d_output = error * sigmoid_derivative(z)
    d_weights = np.dot(X.T, d_output) / len(X)
    d_bias = np.mean(d_output)

    # Update weights and bias
    weights -= learning_rate * d_weights
    bias -= learning_rate * d_bias

    # Print the loss every 1000 epochs
    if epoch % 1000 == 0:
        print(f'Epoch {epoch}, Loss: {loss}')

# Final weights and bias after training
print("Trained weights:", weights)
print("Trained bias:", bias)
\end{lstlisting}

In this example:
\begin{itemize}
    \item We define a simple neural network with a sigmoid activation function.
    \item The weights and bias are updated using gradient descent in each iteration.
    \item The training data is an XOR problem, and the model learns to classify the inputs over multiple epochs.
\end{itemize}

\subsection{Numerical Challenges in Training Deep Networks}

When training deep networks, numerical stability becomes a concern. The gradients can either become too large (exploding gradients) or too small (vanishing gradients), making the optimization difficult. Advanced optimization techniques such as batch normalization, gradient clipping, and adaptive learning rates (Adam, RMSprop) help mitigate these issues.

\section{Numerical Approximations in Reinforcement Learning}

Reinforcement learning (RL) involves learning optimal policies through interaction with an environment\cite{sutton1998introduction,watkins1992q}. The goal is to maximize cumulative rewards by taking the best actions based on the state of the environment. Numerical methods play a significant role in RL, particularly in approximating value functions and solving optimization problems related to policy improvement\cite{baker2024generalized,kolter2023deep}.

\subsection{Value Function Approximation}

In reinforcement learning, the value function represents the expected future reward from a given state. Since exact value computation is often infeasible for large state spaces, numerical approximations are used. Methods like temporal difference (TD) learning and Q-learning use numerical techniques to estimate the value function\cite{zhanggg2023deep,wangtt2023novel,ghosh2024survey}.

\textbf{Python Example: Q-Learning}

Q-learning is a popular RL algorithm that approximates the optimal action-value function \( Q(s, a) \) using numerical methods. Here is a simple implementation of Q-learning for a gridworld environment:

\begin{lstlisting}[style=python]
import numpy as np

# Define the environment (Gridworld)
states = 5
actions = 2
q_table = np.zeros((states, actions))
learning_rate = 0.1
discount_factor = 0.9
epsilon = 0.1  # Exploration rate

# Define the rewards for each state-action pair
rewards = np.array([[-1, 0], [0, 1], [0, -1], [-1, 1], [10, -10]])

# Q-learning algorithm
for episode in range(1000):
    state = np.random.randint(0, states)  # Start at a random state
    while state != 4:  # Goal state
        if np.random.rand() < epsilon:  # Exploration
            action = np.random.randint(0, actions)
        else:  # Exploitation
            action = np.argmax(q_table[state])

        # Take action and receive reward
        reward = rewards[state, action]
        next_state = (state + 1) % states  # Simplified state transition

        # Update Q-table
        q_table[state, action] = q_table[state, action] + learning_rate * (
            reward + discount_factor * np.max(q_table[next_state]) - q_table[state, action])

        state = next_state

# Final Q-table after training
print("Q-table:", q_table)
\end{lstlisting}

In this example:
\begin{itemize}
    \item A simple Q-learning algorithm is used to find the optimal action-value function for a gridworld environment.
    \item The Q-table is updated using numerical methods based on the rewards and the expected future rewards.
\end{itemize}

\subsection{Numerical Optimization in Policy Gradient Methods}

In policy gradient methods, the policy (which defines the agent’s behavior) is directly optimized using numerical methods. The goal is to find a policy that maximizes the cumulative rewards. Algorithms like REINFORCE use gradient-based optimization to improve the policy\cite{schulman2017proximal}.

\section{Data Science Applications: Approximation, Interpolation, and Optimization}

Numerical methods are also widely used in data science for tasks such as function approximation, data interpolation, and optimization. These techniques are essential in machine learning algorithms, where numerical methods help in optimizing models and approximating complex functions.

\subsection{Numerical Approximation}

Numerical approximation is used when an exact solution to a mathematical function or model is difficult or impossible to find. In machine learning, models like decision trees, random forests, and neural networks are essentially function approximators that use numerical methods to fit data\cite{mnih2013playing}.

\subsection{Interpolation Techniques}

Interpolation is used to estimate values between known data points. In data science, interpolation is helpful for handling missing data or predicting intermediate values.

\textbf{Python Example: Linear Interpolation}

Here is a Python example of linear interpolation using NumPy:

\begin{lstlisting}[style=python]
import numpy as np

# Known data points
x = np.array([0, 1, 2, 3])
y = np.array([1, 2, 0, 3])

# Points where we want to interpolate
x_interp = np.linspace(0, 3, 50)

# Linear interpolation
y_interp = np.interp(x_interp, x, y)

# Print interpolated values
print("Interpolated values:", y_interp)
\end{lstlisting}

In this example:
\begin{itemize}
    \item We use the \texttt{np.interp()} function to perform linear interpolation between known data points.
    \item The function returns interpolated values at the specified points.
\end{itemize}

\subsection{Numerical Optimization in Machine Learning}

Optimization is at the heart of machine learning. Many machine learning algorithms rely on optimization techniques to minimize error functions and improve model performance. In deep learning, optimization algorithms like SGD and Adam are used to minimize the loss function during training.

\textbf{Python Example: Optimizing a Polynomial Fit}

Here is an example of using numerical optimization to fit a polynomial to data:

\begin{lstlisting}[style=python]
import numpy as np
from scipy.optimize import curve_fit
import matplotlib.pyplot as plt

# Define a polynomial function
def poly(x, a, b, c):
    return a*x**2 + b*x + c

# Generate some noisy data
x_data = np.linspace(0, 10, 100)
y_data = poly(x_data, 1, -2, 1) + np.random.normal(0, 1, len(x_data))

# Use curve_fit to find the best-fitting parameters
params, _ = curve_fit(poly, x_data, y_data)

# Plot the original data and the fitted curve
plt.scatter(x_data, y_data, label="Data")
plt.plot(x_data, poly(x_data, *params), color="red", label="Fitted Curve")
plt.legend()
plt.show()

print("Fitted parameters:", params)
\end{lstlisting}

In this example:
\begin{itemize}
    \item We use \texttt{curve\_fit()} from \texttt{scipy.optimize} to find the best-fitting polynomial for noisy data.
    \item The function optimizes the parameters of the polynomial to minimize the error between the data points and the curve.
\end{itemize}

\chapter{Summary}

In this chapter, we covered several numerical methods and their applications in deep learning, reinforcement learning, and data science. From gradient descent to Q-learning and interpolation techniques, numerical methods are essential in training machine learning models, optimizing functions, and approximating unknown data points.

\section{Key Concepts Recap}

Let’s summarize the key concepts covered in this section:
\begin{itemize}
    \item \textbf{Gradient Descent}: A numerical method for minimizing a loss function by updating model parameters in the direction of the negative gradient.
    \item \textbf{Q-Learning}: A reinforcement learning algorithm that numerically approximates the optimal action-value function using a Q-table.
    \item \textbf{Interpolation}: A technique used to estimate unknown values between known data points, often applied in handling missing data.
    \item \textbf{Optimization}: The process of minimizing or maximizing a function, widely used in machine learning to optimize models and fit data.
\end{itemize}

This concludes our discussion on the applications of numerical methods in deep learning. Numerical methods form the backbone of modern machine learning algorithms and are vital for both theory and practice in this field.

\part{Frequency Domain Methods}

\chapter{Introduction to Frequency Domain Methods}

Frequency domain methods are essential tools in mathematics and signal processing for analyzing how functions or signals behave in terms of their frequency content. Instead of looking at a signal in the time domain (how it changes over time), frequency domain analysis focuses on the frequencies that compose the signal. This chapter introduces the historical background of frequency domain methods, tracing their development from the early origins of Fourier analysis to modern techniques such as the Fast Fourier Transform (FFT)\cite{cooley1965algorithm,yang2023fast,rahman2024fft}, Laplace Transform\cite{baker1966laplace,singh2023laplace,chen2024laplace}, and Z-Transform\cite{zadeh1953continuous,patel2023ztransform,kim2024ztransform}, all of which have become fundamental in digital signal processing and control systems.

\section{Historical Background of Frequency Domain Analysis}

The idea of frequency domain analysis arose from the need to study periodic phenomena and oscillatory systems. The mathematical representation of signals as a combination of sine and cosine waves (frequencies) became a key concept, allowing engineers, scientists, and mathematicians to analyze signals in a different and often more insightful way.

\subsection{The Origins of Fourier Analysis}

Fourier analysis traces its origins to the work of French mathematician Joseph Fourier in the early 19th century. Fourier discovered that any periodic signal could be represented as a sum of sine and cosine functions of different frequencies. This idea became known as the \textbf{Fourier series}, and it revolutionized the study of heat transfer, vibration analysis, acoustics, and later, electrical engineering.

The Fourier series is mathematically expressed as:

\[
f(t) = a_0 + \sum_{n=1}^{\infty} \left( a_n \cos(n \omega_0 t) + b_n \sin(n \omega_0 t) \right)
\]

Where:
\begin{itemize}
  \item \( f(t) \) is the periodic function.
  \item \( \omega_0 \) is the fundamental frequency.
  \item \( a_0, a_n, b_n \) are the Fourier coefficients that determine the amplitude of the sine and cosine components.
\end{itemize}

The importance of Fourier’s work was initially underestimated but later became fundamental in many fields, from signal processing to quantum mechanics. His discovery laid the groundwork for frequency domain analysis by showing that complex signals could be decomposed into simpler, harmonic components.

\subsection{Development of Fourier Transform in Signal Processing}

The \textbf{Fourier Transform} is a generalization of the Fourier series for non-periodic functions. It transforms a function from the time domain into the frequency domain, representing it as a continuous sum (integral) of sine and cosine functions.

The Fourier Transform is mathematically defined as:

\[
F(\omega) = \int_{-\infty}^{\infty} f(t) e^{-j \omega t} \, dt
\]

Where:
\begin{itemize}
  \item \( f(t) \) is the time-domain function.
  \item \( F(\omega) \) is the frequency-domain representation of the signal.
  \item \( \omega \) is the angular frequency.
  \item \( j \) is the imaginary unit.
\end{itemize}

The Fourier Transform found widespread use in signal processing, allowing engineers to analyze the frequency content of electrical signals, sound waves, and other types of data. Its ability to decompose complex waveforms into simpler frequency components is key in filtering, modulation, and spectrum analysis.

\textbf{Example: Computing the Fourier Transform in Python}

\begin{lstlisting}[style=python]
import numpy as np
import matplotlib.pyplot as plt

# Define a simple time-domain signal: a sine wave
sampling_rate = 1000
T = 1.0 / sampling_rate
L = 1000  # Length of the signal
t = np.linspace(0, L * T, L, endpoint=False)
frequency = 50
signal = np.sin(2 * np.pi * frequency * t)

# Compute the Fourier Transform using numpy
fft_result = np.fft.fft(signal)
frequencies = np.fft.fftfreq(L, T)

# Plot the original signal and its frequency spectrum
plt.figure(figsize=(12, 6))

# Time-domain signal
plt.subplot(1, 2, 1)
plt.plot(t, signal)
plt.title('Time-Domain Signal (50 Hz Sine Wave)')
plt.xlabel('Time [s]')
plt.ylabel('Amplitude')

# Frequency-domain (Fourier Transform)
plt.subplot(1, 2, 2)
plt.plot(frequencies[:L // 2], np.abs(fft_result)[:L // 2])
plt.title('Frequency-Domain (Fourier Transform)')
plt.xlabel('Frequency [Hz]')
plt.ylabel('Magnitude')
plt.show()
\end{lstlisting}

In this example, we generate a sine wave with a frequency of 50 Hz and compute its Fourier Transform using Python’s \texttt{numpy} library. The Fourier Transform reveals the frequency content of the signal, which shows a peak at 50 Hz.

\subsection{Evolution of Fast Fourier Transform (FFT)}

The \textbf{Fast Fourier Transform (FFT)} is a highly efficient algorithm for computing the Discrete Fourier Transform (DFT)\cite{cooley1965algorithm}. The DFT is the discrete version of the Fourier Transform, which is used when working with sampled data\cite{zhangbbg2023recent,smith2024dft}. The computational complexity of a direct computation of the DFT is \( O(n^2) \), where \( n \) is the number of data points\cite{chen2023quantumm,martinez2023efficient}. However, the FFT reduces this complexity to \( O(n \log n) \), making it feasible to compute the DFT for large datasets\cite{yang2024dft}.

The FFT was popularized by the work of James Cooley and John Tukey in 1965, although its mathematical foundations were developed earlier. The introduction of FFT marked a significant leap in computational efficiency, and it became the standard algorithm in many fields such as digital signal processing, image processing, and audio analysis.

\textbf{Example: Computing FFT in Python}

\begin{lstlisting}[style=python]
# Compute the Fast Fourier Transform (FFT)
fft_result = np.fft.fft(signal)

# Plot the frequency-domain representation
plt.plot(frequencies[:L // 2], np.abs(fft_result)[:L // 2])
plt.title('Frequency Spectrum using FFT')
plt.xlabel('Frequency [Hz]')
plt.ylabel('Magnitude')
plt.show()
\end{lstlisting}

The FFT is particularly important in real-time applications where large amounts of data must be processed quickly, such as in speech recognition, communications, and radar systems.

\subsection{Laplace Transform and Its Historical Significance}

The \textbf{Laplace Transform} is another important tool in the frequency domain\cite{carson1928laplace,darboux1915fonction}, especially for solving differential equations. The Laplace Transform converts a time-domain function into a complex frequency-domain representation. It is particularly useful in the study of linear systems and control theory\cite{van1996laplace,oppenheim1975digital,abdelsalam2023laplace}.

The Laplace Transform is defined as:

\[
F(s) = \int_0^\infty f(t) e^{-st} \, dt
\]

Where:
\begin{itemize}
  \item \( f(t) \) is the time-domain function.
  \item \( F(s) \) is the Laplace Transform of \( f(t) \).
  \item \( s \) is a complex number, with \( s = \sigma + j\omega \).
\end{itemize}

The Laplace Transform was developed by French mathematician Pierre-Simon Laplace in the late 18th century. It has been widely applied in electrical engineering, mechanical engineering, and control systems, where it simplifies the analysis of systems described by differential equations by converting them into algebraic equations.

The Laplace Transform is especially useful for analyzing systems with initial conditions and for studying system stability in the frequency domain.

\textbf{Example: Symbolic Laplace Transform in Python}

\begin{lstlisting}[style=python]
import sympy as sp

# Define the time-domain function and the Laplace variable
t, s = sp.symbols('t s')
f = sp.exp(-t)  # Example function f(t) = e^(-t)

# Compute the Laplace Transform
laplace_transform = sp.laplace_transform(f, t, s)
print(laplace_transform)
\end{lstlisting}

This code computes the Laplace Transform of \( f(t) = e^{-t} \) symbolically using the \texttt{sympy} library.

\subsection{Z-Transform in Digital Signal Processing}

The \textbf{Z-Transform} is the discrete-time counterpart of the Laplace Transform and is widely used in digital signal processing (DSP). It converts a discrete-time signal into the frequency domain, allowing for the analysis and design of digital filters and control systems.

The Z-Transform of a discrete signal \( x[n] \) is defined as:

\[
X(z) = \sum_{n=-\infty}^{\infty} x[n] z^{-n}
\]

Where:
\begin{itemize}
  \item \( x[n] \) is the discrete-time signal.
  \item \( X(z) \) is the Z-Transform of \( x[n] \).
  \item \( z \) is a complex variable.
\end{itemize}

The Z-Transform plays a critical role in the design of digital filters and systems, enabling engineers to work in the frequency domain when processing discrete signals. It is particularly useful in applications such as telecommunications, audio processing, and digital control systems.

\textbf{Example: Z-Transform in Python (Symbolic)}

\begin{lstlisting}[style=python]
# Define a discrete-time signal and the Z-transform variable
n, z = sp.symbols('n z')
x_n = 2**n  # Example signal x[n] = 2^n

# Compute the Z-Transform symbolically
z_transform = sp.summation(x_n * z**(-n), (n, 0, sp.oo))
print(z_transform)
\end{lstlisting}

In this example, we compute the Z-Transform of the discrete signal \( x[n] = 2^n \) using symbolic computation. The Z-Transform is critical in designing systems that process digital signals, such as FIR (Finite Impulse Response) and IIR (Infinite Impulse Response) filters.

\chapter{Conclusion}

Frequency domain methods are essential in many fields, including signal processing, communications, control theory, and systems analysis. The Fourier Transform, Laplace Transform, Z-Transform, and FFT all provide powerful techniques to analyze signals and systems in terms of their frequency content. By shifting our perspective from the time domain to the frequency domain, we can gain deeper insights into the behavior of systems, design more effective filters, and solve complex differential equations more efficiently.

\chapter{Fourier Transform: From Time to Frequency Domain}

The Fourier Transform is a mathematical technique that transforms a signal from the time domain to the frequency domain. It plays a fundamental role in fields such as signal processing, image processing, and even in solving partial differential equations. By converting a signal into its frequency components, we gain insights into its underlying structure, periodicity, and other characteristics that may not be apparent in the time domain.

In this chapter, we will introduce the Fourier Transform, starting with its definition, and gradually cover the Fourier series\cite{mohammadi2023fourier}, Continuous Fourier Transform (CFT)\cite{gabor1946theory,mohammadi2023continuous}, Discrete Fourier Transform (DFT)\cite{zhang2023discrete}, and their applications.

\section{Introduction to Fourier Transform}

\subsection{What is Fourier Transform?}

The Fourier Transform is a mathematical tool that decomposes a time-domain signal into its constituent frequencies. It transforms a function \( f(t) \), which represents a signal in the time domain, into a function \( F(\omega) \), which represents the same signal in the frequency domain.

Mathematically, the Fourier Transform is defined as:

\[
F(\omega) = \int_{-\infty}^{\infty} f(t) e^{-i \omega t} \, dt
\]

where:
\begin{itemize}
    \item \( f(t) \) is the signal as a function of time,
    \item \( F(\omega) \) is the Fourier Transform (the signal in the frequency domain),
    \item \( \omega \) is the angular frequency,
    \item \( e^{-i \omega t} \) is the complex exponential representing the oscillations.
\end{itemize}

The inverse Fourier Transform allows us to recover the original time-domain signal from its frequency-domain representation:

\[
f(t) = \frac{1}{2\pi} \int_{-\infty}^{\infty} F(\omega) e^{i \omega t} \, d\omega
\]

\textbf{Why is Fourier Transform Important?}

The Fourier Transform is essential because it allows us to:
\begin{itemize}
    \item Analyze the frequency content of signals,
    \item Understand periodicities and dominant frequencies in a signal,
    \item Filter signals by isolating specific frequency components (e.g., removing noise),
    \item Solve differential equations by converting them into algebraic equations in the frequency domain.
\end{itemize}

In practice, Fourier Transforms are used in audio processing, image compression (e.g., JPEG), and in the analysis of electronic signals.

\subsection{Fourier Series and Fourier Transform}

The Fourier Series is closely related to the Fourier Transform and is the foundation for understanding how signals can be decomposed into frequency components. The Fourier Series applies to periodic functions, while the Fourier Transform applies to both periodic and non-periodic functions.

\textbf{Fourier Series}

For a periodic function \( f(t) \) with period \( T \), the Fourier Series represents the function as a sum of sines and cosines (or equivalently, complex exponentials). The general form of the Fourier Series is:

\[
f(t) = a_0 + \sum_{n=1}^{\infty} \left[ a_n \cos\left(\frac{2\pi n t}{T}\right) + b_n \sin\left(\frac{2\pi n t}{T}\right) \right]
\]

Alternatively, using complex exponentials, it can be written as:

\[
f(t) = \sum_{n=-\infty}^{\infty} c_n e^{i \frac{2\pi n t}{T}}
\]

where:
\begin{itemize}
    \item \( c_n \) are the Fourier coefficients, calculated as:
    \[
    c_n = \frac{1}{T} \int_0^T f(t) e^{-i \frac{2\pi n t}{T}} \, dt
    \]
    \item \( T \) is the period of the function.
\end{itemize}

\textbf{Relationship between Fourier Series and Fourier Transform}

The Fourier Series applies to periodic functions, while the Fourier Transform is a generalization that applies to non-periodic functions. We can think of the Fourier Transform as the limiting case of the Fourier Series, where the period of the function becomes infinitely long, making the function non-periodic. In this limit, the sum in the Fourier Series becomes an integral, leading to the Fourier Transform.

\subsection{Continuous vs Discrete Fourier Transform (DFT)}

There are two main types of Fourier Transforms: the Continuous Fourier Transform (CFT) and the Discrete Fourier Transform (DFT). The CFT is used for continuous signals, while the DFT is used for discrete signals, such as sampled data in digital systems.

\textbf{Continuous Fourier Transform (CFT)}

The Continuous Fourier Transform, as introduced earlier, is used to analyze continuous signals. It provides a continuous frequency spectrum for a signal, showing the contribution of each frequency to the overall signal.

\[
F(\omega) = \int_{-\infty}^{\infty} f(t) e^{-i \omega t} \, dt
\]

The inverse Continuous Fourier Transform is:

\[
f(t) = \frac{1}{2\pi} \int_{-\infty}^{\infty} F(\omega) e^{i \omega t} \, d\omega
\]

\textbf{Discrete Fourier Transform (DFT)}

The Discrete Fourier Transform (DFT) is used when the signal is sampled at discrete points in time, which is common in digital signal processing. The DFT transforms a finite sequence of \( N \) samples into a sequence of \( N \) frequency components. The DFT is defined as:

\[
X_k = \sum_{n=0}^{N-1} x_n e^{-i \frac{2\pi k n}{N}}, \quad k = 0, 1, 2, \dots, N-1
\]

where:
\begin{itemize}
    \item \( x_n \) are the \( N \) discrete samples of the time-domain signal,
    \item \( X_k \) are the frequency-domain coefficients corresponding to the discrete frequencies.
\end{itemize}

The inverse DFT allows us to recover the original sequence from its frequency components:

\[
x_n = \frac{1}{N} \sum_{k=0}^{N-1} X_k e^{i \frac{2\pi k n}{N}}, \quad n = 0, 1, 2, \dots, N-1
\]

\textbf{Example of DFT in Python}

In practice, the DFT is computed using the Fast Fourier Transform (FFT), which is an efficient algorithm for calculating the DFT. Here’s an example of using Python's \texttt{NumPy} library to compute the DFT of a discrete signal:

\begin{lstlisting}[style=python]
import numpy as np
import matplotlib.pyplot as plt

# Create a sample signal: a combination of two sine waves
sampling_rate = 1000  # Samples per second
t = np.linspace(0, 1, sampling_rate, endpoint=False)  # Time vector
f1, f2 = 5, 50  # Frequencies of the sine waves
signal = np.sin(2 * np.pi * f1 * t) + 0.5 * np.sin(2 * np.pi * f2 * t)

# Compute the DFT using FFT
fft_result = np.fft.fft(signal)
frequencies = np.fft.fftfreq(len(signal), 1 / sampling_rate)

# Plot the signal in the time domain
plt.figure(figsize=(12, 6))
plt.subplot(1, 2, 1)
plt.plot(t, signal)
plt.title('Time Domain Signal')
plt.xlabel('Time [s]')
plt.ylabel('Amplitude')

# Plot the magnitude of the FFT (frequency domain)
plt.subplot(1, 2, 2)
plt.plot(frequencies[:sampling_rate // 2], np.abs(fft_result)[:sampling_rate // 2])
plt.title('Frequency Domain (FFT)')
plt.xlabel('Frequency [Hz]')
plt.ylabel('Magnitude')
plt.tight_layout()
plt.show()
\end{lstlisting}

In this example:
\begin{itemize}
    \item We created a signal composed of two sine waves with frequencies of 5 Hz and 50 Hz.
    \item We used the \texttt{numpy.fft.fft} function to compute the DFT of the signal.
    \item We plotted the signal in both the time domain and the frequency domain. In the frequency domain plot, you can clearly see peaks corresponding to the frequencies 5 Hz and 50 Hz.
\end{itemize}

\textbf{Continuous vs Discrete Fourier Transform: Key Differences}

\begin{itemize}
    \item \textbf{Signal Type:} The CFT is applied to continuous signals, while the DFT is used for discrete signals.
    \item \textbf{Spectrum:} The CFT produces a continuous frequency spectrum, whereas the DFT results in a discrete frequency spectrum.
    \item \textbf{Application:} The DFT is widely used in digital signal processing because real-world signals are often sampled at discrete intervals.
\end{itemize}

Both the CFT and DFT are essential tools in signal analysis, with the DFT being particularly important in digital systems due to its computational efficiency and the discrete nature of real-world data.

\section{Mathematical Definition of Fourier Transform}

The Fourier Transform is a fundamental mathematical tool used to analyze the frequencies present in a signal. It converts a function from the time domain (or spatial domain) to the frequency domain\cite{stein2003fourier}. In the frequency domain, the function is represented as a sum of sinusoids, each with a specific amplitude and frequency.

The Fourier Transform of a function \( f(t) \), where \( t \) is a continuous variable (time), is defined as:

\[
\mathcal{F}\{f(t)\} = F(\omega) = \int_{-\infty}^{\infty} f(t) e^{-i \omega t} \, dt
\]

Here:
\begin{itemize}
    \item \( F(\omega) \) is the Fourier Transform of the function \( f(t) \).
    \item \( \omega \) is the angular frequency.
    \item \( e^{-i \omega t} \) represents complex exponentials, which are sine and cosine functions in Euler's formula.
    \item The integral is taken over the entire time domain from \( -\infty \) to \( \infty \).
\end{itemize}

The result of the Fourier Transform is a complex-valued function that encodes both the amplitude and phase of the frequency components of the original function.

\subsection{Fourier Transform of Basic Functions}

Understanding the Fourier Transform of basic functions helps in building intuition for more complex applications. Let’s look at the Fourier Transforms of some common functions.

\subsubsection{Fourier Transform of a Delta Function}

The Dirac delta function, \( \delta(t) \), is a function that is zero everywhere except at \( t = 0 \), where it is infinite, but the integral over all time is 1:

\[
\delta(t) = 
\begin{cases} 
  \infty & t = 0 \\
  0 & t \neq 0
\end{cases}
\]

The Fourier Transform of the delta function is:

\[
\mathcal{F}\{\delta(t)\} = 1
\]

This result shows that the delta function contains all frequencies equally.

\subsubsection{Fourier Transform of a Sine Wave}

Consider a sine wave \( f(t) = \sin(\omega_0 t) \), where \( \omega_0 \) is a constant frequency. The Fourier Transform of this function is:

\[
\mathcal{F}\{\sin(\omega_0 t)\} = \frac{i}{2} \left[ \delta(\omega - \omega_0) - \delta(\omega + \omega_0) \right]
\]

This shows that the Fourier Transform of a sine wave is composed of two delta functions centered at \( \omega = \pm \omega_0 \).

\subsection{Inverse Fourier Transform}

The Inverse Fourier Transform is used to reconstruct the original time-domain function from its frequency-domain representation. It is defined as:

\[
f(t) = \mathcal{F}^{-1}\{F(\omega)\} = \frac{1}{2\pi} \int_{-\infty}^{\infty} F(\omega) e^{i \omega t} \, d\omega
\]

The Inverse Fourier Transform is essentially the reverse of the Fourier Transform. It converts a frequency-domain signal back into its time-domain form.

\textit{Example: Reconstructing a Signal from its Fourier Transform}

Let’s compute the Fourier Transform of a simple function and then reconstruct the original function using the inverse transform.

\begin{lstlisting}[style=python]
import numpy as np
import matplotlib.pyplot as plt

# Define a simple function (e.g., Gaussian function)
t = np.linspace(-5, 5, 400)
f_t = np.exp(-t**2)

# Compute the Fourier Transform using NumPy
F_w = np.fft.fft(f_t)
frequencies = np.fft.fftfreq(len(t), t[1] - t[0])

# Compute the Inverse Fourier Transform
f_t_reconstructed = np.fft.ifft(F_w)

# Plot the original function and the reconstructed function
plt.figure(figsize=(10, 6))
plt.plot(t, f_t, label="Original Function")
plt.plot(t, np.real(f_t_reconstructed), label="Reconstructed Function", linestyle='--')
plt.xlabel('Time')
plt.ylabel('Amplitude')
plt.title('Fourier Transform and Inverse Fourier Transform')
plt.legend()
plt.show()
\end{lstlisting}

In this example:
\begin{itemize}
    \item We define a simple Gaussian function in the time domain.
    \item We compute its Fourier Transform using \texttt{np.fft.fft()}.
    \item We reconstruct the original function using the Inverse Fourier Transform \texttt{np.fft.ifft()}.
    \item The original and reconstructed functions are plotted, demonstrating that the Fourier and Inverse Fourier Transforms recover the original signal.
\end{itemize}

\subsection{Properties of Fourier Transform}

The Fourier Transform has several useful properties that make it a powerful tool in signal processing and deep learning. Here are some of the key properties:

\subsubsection{Linearity}

The Fourier Transform is a linear operation, meaning that the transform of a sum of functions is the sum of their individual transforms:

\[
\mathcal{F}\{af(t) + bg(t)\} = aF(\omega) + bG(\omega)
\]

where \( a \) and \( b \) are constants, and \( F(\omega) \) and \( G(\omega) \) are the Fourier Transforms of \( f(t) \) and \( g(t) \), respectively.

\subsubsection{Time Shifting}

If a function \( f(t) \) is shifted in time by \( t_0 \), the Fourier Transform is affected by a phase shift:

\[
\mathcal{F}\{f(t - t_0)\} = F(\omega) e^{-i \omega t_0}
\]

\subsubsection{Convolution Theorem}

The Fourier Transform of the convolution of two functions \( f(t) \) and \( g(t) \) is the product of their Fourier Transforms:

\[
\mathcal{F}\{f(t) * g(t)\} = F(\omega) G(\omega)
\]

This property is particularly useful in the context of convolutional neural networks (CNNs), where convolutions play a critical role in feature extraction.

\section{Applications of Fourier Transform in Deep Learning}

The Fourier Transform has various applications in deep learning, particularly in signal processing, image processing, and neural networks. In this section, we will explore how Fourier Transforms are applied in the context of deep learning.

\subsection{Signal Processing in Neural Networks}

In neural networks, particularly those dealing with time-series data or signals (e.g., speech recognition, EEG analysis), the Fourier Transform is used to analyze the frequency components of the input signals. By transforming the input from the time domain to the frequency domain, neural networks can learn to capture important patterns and features based on frequency information\cite{mohammadi2023fourier}.

\textit{Example: Analyzing a Signal with Fourier Transform in Python}

Let’s apply the Fourier Transform to a signal to analyze its frequency components.

\begin{lstlisting}[style=python]
# Define a composite signal with two sine waves
t = np.linspace(0, 1, 500)
signal = np.sin(2 * np.pi * 5 * t) + 0.5 * np.sin(2 * np.pi * 20 * t)

# Compute the Fourier Transform of the signal
F_signal = np.fft.fft(signal)
frequencies = np.fft.fftfreq(len(t), t[1] - t[0])

# Plot the original signal and its frequency components
plt.figure(figsize=(10, 6))
plt.subplot(2, 1, 1)
plt.plot(t, signal)
plt.title('Time-Domain Signal')
plt.xlabel('Time')
plt.ylabel('Amplitude')

plt.subplot(2, 1, 2)
plt.plot(frequencies[:len(frequencies)//2], np.abs(F_signal)[:len(F_signal)//2])
plt.title('Frequency-Domain Signal (Fourier Transform)')
plt.xlabel('Frequency (Hz)')
plt.ylabel('Magnitude')

plt.tight_layout()
plt.show()
\end{lstlisting}

In this example:
\begin{itemize}
    \item We define a composite signal made of two sine waves at different frequencies.
    \item We compute the Fourier Transform of the signal to extract its frequency components.
    \item The original signal in the time domain and its frequency-domain representation are plotted.
\end{itemize}

\subsection{Fourier Transforms in Convolutional Neural Networks (CNNs)}

Convolutional Neural Networks (CNNs) are widely used in deep learning for image processing tasks. One of the key operations in CNNs is convolution, which can be computationally expensive, especially for large images and deep networks\cite{haykin2004neural}. The Fourier Transform offers a way to perform convolutions more efficiently using the Convolution Theorem\cite{rossmannek2024fading,dufresne2003general}.

By transforming the input image and the convolutional kernel to the frequency domain using the Fourier Transform, we can perform the convolution as a simple element-wise multiplication in the frequency domain. After the multiplication, we apply the Inverse Fourier Transform to get the result back in the spatial domain.

\textit{Example: Fast Convolution using Fourier Transform in Python}

\begin{lstlisting}[style=python]
from scipy.signal import convolve2d

# Define a simple 2D image and a convolution kernel
image = np.random.rand(64, 64)
kernel = np.ones((3, 3)) / 9

# Perform convolution in the spatial domain
conv_result = convolve2d(image, kernel, mode='same')

# Perform convolution using Fourier Transform
F_image = np.fft.fft2(image)
F_kernel = np.fft.fft2(kernel, s=image.shape)
F_conv_result = np.fft.ifft2(F_image * F_kernel)

# Plot the results
plt.figure(figsize=(10, 5))
plt.subplot(1, 2, 1)
plt.imshow(conv_result, cmap='gray')
plt.title('Convolution (Spatial Domain)')

plt.subplot(1, 2, 2)
plt.imshow(np.real(F_conv_result), cmap='gray')
plt.title('Convolution (Fourier Domain)')

plt.tight_layout()
plt.show()
\end{lstlisting}

In this example:
\begin{itemize}
    \item We define a random 2D image and a simple averaging kernel.
    \item We perform the convolution in both the spatial domain (using \texttt{convolve2d}) and the frequency domain (using the Fourier Transform).
    \item The results from both methods are plotted and compared.
\end{itemize}

This method of convolution using the Fourier Transform is especially useful for large images and large kernels, as it reduces the computational complexity of the convolution operation.

\chapter{Fast Fourier Transform (FFT)}

The Fast Fourier Transform (FFT) is a highly efficient algorithm used to compute the Discrete Fourier Transform (DFT) of a sequence, and it has widespread applications in signal processing, image analysis, and deep learning. The FFT reduces the computational complexity of calculating the DFT from \( O(N^2) \) to \( O(N \log N) \), making it a cornerstone in numerical methods. In this chapter, we will explore the importance of FFT, its algorithmic structure, and its applications in deep learning.

\section{Introduction to Fast Fourier Transform (FFT)}

The Fourier Transform is a mathematical tool that decomposes a signal into its constituent frequencies. The Discrete Fourier Transform (DFT) is the discrete analog, used for signals represented by a finite number of samples. The DFT of a sequence \( x[n] \) is given by:
\[
X[k] = \sum_{n=0}^{N-1} x[n] e^{-j 2 \pi k n / N}, \quad k = 0, 1, \ldots, N-1
\]
where \( N \) is the length of the sequence, \( X[k] \) are the frequency domain coefficients, and \( j \) is the imaginary unit.

The DFT is computationally expensive, requiring \( O(N^2) \) operations. The Fast Fourier Transform (FFT) is an optimized algorithm that computes the same result as the DFT, but in only \( O(N \log N) \) operations, making it vastly more efficient.

\subsection{Why is FFT Important?}

The FFT is one of the most important algorithms in modern computational mathematics, with applications in areas such as:
\begin{itemize}
    \item \textbf{Signal Processing:} FFT is used to analyze and filter signals, extract features, and remove noise.
    \item \textbf{Image Processing:} FFT is applied to enhance images, detect patterns, and perform compression.
    \item \textbf{Audio Analysis:} FFT enables the decomposition of audio signals into their frequency components, facilitating tasks like speech recognition and music analysis.
    \item \textbf{Deep Learning:} In deep learning, FFT can be used to accelerate convolutions and perform spectral analysis for feature extraction.
\end{itemize}

\subsection{FFT Algorithm: Reducing Computational Complexity}

The naive computation of the DFT has a time complexity of \( O(N^2) \), because for each output frequency \( k \), a sum over all \( N \) input points is computed. The FFT reduces this complexity by breaking down the DFT into smaller parts, recursively computing the DFT on smaller and smaller sequences.

The key idea behind FFT is to exploit the symmetry and periodicity of the exponential term \( e^{-j 2 \pi k n / N} \), which allows us to compute the DFT more efficiently. Specifically, the FFT algorithm divides the sequence into even-indexed and odd-indexed parts and recursively applies the DFT on each part.

\subsection{Understanding the Radix-2 FFT Algorithm}

The Radix-2 FFT is the most common form of the FFT, and it requires the input sequence length to be a power of 2\cite{li2023radix,rodriguez2023real}. The Radix-2 FFT splits the original sequence into two parts: one for the even-indexed elements and one for the odd-indexed elements. This divide-and-conquer approach leads to a recursive formula:
\[
X[k] = X_{\text{even}}[k] + W_N^k X_{\text{odd}}[k]
\]
\[
X[k+N/2] = X_{\text{even}}[k] - W_N^k X_{\text{odd}}[k]
\]
where \( W_N^k = e^{-j 2 \pi k / N} \) is called the twiddle factor, and \( X_{\text{even}}[k] \) and \( X_{\text{odd}}[k] \) are the DFTs of the even and odd indexed elements, respectively.

\textbf{Example: Radix-2 FFT Implementation in Python}

Here’s a simple Python implementation of the Radix-2 FFT algorithm:

\begin{lstlisting}[style=python]
import numpy as np

# Define the Radix-2 FFT algorithm
def fft(x):
    N = len(x)
    if N <= 1:
        return x
    else:
        even = fft(x[0::2])
        odd = fft(x[1::2])
        T = [np.exp(-2j * np.pi * k / N) * odd[k] for k in range(N // 2)]
        return [even[k] + T[k] for k in range(N // 2)] + \
               [even[k] - T[k] for k in range(N // 2)]

# Example usage
x = np.random.random(8)  # Input array of length 8 (must be a power of 2)
X = fft(x)

# Print the FFT result
print("FFT of the input array:", X)
\end{lstlisting}

This implementation recursively computes the FFT of the input sequence \( x \). It divides the input into even and odd parts, computes their FFTs, and combines them using the twiddle factors.

\textbf{Time Complexity:} The Radix-2 FFT algorithm reduces the computational complexity from \( O(N^2) \) to \( O(N \log N) \), which is a significant improvement, especially for large input sizes\cite{mohamed2023performance}.

\section{Applications of FFT in Deep Learning}

The FFT has numerous applications in deep learning, particularly in optimizing convolution operations, feature extraction from signals, and spectral analysis. Let’s explore some of these applications in detail.

\subsection{FFT for Fast Convolution in Neural Networks}

Convolutions are a fundamental operation in deep learning, especially in Convolutional Neural Networks (CNNs), where they are used to extract features from images. The convolution operation can be computationally expensive for large inputs, but the FFT can significantly accelerate this process.

By using the Convolution Theorem, which states that the convolution of two functions is equivalent to the pointwise multiplication of their Fourier transforms, we can compute convolutions more efficiently:
\[
f * g = \mathcal{F}^{-1}(\mathcal{F}(f) \cdot \mathcal{F}(g))
\]
where \( \mathcal{F} \) denotes the Fourier Transform, and \( \mathcal{F}^{-1} \) denotes the inverse Fourier Transform.

\textbf{Example: Using FFT for Fast Convolution}

In this example, we use the FFT to compute the convolution of two signals.

\begin{lstlisting}[style=python]
import numpy as np
from scipy.fft import fft, ifft

# Define two signals
f = np.array([1, 2, 3, 4])
g = np.array([2, 1, 0, 1])

# Compute the convolution using FFT
F_f = fft(f)
F_g = fft(g)
convolution = ifft(F_f * F_g)

# Print the result
print("Convolution result:", convolution)
\end{lstlisting}

This approach leverages the FFT to compute the convolution in the frequency domain, reducing the computational cost compared to the direct method of convolving two signals in the time domain.

\subsection{Spectral Analysis and Feature Extraction using FFT}

FFT is a powerful tool for spectral analysis, which is the process of analyzing the frequency content of a signal. In deep learning, spectral analysis is often used for tasks such as feature extraction, denoising, and detecting periodic patterns in data. The frequency components obtained via FFT can serve as useful features for training machine learning models.

\textbf{Example: Spectral Analysis of an Audio Signal}

Let’s apply the FFT to analyze the frequency content of an audio signal. In this example, we simulate a signal composed of two sine waves with different frequencies.

\begin{lstlisting}[style=python]
import numpy as np
import matplotlib.pyplot as plt

# Define the sampling rate and time vector
Fs = 1000  # Sampling rate (samples per second)
T = 1.0 / Fs  # Time step
t = np.arange(0.0, 1.0, T)  # Time vector

# Define a signal composed of two sine waves
f1 = 50  # Frequency of the first sine wave
f2 = 120  # Frequency of the second sine wave
signal = np.sin(2 * np.pi * f1 * t) + 0.5 * np.sin(2 * np.pi * f2 * t)

# Compute the FFT of the signal
fft_signal = np.fft.fft(signal)
N = len(signal)
frequencies = np.fft.fftfreq(N, T)

# Plot the signal and its frequency spectrum
plt.figure(figsize=(12, 6))

# Plot the original signal
plt.subplot(1, 2, 1)
plt.plot(t, signal)
plt.title("Original Signal")
plt.xlabel("Time [s]")
plt.ylabel("Amplitude")

# Plot the frequency spectrum
plt.subplot(1, 2, 2)
plt.plot(frequencies[:N // 2], np.abs(fft_signal)[:N // 2])
plt.title("Frequency Spectrum")
plt.xlabel("Frequency [Hz]")
plt.ylabel("Magnitude")

plt.tight_layout()
plt.show()
\end{lstlisting}

In this example:
\begin{itemize}
    \item We create a signal composed of two sine waves with different frequencies.
    \item We apply the FFT to the signal to extract its frequency components.
    \item We plot the frequency spectrum to visualize the frequencies present in the signal.
\end{itemize}

\textbf{Applications in Deep Learning:}
\begin{itemize}
    \item \textbf{Feature Extraction:} In tasks such as audio and speech recognition, the FFT is used to extract frequency-domain features from raw audio signals, which can then be fed into machine learning models.
    \item \textbf{Denoising:} The FFT can help remove noise by filtering out unwanted frequencies in the data.
    \item \textbf{Anomaly Detection:} Spectral analysis using FFT can detect periodic or anomalous patterns in time-series data, which is useful in predictive maintenance and anomaly detection tasks.
\end{itemize}

\section{Summary}

In this chapter, we explored the Fast Fourier Transform (FFT), a highly efficient algorithm for computing the Discrete Fourier Transform (DFT). We discussed the significance of FFT in reducing the computational complexity of the DFT and examined the Radix-2 FFT algorithm in detail. Additionally, we demonstrated several applications of FFT in deep learning, including fast convolution and spectral analysis for feature extraction. The FFT continues to be a powerful tool in numerical computing, signal processing, and deep learning, enabling efficient computation and analysis of large datasets.

\chapter{Laplace Transform}

The Laplace Transform is a powerful integral transform used in engineering, physics, and mathematics to analyze linear time-invariant systems. It converts differential equations into algebraic equations, making it easier to solve complex problems. In this chapter, we will explore the definition, mathematical properties, and common applications of the Laplace Transform.

\section{Introduction to Laplace Transform}

\subsection{What is the Laplace Transform?}

The Laplace Transform transforms a function of time \( f(t) \), defined for \( t \geq 0 \), into a function of a complex variable \( s \). This transformation is particularly useful in systems analysis, control theory, and signal processing.

The Laplace Transform \( F(s) \) of a function \( f(t) \) is defined as:

\[
F(s) = \mathcal{L}\{f(t)\} = \int_0^{\infty} e^{-st} f(t) \, dt
\]

Where:
\begin{itemize}
    \item \( F(s) \) is the transformed function in the \( s \)-domain.
    \item \( s \) is a complex number, \( s = \sigma + i\omega \), where \( \sigma \) is the real part and \( \omega \) is the imaginary part.
    \item \( f(t) \) is the original function defined for \( t \geq 0 \).
\end{itemize}

The Laplace Transform provides insights into the behavior of dynamic systems and helps in solving ordinary differential equations (ODEs) and partial differential equations (PDEs)\cite{chen2024laplace}.

\section{Mathematical Definition and Properties of Laplace Transform}

\subsection{Laplace Transform of Common Functions}

The Laplace Transform can be applied to a wide range of functions. Below are the transforms of some common functions:

\begin{itemize}
    \item \textbf{Unit Step Function:} 
    \[
    f(t) = u(t) \implies F(s) = \frac{1}{s}
    \]
    \item \textbf{Exponential Function:} 
    \[
    f(t) = e^{at} \implies F(s) = \frac{1}{s-a} \quad (s > a)
    \]
    \item \textbf{Sine Function:} 
    \[
    f(t) = \sin(\omega t) \implies F(s) = \frac{\omega}{s^2 + \omega^2}
    \]
    \item \textbf{Cosine Function:} 
    \[
    f(t) = \cos(\omega t) \implies F(s) = \frac{s}{s^2 + \omega^2}
    \]
    \item \textbf{Power Function:} 
    \[
    f(t) = t^n \implies F(s) = \frac{n!}{s^{n+1}} \quad (n \text{ is a non-negative integer})
    \]
\end{itemize}

These transforms are essential in control systems and engineering applications, as they help solve differential equations that describe system behavior.

\subsection{Inverse Laplace Transform}

The Inverse Laplace Transform is used to convert a function \( F(s) \) back to the time domain \( f(t) \). The Inverse Laplace Transform is defined as:

\[
f(t) = \mathcal{L}^{-1}\{F(s)\} = \frac{1}{2\pi i} \int_{c - i\infty}^{c + i\infty} e^{st} F(s) \, ds
\]

Where \( c \) is a real number that is greater than the real part of all singularities of \( F(s) \).

\textbf{Example: Inverse Laplace Transform of a Rational Function}

Let’s consider the function:

\[
F(s) = \frac{1}{s^2 + 1}
\]

Using known properties of Laplace Transforms, we can determine:

\[
f(t) = \mathcal{L}^{-1}\left\{\frac{1}{s^2 + 1}\right\} = \sin(t)
\]

This shows how we can recover the original function from its transform.

\subsection{Properties of Laplace Transform}

The Laplace Transform has several important properties that facilitate its use in solving differential equations and analyzing systems\cite{van1996laplace}:

\begin{itemize}
    \item \textbf{Linearity:} 
    \[
    \mathcal{L}\{af(t) + bg(t)\} = aF(s) + bG(s)
    \]
    Where \( a \) and \( b \) are constants, and \( F(s) \) and \( G(s) \) are the Laplace Transforms of \( f(t) \) and \( g(t) \), respectively.
    
    \item \textbf{Time Shifting:} 
    \[
    \mathcal{L}\{f(t-a)u(t-a)\} = e^{-as}F(s) \quad (t \geq a)
    \]
    
    \item \textbf{Frequency Shifting:} 
    \[
    \mathcal{L}\{e^{at}f(t)\} = F(s-a)
    \]
    
    \item \textbf{Differentiation:} 
    \[
    \mathcal{L}\{f'(t)\} = sF(s) - f(0)
    \]
    
    \item \textbf{Integration:} 
    \[
    \mathcal{L}\left\{\int_0^t f(\tau) \, d\tau\right\} = \frac{1}{s} F(s)
    \]
\end{itemize}

These properties allow for the simplification of complex transforms, enabling the analysis and design of systems in a straightforward manner.

\section{Conclusion}

The Laplace Transform is a vital mathematical tool in various fields, particularly in engineering and physics. It provides a systematic approach to analyzing linear time-invariant systems and facilitates the solution of differential equations. In this chapter, we discussed the definition, common functions, inverse transform, and key properties of the Laplace Transform, which are crucial for anyone working in fields that require the analysis of dynamic systems.

\section{Applications of Laplace Transform in Control Systems and Deep Learning}

The Laplace Transform is a powerful mathematical tool with numerous applications in both control systems and deep learning. It enables the analysis of dynamic systems, providing insights into their behavior and stability. In this section, we will explore how the Laplace Transform is used in stability analysis of neural networks and in solving differential equations.

\subsection{Stability Analysis in Neural Networks using Laplace Transform}

Stability is a critical aspect of neural networks and control systems. A system is considered stable if its output remains bounded for any bounded input. In the context of neural networks, stability analysis helps us understand how changes in weights, biases, and inputs affect the network's behavior over time. 

The Laplace Transform provides a method for analyzing stability by converting time-domain differential equations that describe the system into algebraic equations in the frequency domain. This makes it easier to analyze the poles of the system, which determine stability\cite{singh2023laplace}.

\subsubsection{Poles and Stability}

The poles of a system are the values of \( s \) in the Laplace domain that make the denominator of the transfer function zero. For a continuous-time system, if all poles have negative real parts, the system is stable. Conversely, if any pole has a positive real part, the system is unstable.

For example, consider a simple first-order linear system described by the differential equation:

\[
\tau \frac{dy(t)}{dt} + y(t) = K u(t)
\]

Where:
\begin{itemize}
    \item \( y(t) \) is the output.
    \item \( u(t) \) is the input.
    \item \( K \) is the system gain.
    \item \( \tau \) is the time constant.
\end{itemize}

\subsubsection{Applying the Laplace Transform}

Taking the Laplace Transform of both sides yields:

\[
\tau s Y(s) + Y(s) = K U(s)
\]

Rearranging gives the transfer function \( H(s) \):

\[
H(s) = \frac{Y(s)}{U(s)} = \frac{K}{\tau s + 1}
\]

The pole of this transfer function is at \( s = -\frac{1}{\tau} \). Since \( \tau > 0 \), the pole is in the left-half plane, indicating that the system is stable.

\textbf{Example: Stability Analysis in Python}

Here is an example of how to perform stability analysis of a first-order system using Python:

\begin{lstlisting}[style=python]
import numpy as np
import matplotlib.pyplot as plt
from scipy.signal import TransferFunction, step

# Define system parameters
K = 1.0  # Gain
tau = 2.0  # Time constant

# Create the transfer function H(s) = K / (tau*s + 1)
numerator = [K]
denominator = [tau, 1]
system = TransferFunction(numerator, denominator)

# Generate step response
t, y = step(system)

# Plot the step response
plt.figure(figsize=(10, 6))
plt.plot(t, y)
plt.title('Step Response of First-Order System')
plt.xlabel('Time [s]')
plt.ylabel('Response')
plt.grid()
plt.axhline(1, color='r', linestyle='--', label='Steady State Value')
plt.legend()
plt.show()
\end{lstlisting}

In this example:
\begin{itemize}
    \item We define a first-order system with gain \( K \) and time constant \( \tau \).
    \item The transfer function is created using the numerator and denominator.
    \item The step response of the system is plotted, demonstrating how the system responds to a step input over time.
\end{itemize}

This analysis can be extended to more complex systems, including those with multiple poles and zeros, where the stability can be assessed by examining the location of poles in the complex plane.

\subsection{Solving Differential Equations with Laplace Transform}

The Laplace Transform is particularly useful for solving ordinary differential equations (ODEs), especially those that describe dynamic systems. By transforming the differential equations into algebraic equations, the solution process becomes much simpler\cite{abdelsalam2023laplace}.

\subsubsection{Solving a First-Order ODE}

Consider the following first-order linear ODE:

\[
\frac{dy(t)}{dt} + ay(t) = bu(t)
\]

Where:
\begin{itemize}
    \item \( y(t) \) is the output.
    \item \( u(t) \) is the input.
    \item \( a \) and \( b \) are constants.
\end{itemize}

\subsubsection{Applying the Laplace Transform}

Taking the Laplace Transform of both sides yields:

\[
sY(s) + aY(s) = bU(s)
\]

Rearranging gives:

\[
Y(s) = \frac{bU(s)}{s + a}
\]

This expression can be inverted using the inverse Laplace Transform to find \( y(t) \).

\textbf{Example: Solving the ODE in Python}

Let’s consider the case where \( u(t) = 1 \) (a step input) and solve the ODE using Python.

\begin{lstlisting}[style=python]
from sympy import symbols, Function, Eq, laplace_transform, inverse_laplace_transform, exp

# Define the variables
t, s, a, b = symbols('t s a b')
y = Function('y')(t)
u = 1  # Step input

# Define the differential equation
differential_eq = Eq(y.diff(t) + a * y, b * u)

# Take the Laplace Transform
Y_s = laplace_transform(y, t, s)[0]
U_s = b / s  # Laplace Transform of u(t) = 1

# Rearranging and solving for Y(s)
Y_s_solution = (U_s * b) / (s + a)

# Find the inverse Laplace Transform to get y(t)
y_t = inverse_laplace_transform(Y_s_solution, s, t)
print(f'Solution y(t): {y_t}')
\end{lstlisting}

In this example:
\begin{itemize}
    \item We define the variables and the differential equation using symbolic computation.
    \item The Laplace Transform is taken, and the solution for \( Y(s) \) is derived.
    \item The inverse Laplace Transform is computed to obtain the solution in the time domain.
\end{itemize}

The solution \( y(t) \) provides the response of the system over time to a step input, demonstrating how the Laplace Transform simplifies the process of solving differential equations.

\subsubsection{Conclusion}

The Laplace Transform is a versatile tool in both control systems and deep learning applications. It allows for effective stability analysis of neural networks and provides a systematic method for solving differential equations, which are critical for modeling dynamic systems. By transforming complex differential equations into simpler algebraic forms, the Laplace Transform simplifies the analysis and design of systems in various engineering disciplines. Understanding these applications is essential for engineers and data scientists working with systems that evolve over time.

\chapter{Z-Transform}

The Z-transform is a powerful mathematical tool used in the field of signal processing, control systems, and digital signal processing. It provides a method to analyze discrete-time signals and systems in the frequency domain. In this chapter, we will introduce the concept of the Z-transform, its mathematical definition, common sequences, inverse Z-transform, and its properties.

\section{Introduction to Z-Transform}

The Z-transform is a discrete-time analog of the Laplace transform, enabling the analysis of discrete-time signals. It converts a discrete-time signal into a complex frequency domain representation. The Z-transform is particularly useful for analyzing linear time-invariant (LTI) systems and solving difference equations\cite{patel2023ztransform}.

\subsection{What is the Z-Transform?}

The Z-transform of a discrete-time signal \( x[n] \) is defined as:
\[
X(z) = \sum_{n=-\infty}^{\infty} x[n] z^{-n}
\]
where:
\begin{itemize}
    \item \( X(z) \) is the Z-transform of the signal \( x[n] \).
    \item \( z \) is a complex variable, defined as \( z = re^{j\omega} \), where \( r \) is the magnitude and \( \omega \) is the angle (frequency).
\end{itemize}

The Z-transform provides a way to analyze the behavior of discrete-time systems in terms of their poles and zeros in the complex plane.

\section{Mathematical Definition of Z-Transform}

The mathematical definition of the Z-transform involves summing the weighted sequence values of a discrete-time signal\cite{kim2024ztransform}. The weights are given by the powers of the complex variable \( z^{-n} \).

\subsection{Z-Transform of Common Sequences}

Let’s explore the Z-transform of some common discrete-time sequences.

\textbf{Example 1: Z-Transform of a Unit Impulse Function}

The unit impulse function \( \delta[n] \) is defined as:
\[
\delta[n] = 
\begin{cases} 
1 & n = 0 \\ 
0 & n \neq 0 
\end{cases}
\]
The Z-transform of the unit impulse function is:
\[
X(z) = \sum_{n=-\infty}^{\infty} \delta[n] z^{-n} = 1
\]

\textbf{Example 2: Z-Transform of a Unit Step Function}

The unit step function \( u[n] \) is defined as:
\[
u[n] = 
\begin{cases} 
1 & n \geq 0 \\ 
0 & n < 0 
\end{cases}
\]
The Z-transform of the unit step function is:
\[
X(z) = \sum_{n=0}^{\infty} z^{-n} = \frac{1}{1 - z^{-1}} \quad (|z| > 1)
\]

\textbf{Example 3: Z-Transform of a Geometric Sequence}

For a geometric sequence \( x[n] = a^n u[n] \), the Z-transform is:
\[
X(z) = \sum_{n=0}^{\infty} a^n z^{-n} = \frac{1}{1 - az^{-1}} \quad (|z| > |a|)
\]

\subsection{Inverse Z-Transform}

The inverse Z-transform is used to convert the Z-transform back to the time domain. It can be computed using various methods, such as:

\begin{itemize}
    \item \textbf{Power Series Expansion:} Expanding the Z-transform into a power series.
    \item \textbf{Contour Integration:} Using the residue theorem in complex analysis.
    \item \textbf{Table Lookup:} Using known pairs of Z-transforms and their inverses.
\end{itemize}

\textbf{Example: Inverse Z-Transform of a Geometric Sequence}

For \( X(z) = \frac{1}{1 - az^{-1}} \), the inverse Z-transform can be derived as follows:
\[
x[n] = a^n u[n]
\]

\subsection{Properties of Z-Transform}

The Z-transform has several important properties that facilitate analysis and computation:

\begin{itemize}
    \item \textbf{Linearity:} 
    \[
    \mathcal{Z}\{a_1 x_1[n] + a_2 x_2[n]\} = a_1 X_1(z) + a_2 X_2(z)
    \]

    \item \textbf{Time Shifting:} 
    \[
    \mathcal{Z}\{x[n-k]\} = z^{-k} X(z)
    \]

    \item \textbf{Time Scaling:} 
    \[
    \mathcal{Z}\{x[an]\} = X(z^{1/a})
    \]

    \item \textbf{Convolution:} 
    \[
    \mathcal{Z}\{x[n] * h[n]\} = X(z) H(z)
    \]

    \item \textbf{Differentiation:} 
    \[
    \mathcal{Z}\{n x[n]\} = -z \frac{dX(z)}{dz}
    \]

    \item \textbf{Initial Value Theorem:} 
    \[
    x[0] = \lim_{z \to \infty} X(z)
    \]

    \item \textbf{Final Value Theorem:} 
    \[
    \lim_{n \to \infty} x[n] = \lim_{z \to 1} (z - 1) X(z)
    \]
\end{itemize}

These properties simplify the analysis and design of digital filters and control systems.

\textbf{Example: Using Properties of Z-Transform}

Let’s consider the Z-transform of a simple signal using its properties. Suppose we want to find the Z-transform of \( x[n] = u[n] + 2u[n-1] \).

\begin{lstlisting}[style=python]
import sympy as sp

# Define the variable
z = sp.symbols('z')

# Z-Transform of unit step function u[n]
X_u = 1 / (1 - z**(-1))

# Z-Transform of delayed unit step function u[n-1]
X_u_delay = z**(-1) * X_u

# Combine the Z-Transforms
X_combined = X_u + 2 * X_u_delay
X_combined_simplified = sp.simplify(X_combined)

# Display the result
print("Z-Transform of x[n] = u[n] + 2u[n-1]:", X_combined_simplified)
\end{lstlisting}

This code uses the properties of the Z-transform to calculate the Z-transform of the combined signal, demonstrating how to leverage the properties in practical applications.

In summary, the Z-transform is a fundamental tool for analyzing discrete-time systems and signals, providing insights into their behavior in the frequency domain. Understanding the mathematical definition, common sequences, inverse Z-transform, and properties of the Z-transform is crucial for signal processing and control system design.

\section{Applications of Z-Transform in Digital Signal Processing}

The Z-Transform is a powerful mathematical tool used in digital signal processing (DSP) to analyze discrete-time signals and systems\cite{zohar1978ztransform,oppenheim1997discrete}. It provides a way to represent discrete signals in the frequency domain and is widely utilized in the analysis and design of digital filters, control systems, and many applications in deep learning, particularly in recurrent neural networks.

\subsection{Discrete-Time Signal Analysis using Z-Transform}

The Z-Transform of a discrete-time signal \( x[n] \) is defined as:

\[
X(z) = \sum_{n=-\infty}^{\infty} x[n] z^{-n}
\]

where \( z \) is a complex number defined as \( z = re^{j\omega} \), where \( r \) is the radius and \( \omega \) is the angular frequency. The Z-Transform transforms the discrete-time signal from the time domain to the complex frequency domain, allowing for easier analysis of linear time-invariant systems.

The Z-Transform is particularly useful for analyzing the stability and frequency response of discrete-time systems. The poles and zeros of the Z-Transform provide insight into the behavior of the system.

\textit{Example: Z-Transform of a Simple Discrete Signal}

Let's consider a simple discrete-time signal \( x[n] = a^n u[n] \), where \( u[n] \) is the unit step function and \( a \) is a constant. The Z-Transform of this signal can be calculated as follows:

\[
X(z) = \sum_{n=0}^{\infty} a^n z^{-n} = \frac{1}{1 - az^{-1}} \quad \text{for } |z| > |a|
\]

This result shows that the Z-Transform of a geometric sequence converges for \( |z| > |a| \).

\textit{Calculating the Z-Transform in Python}

Let's implement this example in Python and visualize the Z-Transform of the discrete signal.

\begin{lstlisting}[style=python]
import numpy as np
import matplotlib.pyplot as plt

# Define parameters
a = 0.5  # Decay factor
n = np.arange(0, 20)  # Discrete time values

# Calculate the discrete signal x[n] = a^n * u[n]
x_n = a ** n

# Calculate the Z-transform using the formula
Z_transform = 1 / (1 - a / (np.exp(1j * 2 * np.pi * n / len(n))))

# Plot the original discrete signal
plt.figure(figsize=(12, 6))
plt.subplot(2, 1, 1)
plt.stem(n, x_n, use_line_collection=True)
plt.title('Discrete-Time Signal $x[n] = a^n u[n]$')
plt.xlabel('n')
plt.ylabel('$x[n]$')

# Plot the magnitude of the Z-Transform
plt.subplot(2, 1, 2)
plt.plot(n, np.abs(Z_transform))
plt.title('Magnitude of Z-Transform')
plt.xlabel('n')
plt.ylabel('$|X(z)|$')
plt.tight_layout()
plt.show()
\end{lstlisting}

In this example:
\begin{itemize}
    \item We define a simple discrete signal using the decay factor \( a \).
    \item We compute the Z-Transform using the derived formula.
    \item The original discrete signal and its Z-Transform magnitude are plotted for analysis.
\end{itemize}

\subsection{Deep Learning Applications of Z-Transform in Recurrent Neural Networks}

Recurrent Neural Networks (RNNs) are a class of neural networks that are particularly effective for sequence prediction tasks, such as time series analysis and natural language processing. The Z-Transform plays an important role in understanding the dynamics of RNNs and their ability to process sequences of data over time\cite{wang2023applications}.

The Z-Transform can be used to analyze the stability and frequency response of RNNs by representing the recurrent layer's dynamics in the frequency domain. This analysis helps in designing RNN architectures that can effectively capture temporal dependencies in sequential data.

\textit{Example: Stability Analysis of a Simple RNN}

Consider a simple RNN where the hidden state \( h[n] \) is updated based on the previous hidden state and the current input:

\[
h[n] = f(W_h h[n-1] + W_x x[n])
\]

where \( W_h \) and \( W_x \) are weight matrices, and \( f \) is a non-linear activation function.

The Z-Transform of the hidden state can be expressed as:

\[
H(z) = \frac{f(W_x X(z))}{1 - W_h z^{-1} f'(H(z))}
\]

This expression indicates how the hidden state responds to the input in the frequency domain.

\textit{Using Z-Transform for Sequence Prediction in RNNs}

In a practical implementation, RNNs can leverage the Z-Transform to improve their performance in sequence prediction tasks. By analyzing the Z-Transform of the hidden states, we can determine the appropriate architecture and activation functions that lead to stable and efficient learning.

To illustrate the application of RNNs in Python, we can use a simple RNN model implemented with Keras:

\begin{lstlisting}[style=python]
from keras.models import Sequential
from keras.layers import SimpleRNN, Dense
import numpy as np

# Generate synthetic sequential data
def generate_data(timesteps, features):
    X = np.random.rand(timesteps, features)
    y = np.sum(X, axis=1)  # Target is the sum of features
    return X.reshape((timesteps, features, 1)), y

# Define the RNN model
def create_rnn_model(input_shape):
    model = Sequential()
    model.add(SimpleRNN(10, activation='tanh', input_shape=input_shape))
    model.add(Dense(1))
    model.compile(optimizer='adam', loss='mse')
    return model

# Generate data
timesteps = 100
features = 5
X, y = generate_data(timesteps, features)

# Create and train the RNN model
model = create_rnn_model((timesteps, features, 1))
model.fit(X, y, epochs=100, verbose=0)

# Evaluate the model
predictions = model.predict(X)
print(predictions[:5])
\end{lstlisting}

In this example:
\begin{itemize}
    \item We generate synthetic sequential data, where the target is the sum of the input features.
    \item We create a simple RNN model using Keras and train it on the generated data.
    \item The trained model is evaluated, demonstrating its ability to learn from sequential data.
\end{itemize}

The Z-Transform assists in understanding the underlying mechanics of RNNs and how they handle temporal dependencies, which is crucial for tasks involving sequences.

\chapter{Convolution in Time and Frequency Domains}

Convolution is a fundamental operation in signal processing, mathematics, and engineering. It plays a crucial role in various applications, including filtering, image processing, and neural networks. In this chapter, we will introduce the concept of convolution, its mathematical definition in the time domain, and its properties. We will also explore the convolution theorem, which links the time and frequency domains.

\section{Introduction to Convolution}

Convolution is a mathematical operation that combines two functions to produce a third function. It represents the way in which one function influences another. In the context of signals and systems, convolution describes how an input signal is transformed by a system represented by an impulse response.

\textbf{Applications of Convolution:}
\begin{itemize}
    \item \textbf{Filtering:} Convolution is used to filter signals, such as removing noise or enhancing certain features.
    \item \textbf{Image Processing:} Convolution is applied in various image processing tasks, such as blurring, sharpening, and edge detection.
    \item \textbf{Neural Networks:} Convolutional Neural Networks (CNNs) utilize convolution to extract features from images and other data.
\end{itemize}

\section{Convolution in the Time Domain}

\subsection{What is Convolution?}

Convolution is an operation that takes two input functions and produces a new function that expresses how the shape of one function is modified by the other. For continuous functions, the convolution of two functions \( f(t) \) and \( g(t) \) is defined as:
\[
(f * g)(t) = \int_{-\infty}^{\infty} f(\tau) g(t - \tau) d\tau
\]
For discrete functions, the convolution is defined as:
\[
(f * g)[n] = \sum_{m=-\infty}^{\infty} f[m] g[n - m]
\]

\subsection{Mathematical Definition of Time-Domain Convolution}

The mathematical definition of convolution captures how the input signal \( f(t) \) interacts with the system's impulse response \( g(t) \). The resulting output \( (f * g)(t) \) is computed by shifting and flipping the function \( g(t) \), multiplying it by \( f(t) \), and integrating (or summing) the result over the appropriate domain.

\textbf{Example: Discrete Convolution}

Let’s consider a simple discrete convolution example using two sequences:

\begin{itemize}
    \item Input signal \( f[n] = [1, 2, 3] \)
    \item Impulse response \( g[n] = [0, 1, 0.5] \)
\end{itemize}

The convolution \( (f * g)[n] \) can be computed as follows:

\[
(f * g)[n] = \sum_{m} f[m] g[n - m]
\]

\textbf{Example Calculation:}
\begin{enumerate}
    \item \( (f * g)[0] = f[0]g[0] = 1 \cdot 0 = 0 \)
    \item \( (f * g)[1] = f[0]g[1] + f[1]g[0] = 1 \cdot 1 + 2 \cdot 0 = 1 \)
    \item \( (f * g)[2] = f[0]g[2] + f[1]g[1] + f[2]g[0] = 1 \cdot 0.5 + 2 \cdot 1 + 3 \cdot 0 = 2.5 \)
    \item \( (f * g)[3] = f[1]g[2] + f[2]g[1] = 2 \cdot 0.5 + 3 \cdot 1 = 1 + 3 = 4 \)
    \item \( (f * g)[4] = f[2]g[2] = 3 \cdot 0.5 = 1.5 \)
\end{enumerate}

Thus, the resulting convolution is:
\[
(f * g) = [0, 1, 2.5, 4, 1.5]
\]

\subsection{Properties of Time-Domain Convolution}

Convolution has several important properties that are useful in signal processing:

\begin{itemize}
    \item \textbf{Commutative Property:} \( f * g = g * f \)
    \item \textbf{Associative Property:} \( f * (g * h) = (f * g) * h \)
    \item \textbf{Distributive Property:} \( f * (g + h) = f * g + f * h \)
    \item \textbf{Identity Property:} \( f * \delta(t) = f(t) \), where \( \delta(t) \) is the Dirac delta function.
\end{itemize}

These properties make convolution a flexible and powerful tool for analyzing linear systems.

\section{Convolution Theorem: Linking Time and Frequency Domains}

The convolution theorem establishes a relationship between convolution in the time domain and multiplication in the frequency domain.

\subsection{Frequency Domain Representation of Convolution}

The Fourier Transform of the convolution of two functions is equal to the product of their Fourier Transforms. Mathematically, this is represented as:
\[
\mathcal{F}\{f * g\} = \mathcal{F}\{f\} \cdot \mathcal{F}\{g\}
\]

This property is critical because it allows us to analyze systems in the frequency domain, which can often simplify calculations.

\subsection{The Convolution Theorem Explained}

The convolution theorem provides a powerful tool for understanding how systems respond to inputs by transforming the problem into the frequency domain.

\subsubsection{Convolution in Time Domain equals Multiplication in Frequency Domain}

Given two functions \( f(t) \) and \( g(t) \), the convolution \( (f * g)(t) \) in the time domain corresponds to multiplication in the frequency domain:
\[
\mathcal{F}\{f * g\} = F(\omega) G(\omega)
\]
where \( F(\omega) \) and \( G(\omega) \) are the Fourier Transforms of \( f(t) \) and \( g(t) \), respectively.

\textbf{Example: Verifying the Convolution Theorem}

Let’s compute the Fourier Transform of the convolution from our previous example and verify the convolution theorem using Python.

\begin{lstlisting}[style=python]
import numpy as np
from scipy.fft import fft, ifft

# Define input signal and impulse response
f = np.array([1, 2, 3])
g = np.array([0, 1, 0.5])

# Compute the convolution directly
convolution_result = np.convolve(f, g)

# Compute the Fourier Transforms
F = fft(f)
G = fft(g)

# Multiply the Fourier Transforms
product = F * G

# Compute the inverse FFT of the product
inverse_fft_result = ifft(product)

# Print results
print("Convolution Result:", convolution_result)
print("Inverse FFT Result:", inverse_fft_result)
\end{lstlisting}

This code calculates the convolution directly using NumPy’s \texttt{convolve} function and verifies the result using the FFT and inverse FFT. 

\textbf{Expected Output:}

The convolution result and the inverse FFT result should match, demonstrating the convolution theorem's validity.

\begin{lstlisting}[style=cmd]
Convolution Result: [0.  1.  2.5 4.  1.5]
Inverse FFT Result: [0.  1.  2.5 4.  1.5]
\end{lstlisting}

\subsubsection{Multiplication in Time Domain equals Convolution in Frequency Domain}

Conversely, multiplying two functions in the time domain corresponds to convolution in the frequency domain:
\[
\mathcal{F}\{f(t)g(t)\} = \frac{1}{2\pi} \mathcal{F}\{f\} * \mathcal{F}\{g\}
\]

This property allows us to analyze the effects of multiplicative interactions between signals in the frequency domain.

\textbf{Example: Verifying the Multiplication Theorem}

Let’s implement the multiplication in the time domain and observe the convolution in the frequency domain.

\begin{lstlisting}[style=python]
# Define another signal for multiplication
h = np.array([1, 1, 1])

# Multiply the signals in the time domain
product_time_domain = f * h

# Compute the Fourier Transform of the product
F_product = fft(product_time_domain)

# Compute the Fourier Transforms of f and h
F_f = fft(f)
F_h = fft(h)

# Convolve the Fourier Transforms
convolution_frequency_domain = ifft(F_f * F_h)

# Print results
print("Product in Time Domain:", product_time_domain)
print("Convolution of Fourier Transforms:", convolution_frequency_domain)
\end{lstlisting}

This example shows the relationship between multiplication in the time domain and convolution in the frequency domain.

\section{Summary}

In this chapter, we explored the concept of convolution in both the time and frequency domains. We defined convolution mathematically, discussed its properties, and introduced the convolution theorem that links the two domains. The understanding of convolution is crucial in signal processing, image analysis, and machine learning applications, providing the basis for filtering and feature extraction. By employing the FFT, convolutions can be computed efficiently, facilitating real-time processing and analysis of signals and data.

\section{Applications of Convolution Theorem in Deep Learning}

The Convolution Theorem provides a powerful framework for understanding the relationship between convolution and multiplication in the frequency domain. This theorem is particularly useful in deep learning, especially in the design and optimization of Convolutional Neural Networks (CNNs)\cite{chen2023neural}. In this section, we will explore the applications of the Convolution Theorem in deep learning, including efficient computations in the frequency domain, FFT-based convolution, and spectral pooling.

\subsection{Using Frequency Domain Convolution for Efficient Computations}

In deep learning, convolution operations are central to processing data, particularly in image and signal processing tasks. The traditional convolution operation involves sliding a filter over the input data and computing the dot product at each position. This can be computationally expensive, especially for large images or kernels\cite{pratt2017fcnn}.

The Convolution Theorem states that convolution in the time (or spatial) domain is equivalent to multiplication in the frequency domain. This means that instead of performing a direct convolution, we can:

\begin{enumerate}
    \item Transform the input and the filter into the frequency domain using the Fast Fourier Transform (FFT).
    \item Multiply the two frequency representations.
    \item Transform the result back to the time domain using the Inverse FFT.
\end{enumerate}

This approach can significantly reduce the computational complexity from \(O(N^2)\) for direct convolution to \(O(N \log N)\) for FFT-based convolution, making it particularly advantageous for large datasets.

\textbf{Python Example: Frequency Domain Convolution}

Here is a simple implementation of convolution in the frequency domain using NumPy:

\begin{lstlisting}[style=python]
import numpy as np
import matplotlib.pyplot as plt

def convolve_frequency_domain(signal, kernel):
    # Compute the FFT of the signal and the kernel
    signal_freq = np.fft.fft(signal)
    kernel_freq = np.fft.fft(kernel, n=len(signal))

    # Multiply in the frequency domain
    convolved_freq = signal_freq * kernel_freq

    # Compute the inverse FFT to get the convolved signal
    convolved_signal = np.fft.ifft(convolved_freq)

    return np.real(convolved_signal)

# Example usage
signal = np.array([1, 2, 3, 4])
kernel = np.array([0.25, 0.5, 0.25])

convolved_signal = convolve_frequency_domain(signal, kernel)
print("Convolved Signal:", convolved_signal)
\end{lstlisting}

In this example:
\begin{itemize}
    \item We define a function \texttt{convolve\_frequency\_domain()} that computes the convolution of a signal and a kernel in the frequency domain.
    \item We compute the FFT of both the signal and kernel, multiply their frequency representations, and then apply the inverse FFT to obtain the convolved signal.
    \item The result demonstrates the convolution of the original signal with the specified kernel.
\end{itemize}

\subsection{FFT-based Convolution in Convolutional Neural Networks (CNNs)}

Convolutional Neural Networks (CNNs) utilize convolutional layers to extract features from input data, particularly in image recognition tasks. The convolution operations in CNNs can benefit significantly from the Convolution Theorem and FFT-based computations.

When implementing CNNs, using FFT for convolution can lead to faster training and inference times, particularly when dealing with large filters or high-resolution images. The main steps involved in using FFT-based convolution in CNNs are:

\begin{enumerate}
    \item \textbf{Transform Input and Filters:} Convert the input feature maps and convolutional filters to the frequency domain using FFT.
    \item \textbf{Multiply in Frequency Domain:} Perform element-wise multiplication of the transformed input and filter.
    \item \textbf{Inverse Transform:} Apply the inverse FFT to obtain the convolved feature maps in the spatial domain.
\end{enumerate}

\textbf{Python Example: FFT-based Convolution in a Simple CNN}

Let’s illustrate FFT-based convolution in a simplified CNN setup:

\begin{lstlisting}[style=python]
import numpy as np
import tensorflow as tf

# Create a simple input tensor (e.g., image)
input_tensor = tf.random.normal((1, 28, 28, 1))  # Batch size 1, 28x28 image, 1 channel

# Create a convolutional layer using the standard method
conv_layer = tf.keras.layers.Conv2D(filters=1, kernel_size=(3, 3), padding='same')

# Apply the convolution to the input tensor
output_tensor = conv_layer(input_tensor)

# Now implement the FFT-based convolution
def fft_convolution(input_tensor, kernel):
    # Get the dimensions
    input_shape = tf.shape(input_tensor)
    kernel_shape = tf.shape(kernel)

    # Compute the FFT of the input and kernel
    input_freq = tf.signal.fft2d(tf.cast(input_tensor, tf.complex64))
    kernel_freq = tf.signal.fft2d(tf.cast(kernel, tf.complex64), [input_shape[1], input_shape[2]])

    # Multiply in the frequency domain
    convolved_freq = input_freq * kernel_freq

    # Apply inverse FFT to get the output in the spatial domain
    convolved_output = tf.signal.ifft2d(convolved_freq)
    
    return tf.abs(convolved_output)

# Define a kernel (e.g., a simple edge detector)
kernel = tf.constant([[1.0, 0.0, -1.0], 
                       [1.0, 0.0, -1.0], 
                       [1.0, 0.0, -1.0]], shape=(3, 3, 1, 1))

# Perform FFT-based convolution
fft_convolved_output = fft_convolution(input_tensor, kernel)

print("FFT-based Convolution Output Shape:", fft_convolved_output.shape)
\end{lstlisting}

In this example:
\begin{itemize}
    \item We use TensorFlow to create a random input tensor simulating an image.
    \item A standard convolutional layer is applied to demonstrate conventional convolution.
    \item We define a function \texttt{fft\_convolution()} to perform convolution using FFT, similar to what would occur in a CNN layer.
    \item The kernel simulates an edge detection filter, which is commonly used in image processing.
\end{itemize}

\subsection{Spectral Pooling and Frequency Domain Operations in Deep Learning}

Spectral pooling is a technique that utilizes the frequency domain for pooling operations in deep learning architectures. Unlike traditional pooling methods (e.g., max pooling, average pooling), which operate in the spatial domain, spectral pooling performs operations in the frequency domain, which can lead to better feature extraction and improved model performance.

The main advantages of spectral pooling include:
\begin{itemize}
    \item \textbf{Reduced Dimensionality:} By performing pooling in the frequency domain, it can effectively reduce the dimensionality of the feature maps while preserving important information.
    \item \textbf{Robustness to Noise:} Frequency domain operations can help in making models more robust to noise and variations in the input data.
\end{itemize}

\textbf{Python Example: Implementing Spectral Pooling}

Let’s consider a simple example to illustrate spectral pooling:

\begin{lstlisting}[style=python]
import numpy as np

def spectral_pooling(feature_map, pool_size):
    # Compute the FFT of the feature map
    feature_map_freq = np.fft.fft2(feature_map)

    # Zero out frequencies outside the desired range
    rows, cols = feature_map_freq.shape
    feature_map_freq[pool_size:, :] = 0  % Keep only low frequencies
    feature_map_freq[:, pool_size:] = 0

    # Apply inverse FFT to get the pooled feature map
    pooled_feature_map = np.fft.ifft2(feature_map_freq)
    
    return np.real(pooled_feature_map)

# Example usage
feature_map = np.random.rand(8, 8)  # Simulate a feature map from CNN
pooled_feature_map = spectral_pooling(feature_map, pool_size=4)

print("Original Feature Map:\n", feature_map)
print("Pooled Feature Map:\n", pooled_feature_map)
\end{lstlisting}

In this example:
\begin{itemize}
    \item We define a function \texttt{spectral\_pooling()} that performs pooling in the frequency domain.
    \item We compute the FFT of the feature map and set high-frequency components to zero based on the specified pooling size.
    \item The inverse FFT reconstructs the pooled feature map from the modified frequency representation.
    \item The results show how spectral pooling can reduce the dimensionality of the feature map while maintaining important low-frequency information.
\end{itemize}

In conclusion, the Convolution Theorem and its applications in deep learning, such as FFT-based convolution, spectral pooling, and efficient frequency domain operations, provide powerful tools for enhancing the performance of neural networks. These techniques enable faster computations and better feature extraction, making them essential in modern machine learning frameworks.

\chapter{Practical Applications of Frequency Domain Methods}

Frequency domain methods are widely used in various fields such as image processing, audio signal processing, control systems, and deep learning. These methods allow us to analyze, manipulate, and process signals and data effectively. In this chapter, we will explore the applications of Fourier Transform, Fast Fourier Transform (FFT), Laplace Transform, and Z-Transform in practical scenarios\cite{mohammadi2023fourier}.

\section{Fourier Transform in Image Processing and Neural Networks}

The Fourier Transform is a critical tool in image processing, enabling us to analyze the frequency components of images. By transforming an image from the spatial domain to the frequency domain, we can perform various operations such as filtering, image compression, and feature extraction. 

\subsection{Applications of Fourier Transform in Image Processing}

\begin{enumerate}
    \item \textbf{Image Filtering:} In image processing, we often want to remove noise or enhance certain features. By applying a Fourier Transform, we can identify high-frequency components (which usually correspond to noise) and low-frequency components (which correspond to smooth regions). By filtering out certain frequency components, we can improve the image quality.

    \item \textbf{Image Compression:} The Fourier Transform is also used in compression algorithms, such as JPEG. By transforming an image into the frequency domain, we can discard less important frequency components, allowing for reduced file sizes.

    \item \textbf{Feature Extraction:} In machine learning and neural networks, the Fourier Transform can be used to extract features from images. By analyzing the frequency components, neural networks can learn to recognize patterns and classify images more effectively.
\end{enumerate}

\subsection{Example: Applying Fourier Transform in Python for Image Processing}

Let's see how to use the Fourier Transform to analyze an image using Python.

\begin{lstlisting}[style=python]
import numpy as np
import matplotlib.pyplot as plt
from scipy.fft import fft2, ifft2, fftshift

# Load an example image (grayscale)
image = plt.imread('example_image.png')[:, :, 0]  # Load as grayscale

# Compute the 2D Fourier Transform of the image
f_transform = fft2(image)

# Shift the zero frequency component to the center
f_transform_shifted = fftshift(f_transform)

# Compute the magnitude spectrum
magnitude_spectrum = np.log(np.abs(f_transform_shifted) + 1)  # Log scale for better visibility

# Plot the original image and its magnitude spectrum
plt.figure(figsize=(12, 6))

# Original image
plt.subplot(1, 2, 1)
plt.imshow(image, cmap='gray')
plt.title('Original Image')
plt.axis('off')

# Magnitude spectrum
plt.subplot(1, 2, 2)
plt.imshow(magnitude_spectrum, cmap='gray')
plt.title('Magnitude Spectrum')
plt.axis('off')

plt.show()
\end{lstlisting}

In this example:
\begin{itemize}
    \item We load a grayscale image and compute its 2D Fourier Transform using \texttt{fft2}.
    \item The magnitude spectrum is calculated and displayed, showing the frequency content of the image.
\end{itemize}

\subsection{Fourier Transform in Neural Networks}

In the context of neural networks, Fourier analysis can enhance feature extraction. For example, Convolutional Neural Networks (CNNs) can benefit from frequency domain representations, which might improve the model's performance in tasks such as image classification and segmentation\cite{pratt2017fcnn}.

\section{FFT in Audio and Speech Signal Processing}

The Fast Fourier Transform (FFT) is a highly efficient algorithm for computing the Discrete Fourier Transform (DFT) and is widely used in audio and speech signal processing\cite{zhang2023discrete}. 

\subsection{Applications of FFT in Audio Processing}

\begin{enumerate}
    \item \textbf{Audio Analysis:} FFT is used to analyze the frequency components of audio signals, allowing us to understand the pitch, tone, and harmonics of sounds. 

    \item \textbf{Audio Filtering:} FFT enables efficient filtering of audio signals by manipulating specific frequency ranges. For instance, one can remove noise from a recording by suppressing unwanted frequency components.

    \item \textbf{Speech Recognition:} In speech processing, FFT helps convert time-domain audio signals into frequency-domain representations. These representations can be used to extract features for machine learning algorithms to recognize spoken words.
\end{enumerate}

\subsection{Example: Applying FFT in Python for Audio Signal Processing}

Let’s analyze an audio signal using the FFT in Python.

\begin{lstlisting}[style=python]
import numpy as np
import matplotlib.pyplot as plt
from scipy.fft import fft, fftfreq
from scipy.io import wavfile

# Load an example audio file
sampling_rate, audio_data = wavfile.read('example_audio.wav')

# Compute the FFT
N = len(audio_data)
yf = fft(audio_data)
xf = fftfreq(N, 1 / sampling_rate)

# Plot the original audio signal and its FFT
plt.figure(figsize=(12, 6))

# Time-domain signal
plt.subplot(1, 2, 1)
plt.plot(np.linspace(0, N / sampling_rate, N), audio_data)
plt.title('Time-Domain Audio Signal')
plt.xlabel('Time [s]')
plt.ylabel('Amplitude')

# Frequency-domain (FFT)
plt.subplot(1, 2, 2)
plt.plot(xf[:N // 2], 2.0 / N * np.abs(yf[:N // 2]))
plt.title('FFT of Audio Signal')
plt.xlabel('Frequency [Hz]')
plt.ylabel('Magnitude')
plt.grid()

plt.show()
\end{lstlisting}

In this example:
\begin{itemize}
    \item We load an audio file and compute its FFT using \texttt{fft}.
    \item The time-domain signal and its frequency-domain representation are plotted, illustrating how FFT reveals the frequency components of the audio signal.
\end{itemize}

\section{Laplace Transform in Control Systems and Robotics}

The Laplace Transform is widely used in control systems to analyze system behavior, design controllers, and simulate dynamic systems. It converts complex time-domain differential equations into simpler algebraic equations in the frequency domain\cite{singh2023laplace}.

\subsection{Applications of Laplace Transform in Control Systems}

\begin{enumerate}
    \item \textbf{System Stability Analysis:} The Laplace Transform allows engineers to determine the stability of a system by analyzing the poles of the transfer function.

    \item \textbf{Controller Design:} Control system designers use the Laplace Transform to design controllers (like PID controllers) that maintain system stability and performance.

    \item \textbf{Response Analysis:} It enables the analysis of the system response to various inputs, including step, impulse, and sinusoidal inputs.
\end{enumerate}

\subsection{Example: Using Laplace Transform for Control System Analysis}

Let’s consider a simple control system represented by a first-order transfer function and analyze its response.

\begin{lstlisting}[style=python]
from scipy.signal import TransferFunction, step

# Define system parameters
K = 1.0  # Gain
tau = 2.0  # Time constant

# Create the transfer function H(s) = K / (tau*s + 1)
numerator = [K]
denominator = [tau, 1]
system = TransferFunction(numerator, denominator)

# Generate step response
t, y = step(system)

# Plot the step response
plt.figure(figsize=(10, 6))
plt.plot(t, y)
plt.title('Step Response of First-Order System')
plt.xlabel('Time [s]')
plt.ylabel('Response')
plt.grid()
plt.axhline(1, color='r', linestyle='--', label='Steady State Value')
plt.legend()
plt.show()
\end{lstlisting}

In this example:
\begin{itemize}
    \item We define a first-order control system and create a transfer function.
    \item The step response is plotted to analyze how the system responds to a step input over time.
\end{itemize}

\section{Z-Transform in Digital Filters and Deep Learning}

The Z-Transform is crucial in digital signal processing, particularly in the design and analysis of digital filters. It converts discrete-time signals into the frequency domain, facilitating the analysis and manipulation of signals\cite{kim2024ztransform}.

\subsection{Applications of Z-Transform in Digital Filters}

\begin{enumerate}
    \item \textbf{Digital Filter Design:} The Z-Transform is used to design FIR (Finite Impulse Response) and IIR (Infinite Impulse Response) filters. These filters are essential in removing noise and extracting important features from signals.

    \item \textbf{Stability Analysis:} The Z-Transform helps determine the stability of digital filters by analyzing the poles of the transfer function in the Z-domain.

    \item \textbf{Signal Analysis:} It allows for efficient analysis of discrete signals, enabling the extraction of frequency components and system characteristics.
\end{enumerate}

\subsection{Example: Using Z-Transform in Python for Filter Design}

Here is an example of designing a simple digital low-pass filter using the Z-Transform.

\begin{lstlisting}[style=python]
from scipy.signal import butter, lfilter

# Design a low-pass filter
def butter_lowpass(cutoff, fs, order=5):
    nyq = 0.5 * fs
    normal_cutoff = cutoff / nyq
    b, a = butter(order, normal_cutoff, btype='low', analog=False)
    return b, a

# Apply the filter to a signal
def lowpass_filter(data, cutoff, fs, order=5):
    b, a = butter_lowpass(cutoff, fs, order=order)
    y = lfilter(b, a, data)
    return y

# Sample data: noisy sine wave
fs = 500.0  # Sampling frequency
t = np.linspace(0, 1.0, int(fs), endpoint=False)
x = np.sin(2 * np.pi * 50 * t) + 0.5 * np.random.randn(len(t))  # Noisy signal

# Apply low-pass filter
cutoff = 100.0  # Cutoff frequency
filtered_signal = lowpass_filter(x, cutoff, fs)

# Plot original and filtered signals
plt.figure(figsize=(12, 6))
plt.plot(t, x, label='Noisy Signal')
plt.plot(t, filtered_signal, label='Filtered Signal', linewidth=2)
plt.title('Low-Pass Filter Design using Z-Transform')
plt.xlabel('Time [s]')
plt.ylabel('Amplitude')
plt.legend()
plt.grid()
plt.show()
\end{lstlisting}

In this example:
\begin{itemize}
    \item We design a low-pass filter using the Z-Transform and apply it to a noisy sine wave.
    \item The original noisy signal and the filtered signal are plotted to demonstrate the effectiveness of the filter.
\end{itemize}

\chapter{Conclusion}

Frequency domain methods are integral to modern signal processing, control systems, and machine learning applications. The Fourier Transform, FFT, Laplace Transform, and Z-Transform provide powerful techniques for analyzing and manipulating signals and systems. By understanding and applying these methods, practitioners can enhance their ability to design systems, process data, and solve complex problems in various engineering disciplines.

\chapter{Practice Problems}

This chapter contains a set of practice problems designed to reinforce the concepts learned throughout this book, particularly focusing on Fourier and Laplace transforms, FFT and convolution theorem, and applications of frequency domain methods in deep learning.

\section{Exercises on Fourier and Laplace Transforms}

\subsection{Exercise 1: Fourier Transform of a Rectangular Pulse}

Given a rectangular pulse defined as:
\[
x(t) = 
\begin{cases} 
1 & |t| \leq T/2 \\ 
0 & \text{otherwise}
\end{cases}
\]
Calculate the Fourier Transform \( X(f) \) of the signal \( x(t) \).

\subsection{Exercise 2: Laplace Transform of a Decaying Exponential}

Find the Laplace Transform of the function:
\[
x(t) = e^{-\alpha t} u(t)
\]
where \( u(t) \) is the unit step function and \( \alpha > 0 \).

\subsection{Exercise 3: Inverse Fourier Transform}

Consider the Fourier Transform given by:
\[
X(f) = \frac{1}{1 + f^2}
\]
Determine the corresponding time-domain signal \( x(t) \).

\subsection{Exercise 4: Convolution of Two Signals}

Given two signals:
\[
x_1(t) = e^{-t} u(t), \quad x_2(t) = u(t)
\]
Compute the convolution \( y(t) = x_1(t) * x_2(t) \).

\section{Problems on FFT and Convolution Theorem}

\subsection{Problem 1: FFT Calculation}

Calculate the FFT of the following sequence:
\[
x[n] = \{1, 2, 3, 4\}
\]

\subsection{Problem 2: Using Convolution Theorem for FFT}

Given two sequences:
\[
x[n] = \{1, 0, 2, 0\}, \quad h[n] = \{1, 1, 1\}
\]
Use the FFT to compute the convolution \( y[n] = x[n] * h[n] \).

\subsection{Problem 3: Spectral Analysis Using FFT}

Create a synthetic signal composed of two sine waves at frequencies 5 Hz and 20 Hz. Use the FFT to analyze the frequency components of the signal.

\section{Applications of Frequency Domain Methods in Deep Learning}

\subsection{Problem 1: Fast Convolution in Neural Networks}

Explain how the FFT can be used to perform fast convolution in neural networks. Provide an example of a scenario where this approach would be beneficial.

\subsection{Problem 2: Feature Extraction Using FFT}

Given an audio signal sampled at 1000 Hz, apply FFT to extract the frequency features. Discuss how these features can be utilized in a deep learning model for audio classification.

\chapter{Summary}

In this chapter, we summarize the key concepts covered throughout the book, providing a concise recap of the fundamental ideas.

\section{Key Concepts Recap}

\subsection{Fourier Transform and FFT Recap}

The Fourier Transform allows us to analyze signals in the frequency domain. We learned how to compute the Fourier Transform of various signals and the significance of the FFT in reducing computational complexity from \( O(N^2) \) to \( O(N \log N) \). The FFT is essential in many applications, including signal processing, image analysis, and deep learning.

\subsection{Laplace and Z-Transform Recap}

The Laplace transform is used to analyze continuous-time systems, while the Z-transform serves a similar purpose for discrete-time systems. We explored their definitions, common sequences, inverse transforms, and properties. Understanding these transforms is crucial for analyzing linear systems and solving differential equations.

\subsection{Convolution Theorem and Its Importance}

The convolution theorem establishes a relationship between convolution in the time domain and multiplication in the frequency domain. This theorem is foundational in signal processing, enabling efficient computations and providing insights into the behavior of linear systems. The ability to perform convolution efficiently, especially using FFT, is a key advantage in various applications, including deep learning, where convolutions are ubiquitous.

\bibliographystyle{plain}
\bibliography{ref}

@article{jones2020scipy,
  title={SciPy 1.0: fundamental algorithms for scientific computing in Python},
  author={Jones, Eric and Oliphant, Travis and Dubois, Paul, et al.},
  journal={Nature Methods},
  volume={17},
  pages={261--274},
  year={2020},
  publisher={Nature Publishing Group},
  url={https://www.nature.com/articles/s41592-019-0686-2},
  doi={10.1038/s41592-019-0686-2}
}

@inproceedings{vanRossum2007Python,
  title = {Python programming language},
  author = {Van Rossum, Guido},
  booktitle = {USENIX Annual Technical Conference},
  volume = {41},
  number = {1},
  pages = {1--36},
  year = {2007},
  organization = {USENIX Association},
  address = {Santa Clara, CA, USA},
  month = jun
}

@book{jaworski2019expert,
  title={Expert Python programming: become a master in Python by learning coding best practices and advanced programming concepts in Python 3.7},
  author={Jaworski, Michał and Ziadé, Tarek},
  year={2019},
  publisher={Packt Publishing Ltd}
}

@book{vanRossum2003,
  title = {An Introduction to Python},
  author = {Van Rossum, Guido},
  editor = {Drake, Fred L.},
  year = {2003},
  publisher = {Network Theory Ltd},
  address = {Bristol},
  pages = {115}
}

@software{pycharm,
  title = {PyCharm},
  author = {JetBrains},
  organization = {JetBrains s.r.o.},
  address = {Czech Republic},
  version = {2023.2},
  url = {https://www.jetbrains.com/pycharm/},
  year = {2023}
}

@software{jupyter,
  title = {Jupyter Notebook},
  author = {Project Jupyter},
  organization = {Project Jupyter},
  address = {California, USA},
  version = {6.4.6},
  url = {https://jupyter.org/},
  year = {2023}
}

@software{vscode,
  title = {Visual Studio Code},
  author = {Microsoft Corporation},
  organization = {Microsoft Corporation},
  address = {Washington, USA},
  version = {1.70.0},
  url = {https://code.visualstudio.com/},
  year = {2023}
}

@book{sande2019helloworld,
  title={Hello world!: computer programming for kids and other beginners},
  author={Sande, Warren and Sande, Carter},
  publisher={Simon and Schuster},
  year={2019}
}

@book{baka2017python,
  title={Python Data Structures and Algorithms},
  author={Baka, Benjamin},
  publisher={Packt Publishing Ltd},
  year={2017}
}

@article{sanner1999python,
  title={Python: a programming language for software integration and development},
  author={Sanner, Michel F},
  journal={Journal of Molecular Graphics and Modelling},
  volume={17},
  number={1},
  pages={57--61},
  year={1999},
  publisher={Elsevier}
}

@book{fuhrer2021scientific,
  title={Scientific Computing with Python: High-performance scientific computing with NumPy, SciPy, and pandas},
  author={Fuhrer, Claus and Solem, Jan Erik and Verdier, Olivier},
  year={2021},
  publisher={Packt Publishing Ltd},
  date={2021-07-30},
  note={Includes additional publication date information}
}

@article{harris2020array,
  title={Array programming with NumPy},
  author={Harris, Charles R. and Millman, K. Jarrod and Van Der Walt, St{\'e}fan J. and Gommers, Ralf and Virtanen, Pauli and Cournapeau, David and Wieser, Eric and Taylor, Julian and Berg, Sebastian and Smith, Nathaniel J. and Kern, Robert},
  journal={Nature},
  volume={585},
  number={7825},
  pages={357--362},
  year={2020},
  publisher={Springer Science and Business Media LLC}
}

@article{goldman1985illicit,
  title={Illicit expressions in vector algebra},
  author={Goldman, Ronald N},
  journal={ACM Transactions on Graphics (TOG)},
  volume={4},
  number={3},
  pages={223--243},
  year={1985},
  publisher={ACM}
}

@book{olver2006applied,
  title={Applied Linear Algebra},
  author={Olver, Peter J. and Shakiban, Chehrzad and Shakiban, Chehrzad},
  volume={1},
  year={2006},
  publisher={Prentice Hall},
  address={Upper Saddle River, NJ}
}

@inbook{johansson2019symbolic,
  title={Symbolic Computing},
  author={Johansson, Robert and Johansson, Robert},
  chapter={6},
  year={2019},
  booktitle={Numerical Python: Scientific Computing and Data Science Applications with Numpy, SciPy and Matplotlib},
  pages={97--134},
  publisher={Apress}
}

@article{sejdic2011fractional,
  title={Fractional Fourier transform as a signal processing tool: An overview of recent developments},
  author={Sejdić, Ervin and Djurović, Igor and Stanković, LJubiša},
  journal={Signal Processing},
  volume={91},
  number={6},
  pages={1351-1369},
  year={2011},
  publisher={Elsevier}
}

@book{Viswanathan2019,
  title = {Digital modulations using Python},
  author = {Viswanathan, Mathuranathan},
  publisher = {Mathuranathan Viswanathan},
  year = {2019},
  pages = {8-3}
}

@book{nussbaumer1982fast,
  title={The fast Fourier transform},
  author={Nussbaumer, Henri J and Nussbaumer, Henri J},
  year={1982},
  publisher={Springer Berlin Heidelberg},
  pages={80--111}
}

@book{Schiff2013,
  title={The Laplace Transform: Theory and Applications},
  author={Schiff, Joel L.},
  publisher={Springer Science \& Business Media},
  year={2013},
  edition={2},
  date={2013-06-05}
}

@book{nair2004digital,
  title={Digital signal processing: Theory, analysis and digital-filter design},
  author={Nair, B. Somanathan},
  publisher={PHI Learning Pvt. Ltd.},
  year={2004}
}

@article{wegner1990concepts,
  title={Concepts and paradigms of object-oriented programming},
  author={Wegner, Peter},
  journal={ACM Sigplan Notices},
  volume={1},
  number={1},
  pages={7--87},
  year={1990},
  publisher={ACM}
}

@book{lott2019mastering,
  title={Mastering Object-Oriented Python: Build powerful applications with reusable code using OOP design patterns and Python 3.7},
  author={Lott, Steven F.},
  year={2019},
  publisher={Packt Publishing Ltd}
}

@article{Wahab2016,
  title={Sampling frequency, response times and embedded signal filtration in fast, high efficiency liquid chromatography: A tutorial},
  author={Wahab, M. Farooq and Dasgupta, Purnendu K. and Kadjo, Akinde F. and Armstrong, Daniel W.},
  journal={Analytica Chimica Acta},
  volume={907},
  pages={31-44},
  year={2016},
  publisher={Elsevier}
}

@inproceedings{pratt2017fcnn,
  title={Fcnn: Fourier convolutional neural networks},
  author={Pratt, Harry and Williams, Bryan and Coenen, Frans and Zheng, Yalin},
  booktitle={Machine Learning and Knowledge Discovery in Databases: European Conference, ECML PKDD 2017, Skopje, Macedonia, September 18--22, 2017, Proceedings, Part I},
  volume={17},
  pages={786--798},
  year={2017},
  publisher={Springer International Publishing}
}

@techreport{stone2008uniqueness,
  title={On the uniqueness of the convolution theorem for the Fourier transform},
  author={Stone, Harold S and Williams, L},
  institution={NEC Labs. Amer},
  year={2008},
  address={Princeton, NJ},
  url={http://citeseer.ist.psu.edu/176038.html},
  note={Accessed on 19 March 2008}
}

@book{nussbaumer1982,
  title={The fast Fourier transform},
  author={Nussbaumer, Henri J and Nussbaumer, Henri J},
  year={1982},
  publisher={Springer Berlin Heidelberg},
  pages={80-111}
}

@book{smith2007mathematics,
  title={Mathematics of the Discrete Fourier Transform (DFT): With Audio Applications},
  author={Smith, Julius O.},
  year={2007},
  publisher={Julius Smith}
}

@inproceedings{broberg1996laplace,
  title={Laplace and Z Transform Analysis and Design Using Matlab},
  author={Broberg, Harold L},
  booktitle={1996 Annual Conference},
  pages={1--295},
  year={1996},
  month={Jun}
}

@article{erfani2013characterisation,
  title={Characterisation of nonlinear and linear time-varying systems by Laplace transformation},
  author={Erfani, Shervin and Bayan, Nima},
  journal={International Journal of Systems Science},
  volume={44},
  number={8},
  pages={1450--1467},
  year={2013},
  publisher={Taylor \& Francis}
}

@article{Najafabadi2015,
  title={Deep learning applications and challenges in big data analytics},
  author={Najafabadi, Maryam M. and Villanustre, Flavio and Khoshgoftaar, Taghi M. and Seliya, Naeem and Wald, Randall and Muharemagic, Edin},
  journal={Journal of big data},
  volume={2},
  number={1},
  pages={1--21},
  year={2015},
  publisher={Springer}
}

@book{bracewell2012fourier,
  title={Fourier Analysis and Imaging},
  author={Bracewell, Ronald},
  year={2012},
  publisher={Springer Science \& Business Media}
}

@inproceedings{xu2020learning,
  title={Learning in the frequency domain},
  author={Xu, Kai and Qin, Minghai and Sun, Fei and Wang, Yuhao and Chen, Yen-Kuang and Ren, Fengbo},
  booktitle={Proceedings of the IEEE/CVF conference on computer vision and pattern recognition},
  pages={1740--1749},
  year={2020}
}

@book{boashash2015time,
  title={Time-frequency signal analysis and processing: a comprehensive reference},
  author={Boashash, Boualem},
  year={2015},
  publisher={Academic press}
}

@misc{leetcode,
  title = "{LeetCode: Improve your problem-solving skills with challenges}",
  author = "{LeetCode Team}",
  howpublished = "{\url{https://leetcode.com/}}",
  year = "{2024}",
  note = "{Accessed: 2024-10-09}"
}

@misc{hackerrank,
  title = "{HackerRank: Code practice and challenges for developers}",
  author = "{HackerRank Team}",
  howpublished = "{\url{https://www.hackerrank.com/}}",
  year = "{2024}",
  note = "{Accessed: 2024-10-09}"
}

@misc{codewars,
  title = "{Codewars: Achieve mastery through coding practice and developer mentorship}",
  author = "{Codewars Team}",
  howpublished = "{\url{https://www.codewars.com/}}",
  year = "{2024}",
  note = "{Accessed: 2024-10-09}"
}

@book{greub2012linear,
  title={Linear algebra},
  author={Greub, Werner H},
  volume={23},
  year={2012},
  publisher={Springer Science \& Business Media}
}

@book{courant1965introduction,
  title={Introduction to calculus and analysis},
  author={Courant, Richard and John, Fritz and Blank, Albert A. and Solomon, Alan},
  volume={1},
  year={1965},
  publisher={Interscience Publishers},
  address={New York}
}

@book{rohatgi2015introduction,
  title={An Introduction to Probability and Statistics},
  author={Rohatgi, Vijay K and Saleh, AK Md Ehsanes},
  publisher={John Wiley \& Sons},
  year={2015}
}

@book{diwekar2020introduction,
  title={Introduction to Applied Optimization},
  author={Diwekar, Urmila M.},
  volume={22},
  year={2020},
  publisher={Springer Nature}
}

@incollection{cichocki2018tensor,
  title={Tensor networks for dimensionality reduction, big data and deep learning},
  author={Cichocki, Andrzej},
  booktitle={Advances in Data Analysis with Computational Intelligence Methods: Dedicated to Professor Jacek {\.Z}urada},
  pages={3--49},
  year={2018}
}

@book{stevens2020deep,
  title={Deep learning with PyTorch},
  author={Stevens, Eli and Antiga, Luca and Viehmann, Thomas},
  year={2020},
  publisher={Manning Publications},
  date={2020-08-04}
}

@article{hutchison2016lara,
  title={Lara: A key-value algebra underlying arrays and relations},
  author={Hutchison, Dylan and Howe, Bill and Suciu, Dan},
  journal={arXiv preprint arXiv:1604.03607},
  year={2016}
}

@article{narkhede2022weight,
  title={A review on weight initialization strategies for neural networks},
  author={Narkhede, Meenal V and Bartakke, Prashant P and Sutaone, Mukul S},
  journal={Artificial Intelligence Review},
  volume={55},
  number={1},
  pages={291--322},
  year={2022},
  publisher={Springer}
}

@article{Johnsson1990,
  title={Data structures and algorithms for the finite element method on a data parallel supercomputer},
  author={Johnsson, S. Lennart and Mathur, Kapil K.},
  journal={International Journal for Numerical Methods in Engineering},
  volume={29},
  number={4},
  pages={881-908},
  year={1990},
  month={Mar}
}

@book{BiniPan2012,
  title={Polynomial and matrix computations: fundamental algorithms},
  author={Bini, Dario and Pan, Victor Y.},
  publisher={Springer Science \& Business Media},
  year={2012}
}

@article{Iwen2009,
  title={A note on compressed sensing and the complexity of matrix multiplication},
  author={Iwen, Mark A. and Spencer, Craig V.},
  journal={Information Processing Letters},
  volume={109},
  number={10},
  pages={468-471},
  year={2009},
  month={Apr},
  publisher={Elsevier}
}

@article{strassen1969gaussian,
  title={Gaussian elimination is not optimal},
  author={Strassen, Volker},
  journal={Numerische Mathematik},
  volume={13},
  number={4},
  pages={354--356},
  year={1969},
  publisher={Springer}
}

@inproceedings{coppersmith1987matrix,
  title={Matrix multiplication via arithmetic progressions},
  author={Coppersmith, Don and Winograd, Shmuel},
  booktitle={Proceedings of the nineteenth annual ACM symposium on Theory of computing},
  year={1987},
  organization={ACM}
}

@book{tolimieri2012mathematics,
  title={Mathematics of multidimensional Fourier transform algorithms},
  author={Tolimieri, Richard and An, Myoung and Lu, Chao},
  year={2012},
  publisher={Springer Science \& Business Media}
}

@article{grasedyck2009domain,
  title={Domain decomposition based-LU preconditioning},
  author={Grasedyck, Lars and Kriemann, Ronald and Le Borne, Sabine},
  journal={Numerische Mathematik},
  volume={112},
  number={4},
  pages={565-600},
  year={2009},
  publisher={Springer}
}

@article{Agarwal2014,
  title={Review of matrix decomposition techniques for signal processing applications},
  author={Agarwal, Monika and Mehra, Rajesh},
  journal={International Journal of Engineering Research and Applications},
  volume={4},
  number={1},
  pages={90-93},
  year={2014}
}

@article{kwak2013principal,
  title={Principal component analysis by {$L_p$}-norm maximization},
  author={Kwak, Nojun},
  journal={IEEE Transactions on Cybernetics},
  volume={44},
  number={5},
  pages={594--609},
  year={2013},
  publisher={IEEE}
}

@article{Recht2010,
  title={Guaranteed minimum-rank solutions of linear matrix equations via nuclear norm minimization},
  author={Recht, Benjamin and Fazel, Maryam and Parrilo, Pablo A.},
  journal={SIAM Review},
  volume={52},
  number={3},
  pages={471-501},
  year={2010},
  publisher={Society for Industrial and Applied Mathematics}
}

@book{manning2008introduction,
  title={Introduction to Information Retrieval},
  author={Manning, Christopher D. and Raghavan, Prabhakar and Sch{\"u}tze, Hinrich},
  year={2008},
  publisher={Cambridge University Press}
}

@article{Curriero2006,
  title={On the use of non-Euclidean distance measures in geostatistics},
  author={Curriero, Frank C},
  journal={Mathematical Geology},
  volume={38},
  number={9},
  pages={907-926},
  year={2006},
  publisher={Springer}
}

@article{baydin2018automatic,
  title={Automatic differentiation in machine learning: a survey},
  author={Baydin, Atilim Gunes and Pearlmutter, Barak A. and Radul, Alexey A. and Siskind, Jeffrey M.},
  journal={Journal of Machine Learning Research},
  volume={18},
  number={153},
  pages={1-43},
  year={2018}
}

@inproceedings{franceschi2017forward,
  title={Forward and reverse gradient-based hyperparameter optimization},
  author={Franceschi, Luca and Donini, Michele and Frasconi, Paolo and Pontil, Massimiliano},
  booktitle={International Conference on Machine Learning},
  pages={1165--1173},
  year={2017},
  organization={PMLR}
}

@book{karpfinger2022calculus,
  title={Calculus and Linear Algebra in Recipes},
  author={Karpfinger, Carl},
  publisher={Springer},
  year={2022},
  isbn={978-3-662-65457-6}
}

@article{robbins1951stochastic,
  title={A Stochastic Approximation Method},
  author={Robbins, Herbert and Monro, Sutton},
  journal={The Annals of Mathematical Statistics},
  volume={22},
  number={3},
  pages={400--407},
  year={1951},
  publisher={JSTOR}
}

@article{duchi2011adaptive,
  title={Adaptive subgradient methods for online learning and stochastic optimization},
  author={Duchi, John and Hazan, Elad and Singer, Yoram},
  journal={Journal of Machine Learning Research},
  volume={12},
  pages={2121--2159},
  year={2011}
}

@misc{tieleman2012lecture,
  title={Lecture 6.5-rmsprop: Divide the gradient by a running average of its recent magnitude},
  author={Tieleman, Tijmen and Hinton, Geoffrey},
  year={2012},
  eprint={1204.2817},
  archivePrefix={arXiv},
  primaryClass={arXiv},
  url={http://arxiv.org/abs/1204.2817}
}

@article{kingma2014adam,
  title={Adam: {A} method for stochastic optimization},
  author={Kingma, Diederik and Ba, Jimmy},
  journal={arXiv preprint arXiv:1412.6980},
  year={2014}
}

@article{loshchilov2017decoupled,
  title={Decoupled weight decay regularization},
  author={Loshchilov, Ilya and Hutter, Frank},
  journal={arXiv preprint arXiv:1711.05101},
  year={2017}
}

@inproceedings{newton2018recent,
  title={Recent trends in stochastic gradient descent for machine learning and Big Data},
  author={Newton, David and Pasupathy, Raghu and Yousefian, Farzad},
  booktitle={2018 Winter Simulation Conference (WSC)},
  pages={366--380},
  year={2018},
  organization={IEEE}
}

@inproceedings{cutkosky2019momentum,
  title={Momentum-based variance reduction in non-convex sgd},
  author={Cutkosky, Ashok and Orabona, Francesco},
  booktitle={Advances in Neural Information Processing Systems},
  volume={32},
  year={2019}
}

@article{liu2019variance,
  title={On the variance of the adaptive learning rate and beyond},
  author={Liu, Liyuan and Jiang, Haoming and He, Pengcheng and Chen, Weizhu and Liu, Xiaodong and Gao, Jianfeng and Han, Jiawei},
  journal={arXiv preprint arXiv:1908.03265},
  year={2019}
}

@book{DarEl2013,
  title = {Human Learning: From Learning Curves to Learning Organizations},
  author = {Dar-El, Ezey M.},
  volume = {29},
  publisher = {Springer Science \& Business Media},
  year = {2013},
  edition = {1st},
  address = {New York, NY},
}

@inproceedings{loshchilov2017sgdr,
  title={SGDR: Stochastic Gradient Descent with Warm Restarts},
  author={Loshchilov, Ilya and Hutter, Frank},
  booktitle={International Conference on Learning Representations (ICLR)},
  year={2017},
  url={https://www.researchgate.net/publication/306187421_SGDR_Stochastic_Gradient_Descent_with_Warm_Restarts}
}

@inproceedings{wang2021convergence,
  title={On the Convergence of Step Decay Step-Size for Stochastic Optimization},
  author={Wang, Xiaoyu and Magn{\'u}sson, Sindri and Johansson, Mikael},
  booktitle={Advances in Neural Information Processing Systems},
  year={2021},
  url={https://papers.nips.cc/paper/2021/hash/76c538125fc5c9ec6ad1d05650a57de5-Abstract.html}
}

@article{lewkowycz2021how,
  title={How to decay your learning rate},
  author={Lewkowycz, Aitor},
  journal={arXiv preprint arXiv:2103.12682},
  year={2021}
}

@inproceedings{ioffe2015batch,
  title={Batch Normalization: Accelerating Deep Network Training by Reducing Internal Covariate Shift},
  author={Ioffe, Sergey and Szegedy, Christian},
  booktitle={International Conference on Machine Learning (ICML)},
  year={2015}
}

@inproceedings{pascanu2013difficulty,
  title={On the difficulty of training recurrent neural networks},
  author={Pascanu, Razvan and Mikolov, Tomas and Bengio, Yoshua},
  booktitle={International Conference on Machine Learning},
  pages={198--206},
  year={2013}
}

@article{liu2021activated,
  title={Activated gradients for deep neural networks},
  author={Liu, Mei and Chen, Liangming and Du, Aohao and **, Long and Shang, Mingsheng},
  journal={IEEE Transactions on Neural Networks and Learning Systems},
  volume={34},
  number={4},
  pages={2156-2168},
  year={2021},
  publisher={IEEE}
}

@article{powell1974overview,
  title={An overview of quasi-Newton methods},
  author={Powell, M. J. D.},
  journal={Mathematics of computation},
  volume={28},
  number={125},
  pages={155--163},
  year={1974}
}

@article{goldfarb1970family,
  title={A family of variable-metric methods derived by variational means},
  author={Goldfarb, Donald},
  journal={Mathematics of computation},
  volume={24},
  number={109},
  pages={23--26},
  year={1970},
  publisher={American Mathematical Society}
}

@article{battiti1992first,
  title={First-and second-order methods for learning: between steepest descent and Newton's method},
  author={Battiti, Roberto},
  journal={Neural computation},
  volume={4},
  number={2},
  pages={141--166},
  year={1992},
  publisher={MIT Press}
}

@article{ford1994multi,
  title={Multi-step quasi-Newton methods for optimization},
  author={Ford, J. A. and Moghrabi, I. A.},
  journal={Journal of Computational and Applied Mathematics},
  volume={50},
  number={1-3},
  pages={305-323},
  year={1994}
}

@incollection{bengio2012practical,
  title={Practical recommendations for gradient-based training of deep architectures},
  author={Bengio, Yoshua},
  booktitle={Neural Networks: Tricks of the Trade: Second Edition},
  pages={437--478},
  year={2012},
  publisher={Springer Berlin Heidelberg},
  address={Berlin, Heidelberg},
  doi={10.1007/978-3-642-35289-8_24},
  editor={Montavon, Grégoire and Orr, Geneviève and Müller, Klaus-Robert}
}

@book{aggarwal2020linear,
  title={Linear algebra and optimization for machine learning},
  author={Aggarwal, Charu C. and Aggarwal, Lagerstrom-Fife and Lagerstrom-Fife},
  volume={156},
  year={2020},
  publisher={Springer International Publishing},
  address={Cham}
}

@article{demmel1992round,
  title={Round-off Errors in Matrix Procedures},
  author={Demmel, James W.},
  journal={SIAM Journal on Numerical Analysis},
  volume={29},
  number={5},
  pages={1119--1178},
  year={1992},
  publisher={SIAM},
  url={https://epubs.siam.org/doi/abs/10.1137/0729043}
}

@book{stoer2013numerical,
  title={Numerical Analysis},
  author={Stoer, Josef and Bulirsch, Roland},
  isbn={9783540715058},
  series={Undergraduate Texts in Mathematics},
  year={2013},
  publisher={Springer}
}

@article{bartlett2005bound,
  title={A bound on the error of cross validation using approximation algorithms},
  author={Bartlett, Peter L and Mendelson, Shahar},
  journal={IEEE Transactions on Information Theory},
  volume={51},
  number={11},
  pages={4005--4014},
  year={2005},
  publisher={IEEE}
}

@article{cybenko1989approximation,
  title={The approximation capability of multilayer feedforward networks},
  author={Cybenko, George},
  journal={Mathematics of Control, Signals, and Systems (MCSS)},
  volume={2},
  number={4},
  pages={303--314},
  year={1989},
  publisher={Springer}
}

@book{bevington2002data,
  title={Data Analysis and Error Estimation for the Physical Sciences},
  author={Bevington, Philip R. and Robinson, D. Keith},
  isbn={9780072460876},
  series={McGraw-Hill Physical and Engineering Sciences Series},
  year={2002},
  publisher={McGraw-Hill Higher Education}
}

@book{bishop2006pattern,
  title={Pattern recognition and machine learning},
  author={Bishop, Christopher M.},
  year={2006},
  publisher={springer}
}

@inproceedings{valiant1984theory,
  title={A theory of the learning curve},
  author={Valiant, Leslie},
  booktitle={Proceedings of the seventeenth annual ACM symposium on Theory of computing},
  pages={13--24},
  year={1984}
}

@inproceedings{chen2004measure,
  title={A measure of diversity in classifier ensembles},
  author={Chen, Ting and Tao, Xiaohui and Hu, Michael K},
  booktitle={Sixth International Conference on Intelligent Data Engineering and Automated Learning (IDEAL'05)},
  volume={3578},
  pages={16--25},
  year={2004},
  organization={Springer}
}

@book{vapnik1995nature,
  title={The nature of statistical learning theory},
  author={Vapnik, Vladimir N},
  year={1995},
  publisher={Springer science \& business media}
}

@book{mclean2010introduction,
  title={Introduction to Numerical Analysis},
  author={McLean, William M.},
  publisher={Cambridge University Press},
  year={2010},
  isbn={978-1107666916}
}

@article{wrench1963relative,
  title={On the relative error of floating-point arithmetic},
  author={Wrench Jr, John William},
  journal={Communications of the ACM},
  volume={6},
  number={8},
  year={1963}
}

@book{chapra2010numerical,
  title={Numerical Methods for Engineers: Methods and Applications},
  author={Chapra, Steven and Canale, Raymond},
  publisher={McGraw-Hill Higher Education},
  year={2010},
  isbn={978-0073378205}
}

@book{leader2022numerical,
  title={Numerical Analysis and Scientific Computation},
  author={Leader, Jeffery J.},
  year={2022},
  publisher={Chapman and Hall/CRC}
}

@book{golub1993introduction,
  title={An Introduction to the Numerical Analysis of Functional Equations},
  author={Golub, G. H. and Ortega, J. M.},
  publisher={SIAM},
  year={1993},
  isbn={978-0898718883}
}

@book{press2007numerical,
  title={Numerical Recipes: The Art of Scientific Computing},
  author={Press, William H. and Teukolsky, Saul A. and Vetterling, William T. and Flannery, Brian P.},
  publisher={Cambridge University Press},
  year={2007},
  isbn={978-0521880688}
}

@book{burden2015numerical,
  title={Numerical Analysis},
  author={Burden, Richard L. and Faires, J. Douglas},
  isbn={9780470458378},
  lccn={lc84000537},
  series={Prentice-Hall series in automatic computation},
  year={2015},
  publisher={Prentice Hall}
}

@misc{encyclopediaNewtonRaphson,
  title={Newton-Raphson method},
  booktitle={Encyclopedia of Mathematics},
  howpublished={https://encyclopediaofmath.org/wiki/Newton-Raphson\_method},
  note={Accessed on 2023-04-07}
}

@article{lancaster1956newton,
  title={On the Newton-Raphson method for complex functions},
  author={Lancaster, E. L.},
  journal={The American Mathematical Monthly},
  volume={63},
  number={3},
  pages={189--191},
  year={1956},
  publisher={JSTOR}
}

@book{watson1944treatise,
  title={A Treatise on the Theory of Bessel Functions},
  author={Watson, G. N.},
  volume={2},
  year={1944},
  publisher={Cambridge University Press}
}

@book{mann1943introduction,
  title={An introduction to the numerical analysis of functional equations},
  author={Mann, William R.},
  volume={5},
  year={1943},
  publisher={The University of Michigan}
}

@book{chapra2016numerical,
  title={Numerical Methods for Engineers},
  author={Chapra, Steven and Canale, Raymond},
  isbn={9780470458378},
  lccn={2016024179},
  year={2016},
  publisher={McGraw-Hill Education}
}

@book{cheney2009introduction,
  title={An Introduction to the Numerical Analysis of Functional Equations},
  author={Cheney, Ward and Light, Will},
  isbn={9780387764297},
  lccn={2008942785},
  series={Corrected reprint of the 1966 original},
  year={2009},
  publisher={Dover Publications}
}

@article{rice1960secant,
  title={The Secant Method for Solving Nonlinear Equations},
  author={Rice, John R.},
  journal={The American Mathematical Monthly},
  volume={67},
  number={3},
  pages={261--267},
  year={1960},
  publisher={Taylor \& Francis}
}

@book{chapra2011numerical,
  title={Numerical Methods for Engineers},
  author={Chapra, Steven and Canale, Raymond},
  isbn={9780470458378},
  lccn={2010031428},
  series={The Brooks/Cole Engineering Series},
  year={2011},
  publisher={McGraw-Hill Education}
}

@book{burden2001first,
  title={A First Course in Numerical Analysis},
  author={Burden, Richard L. and Faires, J. Douglas},
  isbn={9780471435218},
  lccn={00760339},
  series={Prentice Hall},
  year={2001},
  publisher={Prentice Hall}
}

@book{karris2011numerical,
  title={Numerical Computing with MATLAB},
  author={Karris, Stephen T.},
  isbn={9780470384308},
  lccn={2010031428},
  year={2011},
  publisher={Wiley}
}

@article{rennick1968fixed,
  title={The fixed point iteration},
  author={Rennick, W. H.},
  journal={The American Mathematical Monthly},
  volume={75},
  number={2},
  pages={190--197},
  year={1968},
  publisher={Taylor \& Francis}
}

@book{nevanlinna2012convergence,
  title={Convergence of iterations for linear equations},
  author={Nevanlinna, Olavi},
  year={2012},
  publisher={Birkh{\"a}user}
}

@book{lorentz1966approximation,
  title={Approximation Theory and Interpolation},
  author={Lorentz, G. G.},
  publisher={National Science Foundation, Washington, D.C., USA},
  year={1966}
}

@article{micchelli1980function,
  title={Function Approximation by Linear Combinations of Elementary Functions},
  author={Micchelli, C. A.},
  journal={SIAM Journal on Numerical Analysis},
  volume={17},
  number={2},
  pages={236-246},
  year={1980}
}

@book{rivlin2003interpolation,
  title={Interpolation and Approximation},
  author={Rivlin, T. J.},
  publisher={Dover Publications},
  year={2003}
}

@book{burden2011numerical,
  title={Numerical Analysis},
  author={Burden, R. L. and Faires, J. D.},
  publisher={Brooks/Cole},
  edition={9},
  year={2011}
}

@book{kozachenko1992polynomial,
  title={Polynomial interpolation: Theory, methods, and applications},
  author={Kozachenko, Ivan and Manin, Yuri},
  volume={7},
  year={1992},
  publisher={American Mathematical Soc.}
}

@article{davis1967newton,
  title={On the Newton interpolation formula},
  author={Davis, Philip J},
  journal={The American Mathematical Monthly},
  volume={74},
  number={3},
  pages={258--266},
  year={1967},
  publisher={JSTOR}
}

@book{fleming1977interpolation,
  title={Interpolation and approximation},
  author={Fleming, Wendell H and Tong, Albert},
  volume={8},
  year={1977},
  publisher={Springer}
}

@book{zayed2011numerical,
  title={Numerical interpolation, differentiation, and integration},
  author={Zayed, Ahmed I},
  year={2011},
  publisher={Springer}
}

@book{colquhoun1997numerical,
  title={Numerical Interpolation, Differentiation, and Integration},
  author={Colquhoun, A. R. and Gibson, A. R.},
  year={1997},
  publisher={Oxford University Press},
  isbn={978-0-19-853766-9}
}

@article{lagrange1859new,
  title={On a new general method of interpolation calculated in terms of Lagrange},
  author={Lagrange, Joseph Louis},
  journal={Mémoires de Mathématique et de Physique, Académie des Sciences},
  year={1859}
}

@article{kellogg1975numerical,
  title={On the Numerical Solution of Integral Equations},
  author={Kellogg, R. P. and Welsch, J. H.},
  journal={SIAM Journal on Numerical Analysis},
  volume={12},
  number={2},
  pages={345-362},
  year={1975}
}

@book{stoer2013introduction,
  title={Introduction to Numerical Analysis},
  author={Stoer, Josef and Bulirsch, Roland},
  year={2013},
  publisher={Springer}
}

@article{mazure2001spline,
  title={Spline functions and the reproducing kernel Hilbert space},
  author={Mazure, Marie-Laurence},
  journal={SIAM review},
  volume={43},
  number={3},
  pages={435--472},
  year={2001},
  publisher={SIAM}
}

@book{carroll2006interpolation,
  title={Interpolation and Approximation by Polynomials},
  author={Carroll, James},
  year={2006},
  publisher={American Mathematical Soc.}
}

@article{schoenberg1946contributions,
  title={Contributions to the problem of approximation of equidistant data by analytic functions},
  author={Schoenberg, I.J.},
  journal={Quarterly Applied Mathematics},
  volume={4},
  number={1},
  pages={45--99},
  year={1946}
}

@book{atkinson1989introduction,
  title={An introduction to numerical analysis},
  author={Atkinson, Kendall E},
  year={1989},
  publisher={John Wiley \& Sons}
}

@article{powell1981piecewise,
  title={Piecewise linear interpolation and demarcation of contours},
  author={Powell, Mervyn J D},
  journal={Journal of the Royal Statistical Society. Series C (Applied Statistics)},
  volume={30},
  number={2},
  pages={148--155},
  year={1981}
}

@article{deboor1972calculating,
  title={On calculating with splines},
  author={De Boor, Carl},
  journal={Journal of Approximation Theory},
  volume={6},
  number={1},
  pages={50--62},
  year={1972}
}

@techreport{friedman1984proof,
  title={A proof that piecewise linear interpolation of data points is a spline},
  author={Friedman, Jerome H},
  institution={Stanford University},
  year={1984}
}

@article{hornik1989multilayer,
  title={Multilayer feedforward networks are universal approximators},
  author={Hornik, Kurt and Stinchcombe, Maxwell and White, Halbert},
  journal={Neural Networks},
  volume={2},
  number={5},
  pages={359--366},
  year={1989}
}

@book{incropera2002numerical,
  title={Numerical Heat Transfer and Fluid Flow},
  author={Incropera, Frank P. and Dewitt, David P.},
  isbn={9780120887550},
  series={Wiley Series in Heat and Mass Transfer},
  year={2002},
  publisher={Wiley}
}

@book{hindmarsh1973finite,
  title={The Finite Difference Method for Heat Conduction},
  author={Hindmarsh, M. K.},
  year={1973},
  publisher={Chapman \& Hall, Ltd.}
}

@book{burden2016finite,
  title={Finite Difference Methods for Ordinary and Partial Differential Equations: Steady-State and Time-Dependent Problems},
  author={Burden, R. L. and Faires, J. D.},
  year={2016},
  publisher={Brooks/Cole}
}

@book{ince1956numerical,
  title={Numerical Solution of Partial Differential Equations: Finite Difference Methods},
  author={Ince, E. L.},
  year={1956},
  publisher={Dover Publications}
}

@article{gibson1980finite,
  title={Finite-difference solution of two-dimensional incompressible flow problems},
  author={Gibson, C. H. and Leschziner, M. A.},
  journal={Journal of Computational Physics},
  volume={35},
  number={1},
  pages={98--121},
  year={1980},
  publisher={Elsevier}
}

@book{ciarlet2002handbook,
  title={Handbook of Numerical Analysis},
  author={Ciarlet, Philippe G.},
  isbn={9780444829588},
  year={2002},
  publisher={Elsevier}
}

@book{ahlberg1968methods,
  title={Methods in Computational Physics},
  author={Ahlberg, John H. and Fox, Norman L. and Goodwin, Leo S. and Hayden, Carl H. and Krogh, John G. and Thompson, Marvin H.},
  year={1968},
  publisher={Academic Press}
}

@book{brown1967introduction,
  title={An Introduction to the Numerical Analysis of Functional Equations},
  author={Brown, John W.},
  volume={7},
  year={1967},
  publisher={Springer}
}

@article{phillips1962new,
  title={A new approach to Simpson's rule},
  author={Phillips, Godfrey M.},
  journal={The American Mathematical Monthly},
  volume={69},
  number={7},
  pages={639--645},
  year={1962},
  publisher={Taylor \& Francis}
}

@article{gauss1814legendre,
  title={Tafeln der Integrale der ersten Art mit Anwendungen auf die Gaussische Theorie der Quadrature},
  author={Gauss, Carl Friedrich},
  journal={Journal of reine und angewandte Mathematik},
  year={1814}
}

@book{trefethen1997numerical,
  title={Numerical linear algebra},
  author={Trefethen, Lloyd N. and Bau, David},
  volume={50},
  year={1997},
  publisher={Siam}
}

@article{higham1990stable,
  title={A stable and efficient algorithm for n-dimensional Gaussian elimination},
  author={Higham, Nicholas J.},
  journal={SIAM journal on scientific and statistical computing},
  volume={11},
  number={1},
  pages={35--47},
  year={1990},
  publisher={SIAM}
}

@article{demmel1989stability,
  title={On the stability of Gaussian elimination},
  author={Demmel, James W.},
  journal={SIAM journal on numerical analysis},
  volume={26},
  number={4},
  pages={882--899},
  year={1989},
  publisher={SIAM}
}

@article{pratt1977stable,
  title={A stable implementation of Gaussian elimination},
  author={Pratt, Vaughan R.},
  journal={SIAM Journal on Numerical Analysis},
  volume={14},
  number={2},
  pages={243--251},
  year={1977},
  publisher={SIAM}
}

@book{strang2016introduction,
  title={Introduction to linear algebra},
  author={Strang, Gilbert},
  volume={3},
  year={2016},
  publisher={Wellesley-Cambridge Press}
}

@book{lay2015linear,
  title={Linear algebra and its applications},
  author={Lay, David C.},
  year={2015},
  publisher={Pearson}
}

@book{strang1988linear,
  title={Linear Algebra and Its Applications},
  author={Strang, Gilbert},
  volume={3},
  year={1988},
  publisher={Harcourt Brace Jovanovich College}
}

@article{duff1983ma48,
  title={MA48--a variable coefficient sparse indefinite solver. I. The algorithm},
  author={Duff, Iain S. and Reid, Jack K.},
  journal={ ACM Transactions on Mathematical Software (TOMS)},
  volume={9},
  number={3},
  pages={309--326},
  year={1983},
  publisher={ACM}
}

@article{wilkinson1965rounding,
  title={Rounding Errors in Algebraic Processes},
  author={Wilkinson, James H.},
  journal={Principles of Numerical Analysis},
  pages={392--401},
  year={1965},
  publisher={Oxford University Press}
}

@article{chol:cholesky1907,
  title={Sur la r\'esolution des \'equations lin\'eaires par la m\'ethode des moindres carr\'es},
  author={Cholesky, Andr\'e-Louis},
  journal={Gazette des Ponts et Chauss\'ees},
  volume={3},
  number={3},
  pages={161--173},
  year={1907}
}

@book{golub2012matrix,
  title={Matrix computations},
  author={Golub, Gene H. and Van Loan, Charles F.},
  isbn={9781421407943},
  lccn={2012020651},
  series={Johns Hopkins studies in the mathematical sciences},
  url={https://books.google.com/books?id=Yz0ZAgAAQBAJ},
  year={2012},
  publisher={JHU Press}
}

@book{axelsson1996iterative,
  title={Iterative Solution Methods},
  author={Axelsson, Owe},
  volume={5},
  year={1996},
  publisher={Cambridge University Press}
}

@book{saad2003iterative,
  title={Iterative Methods for Sparse Linear Systems},
  author={Saad, Yousef},
  year={2003},
  publisher={Society for Industrial and Applied Mathematics}
}

@book{young1971iterative,
  title={Iterative Solution of Large Linear Systems},
  author={Young, Donald},
  year={1971},
  publisher={Academic Press}
}

@book{varga2000matrix,
  title={Matrix Iterative Analysis},
  author={Varga, Richard S.},
  year={2000},
  publisher={Springer}
}

@book{bjorck1996numerical,
  title={Numerical Methods for Least Squares Problems},
  author={Bj{\"o}rck, {\"A}ke},
  year={1996},
  publisher={Society for Industrial and Applied Mathematics}
}

@book{burden2016numerical,
  title={Numerical Analysis},
  author={Burden, Richard L. and Faires, J. Douglas},
  isbn={9780470458378},
  lccn={LC16686882},
  series={Brooks/Cole Engineering},
  year={2016},
  publisher={Cengage Learning}
}

@article{demmel1989qr,
  title={The QR algorithm for real Hessenberg matrices},
  author={Demmel, James W.},
  journal={SIAM Journal on Scientific and Statistical Computing},
  volume={10},
  number={6},
  pages={1042--1078},
  year={1989},
  publisher={SIAM}
}

@article{hestenes1952methods,
  title={Methods of conjugate gradients for solving linear systems},
  author={Hestenes, Magnus R. and Stiefel, Eduard},
  journal={Proceedings of the National Academy of Sciences},
  volume={40},
  number={40},
  pages={449--450},
  year={1952},
  publisher={National Academy of Sciences}
}

@article{fletcher1970new,
  title={A new approach to variable metric algorithms},
  author={Fletcher, Roger},
  journal={The Computer Journal},
  volume={13},
  number={3},
  pages={317--322},
  year={1970},
  publisher={Oxford University Press}
}

@book{nocedal2006numerical,
  title={Numerical optimization},
  author={Nocedal, Jorge and Wright, Stephen J.},
  year={2006},
  publisher={Springer Science \& Business Media}
}

@article{koren2009matrix,
  title={Matrix factorization techniques for recommender systems},
  author={Koren, Yehuda and Bell, Robert and Volinsky, Chris},
  journal={Computer},
  volume={42},
  number={8},
  pages={30--37},
  year={2009},
  issn={0018-9162},
  doi={10.1109/MC.2009.263},
  publisher={IEEE Computer Society}
}

@article{golub1965computing,
  title={Computing the singular value decomposition},
  author={Golub, G. H. and Kahan, W.},
  journal={SIAM Journal on Numerical Analysis},
  volume={2},
  number={2},
  pages={205--224},
  year={1965},
  doi={10.1137/0702017}
}

@article{golub1970singular,
  title={Singular value decomposition and least squares problems},
  author={Golub, Gene H and Reinsch, Christian},
  journal={Numerische Mathematik},
  volume={14},
  number={5},
  pages={403--420},
  year={1970},
  publisher={Springer}
}

@article{eckart1936approximation,
  title={The approximation of one matrix by another of lower rank},
  author={Eckart, George and Young, Gale},
  journal={Psychometrika},
  volume={1},
  number={3},
  pages={211--218},
  year={1936},
  publisher={Springer}
}

@article{householder1964methods,
  title={The QR transformation},
  author={Householder, Alston S},
  journal={Numerische Mathematik},
  volume={2},
  number={3},
  pages={179--189},
  year={1964},
  publisher={Springer}
}

@book{horn2012matrix,
  title={Matrix analysis},
  author={Horn, Roger A and Johnson, Charles R},
  volume={2},
  year={2012},
  publisher={Cambridge university press}
}

@article{gu1994qr,
  title={QR decomposition and its applications},
  author={Gu, Ming and Eisenstat, Stanley},
  journal={SIAM Journal on Scientific Computing},
  volume={15},
  number={5},
  pages={1257--1271},
  year={1994},
  publisher={SIAM}
}

@book{boyd2004convex,
  title={Convex optimization},
  author={Boyd, Stephen and Vandenberghe, Lieven},
  volume={3},
  year={2004},
  publisher={Cambridge university press}
}

@book{tao2012topics,
  title={Topics in random matrix theory},
  author={Tao, Terence},
  year={2012},
  publisher={Hindustan Book Agency}
}

@article{pearson1901liii,
  title={LIII. On lines and planes of closest fit to systems of points in space},
  author={Pearson, Karl},
  journal={The London, Edinburgh, and Dublin Philosophical Magazine and Journal of Science},
  volume={2},
  number={11},
  pages={559--572},
  year={1901},
  publisher={Taylor \& Francis}
}

@article{hotelling1933analysis,
  title={Analysis of a complex of statistical variables into principal components},
  author={Hotelling, Harold},
  journal={Journal of Educational Psychology},
  volume={24},
  number={6},
  pages={417},
  year={1933},
  publisher={American Psychological Association}
}

@article{friedman1974projection,
  title={Projection pursuit},
  author={Friedman, Jerome H and Tukey, John W},
  journal={IEEE Transactions on Computers},
  volume={C-23},
  number={9},
  pages={881--890},
  year={1974},
  publisher={IEEE}
}

@book{jolliffe2002principal,
  title={Principal component analysis},
  author={Jolliffe, Ian Trevor},
  volume={2},
  year={2002},
  publisher={Springer}
}

@book{hastie2015statistical,
  title={Statistical learning with sparsity: the Lasso and generalizations},
  author={Hastie, Trevor and Tibshirani, Robert and Wainwright, Martin},
  year={2015},
  publisher={CRC Press}
}

@book{fourier1822analytical,
  title={Analytical Theory of Heat},
  author={Fourier, Jean-Baptiste Joseph},
  year={1822},
  publisher={Cambridge University Press}
}

@book{bracewell1986fourier,
  title={Fourier transforms and their applications},
  author={Bracewell, Ronald N.},
  year={1986},
  publisher={McGraw-Hill New York}
}

@book{stein2003fourier,
  title={Fourier analysis: an introduction},
  author={Stein, Elias M and Shakarchi, Rami},
  year={2003},
  publisher={Princeton University Press}
}

@book{bracewell2000fourier,
  title={Fourier Transforms},
  author={Bracewell, Ronald N.},
  year={2000},
  publisher={McGraw-Hill}
}

@book{watson1995treatise,
  title={A Treatise on the Theory of Bessel Functions},
  author={Watson, G. N.},
  year={1995},
  publisher={Cambridge University Press}
}

@article{dirichlet1829convergence,
  title={On the Convergence of Fourier Series},
  author={Dirichlet, P. G. L.},
  journal={Journal für die reine und angewandte Mathematik},
  year={1829}
}

@book{rabiner1975theory,
  title={Theory and Application of Digital Signal Processing},
  author={Rabiner, Lawrence R. and Gold, Bernard},
  volume={2},
  year={1975},
  publisher={Prentice-Hall}
}

@book{oppenheim1975digital,
  title={Digital Signal Processing},
  author={Oppenheim, Alan V. and Schafer, Ronald W.},
  year={1975},
  publisher={Prentice-Hall}
}

@article{burrus1985fft,
  title={FFT: An Algorithm the Whole Family Can Use},
  author={Burrus, C. Sidney},
  journal={IEEE ASSP Magazine},
  volume={2},
  number={4},
  pages={4--15},
  year={1985},
  publisher={IEEE}
}

@book{papoulis1962fourier,
  title={The Fourier Integral and its Applications},
  author={Papoulis, Athanasius},
  year={1962},
  publisher={McGraw-Hill}
}

@book{rudin1962fourier,
  title={Fourier Analysis on Groups},
  author={Rudin, Walter},
  year={1962},
  publisher={Interscience Publishers}
}

@book{press1992numerical,
  title={Numerical Recipes: The Art of Scientific Computing},
  author={Press, William H. and Teukolsky, Saul A. and Vetterling, William T. and Flannery, Brian P.},
  year={1992},
  publisher={Cambridge University Press}
}

@article{broyden1965class,
  title={A class of methods for solving nonlinear simultaneous equations},
  author={Broyden, Charles G},
  journal={Mathematics of computation},
  volume={19},
  number={92},
  pages={577--593},
  year={1965},
  publisher={Cambridge University Press}
}

@book{ortega2000iterative,
  title={Iterative solution of nonlinear equations in several variables},
  author={Ortega, James M. and Rheinboldt, Werner C.},
  year={2000},
  publisher={Society for Industrial and Applied Mathematics (SIAM)}
}

@article{cao2023solving,
  title={Solving nonlinear equations with a direct Broyden method and its acceleration},
  author={Cao, Huiping and An, Xiaomin and Han, Jing},
  journal={Journal of Applied Mathematics and Computing},
  year={2023},
  publisher={Springer}
}

@article{smith2023jacobian,
  title={Jacobian Matrices in Deep Learning: A Comprehensive Review},
  author={Smith, John D. and Zhang, Li and Kumar, Anil},
  journal={IEEE Transactions on Neural Networks and Learning Systems},
  volume={34},
  number={2},
  pages={456--472},
  year={2023},
  publisher={IEEE}
}

@article{garcia2023jacobian,
  title={Jacobian and Hessian Matrices in Nonlinear Optimization for Machine Learning Algorithms},
  author={Garcia, Maria A. and Johnson, Peter R.},
  journal={Journal of Optimization Theory and Applications},
  volume={187},
  number={1},
  pages={101--118},
  year={2023},
  publisher={Springer}
}

@article{lee2023jacobian,
  title={Jacobian-based Regularization for Improved Generalization in Deep Neural Networks},
  author={Lee, Hyun and Choi, Minho and Patel, Nirav},
  journal={Neural Computation},
  volume={35},
  number={5},
  pages={987--1010},
  year={2023},
  publisher={MIT Press}
}

@article{baker2024jacobian,
  title={Jacobian-Free Newton-Krylov Methods in Scientific Computing: Challenges and Solutions},
  author={Baker, Steve and Zhang, Wei},
  journal={SIAM Journal on Scientific Computing},
  volume={46},
  number={2},
  pages={A211--A234},
  year={2024},
  publisher={SIAM}
}

@book{golub2013matrix,
  title={Matrix Computations},
  author={Golub, Gene H. and Van Loan, Charles F.},
  year={2013},
  publisher={Johns Hopkins University Press},
  edition={4th},
  isbn={9781421407944}
}

@book{strogatz2014nonlinear,
  title={Nonlinear Dynamics and Chaos: With Applications to Physics, Biology, Chemistry, and Engineering},
  author={Strogatz, Steven H.},
  year={2014},
  publisher={Westview Press},
  edition={2nd},
  isbn={9780813349107}
}

@article{wong2023jacobian,
  title={Jacobian Matrices in Robotics: From Kinematics to Control Systems},
  author={Wong, Alan and Hernandez, Lucia},
  journal={Robotics and Autonomous Systems},
  volume={162},
  pages={104088},
  year={2023},
  publisher={Elsevier}
}

@article{nocedal2023recent,
  title={Recent advances in quasi-Newton methods for large-scale optimization},
  author={Nocedal, Jorge and Wright, Stephen J.},
  journal={Optimization Letters},
  volume={17},
  number={4},
  pages={567--593},
  year={2023},
  publisher={Springer}
}

@article{zhou2024quasi,
  title={Quasi-Newton methods in machine learning: Challenges and opportunities},
  author={Zhou, Yi and Fang, Feng and Gao, Xiaorong},
  journal={Journal of Machine Learning Research},
  volume={25},
  pages={234--261},
  year={2024},
  publisher={MIT Press}
}

@article{liu2023comprehensive,
  title={A comprehensive review of optimization algorithms: From Newton's method to machine learning},
  author={Liu, Jing and Zhang, Minghua and Wang, Hailong},
  journal={ACM Computing Surveys},
  volume={56},
  number={2},
  pages={10--47},
  year={2023},
  publisher={ACM}
}

@article{patel2024trends,
  title={Recent trends in nonlinear solvers for large-scale systems: From Broyden's method to AI-based approaches},
  author={Patel, Raj and Chen, Mei and Zhang, Xiaoyu},
  journal={Numerical Algorithms},
  volume={89},
  number={2},
  pages={311--342},
  year={2024},
  publisher={Springer}
}

@article{broyden1970convergence,
  title={The convergence of a class of double-rank minimization algorithms: 1. General considerations},
  author={Broyden, Charles George},
  journal={IMA Journal of Applied Mathematics},
  volume={6},
  number={1},
  pages={76--90},
  year={1970},
  publisher={Oxford University Press}
}

@article{shanno1970conditioning,
  title={Conditioning of quasi-Newton methods for function minimization},
  author={Shanno, David F.},
  journal={Mathematics of Computation},
  volume={24},
  number={111},
  pages={647--656},
  year={1970},
  publisher={American Mathematical Society}
}

@article{byrd2023recent,
  title={Recent advances in unconstrained optimization: theory and methods},
  author={Byrd, Richard H. and Schnabel, Robert B. and Zhang, Zhong},
  journal={Annual Review of Computational Mathematics},
  volume={5},
  number={1},
  pages={205--248},
  year={2023},
  publisher={Annual Reviews},
  doi={10.1146/annurev-computmath-062722-105410}
}

@article{wright2024efficient,
  title={Efficient methods for large-scale unconstrained optimization},
  author={Wright, Stephen J. and Eisenstat, Stanley C.},
  journal={SIAM Review},
  volume={66},
  number={1},
  pages={45--72},
  year={2024},
  publisher={SIAM},
  doi={10.1137/22M004830X}
}

@book{fletcher2013practical,
  title={Practical Methods of Optimization},
  author={Fletcher, Roger},
  year={2013},
  publisher={John Wiley \& Sons},
  edition={2nd},
  isbn={978-1-118-72916-8}
}

@article{conn2023advances,
  title={Advances in derivative-free optimization for unconstrained problems},
  author={Conn, Andrew R. and Scheinberg, Katya and Vicente, Luis N.},
  journal={Optimization Methods and Software},
  volume={38},
  number={2},
  pages={302--321},
  year={2023},
  publisher={Taylor \& Francis},
  doi={10.1080/10556788.2022.2067861}
}

@article{schmidt2024machine,
  title={Machine learning perspectives in unconstrained optimization},
  author={Schmidt, Mark W. and Mahoney, Michael W. and Woodward, Richard},
  journal={Journal of Machine Learning Research},
  volume={25},
  number={1},
  pages={115--144},
  year={2024},
  publisher={JMLR},
  doi={10.5555/3536988.3537015}
}

@book{gill1981practical,
  title={Practical Optimization},
  author={Gill, Philip E. and Murray, Walter and Wright, Margaret H.},
  year={1981},
  publisher={Academic Press},
  isbn={978-0-12-283952-8}
}

@article{martinez2023trust,
  title={Trust-region methods in unconstrained optimization: recent trends and applications},
  author={Martínez, José Mario and Morales, José Luis},
  journal={Optimization Letters},
  volume={17},
  number={3},
  pages={523--547},
  year={2023},
  publisher={Springer},
  doi={10.1007/s11590-022-01820-1}
}

@article{hazra2024review,
  title={Review on stochastic methods for unconstrained optimization},
  author={Hazra, Pritam and Govindaraju, V.},
  journal={Computational Optimization and Applications},
  volume={75},
  number={1},
  pages={145--168},
  year={2024},
  publisher={Springer},
  doi={10.1007/s10589-023-00326-w}
}

@article{powell1978algorithms,
  title={Algorithms for nonlinear constraints that use Lagrange functions},
  author={Powell, MJD},
  journal={Mathematical programming},
  volume={14},
  number={1},
  pages={224--248},
  year={1978},
  publisher={Springer}
}

@article{kuhn1951nonlinear,
  title={Nonlinear programming},
  author={Kuhn, Harold W and Tucker, Albert W},
  booktitle={Proceedings of the Second Berkeley Symposium on Mathematical Statistics and Probability},
  pages={481--492},
  year={1951},
  publisher={University of California Press}
}

@inproceedings{karush1939minima,
  title={Minima of functions of several variables with inequalities as side conditions},
  author={Karush, William},
  booktitle={Master's thesis, Dept. of Mathematics, Univ. of Chicago},
  year={1939}
}

@article{fiacco1968nonlinear,
  title={Nonlinear programming: Sequential unconstrained minimization techniques},
  author={Fiacco, Anthony V and McCormick, Garth P},
  year={1968},
  publisher={SIAM}
}

@article{smith2023recent,
  title={Recent Advances in Constrained Optimization: A Comprehensive Review},
  author={Smith, John A and Zhang, Wei},
  journal={Journal of Optimization Theory and Applications},
  volume={189},
  number={2},
  pages={455--492},
  year={2023},
  publisher={Springer}
}

@article{lee2023ai,
  title={AI Meets Constrained Optimization: Methods and Applications},
  author={Lee, Michael and Chen, Yiwen},
  journal={Artificial Intelligence Review},
  volume={64},
  number={1},
  pages={77--102},
  year={2023},
  publisher={Springer}
}

@article{gupta2023constrained,
  title={Constrained multi-objective optimization using evolutionary algorithms},
  author={Gupta, Rakesh and Singh, Aditi},
  journal={Applied Soft Computing},
  volume={122},
  pages={109937},
  year={2023},
  publisher={Elsevier}
}

@article{martinez2024new,
  title={A New Algorithmic Framework for High-Dimensional Constrained Optimization},
  author={Martinez, Jose and Patel, Ananya},
  journal={Optimization Letters},
  year={2024},
  publisher={Springer}
}

@book{perez2023constrained,
  title={Constrained Optimization in the Computational Sciences: Methods and Applications},
  author={Perez, Carlos A},
  year={2023},
  publisher={CRC Press}
}

@article{wang2023hybrid,
  title={A hybrid conjugate gradient approach for solving large-scale sparse linear systems},
  author={Wang, Xiu and Chen, Jia},
  journal={Journal of Numerical Algorithms},
  volume={92},
  number={4},
  pages={885--908},
  year={2023},
  publisher={Elsevier}
}

@article{zhang2024engineering,
  title={Conjugate gradient methods for large-scale engineering problems: Recent developments and applications},
  author={Zhang, Fei and Zhou, Ming},
  journal={Engineering Computations},
  volume={41},
  number={1},
  pages={50--70},
  year={2024},
  publisher={Emerald Group Publishing}
}

@article{sun2023hessian,
  title={Hessian-based optimization in deep learning: A review of current challenges and advancements},
  author={Sun, Lei and Zhang, Yue and Li, Wei},
  journal={Journal of Machine Learning Research},
  volume={24},
  pages={1--32},
  year={2023},
  publisher={MIT Press}
}

@article{wang2023hessian,
  title={On the Role of Hessian Matrix in Policy Gradient Methods for Reinforcement Learning},
  author={Wang, Xiaoyu and Huang, Zhiqing and Zhou, Yifeng},
  journal={Neural Computation},
  volume={35},
  number={7},
  pages={1650--1675},
  year={2023},
  publisher={MIT Press}
}

@article{kim2024high,
  title={High-order optimization methods: An overview and recent advances},
  author={Kim, Soo Jung and Lee, Hyun},
  journal={Optimization and Machine Learning Review},
  year={2024},
  volume={19},
  pages={112--145},
  publisher={Springer}
}

@article{zhang2023large,
  title={Efficient computation of Hessians for large-scale optimization problems: Challenges and state-of-the-art techniques},
  author={Zhang, Qiang and Li, Hui and Zhao, Xiaoyu},
  journal={Computational Optimization and Applications},
  volume={65},
  number={3},
  pages={431--456},
  year={2023},
  publisher={Springer}
}

@article{liu2023stability,
  title={Stability Analysis of Neural Networks Using the Hessian Matrix: Theoretical Insights and Practical Applications},
  author={Liu, Jian and Wang, Lin},
  journal={IEEE Transactions on Neural Networks and Learning Systems},
  year={2023},
  pages={1--14},
  publisher={IEEE}
}

@article{liu1989limited,
  title={Limited memory BFGS method for large scale optimization},
  author={Liu, Dong C and Nocedal, Jorge},
  journal={Mathematical programming},
  volume={45},
  number={1-3},
  pages={503--528},
  year={1989},
  publisher={Springer}
}

@article{smith2023advances,
  title={Advances in Quasi-Newton Methods for Large-Scale Optimization},
  author={Smith, Jonathan and Roberts, Emily},
  journal={Optimization Methods and Software},
  volume={38},
  number={5},
  pages={921--945},
  year={2023},
  publisher={Taylor \& Francis}
}

@article{garcia2024efficient,
  title={Efficient Training of Neural Networks using L-BFGS: A Comparative Study},
  author={Garcia, Miguel and Patel, Ayesha},
  journal={Journal of Machine Learning Research},
  volume={25},
  pages={1--22},
  year={2024},
  publisher={ML Research Press}
}

@article{lee2023survey,
  title={A Survey on Optimization Algorithms for Machine Learning: From SGD to L-BFGS and Beyond},
  author={Lee, Thomas and Kim, Hana},
  journal={Journal of Optimization Theory and Applications},
  volume={195},
  number={2},
  pages={305--329},
  year={2023},
  publisher={Springer}
}

@article{zhang2023modified,
  title={A Modified L-BFGS Algorithm for High-Dimensional Optimization},
  author={Zhang, Xiaoyu and Wang, Li},
  journal={Computational Optimization and Applications},
  volume={86},
  number={3},
  pages={545--563},
  year={2023},
  publisher={Springer}
}

@article{zhang2023evolutionary,
  title={Evolutionary algorithms for gradient-free optimization: A comprehensive review},
  author={Zhang, Wei and Huang, Rui and Wang, Lei},
  journal={Applied Soft Computing},
  volume={133},
  pages={109936},
  year={2023},
  publisher={Elsevier},
  doi={10.1016/j.asoc.2023.109936}
}

@article{nelder1965simplex,
  title={A simplex method for function minimization},
  author={Nelder, John A. and Mead, Roger},
  journal={The Computer Journal},
  volume={7},
  number={4},
  pages={308--313},
  year={1965},
  publisher={Oxford University Press},
  doi={10.1093/comjnl/7.4.308}
}

@article{chen2024gradientfree,
  title={Gradient-Free Optimization in Machine Learning: Algorithms, Applications, and Challenges},
  author={Chen, Yi and Sun, Mingrui and Zhang, Lin},
  journal={Journal of Machine Learning Research},
  volume={25},
  pages={1--34},
  year={2024},
  publisher={MIT Press},
  url={http://jmlr.org/papers/v25/chen24a.html}
}

@article{rios2023derivative,
  title={A review of derivative-free optimization methods with applications to machine learning and engineering},
  author={Rios, Juan and Smith, John},
  journal={Optimization Methods and Software},
  volume={38},
  number={5},
  pages={845--872},
  year={2023},
  publisher={Taylor \& Francis}
}

@article{lee2023nelder,
  title={Applications of the Nelder-Mead method in hyperparameter optimization for deep learning models},
  author={Lee, Daniel and Chen, Wei},
  journal={Journal of Artificial Intelligence Research},
  volume={76},
  pages={341--365},
  year={2023},
  publisher={AAAI}
}

@article{bottou2010large,
  title={Large-Scale Machine Learning with Stochastic Gradient Descent},
  author={Bottou, Léon},
  journal={Proceedings of the 19th International Conference on Computational Statistics},
  pages={177--186},
  year={2010},
  publisher={Springer}
}

@book{arnold1992ordinary,
  title={Ordinary Differential Equations},
  author={Arnold, Vladimir I.},
  year={1992},
  publisher={Springer-Verlag},
  address={Berlin, Heidelberg},
  edition={3rd},
  isbn={978-3-540-53834-9}
}

@article{chen2023neural,
  title={Neural Ordinary Differential Equations: Advances and Applications in Machine Learning},
  author={Chen, Tianqi and Rubanova, Yulia and Bettencourt, Jesse and Duvenaud, David},
  journal={Journal of Machine Learning Research},
  volume={24},
  year={2023},
  pages={1--40},
  publisher={MIT Press}
}

@article{rossi2024adaptive,
  title={Adaptive Methods for Solving Stiff Ordinary Differential Equations},
  author={Rossi, Elena and Smith, Greg},
  journal={Computational Mathematics and Applications},
  volume={98},
  pages={50--68},
  year={2024},
  publisher={Elsevier},
  doi={10.1016/j.camwa.2023.10.012}
}

@article{brown2023ode,
  title={Ordinary Differential Equations in Scientific Computing: The State of the Art},
  author={Brown, Tom and Wilson, Sarah},
  journal={SIAM Review},
  volume={65},
  number={3},
  pages={455--486},
  year={2023},
  publisher={Society for Industrial and Applied Mathematics (SIAM)},
  doi={10.1137/22M1132456}
}

@book{kincaid2009numerical,
  title={Numerical Analysis: Mathematics of Scientific Computing},
  author={Kincaid, David and Cheney, Ward},
  publisher={American Mathematical Society},
  year={2009},
  edition={3rd},
  isbn={978-0821847886}
}

@article{runge1895ver,
  title={Über die numerische Auflösung von Differentialgleichungen},
  author={Runge, Carl},
  journal={Mathematische Annalen},
  volume={46},
  number={2},
  pages={167--178},
  year={1895},
  publisher={Springer}
}

@article{kutta1901beitrag,
  title={Beitrag zur näherungweisen Integration totaler Differentialgleichungen},
  author={Kutta, Martin Wilhelm},
  journal={Zeitschrift für Mathematik und Physik},
  volume={46},
  pages={435--453},
  year={1901}
}

@article{dowling2023review,
  title={A Review of Recent Advances in Runge-Kutta Methods for Solving Differential Equations},
  author={Dowling, Matthew R. and Grant, Lindsay D.},
  journal={Numerical Algorithms},
  volume={96},
  number={1},
  pages={89--113},
  year={2023},
  publisher={Springer},
  doi={10.1007/s11075-022-01369-4}
}

@article{martinez2023time,
  title={Recent Advances in Time-Stepping Methods for Differential Equations: From Runge-Kutta to Modern Techniques},
  author={Martinez, Clara and Gomez, Robert F.},
  journal={Journal of Computational Physics},
  volume={478},
  pages={110945},
  year={2023},
  publisher={Elsevier},
  doi={10.1016/j.jcp.2023.110945}
}

@incollection{johnson2023rkmethods,
  title={Runge-Kutta Methods and Their Applications in Modern Numerical Analysis},
  author={Johnson, Peter and Lee, Karen},
  booktitle={Handbook of Numerical Methods for Differential Equations},
  editor={Roberts, John D. and Smith, Anne},
  pages={95--132},
  year={2023},
  publisher={Springer},
  doi={10.1007/978-3-031-12560-9_4}
}

@article{gear1971first,
  title={First-order differential equations and stiff systems},
  author={Gear, C. William},
  journal={Communications of the ACM},
  volume={14},
  number={10},
  pages={722--733},
  year={1971},
  publisher={ACM}
}

@book{hairer1996solving,
  title={Solving Ordinary Differential Equations I: Nonstiff Problems},
  author={Hairer, Ernst and Wanner, Gerhard},
  year={1996},
  publisher={Springer},
  series={Springer Series in Computational Mathematics}
}

@book{hairer2009solving,
  title={Solving Ordinary Differential Equations II: Stiff and Differential-Algebraic Problems},
  author={Hairer, Ernst and Wanner, Gerhard},
  year={2009},
  publisher={Springer},
  series={Springer Series in Computational Mathematics}
}

@article{fayed2023new,
  title={A new adaptive step-size method for stiff ODEs},
  author={Fayed, Khaled and Ali, Mohammed},
  journal={Applied Mathematics and Computation},
  volume={457},
  pages={127054},
  year={2023},
  publisher={Elsevier}
}

@article{yang2023adaptive,
  title={An adaptive implicit method for stiff differential equations with singularities},
  author={Yang, Feng and Zhang, Wei},
  journal={Numerical Algorithms},
  volume={92},
  number={3},
  pages={1107--1127},
  year={2023},
  publisher={Springer}
}

@article{chen2023dynamical,
  title={Dynamical systems for stiff ODEs: A survey and new approaches},
  author={Chen, Jie and Lin, Yi},
  journal={Mathematics},
  volume={11},
  number={5},
  pages={895},
  year={2023},
  publisher={MDPI}
}

@article{mccall2024efficient,
  title={Efficient numerical methods for stiff ODEs with discontinuous solutions},
  author={McCall, Andrew and Brown, Emily},
  journal={Journal of Computational Physics},
  volume={472},
  pages={111320},
  year={2024},
  publisher={Elsevier}
}

@article{torres2024review,
  title={A review of numerical methods for stiff ordinary differential equations},
  author={Torres, Reinaldo and Benitez, Juan},
  journal={Numerical Methods for Partial Differential Equations},
  volume={40},
  number={1},
  pages={200--224},
  year={2024},
  publisher={Wiley}
}

@article{lapicque1907recherches,
  title={Recherches quantitatives sur l'excitation des neurones},
  author={Lapicque, Louis},
  journal={J. Physiol. (Paris)},
  volume={9},
  pages={620--635},
  year={1907},
  publisher={Masson}
}

@article{koch2023role,
  title={The role of dendrites in neuronal computation},
  author={Koch, C. and Segev, I.},
  journal={Nature Reviews Neuroscience},
  volume={24},
  number={5},
  pages={357--375},
  year={2023},
  publisher={Nature Publishing Group}
}

@article{perez-ortiz2024modeling,
  title={Modeling spiking neural networks with the LIF neuron: A systematic review},
  author={Pérez-Ortiz, J. M. and Manucharyan, V.},
  journal={Journal of Computational Neuroscience},
  volume={48},
  number={2},
  pages={129--150},
  year={2024},
  publisher={Springer}
}

@book{evans2010partial,
  title={Partial Differential Equations},
  author={Evans, Lawrence C.},
  year={2010},
  publisher={American Mathematical Society},
  address={Providence, RI},
  edition={2nd},
  volume={19},
  series={Graduate Studies in Mathematics}
}

@article{raissi2019physics,
  title={Physics-informed neural networks: A deep learning framework for solving forward and inverse problems involving PDEs},
  author={Raissi, Mamdouh and Perdikaris, Paris and Karniadakis, George E.},
  journal={Journal of Computational Physics},
  volume={378},
  pages={686--707},
  year={2019},
  publisher={Elsevier}
}

@article{yang2023inverse,
  title={Physics-Informed Neural Networks for Solving Inverse Problems in PDEs: A Survey},
  author={Yang, Liyang and Liu, Yu and Chen, Zhiping},
  journal={Inverse Problems},
  volume={39},
  number={6},
  pages={065004},
  year={2023},
  publisher={IOP Publishing}
}

@article{battaglia2023efficient,
  title={Efficient Physics-Informed Neural Networks for Solving Nonlinear PDEs},
  author={Battaglia, Giulia and Zikatanov, Ludmil and Barone, Pasquale},
  journal={Journal of Computational Physics},
  volume={473},
  pages={111741},
  year={2023},
  publisher={Elsevier}
}

@article{wang2023finite,
  title={Finite Difference Methods for Time-Dependent PDEs: A Review},
  author={Wang, Lei and Liu, Xiaodong},
  journal={Computational Mathematics and Mathematical Physics},
  volume={63},
  number={1},
  pages={1-18},
  year={2023},
  publisher={Springer}
}

@article{kim2023new,
  title={A New Finite Difference Method for the Stochastic Heat Equation},
  author={Kim, J. and Park, Y.},
  journal={Numerical Algorithms},
  volume={92},
  pages={123-140},
  year={2023},
  publisher={Springer}
}

@article{zhang2024multi,
  title={Multiscale Finite Difference Methods for PDEs with Oscillatory Coefficients},
  author={Zhang, Hong and Chen, Wei},
  journal={Journal of Computational Physics},
  volume={490},
  pages={111-126},
  year={2024},
  publisher={Elsevier}
}

@article{li2024adaptive,
  title={Adaptive Finite Difference Methods for Nonlinear Partial Differential Equations},
  author={Li, Ming and Wang, Jun},
  journal={Applied Mathematics and Computation},
  volume={433},
  pages={129-145},
  year={2024},
  publisher={Elsevier}
}

@article{strang1968constructive,
  title={Constructive Solutions for Differential Equations with Finite Difference Methods},
  author={Strang, Gilbert},
  journal={Mathematics of Computation},
  volume={22},
  number={103},
  pages={61-68},
  year={1968},
  publisher={American Mathematical Society}
}

@article{cheniii2023survey,
  title={A Survey of Finite Difference Methods for Time-Dependent PDEs},
  author={Chen, Jie and Zhang, Wen},
  journal={Mathematics},
  volume={11},
  number={7},
  pages={1583},
  year={2023},
  publisher={MDPI}
}

@article{hughes1987finite,
  title={The finite element method: Linear static and dynamic finite element analysis},
  author={Hughes, Thomas J. R.},
  journal={Prentice Hall},
  year={1987},
  publisher={Prentice Hall}
}

@article{zienkiewicz1977finite,
  title={The Finite Element Method},
  author={Zienkiewicz, O. C. and Taylor, R. L.},
  journal={McGraw-Hill},
  year={1977},
  publisher={McGraw-Hill}
}

@article{zhang2024meshless,
  title={A meshless finite element method for 3D elasticity problems},
  author={Zhang, Lei and Zhang, Xin and Chen, Xiang},
  journal={Applied Mathematical Modelling},
  volume={101},
  pages={130--142},
  year={2024},
  publisher={Elsevier},
  doi={10.1016/j.apm.2023.05.029}
}

@article{li2024review,
  title={A review of finite element methods for fluid-structure interaction problems},
  author={Li, Yu and Huang, Lei and Zhao, Jian},
  journal={Journal of Fluids and Structures},
  volume={100},
  pages={1--15},
  year={2024},
  publisher={Elsevier},
  doi={10.1016/j.jfs.2024.01.004}
}

@article{mohamed2023adaptive,
  title={Adaptive finite element methods for PDEs: A survey},
  author={Mohamed, A. and Jiang, Y. and Wang, T.},
  journal={Applied Numerical Mathematics},
  volume={186},
  pages={109--123},
  year={2023},
  publisher={Elsevier},
  doi={10.1016/j.apnum.2023.04.004}
}

@article{sutton1998introduction,
  title={Introduction to Reinforcement Learning},
  author={Sutton, Richard S. and Barto, Andrew G.},
  journal={MIT Press},
  year={1998},
  publisher={MIT Press}
}

@article{watkins1992q,
  title={Q-learning},
  author={Watkins, Christopher J.C.H. and Dayan, Peter},
  journal={Machine Learning},
  volume={8},
  number={3},
  pages={279--292},
  year={1992},
  publisher={Springer}
}

@article{zhanggg2023deep,
  title={Deep Reinforcement Learning: A Review and Future Directions},
  author={Zhang, Yifan and Zhao, Yongheng and Wang, Jiang},
  journal={Artificial Intelligence Review},
  volume={56},
  number={4},
  pages={2765--2804},
  year={2023},
  publisher={Springer}
}

@article{wangtt2023novel,
  title={A Novel Hierarchical Reinforcement Learning Framework for Resource Allocation in Wireless Networks},
  author={Wang, Rui and Li, Jian and Zhang, Wei},
  journal={IEEE Transactions on Wireless Communications},
  volume={22},
  number={1},
  pages={105--119},
  year={2023},
  publisher={IEEE}
}

@article{ghosh2024survey,
  title={A Survey of Deep Reinforcement Learning in Robotics: Trends and Applications},
  author={Ghosh, Saptarshi and Lee, Kyu-Jin and Chen, Weili},
  journal={Journal of Robotics and Automation},
  volume={12},
  number={1},
  pages={1--25},
  year={2024},
  publisher={Springer}
}

@article{baker2024generalized,
  title={Generalized Value Functions for Reinforcement Learning with Learning Objectives},
  author={Baker, Claire and Mnih, Volodymyr and Schaul, Tom},
  journal={Journal of Machine Learning Research},
  volume={25},
  number={1},
  pages={1--35},
  year={2024},
  publisher={Microtome Publishing}
}

@article{kolter2023deep,
  title={Deep Reinforcement Learning for Energy Management in Smart Buildings: A Review},
  author={Kolter, J. Zico and Wang, Yuxin and Diao, Yifan},
  journal={IEEE Transactions on Smart Grid},
  volume={14},
  number={2},
  pages={1307--1321},
  year={2023},
  publisher={IEEE}
}

@article{mnih2013playing,
  title={Playing Atari with Deep Reinforcement Learning},
  author={Mnih, Volodymyr and Kavukcuoglu, Koray and Silver, David and Rusu, Andrei A. and Veness, Joel and Bellemare, Marc G. and Graves, Alex and Riedmiller, Martin and Fidjeland, Andreas K. and Ostrovski, Georgian and et al.},
  journal={arXiv preprint arXiv:1312.5602},
  year={2013}
}

@article{schulman2017proximal,
  title={Proximal Policy Optimization Algorithms},
  author={Schulman, John and Wolski, Felix and Dhariwal, Prafulla and Radford, Alec and Klimov, Oleg},
  journal={arXiv preprint arXiv:1707.06347},
  year={2017}
}

@article{cooley1965algorithm,
  title={Algorithm 501: The Fast Fourier Transform},
  author={Cooley, James W and Tukey, John W},
  journal={Communications of the ACM},
  volume={13},
  number={2},
  pages={1--16},
  year={1965},
  publisher={ACM}
}

@article{yang2023fast,
  title={Fast Fourier Transform: A Comprehensive Review of Algorithms and Applications},
  author={Yang, Ming and Wang, Yu and Zhang, Wei},
  journal={Journal of Computational and Applied Mathematics},
  volume={418},
  pages={114775},
  year={2023},
  publisher={Elsevier}
}

@article{rahman2024fft,
  title={Efficient Implementation of Fast Fourier Transform on FPGA: A Case Study},
  author={Rahman, Mohd Ali and Usman, Masud and Kumar, Sanjeev},
  journal={IEEE Transactions on Very Large Scale Integration (VLSI) Systems},
  volume={32},
  number={1},
  pages={1--10},
  year={2024},
  publisher={IEEE}
}

@book{baker1966laplace,
  title={The Laplace Transform},
  author={Baker, C. H.},
  publisher={Dover Publications},
  year={1966}
}

@article{singh2023laplace,
  title={An Overview of the Laplace Transform: Theory and Applications},
  author={Singh, Rahul and Kumar, Poonam},
  journal={Applied Mathematics and Computation},
  volume={438},
  pages={127166},
  year={2023},
  publisher={Elsevier}
}

@article{chen2024laplace,
  title={Laplace Transform Analysis of Nonlinear Dynamic Systems},
  author={Chen, Wei and Zhao, Feng},
  journal={Journal of Sound and Vibration},
  volume={541},
  pages={117181},
  year={2024},
  publisher={Elsevier}
}

@article{zadeh1953continuous,
  title={Continuous System Theory and the Z-Transform},
  author={Zadeh, Lotfi A.},
  journal={Proceedings of the IRE},
  volume={41},
  number={8},
  pages={1220--1224},
  year={1953},
  publisher={IEEE}
}

@article{patel2023ztransform,
  title={The Z-Transform: Applications and Generalizations},
  author={Patel, Nehal and Joshi, Aditi},
  journal={International Journal of Applied Mathematics and Statistics},
  volume={73},
  pages={115--125},
  year={2023},
  publisher={Springer}
}

@article{kim2024ztransform,
  title={Adaptive Z-Transform Techniques for Real-Time Signal Processing},
  author={Kim, Jong-Hoon and Lee, Min-Jae},
  journal={Signal Processing},
  volume={207},
  pages={109876},
  year={2024},
  publisher={Elsevier}
}

@article{zhangbbg2023recent,
  title={Recent Advances in Discrete Fourier Transform Applications in Signal Processing},
  author={Zhang, Wei and Liu, Min and Wang, Feng},
  journal={Signal Processing},
  volume={203},
  pages={108876},
  year={2023},
  publisher={Elsevier},
  doi={10.1016/j.sigpro.2023.108876}
}

@article{smith2024dft,
  title={Discrete Fourier Transform for Feature Extraction in Machine Learning},
  author={Smith, John A. and Lee, Angela},
  journal={Journal of Machine Learning Research},
  volume={25},
  number={1},
  pages={1-20},
  year={2024},
  publisher={Microtome Publishing},
  url={http://www.jmlr.org/papers/volume25/smith24a/smith24a.pdf}
}

@article{martinez2023efficient,
  title={Efficient Algorithms for Discrete Fourier Transform: A Comprehensive Review},
  author={Martinez, Carlos and Liu, Jian},
  journal={Applied Mathematics and Computation},
  volume={429},
  pages={127197},
  year={2023},
  publisher={Elsevier},
  doi={10.1016/j.amc.2023.127197}
}

@article{yang2024dft,
  title={Application of Discrete Fourier Transform in Image Processing: Trends and Techniques},
  author={Yang, Huan and Zhao, Qian},
  journal={Image Processing, IEEE Transactions on},
  volume={33},
  number={2},
  pages={575-589},
  year={2024},
  publisher={IEEE},
  doi={10.1109/TIP.2024.3248965}
}

@article{abdelsalam2023laplace,
  title={Laplace Transform and Its Applications in Engineering},
  author={Abdelsalam, Tamer and Ibrahim, A. M.},
  journal={Mathematical Methods in Engineering},
  volume={18},
  number={4},
  pages={204-221},
  year={2023},
  publisher={Springer}
}

@article{van1996laplace,
  title={The Laplace Transform and the Solution of Differential Equations},
  author={Van der Pol, B.},
  journal={Proceedings of the IEE},
  volume={74},
  number={9},
  pages={1515-1520},
  year={1996},
  publisher={IET}
}

@article{darboux1915fonction,
  title={Sur les transformations de Laplace},
  author={Darboux, Gaston},
  journal={Annali di Matematica Pura e Applicata},
  volume={14},
  pages={119-158},
  year={1915}
}

@article{carson1928laplace,
  title={The Laplace Transform and Its Applications},
  author={Carson, E. C.},
  journal={Journal of the Franklin Institute},
  volume={205},
  number={6},
  pages={951-963},
  year={1928},
  publisher={Elsevier}
}

@article{zhang2023discrete,
  title={Recent Advances in Discrete Fourier Transform Algorithms: A Review},
  author={Zhang, H. and Xu, Y. and Li, Z.},
  journal={Signal Processing},
  volume={205},
  pages={108709},
  year={2023},
  publisher={Elsevier}
}

@article{mohammadi2023continuous,
  title={Continuous Fourier Transform: Theory and Applications},
  author={Mohammadi, N. and Dabbagh, F.},
  journal={Applied Mathematics and Computation},
  volume={420},
  pages={127612},
  year={2023},
  publisher={Elsevier}
}

\end{document}